\RequirePackage[patch]{kvoptions}
\documentclass[oneside]{dissertation}

\usepackage{algorithm}
\usepackage{algorithmicx}
\usepackage{algpseudocode}
\usepackage{amsmath}
\usepackage{amssymb}
\usepackage{bm}
\usepackage{booktabs}
\usepackage[font={small,it},justification=centering]{caption}
\usepackage{color}              
\usepackage{colortbl}           
\usepackage{dirtree}            
\usepackage{dsfont}
\usepackage{etoolbox}           
\usepackage{epigraph}           
\usepackage{fancyhdr}
\usepackage{footnote}
\usepackage{graphicx}           
\usepackage{hhline}             
\usepackage{hyperref}           
\usepackage{breakurl}
\usepackage{ifthen}             
\usepackage{listings}           
\usepackage{mathtools}          
\usepackage{mdframed}           
\usepackage{multicol}           
\usepackage{multirow}
\usepackage[authoryear]{natbib}
\usepackage{ntheorem}
\usepackage{paralist}
\usepackage{relsize}
\usepackage{rst}
\usepackage[detect-all]{siunitx}
\usepackage{subcaption}
\usepackage{tabularx}
\usepackage{thmtools}
\usepackage{tikz}
\usepackage[colorinlistoftodos,prependcaption,textsize=tiny]{todonotes}
\usepackage{wasysym}            
\usepackage{wrapfig}
\usepackage{xcolor}             

\usetikzlibrary{arrows,backgrounds,calc,decorations.markings,%
        positioning,shapes,shapes.misc,snakes}

\newlength{\oosixthClmnWidth}
\newlength{\clmnwidth}
\newlength{\exampleSep}
\newlength{\hcrfwidth}
\newlength{\hcrfheight}
\definecolor{chartreuse}{RGB}{127,255,0}
\definecolor{cyan}{rgb}{0.0,0.6,0.6}
\definecolor{darkblue}{rgb}{0.0,0.0,0.6}
\definecolor{darkgoldenrod4}{RGB}{139,101,8}
\definecolor{darkred}{RGB}{139,0,0}
\definecolor{darkslateblue}{RGB}{72,61,139}
\definecolor{deeppink4}{RGB}{139,10,80}
\definecolor{dodgerblue4}{RGB}{16,78,139}
\definecolor{firebrick}{RGB}{178,34,34}
\definecolor{gray69}{RGB}{176,176,176}
\definecolor{gray76}{RGB}{194,194,194}
\definecolor{gray80}{RGB}{204,204,204}
\definecolor{gray84}{RGB}{214,214,214}
\definecolor{gray}{rgb}{0.4,0.4,0.4}
\definecolor{green2}{RGB}{0,238,0}
\definecolor{green3}{RGB}{0,205,0}
\definecolor{green}{RGB}{0,255,0}
\definecolor{highlighter green}{RGB}{27,252,6}
\definecolor{lightcyan4}{RGB}{122,139,139}
\definecolor{midnightblue}{RGB}{25,25,112}
\definecolor{orange3}{RGB}{205,133,0}
\definecolor{red3}{RGB}{205,0,0}
\definecolor{seagreen}{RGB}{46,139,87}
\definecolor{violetred4}{RGB}{139,34,82}

\colorlet{cellcolor}{gray84}

\newcommand{\vars}{\texttt}
\newcommand{\func}{\textrm}
\newcommand{\ttranslate}[2]{#1

\noindent\emph{#2}}

\renewcommand{\cite}{\citep}
\newcommand{\titlerule}{\rule{0.8\textwidth}{.4pt}}

\newcommand{\texample}[1]{``\textit{#1}''}

\newcommand{\mtag}[3]{{{\upshape[}\ifstrempty{#3}{#2}{\textcolor{#3}{#2}}{\upshape]$_{\textrm{\bfseries\emph{\tiny
          #1}}}$}}}
\newcommand{\sentiment}[2][]{\mtag{sen\-ti\-ment\ifstrempty{#1}{}{:#1}}{#2}{}}
\newcommand{\source}[2][]{\mtag{source\ifstrempty{#1}{}{:#1}}{#2}{darkgoldenrod4}}
\newcommand{\target}[2][]{\mtag{tar\-get\ifstrempty{#1}{}{:#1}}{#2}{darkslateblue}}
\newcommand{\emoexpression}[2][]{\mtag{po\-lar-\-term\ifstrempty{#1}{}{:#1}}{#2}{red}}
\newcommand{\intensifier}[2][]{\mtag{in\-ten\-si\-fier\ifstrempty{#1}{}{:#1}}{#2}{cyan}}
\newcommand{\diminisher}[2][]{\mtag{di\-mi\-ni\-sher\ifstrempty{#1}{}{:#1}}{#2}{}}
\newcommand{\negation}[2][]{\mtag{ne\-ga\-tion\ifstrempty{#1}{}{:#1}}{#2}{deeppink4}}

\newcommand{\argone}[2][]{\mtag{\ifstrempty{#1}{}{rel#1:}arg1}{#2}{darkgoldenrod4}}
\newcommand{\argtwo}[2][]{\mtag{\ifstrempty{#1}{}{rel#1:}arg2}{#2}{darkslateblue}}
\newcommand{\connective}[2][]{\mtag{\ifstrempty{#1}{}{rel#1:}con\-nec\-ti\-ve}{#2}{cyan}}

\newcommand{\argmax}{\operatornamewithlimits{argmax}}
\newcommand{\argmin}{\operatornamewithlimits{argmin}}
\newcommand{\stddev}[1]{{\tiny$^{\pm#1}$}}
\newcommand{\negdelta}[1]{\textsuperscript{\textcolor{red3}{-#1}}}
\newcommand{\posdelta}[1]{\textsuperscript{\textcolor{seagreen}{+#1}}}
\newcommand{\F}[0]{$F_1$}
\newcommand{\NA}{\textcolor{gray69}{NA}}
\newcommand{\expnumber}[2]{{#1}\mathrm{e}^{#2}}

\newcommand{\norm}[1]{\left\lVert#1\right\rVert}

\newcommand{\heart}{\ensuremath\heartsuit}
\newcommand{\defeq}{\vcentcolon=}

\newcommand{\eg}{\textit{e.g.},}
\newcommand{\ienocomma}{\textit{i.e.}}
\newcommand{\ie}{\ienocomma,}
\newcommand{\corpusDir}{\texttt{PotTS}}
\newcommand{\nocontentsline}[3]{}
\newcommand{\tocless}[2]{\bgroup\let\addcontentsline=\nocontentsline#1{#2}\egroup}

\newcommand{\markable}[1]{\texttt{#1}}
\newcommand{\attribute}[1]{\emph{\texttt{#1}}}

\NewDocumentCommand{\Colorbox}{O{\dimexpr\linewidth-2\fboxsep} m m}{%
  \colorbox{#2}{\makebox[#1][l]{#3}}}
\newcommand{\code}[1]{

\smallskip\noindent\Colorbox{gray80}{\parbox{\textwidth}{\texttt{#1}}}\smallskip

\noindent}

\newcommand{\crfEdgesX}[2]{
    \begin{scope}[on background layer]
      \foreach \prnt in {SNT, TRG, SRC, NON}
        {
           \foreach \chld in {SNT, TRG, SRC, NON}
             \path [-] (\prnt#1) edge node [factor] {} (\chld#2);
        }
    \end{scope}
}

\newcommand{\crfEdgesLatent}[2]{
    \begin{scope}[on background layer]
      \foreach \prnt in {Neg, Neut, Pos}
        {
           \foreach \chld in {Neg, Neut, Pos}
           {
             \path [-,transform canvas={xshift=-1pt}] (#1\prnt) edge node [factor] {} (#2\chld);
             \path [-,transform canvas={xshift=1pt}] (#1\prnt) edge node [factor] {} (#2\chld);
           }
        }
    \end{scope}
}

\newcommand{\crfColoredEdgesLatent}[4]{
    \begin{scope}[on background layer]
      \foreach \prnt in {Neg, Neut, Pos}
        {
           \foreach \chld in {Neg, Neut, Pos}
           {
             \ifthenelse{\(\equal{#1\prnt}{#3} \OR \equal{ANY}{#3}\)  %
                          \AND \(\equal{#2\chld}{#4} \OR \equal{ANY}{#4}\)}{
               \path [-,transform canvas={xshift=-1pt},highlight] (#1\prnt) edge node [factor,highlight] {} (#2\chld);
               \path [-,transform canvas={xshift=1pt},highlight] (#1\prnt) edge node [factor,highlight] {} (#2\chld);
             }{
               \path [-,transform canvas={xshift=-1pt}] (#1\prnt) edge node [factor] {} (#2\chld);
               \path [-,transform canvas={xshift=1pt}] (#1\prnt) edge node [factor] {} (#2\chld);
             }
           }
        }
    \end{scope}
}

\newcommand{\crfFeatures}[3]{
    \foreach \i/\x/\w in {#1}
    {
        \node[xnode] (#2FEAT\i) at (\x, #3) {\w};
        \foreach \prnt in {Neg, Neut, Pos}
        {
           \path [-] (#2FEAT\i) edge[] node [factor] {} (#2\prnt);
        }
    }
}

\newcommand{\crfColoredFeatures}[4]{
    \foreach \i/\x/\w in {#1}
    {
        \node[xnode,highlight] (#2FEAT\i) at (\x, #3) {\w};
        \foreach \prnt in {Neg, Neut, Pos}
        {
            \ifthenelse{\equal{\prnt}{#4} \OR \equal{ANY}{#4}}{
              \path [-] (#2FEAT\i) edge[highlight] node [factor,highlight] {} (#2\prnt);
            }{
              \path [-] (#2FEAT\i) edge[] node [factor] {} (#2\prnt);
            }
        }
    }
}

\newcommand{\crfFeaturesX}[3]{
    \foreach \i/\x in {#1}
    {
        \node[xnode] (FEAT\i) at (\x, #3) {};
        \begin{scope}[on background layer]
          \foreach \prnt in {SNT, TRG, SRC, NON}
          {
            \path [-] (FEAT\i) edge node [factor] {} (\prnt#2);
          }
        \end{scope}
    }
}

\newcommand{\crfFeaturesXX}[3]{
    \foreach \i/\x in {#1}
    {
        \node[xnode] (FEAT\i) at (\x, #3) {};
        \begin{scope}[on background layer]
          \foreach \prnt in {SNT, TRG, SRC, DOTS, NON}
          {
            \path [-] (FEAT\i) edge node [factor] {} (\prnt#2);
          }
        \end{scope}
    }
}

\newcommand{\crfFeaturesSemiMarkov}[3]{
    \foreach \i/\x/\d in {#1}
    {
      \node[xnode] (FEAT\i) at (\x, #3) {};
      \begin{scope}[on background layer]
        \foreach \prnt in {SNT, TRG, SRC, NON}
        {
          \path [-] (FEAT\i) edge node [factor] {} (\prnt\d);
        }
      \end{scope}
    }
    \foreach \i/\x/\d in {#2}
    {
      \node[xnode] (FEAT\i) at (\x, #3) {};
      \begin{scope}[on background layer]
        \foreach \prnt in {SNT, TRG, SRC, NON}
        {
          \path [-] (FEAT\i) edge node [factor] {} (\prnt\d);
        }
      \end{scope}
    }
}

\newcommand{\hyperNode}[4]{
    \begin{scope}[on background layer,fill opacity=0.4]
     \filldraw[fill=darkslateblue!45] ($(#1)+(-0.49,0.)$)
        to[out=90,in=180] ($(#2) + (0.,0.47)$)
        to[out=0,in=90] ($(#3) + (0.49,0.)$)
        to[out=270,in=0] ($(#2) + (0.,-0.47)$)
        to[out=180,in=270] ($(#1) + (-0.49,0.)$);
     \node[label] (Root) [above right=0em and 0.3em of #3] {%
                  \textcolor{darkslateblue}{\textsc{\bfseries #4}}};
    \end{scope}
}

\newcommand{\hyperNodeX}[5]{
    \begin{scope}[on background layer,fill opacity=0.4]
     \filldraw[fill=darkslateblue!45] ($(#1)+(0.,0.49)$)
        to[out=0,in=90] ($(#1) + (0.47,0.)$)
        to[out=270,in=90] ($(#2) + (0.47,0.)$)
        to[out=270,in=90] ($(#3) + (0.47,0.)$)
        to[out=270,in=90] ($(#4) + (0.47,0.)$)
        to[out=270,in=0] ($(#4) + (0.,-0.49)$)
        to[out=180,in=270] ($(#4) + (-0.47,0.)$)
        to[out=90,in=270] ($(#3) + (-0.47,0.)$)
        to[out=90,in=270] ($(#2) + (-0.47,0.)$)
        to[out=90,in=270] ($(#1) + (-0.47,0.)$)
        to[out=90,in=180] ($(#1) + (0.,0.49)$);
     \node[label] (Root) [above right=0em and 0.3em of #1] {%
                  \textcolor{darkslateblue}{\textsc{\bfseries #5}}};
    \end{scope}
}

\newcommand{\hyperNodeXX}[6]{
    \begin{scope}[on background layer,fill opacity=0.4]
     \filldraw[fill=darkslateblue!45] ($(#1)+(0.,0.49)$)
        to[out=0,in=90] ($(#1) + (0.87,0.)$)
        to[out=270,in=90] ($(#2) + (0.87,0.)$)
        to[out=270,in=90] ($(#3) + (0.87,0.)$)
        to[out=270,in=90] ($(#4) + (0.87,0.)$)
        to[out=270,in=90] ($(#5) + (0.87,0.)$)
        to[out=270,in=0] ($(#5) + (0,-0.49)$)
        to[out=180,in=270] ($(#5) + (-0.87,0.)$)
        to[out=90,in=270] ($(#4) + (-0.87,0.)$)
        to[out=90,in=270] ($(#3) + (-0.87,0.)$)
        to[out=90,in=270] ($(#2) + (-0.87,0.)$)
        to[out=90,in=270] ($(#1) + (-0.87,0.)$)
        to[out=90,in=180] ($(#1) + (0.,0.49)$);
     \node[label] (Root) [above right=0em and 0.3em of #1] {%
                  \textcolor{darkslateblue}{\textsc{\bfseries #6}}};
    \end{scope}
}

\newcommand{\hyperNodeTreeCRF}[5]{
    \begin{scope}[on background layer,fill opacity=0.4]
     \filldraw[fill=darkslateblue!45] ($(#1)+(-0.49,0.)$)
        to[out=90,in=180] ($(#2) + (0.,0.47)$)
        to[out=0,in=180] ($(#3) + (0.,0.47)$)
        to[out=0,in=90] ($(#4) + (0.49,0.)$)
        to[out=270,in=0] ($(#3) + (0.,-0.47)$)
        to[out=180,in=0] ($(#2) + (0.,-0.47)$)
        to[out=180,in=270] ($(#1) + (-0.49,0.)$);
     \node[label] (Root) [above right=0.3em and 0.em of #4] {%
                  \textcolor{darkslateblue}{\textsc{\bfseries #5}}};
    \end{scope}
}

\makeatletter
\tikzset{nomorepostaction/.code={\let\tikz@postactions\pgfutil@empty}}
\tikzset{circle split part fill/.style  args={#1,#2}{%
alias=tmp@name, 
postaction={%
insert path={
\pgfextra{%
\pgfpointdiff{\pgfpointanchor{\pgf@node@name}{center}}%
{\pgfpointanchor{\pgf@node@name}{east}}%
\pgfmathsetmacro\insiderad{\pgf@x}
\fill[#1] (\pgf@node@name.base) ([xshift=-\pgflinewidth]\pgf@node@name.east) arc
(0:180:\insiderad-\pgflinewidth)--cycle;
\fill[#2] (\pgf@node@name.base) ([xshift=\pgflinewidth]\pgf@node@name.west)  arc
(180:360:\insiderad-\pgflinewidth)--cycle;            
}}}}}
\makeatother

\DeclareMathOperator{\E}{\mathop{\mathbb{E}}}

\makeatletter
\renewcommand*\env@matrix[1][\arraystretch]{%
  \edef\arraystretch{#1}%
  \hskip -\arraycolsep
  \let\@ifnextchar\new@ifnextchar
  \array{*\c@MaxMatrixCols c}}
\makeatother

\theoremstyle{break}
\theoremheaderfont{\bfseries\upshape}
\newmdtheoremenv[%
linecolor=gray,%
leftmargin=35,%
rightmargin=35,%
backgroundcolor=gray!20,%
innertopmargin=5pt]{example}{Example}[section]

\newtheorem*{Example*}{Example}
\makesavenoteenv{tabular}

\hypersetup{
    colorlinks=true,
    linktoc=all,
    citecolor=black,
    filecolor=black,
    linkcolor=midnightblue,
    urlcolor=black
}

\lstdefinelanguage{XML}
{
  morestring=[b]",
  morestring=[s]{>}{<},
  morecomment=[s]{<?}{?>},
  stringstyle=\color{black},
  identifierstyle=\color{darkblue},
  keywordstyle=\color{cyan},
  morekeywords={xmlns,id,intensity,mmax_level,span,sarcasm,polarity,version,encoding}%
}

\begin{document}

\pagenumbering{roman}
\pagestyle{empty}



\makeatletter 
\title{
\titlerule\\
\Huge \@Title \\
\titlerule\\
\vspace{0.03\paperheight}
\large\@Fname\ \@Surname\\
\vspace{10em}
\begin{figure*}[htb]
  \centering \includegraphics[height=10 em]{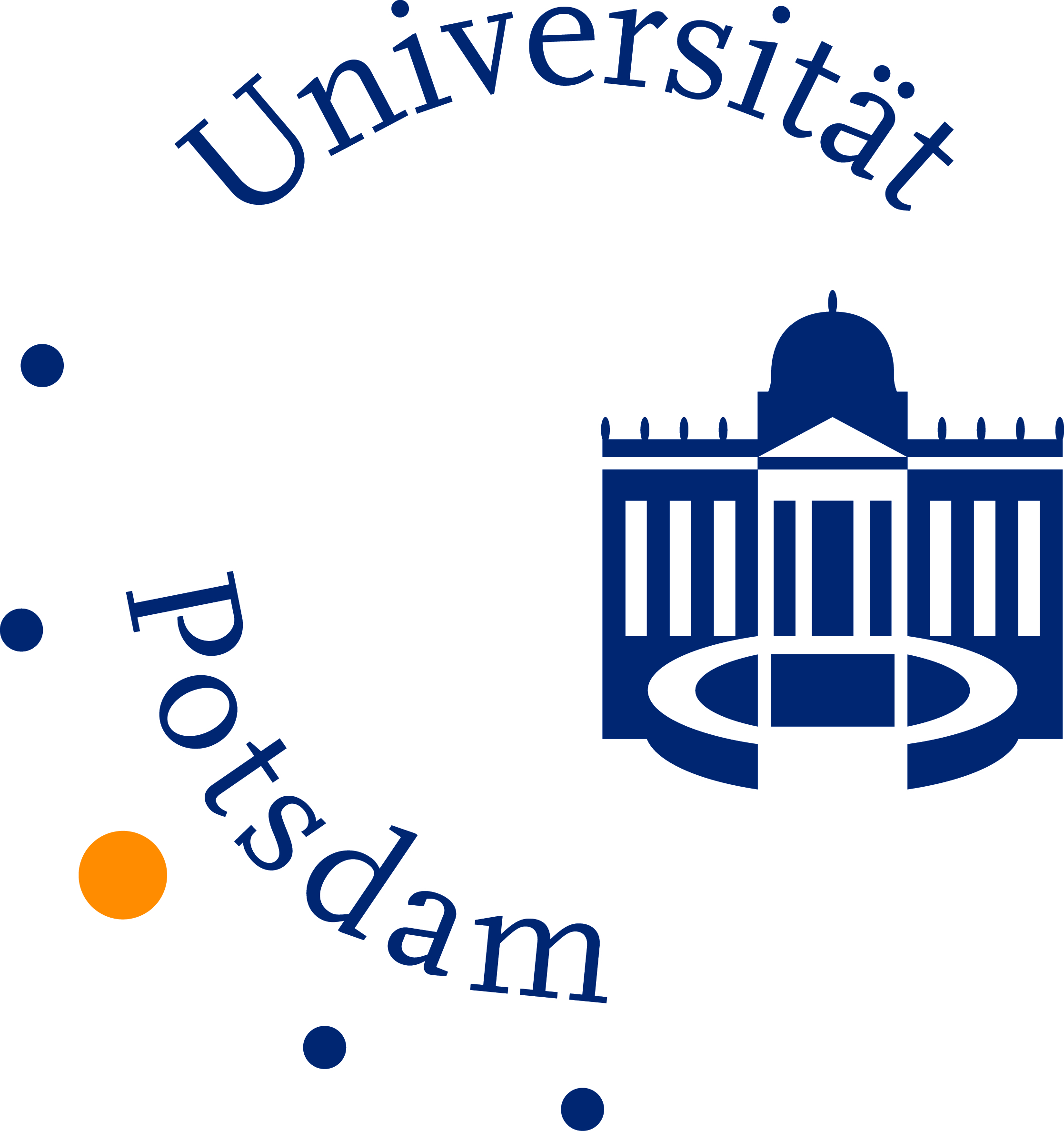}
\end{figure*}
\vspace{1em}
\large{\textit{\@Type\ eingereicht\\ bei der \@Faculty\\ der \@University}}\\
\vspace{1em}
\the\year
}
\author{}
\date{}

\maketitle

\par\vspace*{\fill}
{
\raggedright
Datum der Einreichung:~\@SubmissionDate \\
\vspace{1cm}
Wissenschaftlicher Betreuer:~\@Supervisor\\
Gutachter: \@GutachterA\\
\vspace{1cm}
Tag der m\"undlichen Pr\"ufung:~\@Pruefungsdatum
}
\makeatother 

\vspace*{\fill}

\textit{Ich erkl\"are hiermit, dass die Arbeit selbst\"andig und ohne
  unzul\"assige Hilfe Dritter verfasst wurde und bei der Abfassung nur
  die in der Dissertation angegebenen Hilfsmittel benutzt sowie alle
  w\"ortlich oder inhaltlich \"ubernommenen Stellen als solche
  gekennzeichnet wurden.}

\begin{flushright}
Potsdam, \the\year\\
Vladimir Sidorenko\\
(Uladzimir Sidarenka)
\end{flushright}
\vspace*{\fill}


\thispagestyle{empty}
\vspace*{\fill}
\begin{center}
To my family
\end{center}
\vspace*{\fill}

\selectlanguage{english}
\section*{Abstract}
\addcontentsline{toc}{section}{Abstract}

The immense popularity of online communication services in the last
decade has not only upended our lives (with news spreading like
wildfire on the Web, presidents announcing their decisions on Twitter,
and the outcome of political elections being determined on Facebook)
but also dramatically increased the amount of data exchanged on these
platforms.  Therefore, if we wish to understand the needs of modern
society better and want to protect it from new threats, we urgently
need more robust, higher-quality natural language processing (NLP)
applications that can recognize such necessities and menaces
automatically, by analyzing uncensored texts.  Unfortunately, most NLP
programs today have been created for standard language, as we know it
from newspapers, or, in the best case, adapted to the specifics of
English social media.

\noindent This thesis reduces the existing deficit by entering the new frontier of
German online communication and addressing one of its most prolific
forms---users' conversations on Twitter.  In particular, it explores
the ways and means by how people express their opinions on this
service, examines current approaches to automatic mining of these
feelings, and proposes novel methods, which outperform
state-of-the-art techniques.  For this purpose, I introduce a new
corpus of German tweets that have been manually annotated with
sentiments, their targets and holders, as well as lexical polarity
items and their contextual modifiers.
Using these data, I explore four major areas of sentiment research:
\begin{inparaenum}[(i)]
\item generation of sentiment lexicons,
\item fine-grained opinion mining,
\item message-level polarity classification, and
\item discourse-aware sentiment analysis.
\end{inparaenum}
In the first task, I compare three popular groups of lexicon
generation methods: dictionary-, corpus-, and word-embedding--based
ones, finding that dictionary-based systems generally yield better
polarity lists than the last two groups.  Apart from this, I propose a
linear projection algorithm, whose results surpass many existing
automatically-generated lexicons.  Afterwords, in the second task, I
examine two common approaches to automatic prediction of sentiment
spans, their sources, and targets: conditional random fields (CRFs)
and recurrent neural networks, obtaining higher scores with the former
model and improving these results even further by redefining the
structure of CRF graphs.  When dealing with message-level polarity
classification, I juxtapose three major sentiment paradigms: lexicon-,
machine-learning--, and deep-learning--based systems, and try to unite
the first and last of these method groups by introducing a
bidirectional neural network with lexicon-based attention. Finally, in
order to make the new classifier aware of microblogs' discourse
structure, I let it separately analyze the elementary discourse units
of each tweet and infer the overall polarity of a message from the
scores of its EDUs with the help of two new approaches:
latent-marginalized CRFs and Recursive Dirichlet Process.


\section*{Acknowledgments}
\addcontentsline{toc}{section}{Acknowledgments}

At the beginning of this thesis, I would like to say a big thank you
to all people who faithfully supported me throughout the years of
working on my dissertation: my colleagues, my friends, my teachers, my
advisor and, first of all, my family.  I am heavily indebted to all
these supporters and am absolutely convinced that this work would have
never come into existence without them.  I am also sure that it would
take me another thesis to express the whole debt of gratitude that I
owe each of my helpers, but am afraid that none of them would like to
wait that long for me to finish it.  Therefore, as hard as it is, I
have to be brief in these acknowledgments and will describe all the
infinite gratefulness that I feel with just few, but utterly sincere
words.

First and foremost, I would like to thank my family, who were
incessantly standing by my side throughout this difficult and
exhausting endeavor.  Unfortunately, they also were those who suffered
most from my absence at home and the lack of spare time.  So, taking
the opportunity to speak to them from these pages, I would like to
apologize for my heavily neglected family duties and promise that I
will make up for them twofold after the defense.

Second, I would like to express my gratitude to my advisor, Manfred
Stede, for his courage in accepting me as a Ph.D. student, for the
infinite patience during the course of this work, for the numerous
valuable suggestions, and, of course, for the time he invested in
discussing and reviewing parts of this work.

Using the chance, I would also like to thank Jacob Eisenstein for his
consent to be the reviewer of my dissertation and that he unknowingly,
through his papers, has become my second, imaginary advisor (as a
complex person I allow myself to have imaginary mentors).

For sure, writing this thesis would not have been half as fun if there
had not been such wonderful colleagues and friends as Arne Neumann,
Andr\'e Herzog, Tatjana Scheffler, Thomas Hanneforth, Yulia Grishina,
Andreas Peldszus, Sarah Hemmen, Daniel Schultheis, Misha Lukashyk,
Polina Dovnar, Alexey Cheusov, and Nikolai Krot, who not only
supported me in times of despair but also shared with me occasional
moments of joy.  There are definitely many others who should have been
mentioned here, but as I said, I would like to be brief, and if I
failed to name someone explicitly in these acknowledgments, please be
assured that I did it not for lack of gratitude but because of the
limited writing space and time.  And at least to mitigate this
omission, I would like to address my final big thank you to all those
who remained unnamed on these pages, but feel that they have
facilitated this work.

\begin{flushright}
  \textit{Vladimir Sidorenko\\ (Uladzimir Sidarenka)}
\end{flushright}

\setcounter{page}{1}
\tableofcontents
\newpage
\listoffigures
\newpage
\listoftables
\newpage

\pagenumbering{arabic}
\pagestyle{fancy}



\chapter*{Foreword}
\markboth{\textsc{FOREWORD}}{}
\addcontentsline{toc}{chapter}{Foreword}

\hspace*{\fill}\epigraph{\itshape{}Das Internet ist f\"ur uns alle
  Neuland.}{---Angela Merkel, 2013}

As social media become more and more popular, the need for automatic
analysis of their data rises.  This analysis, however, is greatly
complicated by the fact that the language style used on the Web is
fundamentally different from the style of official documents and
newspaper articles.  Indeed, sentences like the ones shown in
Example~\ref{exmp:intro:tweets:en} \cite[provided by][]{HanBaldwin:11}
are very unlikely to appear in the transcript of an Oval Office
address or in an editorial of The New York Times, even though such
wording is commonplace on English Twitter.
\begin{example}\label{exmp:intro:tweets:en}
u must be talkin bout the paper but I was thinkin movies\\
\dots so hw many time remaining so I can calculate it?
\end{example}

These differences become even more marked when it comes to emotional
speech, where people express their excitement, sadness, happiness,
approval or disapproval.  Compare, for instance, the following
passages from Example~\ref{exmp:intro:telegraph-twitter}, in which a
Telegraph reporter and a Twitter user describe their feelings about
the resignation of Boris Johnson, UK Foreign Secretary, who gave up
his office in criticism of the government's Brexit plan.
\begin{example}\label{exmp:intro:telegraph-twitter}
Je regrette. I cannot express how horrified I am that Boris Johnson
stepped down. He was the standard-bearer of those who wanted not to
get out of the single market, but to curtail the move to political
union in a federal state run by the likes of Juncker. \emph{(Ayesha
  Vardag, The Telegraph)}

\noindent{}That muffled sound is Boris Johnson kicking himself that he
didn't resign before David Davis. Two down and he's the second
\emph{(@Kevin\_Maguire, Twitter)}
\end{example}
As you can see, not only the ways of expression are different, but the
attitudes of the authors are contradictory as well. And nowadays it is
the domain of social media that is steadily gaining popularity, and
that wields more and more influence on the opinions of common people,
predetermining their preferences, choices, and political views.  This
trend is inexorable; this trend is global; and, unfortunately, this
trend opens up new possibilities for misuse of online services as an
instrument of political deception.

One way to avert the looming danger of deliberate manipulation of
public opinion is to monitor social networks in real time in order to
discover suspicious activities or unexplainable fluctuations of
people's attitudes.  A crucial prerequisite for such monitoring though
is reliable, high-quality NLP tools that can analyze users'
dispositions automatically in a split second.

\section*{Motivation}

Automatic mining of people's opinions from text is exactly what the
field of knowledge called \emph{sentiment analysis} or \emph{opinion
  mining}\footnote{Following \citet{Liu:12}, I consider the terms
  \emph{sentiment analysis} and \emph{opinion mining} as synonyms.} is
concerned with, and what we\footnote{Throughout this dissertation, I
  will use the pronoun ``we'' in recognition of the efforts made by
  all people mentioned in the acknowledgments, and in recognition of
  your efforts as a reader who will struggle with me through the pages
  of this work.  This usage, however, does not imply that either you
  or any of my supporters share the same opinions or are responsible
  for any of the claims.} will work on in this dissertation.  In
particular, we are going to analyze users' attitudes on German
Twitter---a linguistic register whose natural language processing is
aggravated not only by the specifics of social media but also by the
scarceness of resources, systems, and established baselines.
Nevertheless, we decided to address precisely this domain because:
\begin{itemize}
  \item German is the most spoken first language in the European
    Union, being the mother-tongue for 18\% of EU
    citizens;\footnote{\url{https://en.wikipedia.org/wiki/Languages_of_the_European_Union}}
  \item Germany has traditionally played a major role in the European
    Government, and, as such, it was one of the main driving forces in
    solving several European crises, including the Ukrainian conflict,
    the prevention of Greek sovereign default, and Brexit;
  \item Numerous internal problems (refugee crisis, rise of right-wing
    populism, and unstable ratings of political parties) make German
    politics susceptible to external influence.
\end{itemize}

Our choice of the Twitter platform was motivated by the following
factors:
\begin{itemize}
  \item First of all, Twitter is the second most popular social
    network in
    Germany,\footnote{\url{https://digiday.com/marketing/state-social-platform-use-germany-5-charts/}}
    with 4.9 million monthly active users (as of
    2017);\footnote{\url{https://luckyshareman.com/blog/die-twitter-nutzung-in-deutschland/}}
  \item Second, Twitter's sociolect is at the cutting edge of modern
    language development, and new linguistic phenomena introduced on
    this service are likely to percolate into other social media and
    might even find their way into the standard language as well;
  \item Finally, the abundance and accessibility of data on this
    platform allows the researchers to analyze virtually any topic,
    from North Korean nuclear weapons to Lady Gaga's dress, getting
    messages (and opinions) from users of different income, gender,
    and age.
\end{itemize}

\section*{Research Questions}

Unfortunately, despite its popularity and social importance, German
Twitter has largely been ignored by computational linguistics in
general, and in particular by its opinion mining branch.  With this
dissertation, we hope to make up this leeway by presenting a new
sentiment corpus of German microblogs and conducting an extensive
study of existing and novel opinion mining methods on these data.  By
doing so, we want to answer the following questions:

\begin{itemize}
\item\textbf{Can we apply opinion mining methods devised for
  standard English to German Twitter?}

  Since there had been literally no attempts to analyze sentiments in
  German social media when we started working on this thesis, as a
  first step, we decided to check whether we could reuse existing
  English solutions without further ado.

\item\textbf{Which groups of approaches are best suited for which
  sentiment tasks?}

  Because sentiment analysis is a wide research field, which operates
  on various linguistic levels and addresses many different problems
  with their own approaches and evaluation metrics, we want to know
  which approaches (rule-based or machine-learning ones, systems that
  operate on lexical taxonomies or those that utilize corpus data)
  work best for specific sentiment tasks;

\item\textbf{How much do word- and discourse-level analyses affect
  message-level sentiment classification?}

  Despite the wide variety of problems addressed by opinion mining,
  one of them---message-level polarity classification---is commonly
  considered as the central task in sentiment analysis of social
  media.  Due to its importance and central role, we would like to see
  which linguistic level (subsentential [\ie{} the level of word] or
  suprasentential [\ie{} the level of discourse]) contributes more to
  determining the overall polarity of a microblog.

\item\textbf{Does text normalization help analyze sentiments?}

  Although many NLP researchers consider social media specifics as a
  hindrance and suggest converting them to the standard-language form,
  other scientists object that a straightforward conversion might
  loose many important details and consequently worsen classification.
  \citet{Brody:11}, for instance, claim that intentional prosodic
  lengthening of words, such as \texample{sooooooo strong} or
  \texample{coooolllllll}, serves as a vivid indicator of opinionated
  sentences, so that keeping these elongations in text would result in
  better predictions.  \citet{Eisenstein:13}, in part, agrees with
  these claims by noting that a straightforward replacement of
  colloquial variants with their standard-language equivalents can
  considerably shift the original meaning.  We admit that the
  arguments of these authors are correct, but it apparently depends on
  the magnitude by which non-standard language helps or hampers NLP
  applications.  So, in this work, we would like to test whether text
  normalization does more harm than good to the analysis of opinions.

\item\textbf{Can we do better than existing approaches?}

  Of course, simply evaluating existing methods on a new dataset would
  not be of much novelty and would not accelerate the progress of the
  research field, therefore, we are going to improve on existing
  results by suggesting our own solutions to various sentiment
  objectives.
\end{itemize}

\section*{Outline of this Work}

We will answer these questions by proceeding in the following way:

\begin{itemize}
\item In Chapter~\ref{chap:introduction}, we will give a short
  introduction to sentiment analysis and make a digression into the
  history of this field;

\item In Chapter~\ref{chap:corpus}, we will present the Potsdam
  Twitter Sentiment Corpus (PotTS), define selection criteria that we
  used in order to collect tweets for this dataset, describe its
  annotation scheme and labeling procedure, and also conduct an
  extensive inter-annotator agreement study, looking for messages that
  were most difficult to analyze for human experts;

\item Afterwards, in Chapter~\ref{chap:snt:lex}, we will turn our
  attention to the first subsentential sentiment task---sentiment
  lexicon generation---in which we will compare three major paradigms:
  dictionary-, corpus-, and word-embedding--based methods, and also
  propose our own linear-projection solution;

\item Chapter~\ref{chap:fgsa} will address the problem of fine-grained
  opinion mining, whose goal is to predict text spans of sentiments,
  sources, and targets.  In particular, we will evaluate three popular
  approaches to this challenging task: conditional random fields
  (CRFs), long-short term memory (LSTM), and gated recurrent unit
  (GRU), checking the effect of various features on the first
  classifier and estimating the results of the last two systems with
  different word-embedding types;

\item In Chapter~\ref{chap:cgsa}, we will deal with one of the most
  prominent sentiment analysis tasks---message-level polarity
  classification.  This time, again, we will juxtapose three main
  classes of methods: lexicon-based, machine-learning--based, and
  deep-learning ones, and will try to unite the first and the last of
  these groups by devising a recurrent neural network with
  lexicon-based attention;

\item Finally, in Chapter~\ref{chap:discourse}, we will enhance the
  proposed system by making it aware of the microblogs' discourse
  structure.  For this purpose, we will let the classifier predict the
  polarity scores of the elementary discourse units of each tweet and
  will then unite these scores using novel techniques: latent
  conditional and conditional-marginalized random fields and Recursive
  Dirichlet Process.
\end{itemize}






\chapter{Introduction to Sentiment Analysis}\label{chap:introduction}

Interpersonal communication is not only a way to share objective
information with other people but also a vibrant channel to convey
one's subjective feelings, impressions, and attitudes.  It is, in
fact, this latter use that provides a personal touch to our
conversations, making them more grasping, more entertaining, and more
living.  And it is often this use that significantly influences our
preferences, decisions, and choices in everyday life.  Therefore, a
high-quality mining of people's opinions is often as important as
retrieval of objective facts.

The field of knowledge that deals with the analysis of emotions,
sentiments, evaluations, and attitudes is called \emph{sentiment
  analysis} (SA) \citep{Liu:12}.  The definition of this discipline,
however, much like the definition of the term \emph{sentiment} itself,
is neither complete nor universally accepted.  The main reasons for
this are
\begin{inparaenum}[(i)]
  \item a frequently blurred boundary between subjective and objective
    parts of information and
  \item the heterogeneity of the language system itself, to which the
    SA methods are applied.
\end{inparaenum}

The first factor, for instance, makes it difficult to delimit which
statements actually belong to the jurisdiction of opinion mining and
which ones should be ignored by its systems.  A prominent example of
such borderline cases are the so-called subjective facts, such as
``terrorist attacks'' or ``anti-cancer drugs,'' which some people
consider as polar terms, while others regard them as objective
expressions.

The second factor complicates a precise definition of sentiment
analysis because different language levels have their own notions of
subjectivity (\eg{} a positive word is not the same as a positive
text), which in turn necessitate different approaches.  Depending on
the analyzed linguistic level, researchers typically distinguish three
main types of SA\@:
\begin{itemize}
  \item\emph{subsentential}, whose task is to determine polarities of
    single words and find opinions within a sentence,
  \item\emph{sentential}, which tries to predict the semantic
    orientation of a statement,
  \item and, finally, \emph{suprasentential}, which analyzes the
    polarity of the whole text.
\end{itemize}
Each of these types has its own specifics, and each of them needs to
be addressed with its unique methods.  Therefore, speaking of general
difficulty of opinion mining for a specific domain is in the same way
wrong as judging about the amenability of this domain to the whole
natural language processing: one needs to specify a particular task
and evaluate it with its own metrics.  Thus, to estimate the
complexity of sentiment analysis for German Twitter, we will address
all three levels of SA\@: subsentential, sentential, and
suprasentential.

\section{Prehistory of the Field}

But before we delve into the depths of contemporary sentiment
research, let us first make a digression into the history of this
field in order to understand its modern trends and theories better.

Like many other scientific disciplines, opinion mining has emerged
from several other areas of research including philosophy, psychology,
cognitive sciences, narratology, and linguistics.  In
\emph{philosophy}, the questions about the nature of emotions, their
interaction with human consciousness, and the influence on people's
deeds have occupied the minds of many great scholars, starting from
Plato and Aristotle.  Plato, for instance, argued that the human soul
consists of three fundamental parts: the rational, the appetitive, and
the passionate \citep[see][Book~IV]{Plato:91}.  The last part (the one
by which we become angry or fly into a temper) determines our notion
of justice by favoring either the rational or the appetitive aspect.
\citet{Aristotle:54}, the most prominent student of Plato, extended
this idea by providing a precise taxonomy of feelings that, in his
opinion, constitute the passionate part of the mind.

As noted by \citet{Sousa:14}, the variety and complexity of phenomena
covered by the term ``emotion'' discouraged tidy philosophical ideas
and was daunting the researchers for many hundred years since
antiquity.  A real renaissance of emotional studies happened in the
late 19-th century in \emph{psychology} with the introduction of the
James-Lange theory~\cite{James:1884,Lange:1885}, which argued that
biological processes were the main and only reason for people's
subjective opinions.  This theory, however, was later criticized by
\citet{Schachter:62}, who objected that bodily means alone were
insufficient to express the full range of possible feelings, and that
cognitive factors were a key determinant of emotional states.


These advances in psychology, reinforced by the nascent appraisal
theory \citep{Arnold:60}, have significantly influenced many other
scientific fields including \emph{literary studies} and
\emph{linguistics}.  Among the most prominent representatives of this
direction in the former discipline were \citet{Rorty:80} and
\citet{Banfield:82}, who analyzed how opinions were expressed by
direct and indirect speech.  \citet{Wiebe:90a,Wiebe:94} adapted
\citeauthor{Banfield:82}'s theory to the needs of \emph{computational
  linguistics} by proposing an algorithm that identified subjective
sentences in text and inferred the main characters of a narrative from
such sentences.  This work was presumably the first attempt to
automatically detect sentiments on the sentential level.

A real breakthrough in the opinion mining field, however, happened
with the introduction of the first sufficiently big corpora.
Important contributions in this regard were made by
\citet{Pang:04,Pang:05}, who released a dataset of $\approx2,000$
movie reviews with their star ratings; \citet{Hu:04}, who presented a
manually labeled set of Amazon and C|Net product comments;
\citet{Thomas:06}, who automatically labeled a collection of
congressional debates; and, finally, \citet{Wiebe:05}, who developed a
manually annotated sentiment corpus of 535 news articles.

The availability of these resources has given rise to a plethora of
new methods for both subsentential and sentential SA, making opinion
mining one of the most challenging and competitive branches of
computational linguistics.  Fundamental cornerstones in this field
have already been set by the works of \citet{Pang:02},
\citet{Wiebe:05}, \citet{Wilson:05}, \citet{Breck:07},
\citet{Choi:09,Choi:10}, and \citet{Socher:11, Socher:12}.
Nevertheless, many challenges of sentiment research, such as domain
adaptation or analysis of non-English texts, still pose considerable
difficulties.

\section{Sentiment Analysis of Social Media}

One of the main problems that people working on opinion mining are
usually confronted with in the first place is the choice of the domain
to deal with. Since sentiment analysis is a highly domain-dependent
task \citep[see][]{Aue:05,Blitzer:07,Li:08}---\ie{} systems trained on
one text genre can hardly generalize to other linguistic
variations---a natural question that arises in this context is which
of the domains should be addressed first in this case.

While earlier sentiment works were primarily concerned with narratives
\citep{Wiebe:90a,Wiebe:94} or newspaper texts
\citep{Wiebe:03,Wiebe:05,Bautin:08}, it soon became clear that social
media provide a much more fertile ground for mining people's
attitudes.  Among the first who tried to extract users' opinions from
online forums were \citet{Das:01}. For this purpose, they manually
annotated a collection of 500 messages from an economic chat board
with three polarity classes (\emph{buy}, \emph{sell}, and \emph{null})
and then trained five different classifiers on these data.  Using the
trained systems, the authors classified the semantic orientation of
the remaining 25,000 forum posts, trying to predict stock prices based
on the polarities of these snippets.

Automatic business intelligence has rapidly won the ground in the
opinion mining field, with further notable works introduced by
\citet{Glance:05}, who analyzed users' opinions on Usenet with
hand-crafted rules; \citet{Antweiler:04}, who investigated how
postings on message boards correlated with stock volatility;
\citet{Ghose:07}, who examined the effect of opinions on pricing in
online marketplaces; and, finally, \citet{Turney:02}, who classified
Epinions reviews into \emph{recommended} (thumbs up) and \emph{not
  recommended} (thumbs down) ones based on the pointwise mutual
information of their adjectives.

Due to its high commercial impact, sentiment analysis of customer
feedback soon became one of the most popular topics in natural
language processing.  \citet{Dave:03}, for example, classified users'
comments on Amazon as positive or negative with the help of SVM and
Na\"{\i}ve Bayes systems.  \citet{Hu:04} developed a three-stage
application that produced concise summaries of positive and negative
opinions about each particular product feature.  \citet{Funk:08}
proposed a supervised SVM classifier that predicted the polarity of
product reviews and then used these results in a business intelligence
application.  Other important contributions were made
\citet{Popescu:05}, \citet{Ding:09}, \citet{Wei:10},
\citet{Mukherjee:12}, etc.


Although opinion analysis of product reviews still plays an important
role in e-commerce, the increased popularity of the blogosphere and
social networks has motivated many sentiment researchers to shift the
focus of their work to these new Web genres.  Among the first who
followed the new trend were \citet{Mishne:05} and \citet{Mishne:07},
who tried to predict the moods of LiveJournal blogs (\eg{}
\emph{amused}, \emph{tired}, \emph{happy}), achieving 58\% accuracy
with an SVM classifier.
Another SVM system was used by \citet{Chesley:06}, who classified
users' blogs into positive, negative, and objective ones,
outperforming the majority class baseline by 15\% with this method.
Drawing on these works, \citet{Gill:08} analyzed the agreement of
human experts on blogs' moods, finding that the annotators had a much
better consensus about feelings that were described in longer blogs.


Speaking of text length, we should certainly say that the inception of
the micro-blogging service Twitter in 2006 was a real game changer to
the whole opinion mining field.  The sudden availability of huge
amounts of data, the presence of all possible social and national
groups, combined with the uniqueness of the language on this service,
have given rise to numerous scientific projects, studies, and
publications, which we will briefly summarize in the next section.

\section{Sentiment Analysis of Twitter}\label{snt:subsec:intro:saot}

One of the first attempts to automatically classify users' opinions on
Twitter was made by \citet{Go:09}, who collected a set of 1.6~M
microblogs containing emoticons.  Considering these smileys as noisy
labels (positive or negative), the authors trained three different
classifiers (Na\"{\i}ve Bayes, Maximum Entropy, and SVM), obtaining
the best results (0.82~\F{}) with the SVM system.  Similar approaches
were taken by \citet{Pak:10}, who performed a three-class prediction
with a Na\"{\i}ve Bayes system, and \citet{Davidov:10}, who trained a
$k$-NN--like classifier on weakly supervised data. Another way to
create a sentiment corpus was proposed by \citet{Barbosa:10}, who
analyzed a set of 200,000 microblogs with three publicly available
opinion mining services and then used the majority votes of these
systems as silver labels for their dataset.

Some time later, \citet{Kouloumpis:11} experimented with the AdaBoost
classifier on the noisy collection of \citet{Go:09} and the Edinburgh
Twitter corpus of \citet{Petrovic:10}, coming to the conclusion that
microblog-specific features (such as presence of intensifiers,
abbreviations, or emoticons) were the most reliable attributes for
this classification.  \citet{Agarwal:11}, however, questioned this
finding, arguing that POS-specific polarity features were a better
alternative.

As in the case of opinion mining of product reviews, sentiment
analysis of Twitter could not go unnoticed by the economic and
sociological communities.  An attempt to address political issues
using this platform was made by~\citet{Tumasjan:10}, who analyzed
users' opinions about German federal elections in 2009 by
automatically translating 100,000 tweets into English and subsequently
classifying these messages with the proprietary LIWC
software~\cite{Pannebaker:07}.  This study showed that not only
sentiments but even the mere numbers of microblogs mentioning
political parties strongly correlated with the results of election
polls.


Nevertheless, the real rise of interest in this domain happened with
the release of the Sem\-Eval corpus~\cite{Nakov:13}, a collection of
15,000 tweets that have been manually annotated by Amazon mechanical
turkers with their message-level polarities and contextual semantic
orientations of their polar terms. The best performing system in the
first run of the SemEval competition was a supervised SVM classifier
of \citet{Mohammad:13}, which won three out of four subtasks
(message-level classification of SMSs and tweets and contextual
polarity prediction of polar terms in Twitter).  This solution relied
on an extensive set of hand-crafted features, powered by multiple
manually- and automatically-generated sentiment lexicons.  The authors
emphasized the crucial importance of lexical resources, which
increased the classification scores by almost $8.5\%$.  Other
competing submissions~\cite{Becker:13,Guenther:13,Kokciyan:13} used a
similar approach, but had considerably fewer feature attributes.

The success of this shared task, which had more than 40 participants,
motivated the organizers to continue the competition.  With slight
modifications (addition of new tweets, inclusion of sarcastic
microblogs and LiveJournal sentences), they rerun both subtasks in the
following four
years~\cite{Rosenthal:14,Rosenthal:15,Nakov:16,Rosenthal:17},
attracting more and more competitors every time.\footnote{A detailed
  overview of these iterations is provided in
  Chapter~\ref{chap:cgsa}.}

As you can see, despite its relatively short history, sentiment
analysis of Twitter has already received much attention from NLP
researchers.  But, with a few
exceptions~\cite[\eg{}][]{Basile:13,Bosco:13,Araque:15,Cesteros:15},
most of existing works were primarily concerned either with English
messages or with automatically translated microblogs.  In the
following chapters, we will explore whether conclusions that have been
drawn from English data (the difficulty of the Twitter domain for
automatic SA, the utility of sentiment lexicons, and the need to
normalize the text input) apply to German tweets as well.

\chapter{Sentiment Corpus}\label{chap:corpus}

A crucial prerequisite for proving any hypotheses in computational
linguistics is the existence of sufficiently big manually annotated
datasets, on which these conjectures could be tested.  Since there
were no human-labeled sentiment data for German Twitter that we were
aware of at the time of writing this chapter, we decided to create our
own corpus, which we will introduce in this part of the thesis.

We begin our introduction by describing the selection criteria and
tracking procedure that we used to collect the initial corpus data.
After presenting the annotation scheme, we perform an extensive
analysis of the inter-annotator agreement.  For this purpose, we
introduce two new versions of the popular $\kappa$
metric~\cite{Cohen:60}---binary and proportional kappa---which have
been specifically adjusted to the peculiarities of our annotation
task.  Using these measures, we check the inter-coder reliability of
annotated sentiments, their targets and sources, polar terms and their
modifying elements (intensifiers, diminishers, and negations).  In the
final step, we estimate the correlation between the initial selection
rules and the number of labeled elements as well as the difficulty of
their annotation.

\section{Data Collection}

A common question that typically arises first when one starts creating
a new dataset is which selection criteria should be used in order to
collect the initial data.  Whereas for low-level NLP applications,
such as part-of-speech tagging or syntactic parsing, it typically
suffices to define the language domain to sample from (since the
phenomena of interest are usually frequent and uniformly spread), for
semantically demanding tasks with many diverse ways of expression one
also needs to consider various in-domain factors, which might
significantly affect the final distribution, making the resulting
corpus either utterly sparse or excessively biased.

In order to minimize both of these risks (sparseness and bias), we
decided to use a compromise approach by gathering one part of the new
dataset from microblogs that were a priori more likely to have
sentiments (thus increasing the recall) and sampling the rest of the
corpus uniformly at random (thus reducing the bias).

As criteria that could help us get more opinions, we considered topic
and form of the tweets, assuming that some subjects, especially social
or political issues, would be more amenable to subjective statements.
Because we started creating the corpus in spring 2013, obvious choices
of opinion-rich topics to us were \emph{the papal conclave}, which
took place in March of that year, and \emph{the German federal
  elections}, which were held in autumn.  Since both of these events
implied some form of voting, we decided to counterbalance the election
specifics by including \emph{general political discussions} as the
third subject in our dataset.  Finally, to obey the second principle,
\ie{} to keep the corpus bias low, we sampled the rest of the data
from \emph{casual everyday conversations} without any prefiltering.

We collected messages for the first and third groups by tracking
German microblogs between March and September 2013 via the public
Twitter API\footnote{\url{https://pypi.python.org/pypi/tweetstream}}
with the help of extensive keyword lists describing these
topics.\footnote{A full list of tracking keywords is available at
  \url{https://github.com/WladimirSidorenko/PotTS/blob/master/docs/tracking_keywords.pdf}.}
Tweets for the second topic (German federal elections) were provided
to us by a research group of communication scientists from the
University of M\"unster, who were our cooperation partner in a joint
BMBF project ``Discourse Analysis in Social Media.''  Finally, for the
fourth category (casual everyday conversations), we used the complete
German Twitter snapshot of~\citet{Scheffler:14}, which includes
$\approx97\%$ of all German microblogs posted in April 2013.  This
way, we obtained a total of 27.4~M messages, with the snapshot corpus
being by far the most prolific source of the data.






In the next step, we divided all tweets of the same topic into three
groups based on the following formal criteria:
\begin{itemize}
\item We put all messages that contained at least one polar term from
  the sentiment lexicon of \citet{Remus:10} into the first group;
\item Microblogs that did not satisfy the first condition, but had at
  least one exclamation mark or emoticon were allocated to the second
  group;
\item All remaining microblogs were assigned to the third category.
\end{itemize}
A detailed breakdown of the resulting distribution across topics and
formal groups is given in Table~\ref{snt:tbl:corp:topic-bins}.
\begin{table}[hbt!]\small
  \begin{center}
    \bgroup\setlength\tabcolsep{0.13\tabcolsep}\scriptsize
    \begin{tabular}{l*{5}{>{\centering\arraybackslash}p{0.155\textwidth}}}
      \toprule
      & \multicolumn{4}{c}{\bfseries Formal Criterion} & \\\cmidrule{2-5}

      \multirow{-2}{0.2\columnwidth}{\centering\bfseries
      Topic} & Polar Terms & Emoticons & Remaining Tweets & Total &\multirow{-2}{0.12\textwidth}{\centering\bfseries Sample\\ Keywords}\\\midrule

      Federal Elections & 537,083 (22.38\%) & 50,567 (2.1\%) & 1,811,742 (75.5\%) & 2,399,392 & \tiny\emph{Abgeordnete} (\emph{representative}), \emph{Regierung} (\emph{government})\\

      Papal Conclave & 7,859 (15.11\%) & 1,260 (2.42\%) & 42,879 (82.46\%) & 51,998 & \tiny\emph{Papst} (\emph{pope}), \emph{Pabst} (\emph{pobe})\\

      Political Discussions & 10,552 (25.8\%) & 777 (1.9\%) & 29,555 (72.29\%) & 40,884 & \tiny\emph{Politik} (\emph{politics}), \emph{Minister} (\emph{minister})\\

      General Conversations & 3,201,847 (18.7\%) & 813,478 (4.7\%) & 13,088,008 (76.5\%) & 17,103,333 & \tiny\emph{den} (\emph{the}), \emph{sie} (\emph{she})\\

      \bottomrule
    \end{tabular}
    \egroup{}
    \caption[Distribution of downloaded messages across topics and
      formal groups]{Distribution of downloaded messages across topics
      and formal groups\newline (percentages are given with respect to
      the total number of tweets pertaining to the given
      topic)\label{snt:tbl:corp:topic-bins}}
  \end{center}
\end{table}

To create the final corpus, we randomly sampled 666 tweets from each
of the three formal classes for each of the four topics, getting a
total of 7,992 messages ($666\text{ microblogs} \times 3\text{ formal
  criteria} \times 4\text{ topics}$).

\section{Annotation Scheme}\label{subsec:snt:ascheme}

In the next step, we devised an annotation scheme for our data.  To
maximally cover all relevant sentiment aspects, we came up with an
extensive list of elements that had to be annotated by our experts.
This list included:

\begin{itemize}
\item
  \textbf{\markable{sentiment}}s, which were defined as \emph{polar
    subjective evaluative opinions about people, entities, or events}.
  According to our definition, a \markable{sentiment} always had to
  evaluate an entity that was explicitly mentioned in text---the
  target; and the annotators had to label both the target and its
  respective evaluative expression with the \markable{sentiment}
  tag. Apart from tagging the text span, they also had to specify the
  following attributes of opinions:
  \begin{itemize}
  \item\attribute{polarity}, which reflected the attitude of opinion's
    holder to the evaluated entity.  Following
    \citet{Jindal:06a,Jindal:06b}, we distinguished between
    \emph{positive}, \emph{negative}, and \emph{comparative}
    sentiments;
  \item\attribute{intensity}, which showed the emotional strength of
    an opinion.  Possible values for this attribute were: \emph{weak},
    \emph{medium}, and \emph{strong};
  \item finally, drawing on the works of~\citet{Bosco:13} and
    \citet{Rosenthal:14}, we introduced a special boolean attribute
    \attribute{sarcasm} in order to distinguish sarcastically meant
    statements;
  \end{itemize}


\item
  we specified \textbf{\markable{target}}s as \emph{(real,
    hypothetical, or collective) entities, properties, or propositions
    (states or events) evaluated by opinions}.  For this item, we
  introduced the following three attributes:
  \begin{itemize}
    \item
      a boolean property \attribute{preferred}, which distinguished
      entities that were favored in comparisons;
    \item
      a link attribute \attribute{anaphref}, which had to point to the
      antecedent of a pronominal target;
    \item and, finally, another edge feature,
      \attribute{sentiment-ref}, which had to link \markable{target}s
      to their respective \markable{sentiment}s in the cases when the
      \markable{target} span was located at intersection of two
      opinions;
  \end{itemize}

\item
  another important component of \markable{sentiment}s were
  \textbf{\markable{source}}s, which denoted \emph{the immediate
    author(s) or holder(s) of opinions}.  The only property associated
  with this element was \attribute{sentiment-ref}, which was defined
  the same way as for \markable{target}s.
\end{itemize}
To help our annotators identify exact boundaries of these elements, we
explicitly asked them to annotate \emph{smallest complete syntactic or
  discourse-level units}, \ie{} noun phrases or sentences with all
their grammatical dependents.

A sample tweet analyzed according to this rule is shown in
Example~\ref{snt:exmp:sent-anno1}.
\begin{example}\label{snt:exmp:sent-anno1}
  \upshape\sentiment{\target{Diese Milliardeneinnahmen} sind selbst
    \source{Sch\"auble} peinlich}\\[0.8em]
  \noindent\sentiment{\target{\itshape{}These billions of
      revenue\upshape{}}\itshape{} are embarrassing even for\\
    \upshape{}\source{\itshape{}Sch\"auble\upshape{}}}
\end{example}
In this message, we assigned the \markable{sentiment} tag to the
complete sentence because this grammatical unit is the smallest
syntactic constituent that simultaneously includes both the target of
the opinion (``Milliardeneinnahmen'' [\emph{billions of revenue}]) and
its evaluation (``peinlich'' [\emph{embarrassing}]).  Furthermore, we
also labeled the whole noun phrase ``diese Milliardeneinnahmen''
(\emph{these billions of revenue}), including the demonstrative
pronoun ``diese'' (\emph{these}), as \markable{target}, since this
pronoun syntactically depends on the main target word
``Milliardeneinnahmen'' (\emph{billions of revenue}).

Apart from \markable{sentiment}s, \markable{target}s,
\markable{source}s, we also asked the annotators to label elements
that could significantly affect the intensity and polarity of an
opinion.  These elements were:

\begin{itemize}
\item
  \textbf{\markable{polar term}}s, which we defined as \emph{words or
    idioms that had a distinguishable evaluative lexical meaning}.
  Typical examples of such terms were lexemes or set phrases such as
  ``ekelhaft'' (\emph{disgusting}), ``lieben'' (\emph{to love}),
  ``Held'' (\emph{hero}), ``wie die Pest meiden'' (\emph{to avoid like
    the pest}).  In contrast to \markable{target}s and
  \markable{source}s, which only could occur in the presence of a
  \markable{sentiment}, \markable{polar term}s were independent of
  other tags and always had to be labeled in the corpus.

  The main attributes of this element (\attribute{polarity},
  \attribute{intensity}, and \attribute{sarcasm}) largely coincided
  with the corresponding properties of \markable{sentiment}s, with the
  only difference that, in the case of \markable{polar term}s, these
  features had to reflect the lexical meaning of a word without taking
  into account its context (\ie{} \emph{prior} polarity and
  intensity), whereas for \markable{sentiment}s, they had to show the
  compositional meaning of the whole opinion (\ie{} its
  \emph{contextual} polarity and intensity).

  Besides these common properties, \markable{polar term}s also had
  their specific attributes: two boolean features
  (\attribute{subjective-fact} and \attribute{uncertain}) and a link
  attribute (\attribute{sen\-ti\-ment\--ref}).  The first feature
  showed whether a polar term denoted a factual entity with a clear
  emotional connotation, \eg{} ``Atombombe'' (\emph{A-bomb}) or
  ``Naturschutz'' (\emph{nature protection}); the second property
  signified cases in which the annotators were unsure about their
  decisions; finally, the last attribute was defined in the same way
  as it was previously specified for \markable{target}s and
  \markable{source}s;

\item
  \emph{elements that increased the expressivity and subjective sense
    of polar terms} had to be labeled as
  \textbf{\markable{intensifier}}s.  Typical examples of such
  expressions were adverbial modifiers such as ``sehr'' (\emph{very}),
  ``super'' (\emph{super}), ``stark'' (\emph{strongly});

\item
  \textbf{\markable{diminisher}}s, on the contrary, were \emph{words
    or phrases that reduced the strength of a polar term}.  Like
  \markable{intensifier}s, these elements were usually expressed by
  adverbs, \eg{} ``weniger'' (\emph{less}), ``kaum'' (\emph{hardly}),
  ``fast'' (\emph{almost}).

  Both of these tags (\markable{intensifier}s and
  \markable{diminisher}s) only had two attributes: a binary feature
  \attribute{degree} with two possible values: \emph{medium} and
  \emph{strong}; and a link attribute \attribute{polar-term-ref},
  which connected the modifier to its \markable{polar-term};

\item the final element, \textbf{\markable{negation}}s, was defined as
  \emph{grammatical or lexical means that reversed the semantic
    orientation of a polar term}.  These were typically represented by
  the negative particle ``nicht'' (\emph{not}) or indefinite pronoun
  ``keine'' (\emph{no}).  The only attribute associated with this tag
  was a mandatory link \attribute{polar-term-ref}.
\end{itemize}

In contrast to sentiment-level tags, which had to be assigned to
syntactic or discourse-level units, \markable{polar term}s and their
modifiers were defined as lexemes and, correspondingly, had to mark
only single words or set phrases without their grammatical dependents.

A complete tweet annotated with sentiment- and term-level elements is
shown in Example~\ref{snt:exmp:sent-anno2}. In this case, we again
labeled the whole sentence as \markable{sentiment} because only the
main verb-phrase simultaneously covers both the evaluated target
(``Die Nazi-Vergangenheit'' [\emph{The Nazi history}]) and its
respective polar expression (``nicht sehr r\"uhmlich'' [\emph{not very
    laudable}]).  The boundaries of \markable{sentiment} and
\markable{target} are determined on the syntactic level, spanning the
whole clause in the former case and including the complete noun phrase
in the latter.  The polarity of the opinion is set to \emph{negative}.
The polar term ``r\"uhmlich'' (\emph{laudable}), its intensifier
``sehr'' (\emph{very}), and negation ``nicht'' (\emph{not}), on the
other hand, only mark single words.  The polarity of the term, \ie{}
its primary semantic orientation without the context, is
\emph{positive}.
\begin{example}\label{snt:exmp:sent-anno2}
  \tikzstyle{every picture}+=[remember picture]
  \tikzstyle{na} = [shape=rectangle,inner sep=0pt]
  \upshape\sentiment{\target{Die Nazi-Vergangenheit} ist
    \negation{\tikz\node[na](word0){nicht};}
    \intensifier{\tikz\node[na](word1){sehr};}
    \emoexpression{\tikz\node[na](word2){r\"uhmlich};}}\\[2.2em]
  \noindent\sentiment{\target{\itshape{}The Nazi
      history\upshape{}}\itshape{} is
    \negation{\tikz\node[na](word3){not};}
    \upshape{}\intensifier{\tikz\node[na](word4){very};}
    \upshape{}
    \emoexpression{\itshape{}\tikz\node[na](word5){laudable};\upshape{}}}

  \begin{tikzpicture}[overlay]
    \path[->,deeppink4,thick](word0) edge [in=145, out=35] node
         [above] {\tiny polar-term-ref} (word2);
    \path[->,cyan,thick](word1) edge [in=145, out=30] node
         [above] {\tiny polar-term-ref} (word2);

    \path[->,deeppink4,thick](word3) edge [in=145, out=35] node
         [above] {\tiny polar-term-ref} (word5);
    \path[->,cyan,thick](word4) edge [in=145, out=30] node
         [above] {\tiny polar-term-ref} (word5);
  \end{tikzpicture}
\end{example}
A more detailed description of all annotation elements and their
possible attributes is given in the original annotation guidelines in
Appendix~\ref{chap:apdx:corp-guidelines} of this thesis.

\section{Annotation Tool and Format}\label{subsec:snt:tformat}

For annotating the collected data, we used \texttt{MMAX2}, a freely
available text-markup
tool.\footnote{\url{http://mmax2.sourceforge.net/}} Because this
program uses a token-oriented stand-off format, where all annotated
spans are stored in a separate file and only refer to the ids of words
in the original text, we first had to split all corpus messages into
tokens.  To this end, we applied a minimally modified version of
Christopher Potts' social media
tokenizer,\footnote{\url{http://sentiment.christopherpotts.net/code-data/happyfuntokenizing.py}}
which had been slightly adjusted to the peculiarities of the German
spelling (we allowed for the capitalized form of common nouns, \eg{}
``Freude'' [\emph{joy}], and the period at the end of ordinal numbers,
\eg{} ``7.''  [\emph{7th}]).



To ease the annotation process and minimize possible data loss, we
split the corpus into 80 smaller project files with \numrange{99}{109}
tweets each.  In each such file, we put microblogs pertaining to the
same topic, ensuring an equal proportion of formal groups.

\section{Inter-Annotator Agreement Metrics}\label{sec:snt:iaa}

For estimating the inter-annotator agreement (IAA), we adopted the
popular $\kappa$ metric \cite{Cohen:60}.  Following the standard
practice, we computed this term as:
\begin{equation*}
  \kappa = \frac{p_o - p_c}{1 - p_c},
\end{equation*}
where $p_o$ denotes the observed agreement, and $p_c$ stands for the
agreement by chance.  We estimated the observed reliability in the
normal way as the ratio of tokens with matching annotations to the
total number of tokens:
\begin{equation*}
  p_o = \frac{T - A_1 + M_1 - A_2 + M_2}{T},
\end{equation*}
where $T$ represents the total token count, $A_1$ and $A_2$ are the
numbers of tokens annotated with the given class by the first and
second annotators respectively, and the $M$ terms mean the numbers of
tokens with matching annotations.  As usual, we computed the chance
agreement $p_c$ as:
\begin{equation*}\textstyle
  p_c = c_1 \times c_2 + (1.0 - c_1) \times (1.0 - c_2).
\end{equation*}
where $c_1$ and $c_2$ are the proportions of tokens annotated with the
given class in the first and second annotations, respectively, \ie{}
$c_1 = \frac{A_1}{T}$ and $c_2 = \frac{A_2}{T}$.

Two questions that arose during this computation though were
\begin{inparaenum}[(i)]
  \item whether tokens belonging to multiple overlapping annotation
    spans of the same class had to be counted several times in one
    annotation when computing the $A$ scores (for instance, whether we
    had to count the words ``dieses'' [\textit{this}], ``sch\"one''
    [\textit{nice}], and ``Buch'' [\textit{book}] in Example
    \ref{example:snt:iaa} twice as sentiments when computing $A_1$ and
    $A_2$), and
  \item whether we had to assume that two annotated spans from
    different experts agreed on all of their tokens if these spans had
    at least one word in common (\eg{} whether we had to consider the
    annotation of the token ``Mein'' [\textit{My}] in the example as
    matching, regarding that the rest of the corresponding
    \markable{sentiment}s agreed).
\end{inparaenum}

\begin{example}\label{example:snt:iaa}
\textcolor{red3}{\textbf{Annotation 1:}}\\
\upshape\sentiment{Mein Vater hasst \sentiment{dieses sch\"one Buch}.}\\
\sentiment{\itshape My father hates \upshape\sentiment{\itshape this
    nice book\upshape}.}

\noindent\textcolor{darkslateblue}{\textbf{\itshape Annotation 2:}}\\
Mein \sentiment{Vater hasst \sentiment{dieses sch\"one Buch}.}\\
\itshape My \upshape\sentiment{\itshape{}father hates \upshape\sentiment{\itshape this
    nice book\upshape}.}
\end{example}

To address these issues, we introduced two different agreement
metrics---\emph{binary} and \emph{proportional} kappa.  With the
former variant, we counted tokens belonging to overlapping annotation
spans of the same class multiple times (\ie{} $A_1$ and $A_2$ would
amount to $10$ and $9$, respectively, in the above tweet) and
considered all tokens belonging to the given annotated element as
matching if this span agreed with the annotation from the other expert
on at least one token (\ie{} $M_1$ and $M_2$ would have the same
values as $A_1$ and $A_2$ in this case).  With the latter metric,
every labeled token was counted only once (\ie{} the numbers of
labeled words in the first and second annotations would be $7$ and
$6$, respectively), and we only calculated the actual number of tokens
with matching labels when computing the $M$ scores (\ie{} both $M_1$
and $M_2$ would be equal to $6$).  The final value of the binary kappa
in Example~\ref{example:snt:iaa} would consequently run up to~1.0
because this metric would consider both annotations as perfectly
matching, since every labeled \markable{sentiment} agreed with the
other annotation on at least one token.  The proportional kappa,
however, would be equal to~0.0, since this metric would emphasize the
fact that the observed reliability $p_o$ is the same as the agreement
by chance $p_c$, and would therefore deem both labelings as
fortuitous.

\section{Annotation Procedure}\label{sec:astages}

After defining the agreement metrics, we finally let our experts
annotate the data.  The annotation procedure was performed in three
steps:
\begin{itemize}
  \item At the beginning, both annotators labeled one half of the
    corpus after only minimal training.  Unfortunately, their mutual
    agreement at this stage was relatively low, reaching only 31.21\%
    proportional-$\kappa$ for \markable{sentiment}s;
  \item In the second step, in order to improve the inter-rater
    reliability, we automatically determined all differences between
    the two annotations and highlighted non-matching tokens with a
    separate class of tags.  Then, we let the experts resolve these
    discrepancies by either correcting their own decisions or
    rejecting the variants of the other coder.  As in the previous
    stage, we allowed the annotators to consult their supervisor (the
    author of this thesis), also updating the FAQ section of the
    guidelines based on their questions, but did not let them
    communicate with each other directly.  This adjudication step
    significantly improved all annotations: The agreement on
    \markable{sentiment}s increased by 30.73\%, reaching 61.94\%.
    Similar effects were observed for \markable{target}s,
    \markable{source}s, \markable{polar term}s, and their modifiers;
  \item After resolving all differences, our assistants proceeded with
    the annotation of remaining files.  Working completely
    independently, one of the experts has annotated 78.8\% of the
    corpus, whereas the second annotator has labeled the complete
    dataset.
\end{itemize}

\section{Evaluation}\label{sec:eval}

\subsection{Initial Annotation Stage}\label{subsec:eval-initial-stage}

The agreement results of the initial annotation stage are shown in
Table~\ref{tbl:snt:agrmnt-init}.
\begin{table*}[thb!]
  \begin{center}
    \bgroup \setlength\tabcolsep{0.7\tabcolsep} \scriptsize
    \begin{tabular}{p{0.154\textwidth} 
        *{10}{>{\centering\arraybackslash}p{0.065\textwidth}}} 
      \toprule
          \multirow{2}{0.2\textwidth}{\bfseries Element} &
          \multicolumn{5}{c}{\bfseries Binary $\kappa$} & %
          \multicolumn{5}{c}{\bfseries Proportional $\kappa$}\\
          \cmidrule(r){2-6}\cmidrule(l){7-11}
          & $M_1$ & $A_1$ & $M_2$ & $A_2$ & $\mathbf{\kappa}$ %
          & $M_1$ & $A_1$ & $M_2$ & $A_2$ & $\mathbf{\kappa}$\\\midrule

          Sentiment & 4,215 & 7,070 & 3,484 & 9,827 & \textbf{38.05} &
          3,269 & 6,812 & 3,269 & 9,796 & \textbf{31.21}\\
          Target & 1,103 & 1,943 & 1,217 & 4,162 & \textbf{35.48} &
          898 & 1,905 & 898 & 4,148 & \textbf{26.85}\\
          Source & 159 & 445 & 156 & 456 & \textbf{34.53} &
          153 & 439 & 153 & 456 & \textbf{33.75}\\
          Polar Term & 1,951 & 2,854 & 2,029 & 3,188 & \textbf{64.29} &
          1,902 & 2,851 & 1,902 & 3,180 & \textbf{61.36}\\
          Intensifier & 57 & 101 & 59 & 123 & \textbf{51.71} &
          57 & 101 & 57 & 123 & \textbf{50.81}\\
          Diminisher & 3 & 10 & 3 & 8 & \textbf{33.32} &
          3 & 10 & 3 & 8 & \textbf{33.32}\\
          Negation & 21 & 63 & 21 & 83 & \textbf{28.69} &
          21 & 63 & 21 & 83 & \textbf{28.69}\\\bottomrule
    \end{tabular}
    \egroup
  \end{center}
  \captionof{table}[Inter-annotator agreement after the initial
    annotation stage]{Inter-annotator agreement after the initial
    annotation stage\\ {\small ($M1$ -- number of tokens with matching
      labels in the first annotation, $A1$ -- total number of tokens
      labeled with that class in the first annotation, $M2$ -- number
      of tokens with matching labels in the second annotation, $A2$ --
      total number of tokens labeled with that class in the second
      annotation)}}
  \label{tbl:snt:agrmnt-init}
\end{table*}

As we can see from the table, the inter-rater reliability of
\markable{sentiment}s strongly correlates with the inter-annotator
agreement on \markable{target}s and \markable{source}s, setting an
upper bound for these elements in the binary-$\kappa$ case.  With the
proportional metric, however, both \markable{sentiment}s and
\markable{target}s show worse results than \markable{source}s:
$31.21\%$ and $26.85\%$ versus $33.75\%$.  We explain this difference
by the fact, that \markable{sentiment}s and \markable{target}s are
typically represented by syntactic or discourse-level constituents
(noun phrases or clauses) and, even though the experts agreed on the
presence of these elements more often (as suggested by the
binary-$\kappa$ metric), reaching a consensus about the exact
boundaries of these elements was still a challenging task for them
despite an explicit clarification of this problem in the annotation
guidelines; \markable{source}s, on the other hand, are usually
expressed by pronouns, which rarely accept syntactic attributes, so
that their boundaries were easier to determine.  Nevertheless, even
with the binary metric, the agreement of all sentiment-level elements
is significantly below the $40\%$ threshold, which means only a slight
reliability according to the \citeauthor{Landis:77} scale
\cite{Landis:77}.

A different situation is observed for \markable{polar terms} and
\markable{intensifiers}.  The inter-annotator agreement of these
elements is above 50\%, for both $\kappa$-measures.  Obviously,
defining these entities as lexical units has significantly eased the
detection of their boundaries.  This effect becomes even more evident
if we look at \markable{diminisher}s and \markable{negation}s, where
the $A$ and $M$ scores are absolutely identical for both metrics.  It
means that both annotators always agreed on the boundaries of these
elements when they agreed on their presence.  Unfortunately, due to a
rather small number of these tags in the corpus (with only 3 cases of
\markable{diminisher}s and 21 cases of \markable{negation}s), the
overall agreement on these labels is relatively small too, amounting
to $33.32\%$ and $28.69\%$, respectively.

\subsection{Adjudication Step}\label{subsec:eval-adjudication-step}

Since these scores were unacceptable for running further experiments,
we decided to revise diverging annotations by letting our experts
recheck each other's decisions.
\begin{table*}[htb!]
  \begin{center}
    \bgroup \setlength\tabcolsep{0.7\tabcolsep} \scriptsize
    \begin{tabular}{p{0.155\textwidth} 
        *{10}{>{\centering\arraybackslash}p{0.065\textwidth}}} 
      \toprule
          \multirow{2}{0.2\textwidth}{\bfseries Element} &
          \multicolumn{5}{c}{\bfseries Binary $\kappa$} & %
          \multicolumn{5}{c}{\bfseries Proportional $\kappa$}\\
          \cmidrule(r){2-6}\cmidrule(l){7-11}
          & $M_1$ & $A_1$ & $M_2$ & $A_2$ & $\mathbf{\kappa}$ %
          & $M_1$ & $A_1$ & $M_2$ & $A_2$ & $\mathbf{\kappa}$\\
          \midrule

          Sentiment & 8,198 & 8,530 & 8,260 & 14,034 & \textbf{67.92} &
          7,435 & 8,243 & 7,435 & 13,714 & \textbf{61.94}\\

          Target & 3,088 & 3,407 & 2,814 & 5,303 & \textbf{65.66} &
          2,554 & 3,326 & 2,554 & 5,212 & \textbf{57.27}\\

          Source & 573 & 690 & 545 & 837 & \textbf{72.91} &
          539 & 676 & 539 & 833 & \textbf{71.12}\\

          Polar Term & 3,164 & 3,298 & 3,261 & 4,134 & \textbf{85.68} &
          3,097 & 3,290 & 3,097 & 4,121 & \textbf{82.64}\\

          Intensifier & 111 & 219 & 113 & 180 & \textbf{56.01} &
          111 & 219 & 111 & 180 & \textbf{55.51}\\

          Diminisher & 9 & 16 & 10 & 16 & \textbf{59.37} &
          9 & 16 & 9 & 15 & \textbf{58.05}\\

          Negation & 68 & 84 & 67 & 140 & \textbf{60.21} &
          67 & 83 & 67 & 140 & \textbf{60.03}\\\bottomrule
    \end{tabular}
    \egroup
  \end{center}
  \captionof{table}[Inter-annotator agreement after the adjudication
  step]{Inter-annotator agreement after the adjudication step\\
    {\small ($M1$ -- number of tokens with matching labels in the
      first annotation, $A1$ -- total number of labeled tokens in the
      first annotation, $M2$ -- number of tokens with matching labels
      in the second annotation, $A2$ -- total number of labeled tokens
      in the second annotation)}}
  \label{tbl:snt:agrmnt-adjud}
\end{table*}
As we can see from the results in Table~\ref{tbl:snt:agrmnt-adjud},
this procedure has significantly improved the inter-rater reliability
of all annotated elements: the binary scores of \markable{sentiment}s
and \markable{target}s increased by $29.87\%$ and $30.18\%$,
respectively.  An even greater improvement is observed for
\markable{source}s, whose binary kappa improved by remarkable
$38.38\%$.  A similar tendency applies to the proportional metric,
where the agreement of \markable{sentiment}s gained $30.73\%$,
reaching $61.94\%$.  Likewise, the reliability of opinion targets and
holders improved by $30.42\%$ and $37.37\%$, running up to $57.27\%$
and $71.12\%$.



As in the previous step, the highest agreement scores are attained by
\markable{polar term}s, whose reliability notably surpasses the 80\%
benchmark, which means an almost perfect agreement.  Interestingly
enough, only 193 out of 3,290 terms annotated by the first expert did
not match the labelings of the second annotator.  Another interesting
observation is that the difference between the binary and proportional
scores of \markable{polar terms} only amounts to 3.04\%, which implies
that the assistants could unproblematically determine the boundaries
of these elements in most of the cases.

Somewhat surprisingly, the agreement of \markable{intensifier}s
improved notably less.  A closer look at the annotated cases revealed
that the majority of their disagreements stemmed from different takes
of exclamation marks: the first expert ignored these punctuation
marks, whereas the second annotator considered them as valid
intensifying elements.  Nevertheless, even despite these diverging
interpretations, the reliability of \markable{intensifier}s is above
$55\%$, which means a moderate level.

\subsection{Final Annotation Stage}\label{subsec:eval-final-annotation}

After ensuring that our annotators could reach an acceptable quality
of annotation, we eventually let them label the remaining part of the
data.  The agreement results of the final stage computed on the files
annotated by both experts are given in
Table~\ref{tbl:snt:agrmnt-final}.
\begin{table*}[thb!]
  \begin{center}
    \bgroup \setlength\tabcolsep{0.7\tabcolsep} \scriptsize
    \begin{tabular}{p{0.155\textwidth} 
        *{10}{>{\centering\arraybackslash}p{0.065\textwidth}}} 
      \toprule
          \multirow{2}{0.2\textwidth}{\bfseries Element} &
          \multicolumn{5}{c}{\bfseries Binary $\kappa$} & %
          \multicolumn{5}{c}{\bfseries Proportional $\kappa$}\\
          \cmidrule(r){2-6}\cmidrule(l){7-11}
          & $M_1$ & $A_1$ & $M_2$ & $A_2$ & $\mathbf{\kappa}$ %
          & $M_1$ & $A_1$ & $M_2$ & $A_2$ & $\mathbf{\kappa}$\\
          \midrule

          Sentiment & 14,748 & 15,929 & 14,969 & 26,047 & \textbf{65.03} &
          13,316 & 15,375 & 13,316 & 25,352 & \textbf{58.82}\\

          Target & 5,765 & 6,629 & 5,292 & 9,852 & \textbf{64.76} &
          4,789 & 6,462 & 4,789 & 9,659 & \textbf{56.61}\\

          Source & 966 & 1,207 & 910 & 1,619 & \textbf{65.99} &
          898 & 1,180 & 898 & 1,604 & \textbf{64.1}\\

          Polar Term & 5,574 & 5,989 & 5,659 & 7,419 & \textbf{82.83} &
          5,441 & 5,977 & 5,441 & 7,395 & \textbf{80.29}\\

          Intensifier & 192 & 432 & 194 & 338 & \textbf{49.97} & 192 &
          432 & 192 & 338 & \textbf{49.71}\\

          Diminisher & 16 & 30 & 17 & 34 & \textbf{51.55} & 16 & 30 &
          16 & 33 & \textbf{50.78}\\

          Negation & 111 & 132 & 110 & 243 & \textbf{58.87} & 110 &
          131 & 110 & 242 & \textbf{58.92}\\\bottomrule
    \end{tabular}
    \egroup
  \end{center}
  \captionof{table}[Inter-annotator agreement of the final
  corpus]{Inter-annotator agreement of the final corpus\\ {\small
      ($M1$ -- number of tokens with matching labels in the first
      annotation, $A1$ -- total number of labeled tokens in the first
      annotation, $M2$ -- number of tokens with matching labels in the
      second annotation, $A2$ -- total number of labeled tokens in the
      second annotation)}}
  \label{tbl:snt:agrmnt-final}
\end{table*}

This time, we can observe a slight decrease of the results: the
proportional score for \markable{sentiment}s dropped by $3.12\%$,
whereas the agreement on \markable{target}s was more persistent and
lost only $0.66\%$, going down to $56.61\%$.  The most dramatic
changes occurred for \markable{source}s, whose proportional value
deteriorated by notable $7.02\%$, sinking to $64.1\%$.  Nonetheless,
the average proportional agreement of all these elements is around
$60.5\%$, which is almost twice as high as the mean reliability
achieved in the first stage.

As before, the scores of \markable{polar term}s are in the ballpark of
almost perfect results.  Their modifying elements, however, show a
decrease: the agreement of \markable{intensifier}s deteriorated by
5.8\%, sinking to 49.71\% proportional kappa.  A similar situation is
observed for \markable{diminisher}s, whose kappa worsened from
$58.05\%$ to $50.78\%$.  The best persistence in this regard is shown
by \markable{negation}s, where the quality dropped by only $1.11\%$,
which can be considered as a very good result, regarding the small
number of these elements in the corpus.

In general, we can see that the reliability of all elements in the
final dataset is at least moderate, with \markable{polar term}s being
the most reliably annotated elements ($\kappa_{\textrm{p}}=80.29\%$),
and \markable{intensifier}s setting a lower bound on the agreement
($\kappa_{\textrm{p}}=49.71\%$).

\subsection{Qualitative Analysis}\label{subsec:eval-qualitative-analysis}

In order to understand the reasons for remaining conflicts, we decided
to have a closer look at the diverging cases.  A sample sentence with
different analyses of \markable{sentiment}s is shown in
Example~\ref{snt:exmp:sent-disagr}:
\begin{example}\label{snt:exmp:sent-disagr}
  \textcolor{red3}{\textbf{Annotation 1:}}\\ \upshape{}@TinaPannes
  immerhin ist die \#afd nicht dabei \smiley{}\\[0.8em]\itshape
  \noindent\textcolor{darkslateblue}{\textbf{\itshape Annotation
      2:}}\\ \upshape{}@TinaPannes
  \sentiment{\textcolor{red}{\target{immerhin ist die \#afd nicht
      dabei} \smiley{}}}\\[0.8em]
  \noindent\itshape{}@TinaPannes
  \upshape\sentiment{\textcolor{red}{\itshape{}\target{anyway the
        \#afd is not there} \smiley{}}\upshape{}}
\end{example}
In this tweet, the first annotator obviously overlooked the emoticon
\smiley{} at the end of the message, whereas the second expert
correctly recognized it as an evaluation of the previous sentence.
Because the first assistant did not label any \markable{sentiment} at
all, she also automatically disagreed on the \markable{target} of this
opinion.


A much rarer case of diverging \markable{target} annotations was when
both experts actually marked a \markable{sentiment} span.  An example
of such situation is shown in the following message:
\begin{example}\label{snt:exmp:targt-disagr}
  \textcolor{red3}{\textbf{Annotation
      1:}}\\
  \upshape{}\sentiment{Koalition wirft der SPD
    \target{\textcolor{red}{Blockadehaltung}} vor}\\[0.5em]
  \noindent\itshape{}\sentiment{Coalition accuses the SPD of
    \target{\textcolor{red}{blocking politics}}}\\[0.6em]\itshape

  \noindent\textcolor{darkslateblue}{\textbf{\itshape Annotation
      2:}}\\
  \upshape{}\sentiment{Koalition wirft \target{\textcolor{red}{der SPD}}
    Blockadehaltung vor}\\[0.5em]
  \noindent\itshape{}\sentiment{Coalition accuses
    \target{\textcolor{red}{the SPD}} of blocking politics}
\end{example}
In this sentence, the first expert considered \emph{blocking politics}
as the main object of criticism, whereas the second annotator regarded
the political party accused of such behavior as sentiment's target.
In our opinion, both of these interpretations are correct and,
ideally, two \markable{sentiment}s had to be labeled in this message:
one with the target ``Blockadehaltung'' (\emph{blocking politics}) and
another one with the target ``die SPD'' (\emph{the SPD}).

Although our annotators were much more consistent about the analysis
of \markable{polar term}s, we still decided to have a look at
disagreeing labels of these elements.  A sample case of differently
annotated \markable{polar term}s is given in
Example~\ref{snt:exmp:emo-disagr}
\begin{example}\label{snt:exmp:emo-disagr}
  \textcolor{red3}{\textbf{Annotation 1:}}\\ \upshape{}Syrien vor dem
  Angriff---bringen diese Bomben den Frieden?\\[0.3em]\itshape
  \noindent\itshape{}Syria facing an attack---will these bombs bring
  peace?\\

  \noindent\textcolor{darkslateblue}{\textbf{\itshape Annotation
      2:}}\\ \upshape{}Syrien vor dem
  \emoexpression{\textcolor{red}{Angriff}}---bringen diese
  \emoexpression{\textcolor{red}{Bomben}} den\\
  \emoexpression{\textcolor{red}{Frieden}}?\\[0.3em]
  \noindent\itshape{}Syria facing an
  \upshape\emoexpression{\textcolor{red}{\itshape{}attack}\upshape}\itshape{}---will
  these
  \upshape\emoexpression{\textcolor{red}{\itshape{}bombs}\upshape}\itshape{}
  bring
  \upshape\emoexpression{\textcolor{red}{\itshape{}peace}\upshape}\itshape{}?
\end{example}
The obvious reason for the misclassifications in this message is the
notorious subjective facts: As you can see, the first assistant
ignored the words ``Angriff'' (\emph{attack}), ``Bombe''
(\emph{bomb}), and ``Frieden'' (\emph{peace}), while the second
annotator considered them as polar items.
We should, however, admit that this difference is partially due to the
adjudication procedure that we used in step two; because at the
initial stage, our experts had had opposite preferences regarding
these entities: the first annotator had labeled these terms, whereas
the second assistant had usually skipped them.  During the revision,
however, both assistants have changed their minds after looking at the
decisions of the other linguist.  Therefore, one needs to keep in mind
the risk of mutual concession when applying the adjudication method in
the future.


\subsection{Attributes Agreement}\label{subsec:eval-qualitative-analysis}

In order to see whether our annotators also agreed on the attributes
of the tags, we estimated the Cohen's kappa for the polarity and the
Krippendorff's alpha~\cite{Krippendorff:07} for the intensity of
matching \markable{sentiment}s and \markable{polar term}s.  The reason
for choosing two different metrics in this case is that
\attribute{polarity} is a categorical feature, whose value takes on
one of the predefined classes (\emph{positive}, \emph{negative}, or
\emph{comparison}), whereas \attribute{intensity} is an ordinal
attribute, whose value can range on a scale from zero (\emph{weak}) to
two (\emph{strong}).  Since disagreements that are further apart on
the scale need to be penalized more strongly than small divergences,
we decided to use the $\alpha$-measure for this attribute, as it
explicitly addresses this problem.
\begin{table}[thb!]
  \begin{center}
    \bgroup \setlength\tabcolsep{0.47\tabcolsep} \scriptsize
    \begin{tabular}{p{0.23\columnwidth}%
          *{2}{>{\centering\arraybackslash}p{0.2\columnwidth}}} 
      \toprule
          {\bfseries Element} & {\bfseries Polarity $\kappa$} & %
          {\bfseries Intensity $\alpha$}\\\midrule
          Sentiment & 58.8 & 73.54\\
          Polar Term & 87.12 & 78.79\\
          \bottomrule
    \end{tabular}
    \egroup
    \caption{Inter-annotator agreement on polarity and intensity of
      sentiments and polar terms}
    \label{tbl:attr-agrmnt}
  \end{center}
\end{table}

As we can see from the results in Table~\ref{tbl:attr-agrmnt},
reaching a consensus about the polarities of \markable{polar term}s is
a much easier task than agreeing on the semantic orientation of
\markable{sentiment}s.  As in Example~\ref{snt:exmp:sent-disagr}, one
of the main reasons for these disagreements is opinions containing
emoticons, especially in the cases when the polarity of the smiley
contradicts the polarity of the preceding text, \eg{} ``Ich hasse die
Piratenpartei \smiley{}'' (\emph{I hate the Pirate Party {\upshape
    \smiley{}}}).

Interestingly enough, the inter-rater agreement on the intensity of
\markable{sentiment}s ($\alpha = 73.54$) is notably higher than the
corresponding score for their polarity ($\kappa = 58.8$), although the
opposite situation is observed for \markable{polar term}s, whose
$\alpha$-value ($78.79$) is almost ten percent lower than $\kappa$
(87.12).  This means that the annotators could easily determine the
semantic orientation of a single word, but had difficulties with
agreeing on the strength of its meaning.  Vice versa, when dealing
with targeted opinions, they usually assigned the \emph{medium}
intensity to most \markable{sentiment}s, but could disagree on the
polarity of these statements.

For the sake of completeness, we compared these results with the
scores obtained on the MPQA dataset \cite[see][pp. 38, 80]{Wilson:07}.
The average $\alpha$-agreement on the intensity of direct subjective
and objective speech events (a rough counterpart of our
\markable{sentiment}s) in this corpus was around 79\%; the
corresponding results for the intensity of expressive subjective
elements (\markable{polar term}s in our case), however, were much
worse, amounting to only 46\%, even though the $\kappa$-value for
their polarity run up to 72\%.  Hence, the reliability of annotated
attributes in our corpus still outperforms the respective agreement in
MPQA on almost all aspects except for sentiment intensity.


\subsection{Effect of the Selection Criteria}\label{subsec:eval-selection-criteria}

\begin{figure*}[htbp!]
{
\centering
\begin{subfigure}{.5\textwidth}
  \centering
  \includegraphics[width=\linewidth]{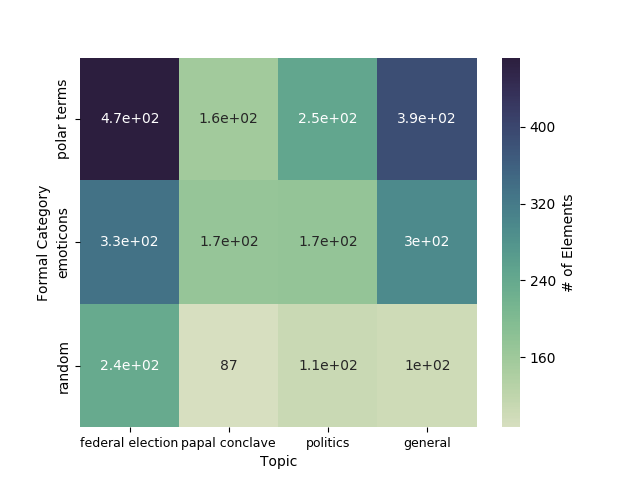}
  \caption{\texttt{Sentiments}}\label{snt:fig:crp-sent-emo-distr-a}
\end{subfigure}%
\begin{subfigure}{.5\textwidth}
  \centering
  \includegraphics[width=\linewidth]{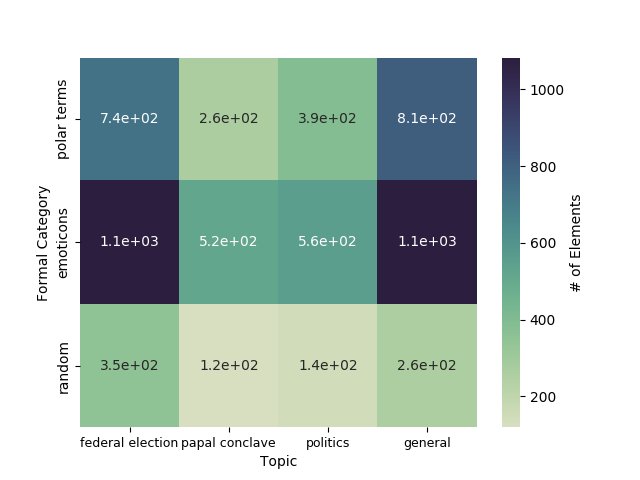}
  \caption{\texttt{Polar Terms}}\label{snt:fig:crp-sent-emo-distr-b}
\end{subfigure}
}
\caption{Distribution of sentiments and polar terms across topics and
  formal groups}\label{snt:fig:crp-sent-emo-distr}
\end{figure*}

\begin{figure*}[htbp!]
{
\centering
\begin{subfigure}{.5\textwidth}
  \centering
  \includegraphics[width=\linewidth]{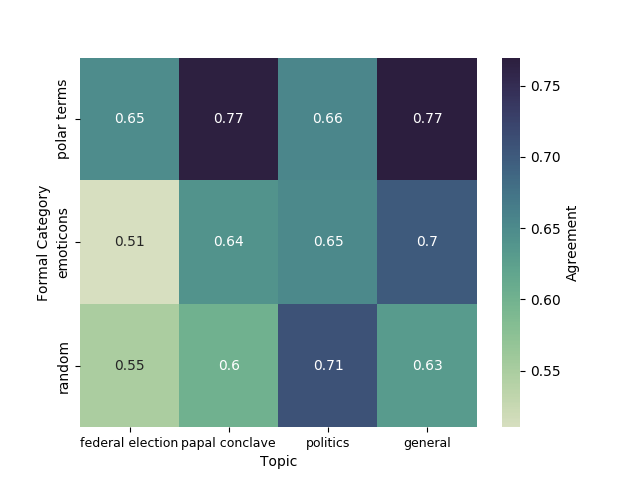}
  \caption{\texttt{Sentiments}}
\end{subfigure}%
\begin{subfigure}{.5\textwidth}
  \centering
  \includegraphics[width=\linewidth]{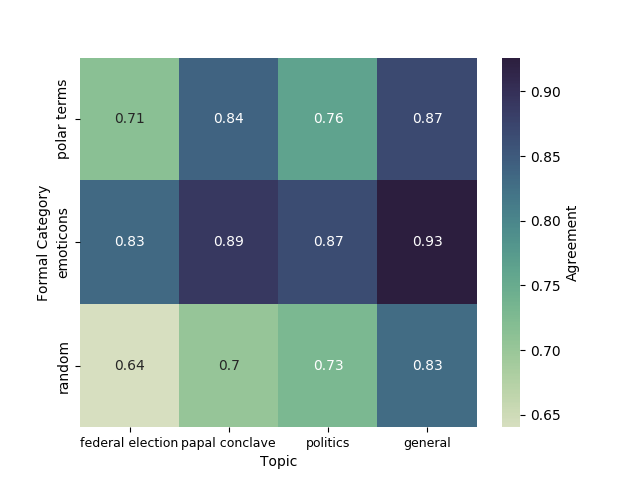}
  \caption{\texttt{Polar Terms}}
\end{subfigure}
}
\caption{Inter-annotator agreement on sentiments and polar terms
  across topics and formal groups}\label{snt:fig:crp-sent-emo-agr}
\end{figure*}

Finally, in order to check how the selection criteria that we applied
initially when sampling the corpus data affected the resulting
distribution of \markable{sentiment}s and \markable{polar term}s in
the final dataset, we plotted the frequencies and agreement scores of
these elements across topics and formal groups, and present these
statistics in Figures~\ref{snt:fig:crp-sent-emo-distr}
and~\ref{snt:fig:crp-sent-emo-agr}.

As we can see from the plots, topics and form clearly affect both the
number of opinions and the difficulty of their interpretation.
According to Figure~\ref{snt:fig:crp-sent-emo-distr}, the greatest
number of \markable{sentiment}s occur in tweets pertaining to the
federal elections and in messages representing casual everyday
conversations.  A similar tendency is observed for \markable{polar
  term}s, but in this case, the form of the microblogs seems to have
more impact on the elements' distribution than their topics.


Regarding the inter-annotator agreement, we can see that
\markable{sentiment}s and \markable{polar term}s are most reliably
annotated in messages from the German Twitter Snapshot.  Moreover, the
former elements are apparently easiest to annotate in tweets that were
preselected using a sentiment lexicon, whereas \markable{polar term}s
are easiest to analyze in microblogs that contain emoticons.

To confirm the correlation between the topics and formal groups on the
one hand and the number and reliability of \markable{sentiment}s and
\markable{polar term}s on the other hand, we computed the correlation
coefficients ($\rho$) of these factors, considering each particular
topic and formal group as a binary variable and measuring the
association of this variable with the number and agreement of
annotated elements.
\begin{table}[thb!]
  \begin{center}
    \bgroup\setlength\tabcolsep{0.47\tabcolsep}\scriptsize
    \begin{tabular}{p{0.2\columnwidth}%
          *{4}{>{\centering\arraybackslash}p{0.185\columnwidth}}} 
      \toprule

      \multirow{3}{0.2\columnwidth}{\centering\bfseries Selection Criteria} & %
      \multicolumn{4}{c}{\bfseries Correlation Coefficients}\\\cmidrule(lr){2-3}\cmidrule(lr){4-5}

      & \multicolumn{2}{c}{\bfseries Sentiment}& %
      \multicolumn{2}{c}{\bfseries Polar Term}\\\cmidrule(lr){2-2}\cmidrule(lr){3-3}%
      \cmidrule(lr){4-4}\cmidrule(lr){5-5}

      & \# of elements & agreement & \# of elements & agreement\\\midrule

      \multicolumn{5}{c}{\cellcolor{cellcolor}Topical Groups}\\
      Federal Elections & \textbf{0.312} & 0.169 & 0.356 & 0.289\\
      Papal Conclave & 0.149 & 0.124 & 0.182 & 0.264\\
      Political Discussions & 0.195 & 0.148 & 0.218 & 0.244\\
      General Conversations & 0.183 & \textbf{0.19} & \textbf{0.372} & \textbf{0.452}\\
      \multicolumn{5}{c}{\cellcolor{cellcolor}Formal Categories}\\
      Polar Terms & \textbf{0.445} & \textbf{0.352} & 0.38 & 0.301\\
      Emoticons & 0.127 & 0.096 & \textbf{0.47} & \textbf{0.615}\\
      Random & 0.216 & 0.134 & 0.143 & 0.138\\
      \bottomrule
    \end{tabular}
    \egroup
    \caption[Correlation coefficients of topics and selection criteria
      with the number and agreement of sentiments and polar
      terms]{Correlation coefficients of topics and formal selection
      criteria with the number and agreement scores of sentiments and
      polar terms}
    \label{sent:tbl:corr-coeff}
  \end{center}
\end{table}

As we can see from the results in Table~\ref{sent:tbl:corr-coeff},
both criteria (topics and form) have a positive correlation with the
number of annotated elements and their reliability.  The highest
$\rho$-score for \markable{sentiment}s is achieved by tweets
describing federal elections and messages containing polar terms,
where it amounts to 0.312 and 0.445, respectively.  A slightly
different situation is observed for \markable{polar term}s: the
highest scores for this element both in terms of the number of
annotated items and their reliability are achieved by casual everyday
conversations and tweets that contain emoticons.

\section{Summary and Conclusions}

Now that we have reached the end of the second chapter, we would like
to remind the reader that in this part of the thesis we have presented
the Potsdam Twitter Sentiment Corpus (PotTS), a collection of 7,992
German microblogs that had been manually annotated by two human
experts with sentiments, targets, sources, polar terms, and their
modifying elements.

We obtained initial data for this corpus by tracking tweets about
German federal elections, papal conclave, discussions of general
political topics, and casual everyday conversations between spring and
autumn 2013.  Afterwards, we grouped these messages into three classes
(tweets containing polar terms, microblogs containing exclamation
marks or emoticons, and the rest of the messages) and randomly sampled
666 posts from each of these classes for each topic.


The annotation process was performed in three steps: first, the
annotators labeled one half of the data after minimal training; then,
we automatically highlighted their divergent analyses and asked them
to resolve these differences; finally, our assistants continued with
the analysis of the remaining files.

To estimate the inter-rater reliability, we introduced two modified
versions of the established $\kappa$-metric---binary and proportional
kappa---which differ in the way how they treat overlapping annotations
and partial matches.  Using these measures, we estimated the
inter-annotator agreement of our experts at different stages of their
work.  This study showed that, initially, our assistants could hardly
agree on the mere notion of targeted opinions, but their disagreements
could be resolved with the help of the adjudication procedure that we
applied in step two.  Despite a small drop of the IAA scores in the
final stage, all $\kappa$-values still remained at the level of at
least moderate reliability.

Finally, we demonstrated that our initial selection criteria had a
strong impact on the number and agreement of annotated sentiments and
polar terms, with tweets about federal elections and messages without
prefiltered topics being the most prolific sources of these elements.

That way, we not only contributed to the inventory of available
sentiment and social-media resources for German but also provided new
insights into different sampling methods that could be used to create
an opinion dataset and described the consequences of applying these
methods in practice.  A detailed inter-annotator agreement study
showed precisely which topics yield most subjective opinions
(elections and casual conversations) and which groups of messages are
especially difficult to annotate (tweets containing emoticons and
microblogs without polar terms or emoticons).  In the next step, we
are going to check whether our dataset can also serve as a basis for
building and evaluating automatic opinion mining applications.



\chapter{Sentiment Lexicons}\label{chap:snt:lex}

The first avenue that we are going to explore using the obtained data
is automatic prediction of polar terms.
For this purpose, we will first evaluate existing German sentiment
lexicons on our corpus.  Since most of these resources, however, were
created semi-automatically by translating English polarity lists and
then manually post-editing and expanding these translations, we will
also look whether methods that were used to produce original English
lexicons would yield comparable results when applied to German data
directly.  Finally, we will analyze whether one of the most popular
areas of research in contemporary computational linguistics,
distributed vector representations of words~\cite{Mikolov:13}, can
produce better polarity lists than previous approaches.  In the
concluding step, we will investigate the effect of different
hyper-parameters and seed sets on these systems, summarizing and
concluding our findings in the last part of this chapter.

\section{Data}\label{sec:snt-lex:data}

As \emph{development set} for our experiments, we will use \emph{400
  randomly selected tweets} annotated by the first expert.  As gold
\emph{test set} for evaluating the lexicons, we will utilize the
\emph{complete corpus labeled by the second linguist}.  These test
data comprise a total of 6,040 positive and 3,055 negative terms.  But
because many of these expressions represent emoticons, which, on the
one hand, are a priori absent in common lexical taxonomies such as
\textsc{WordNet}~\cite{Miller:95,Miller:07} or
\textsc{GermaNet}~\cite{Hamp:97} and therefore not amenable to methods
that rely on these resources, but on the other hand, can be easily
captured by regular expressions, we decided to exclude non-alphabetic
smileys altogether from our study.  This left us with a set of 3,459
positive and 2,755 negative labeled terms (1,738 and 1,943 unique
expressions, respectively), whose $\kappa$-agreement run up to 59\%.

\section{Evaluation Metrics}\label{sec:snt-lex:eval-metrics}

An important question that needs to be addressed before we proceed
with the experiments is which evaluation metrics we should use to
measure the quality of sentiment lexicons.  Usually, this quality is
estimated either \textit{intrinsically} (by taking a lexicon in
isolation and immediately assessing its accuracy) or
\textit{extrinsically} (by considering the lexicon within the scope of
a bigger application, \eg{} a supervised classifier that uses
lexicon's entries as features).

Traditionally, intrinsic evaluation of English polarity lists amounts
to comparing these resources with the General Inquirer
lexicon~\cite[GI; ][]{Stone:66}, a manually compiled list of 11,895
words annotated with their semantic categories.  For this purpose,
researchers usually take the intersection of the two sets and estimate
the percentage of matches in which automatically induced polar terms
have the same polarity as corresponding GI entries.  This evaluation,
however, is somewhat problematic: First of all, it is not easily
transferable to other languages because even a manual translation of
GI is not guaranteed to cover all language- and domain-specific polar
expressions.  Second, since it only considers the intersection of the
two sets, it does not penalize for low recall, so that a polarity list
that consists of just two terms \textit{good}$^+$ and \textit{bad}$^-$
will always have the highest possible score, often surpassing other
lexicons with a greater number of entries.  Finally, such comparison
does not account for polysemy.  As a result, an ambiguous word only
one of whose (possibly rare) senses is subjective will always be
ranked the same as an obvious polar term.

Unfortunately, an extrinsic evaluation does not always provide a
remedy in this case because different extrinsic applications might
yield different results, and a polarity list that performs best with
one system can produce fairly low scores with another application.

Instead of using these methods, we decided to evaluate sentiment
lexicons directly on our corpus by comparing their entries with the
annotated polar terms, since such approach would allow us to solve at
least three of the aforementioned problems, namely,
\begin{inparaenum}[(i)]
  \item it would account for recall,
  \item it would distinguish between different senses of polysemous
    words,\footnote{The annotators of our corpus were asked to label a
      polar term iff the actual sense of this term in the given
      context was polar.} and
  \item it would preclude intermediate modules that could artificially
    improve or worsen the results.
\end{inparaenum}

In particular, in order to evaluate a lexicon on our dataset, we
represent this polarity list as a case-insensitive
trie~\cite[pp. 492--512]{Knuth:98} and compare this trie with the
original and lemmatized\footnote{All lemmatizations in our experiments
  were performed using the \textsc{TreeTagger} of \citet{Schmid:95}.}
corpus tokens, successively matching them from left to right.  We
consider a match as correct if a lexicon term completely agrees with
the (original or lemmatized) tokens of an annotated \markable{polar
  term} and has the same polarity as the labeled element.  All corpus
tokens that are not marked as \markable{polar term}s in the corpus are
considered as gold neutral words; similarly, all terms that are absent
from the lexicon, but present in the corpus are assumed to have a
predicted neutral polarity.

\section{Semi-Automatic Lexicons}

Using this metric, we first estimated the quality of existing German
polarity lists:
\begin{itemize}
\item\textbf{German Polarity Clues}~\cite[GPC;][]{Waltinger:10}, which
  contains 10,141 polar terms from the English sentiment lexicons
  Subjectivity Clues~\cite{Wilson:05} and SentiSpin~\cite{Takamura:05}
  that were automatically translated into German and then manually
  revised by the author.  Apart from that, \citeauthor{Waltinger:10}
  also manually enriched these translations with their frequent
  synonyms and 290 negated phrases;\footnote{In our experiments, we
    excluded the auxiliary words ``aus'' (\emph{from}), ``der''
    (\emph{the}), ``keine'' (\emph{no}), ``nicht'' (\emph{not}),
    ``sein'' (\emph{to be}), ``was'' (\emph{what}), and ``wer''
    (\emph{who}) with their inflection forms from the German Polarity
    Clues, because these entries significantly worsened the evaluation
    results.}

\item\textbf{SentiWS}~\cite[SWS;][]{Remus:10}, which includes 1,818
  positively and 1,650 negatively connoted terms along with their
  part-of-speech tags and inflections, which results in a total of
  32,734 word forms.  As in the previous case, the authors obtained
  the initial entries for their resource by translating an English
  polarity list (the General Inquirer lexicon) and then manually
  correcting these translations.  In addition to this, they expanded
  the translated set with words and phrases that frequently
  co-occurred with positive and negative seed terms in a corpus of
  10,200 customer reviews or in the German Collocation
  Dictionary~\cite{Quasthoff:10};\footnote{Unfortunately, the authors
    do not provide a breakdown of how many terms were obtained through
    translation and how many of them were added during the expansion.}

\item and, finally, the only the lexicon that was not obtained through
  translation---the \textbf{Zurich Polarity
    List}~\cite[ZPL;][]{Clematide:10}, which features 8,000 subjective
  entries extracted from \textsc{GermaNet} synsets~\cite{Hamp:97}.
  These synsets had been manually annotated by human experts with
  their prior polarities.  Since the authors, however, found the
  number of polar adjectives obtained this way to be insufficient for
  their classification experiments, they automatically enriched this
  lexicon with more attributive terms, using the collocation method of
  \citet{Hatzivassi:97}.
\end{itemize}

For our evaluation, we tested each of the three lexicons in isolation,
and also evaluated their union and intersection in order to check for
possible ``synergy'' effects.  The results of this computation are
shown in Table~\ref{snt-lex:tbl:gsl-res}.

\begin{table}[h]
  \begin{center}
    \bgroup\setlength\tabcolsep{0.1\tabcolsep}\scriptsize
    \begin{tabular}{p{0.167\columnwidth} 
        *{9}{>{\centering\arraybackslash}p{0.074\columnwidth}} 
        *{2}{>{\centering\arraybackslash}p{0.068\columnwidth}}} 
      \toprule
      \multirow{2}*{\bfseries Lexicon} & %
      \multicolumn{3}{c}{\bfseries Positive Expressions} & %
      \multicolumn{3}{c}{\bfseries Negative Expressions} & %
      \multicolumn{3}{c}{\bfseries Neutral Terms} & %
      \multirow{2}{0.068\columnwidth}{\bfseries\centering Macro\newline \F{}} & %
      \multirow{2}{0.068\columnwidth}{\bfseries\centering Micro\newline \F{}}\\
      \cmidrule(lr){2-4}\cmidrule(lr){5-7}\cmidrule(lr){8-10}

      & Precision & Recall & \F{} & %
      Precision & Recall & \F{} & %
      Precision & Recall & \F{} & & \\\midrule


      GPC & 0.209 & 0.535 & 0.301 & %
      0.195 & 0.466 & 0.275 & %
      0.983 & 0.923 & 0.952 & %
      0.509 & 0.906 \\


      SWS & 0.335 & 0.435 & 0.379 & %
      0.484 & 0.344 & \textbf{0.402} & %
      0.977 & 0.975 & 0.976 & %
      0.586 & 0.952\\


      ZPL & 0.411 & 0.424 & 0.417 & %
      0.38 & 0.352 & 0.366 & %
      0.977 & 0.979 & 0.978 & %
      0.587 & 0.955 \\


      GPC $\cap$ SWS $\cap$ ZPL & \textbf{0.527} & 0.372 & \textbf{0.436} & %
      \textbf{0.618} & 0.244 & 0.35 & %
      0.973 & \textbf{0.99} & \textbf{0.982} & %
      \textbf{0.589} & \textbf{0.964} \\


      GPC $\cup$ SWS $\cup$ ZPL & 0.202 & \textbf{0.562} & 0.297 & %
      0.195 & \textbf{0.532} & 0.286 & %
      \textbf{0.985} & 0.917 & 0.95 & %
      0.51 & 0.901 \\\bottomrule
    \end{tabular}
    \egroup{}
    \caption[Evaluation of semi-automatic German sentiment lexicons]{
      Evaluation of semi-automatic German sentiment lexicons\\ {\small
        GPC --- German Polarity Clues, SWS --- SentiWS, ZPL --- Zurich
        Polarity List}}\label{snt-lex:tbl:gsl-res}
  \end{center}
\end{table}

As we can see from the table, the intersection of all three polarity
lists achieves the best results for the positive and neutral classes,
and also attains the highest macro- and micro-averaged \F{}-scores.
One of the main reasons for this success is a relatively high
precision of this set for all polarities except neutral, where the
intersection is outperformed by the union of three lexicons.  The last
fact is also not surprising, as the union has the highest recall of
positive and negative terms.  Among all compared lexicons, the results
of the Zurich Polarity List come closest to the scores of the
intersected set: its macro-\F{} is lower by 0.002, and its
micro-average is less by 0.009.  The second-best lexicon is SentiWS,
which reaches the highest \F-score for the negative class, but has a
lower precision of positive entries.  Finally, German Polarity Clues
is the least reliable sentiment resource, which is also mainly due to
the low precision of its polar terms.

\section{Automatic Lexicons}

A natural question that arises upon evaluation of existing
semi-automatic lexicons is how well fully automatic methods can
perform in comparison with these resources.  According to
\citet[p. 79]{Liu:12}, most automatic sentiment lexicon generation
(SLG) algorithms can be grouped into two main classes: dictionary- and
corpus-based ones.  The former systems induce polarity lists from
monolingual thesauri or lexical databases such as the Macquarie
Dictionary~\cite{Bernard:86} or \textsc{WordNet}~\cite{Miller:95}.  A
clear advantage of these methods is their relatively high precision,
as they operate on manually annotated, carefully verified data.  At
the same time, this precision might come at the price of reduced
recall, especially in domains whose language changes very rapidly and
where new terms are coined in a flash.  In contrast to this,
corpus-based systems operate directly on unlabeled in-domain texts
and, consequently, have access to all neologisms; but the downside of
these approaches is that they often have to deal with extremely noisy
input and might therefore have low accuracy.  Since it was unclear to
us which of these pros and cons would have a stronger influence on the
net results, we decided to reimplement the most popular algorithms
from both of these groups and evaluate them on our corpus.

\subsection{Dictionary-Based Methods}

The presumably first SLG system that inferred a sentiment lexicon from
a lexical database was proposed by \citet{Hu:04}.  In their work on
automatic classification of customer reviews, the authors
automatically compiled a list of polar adjectives (which were supposed
to be the most relevant part of speech for mining people's opinions)
by taking a set of seed terms with known semantic orientations and
propagating the polarity scores of these seeds to their
\textsc{WordNet} synonyms.  A similar procedure was also applied to
antonyms, but the polarity values were reversed in this case.  This
expansion continued until no more adjectives could be reached via
synonymy-antonymy links.

\citet{Blair-Goldensohn:08} refined this approach by considering
polarity scores of all \textsc{WordNet} terms as a single vector
$\vec{v}$; the values of all negative seeds in this vector were set to
$-1$, and the scores of all positive seed terms were fixed to $+1$.
To derive their polarity list, the authors multiplied $\vec{v}$ with
an adjacency matrix $A$.  Each cell $a_{ij}$ in this matrix was set to
$\lambda=0.2$, if there was a synonymy link between synsets $i$ and
$j$, and to $-\lambda$, if these synsets were antonymous to each
other.  By performing this multiplication multiple times and storing
the results of the previous iterations in the $\vec{v}$ vector, the
authors ensured that all polarity scores were propagated transitively
through the network, decaying by a constant factor ($\lambda$) as the
length of the paths starting from the original seeds increased.


With various modifications, the core idea of \citet{Hu:04} was
adopted by almost all dictionary-based works: For example,
\citet{Kim:04,Kim:06} estimated the probability of word $w$ belonging
to polarity class $c \in \{\textrm{positive, negative, neutral}\}$ as:
\begin{equation*}
  P(c|w) = P(c)P(w|c) = P(c)\frac{\sum\limits_{i=1}^{n}count(syn_i,
    c)}{count(c)},
\end{equation*}
where $P(c)$ is the prior probability of that class (estimated as the
number of words belonging to class $c$ divided by the total number of
words); $count(syn_i, c)$ denotes the number of times a seed term with
polarity $c$ appeared in a synset of $w$; and $count(c)$ means the
total number of synsets that contain seeds with this polarity.  Using
this formula, the authors successively expanded their initial set of
34 adjectives and 44 verbs to a list of 18,192 polar terms.

Another popular dictionary-based resource, \textsc{SentiWordNet}, was
created by \citet{Esuli:06c}, who enriched a small set of positive and
negative seed adjectives with their \textsc{WordNet} synonyms and
antonyms in $k \in \{0, 2, 4, 6\}$ iterations, considering the rest of
the terms as neutral if they did not have a subjective tag in the
General Inquirer lexicon.  In each of these $k$ steps, the authors
optimized two ternary classifiers (Rocchio and SVM) that used
tf-idf--vectors of synset glosses as features.  Afterwards, they
predicted polarity scores for all \textsc{WordNet} synsets using an
ensemble of all trained classifiers.


Graph-based SLG algorithms were proposed by \citet{Rao:09}, who
experimented with three different methods:
\begin{itemize}
\item\emph{deterministic min-cut}, in which the authors propagated the
  polarity values of seeds to their \textsc{WordNet} synonyms and
  hypernyms and then determined a minimum cut between the polarity
  clusters using the algorithm of~\citet{Blum:01};
\item since this approach, however, always partitioned the graph in
  the same way even if there were multiple possible splits with the
  same cost, the authors also proposed a \emph{randomized} version of
  this method, in which they randomly perturbed edge weights;
\item finally, they compared both min-cut systems with the \emph{label
  propagation algorithm} of~\citet{Zhu:02}, which can be considered as
  a probabilistic variant of \citeauthor{Blair-Goldensohn:08}'s
  approach.
\end{itemize}

Further notable contributions to dictionary-based methods were made
by~\citet{Mohammad:09}, who compiled an initial set of polar terms by
using antonymous morphological patterns (\eg{} \emph{logical} ---
\emph{illogical}, \emph{honest} --- \emph{dishonest}, \emph{happy} ---
\emph{unhappy}) and then expanded this set with the help of the
Macquarie Thesaurus~\cite{Bernard:86}; \citet{Awadallah:10}, who
adopted a random walk approach, estimating word's polarity as a
difference between the average number of steps a random walker had to
make in order to reach a seed term from the positive or negative set;
and \citet{Dragut:10}, who computed words' polarities using manually
specified inference rules.


For our experiments, we reimplemented the approaches of~\citet{Hu:04},
\citet{Blair-Goldensohn:08}, \citet{Kim:04,Kim:06}, \citet{Esuli:06c},
\citet{Rao:09}, and \citet{Awadallah:10}, and applied these methods to
\textsc{GermaNet}\footnote{Throughout our experiments, we will use
  \textsc{GermaNet} Version 9.}~\cite{Hamp:97}, the German equivalent
of the \textsc{WordNet} taxonomy.

In order to make this comparison more fair, we used the same set of
initial seeds for all tested methods.  For this purpose, we translated
the list of 14 polar English adjectives proposed by \citet{Turney:03}
(\emph{good}$^+$, \emph{nice}$^+$, \emph{excellent}$^+$,
\emph{positive}$^+$, \emph{fortunate}$^+$, \emph{correct}$^+$,
\emph{superior}$^+$, \emph{bad}$^-$, \emph{nasty}$^-$,
\emph{poor}$^-$, \emph{negative}$^-$, \emph{unfortunate}$^-$,
\emph{wrong}$^-$, and \emph{inferior}$^-$) into German, getting a
total of 20 terms (10 positive and 10 negative adjectives) due to
multiple possible translations of the same words.  Furthermore, to
settle the differences between binary and ternary approaches (\ie{}
methods that only distinguished between positive and negative terms
and systems that could also predict the neutral class), we extended
the translated seeds with 10 neutral adjectives (\emph{neutral}$^0$,
\emph{objective}$^0$, \emph{technical}$^0$, \emph{chemical}$^0$,
\emph{physical}$^0$, \emph{material}$^0$, \emph{bodily}$^0$,
\emph{financial}$^0$, \emph{theoretical}$^0$, and
\emph{practical}$^0$), letting all classifiers work in the ternary
mode.  Finally, since several algorithms had different takes of
synonymous relations (\eg{} \citeauthor{Hu:04} only considered two
words as synonyms if they appeared in the same synset, whereas
\citeauthor{Esuli:06c}, \citeauthor{Rao:09}, and
\citeauthor{Awadallah:10} also considered hypernyms and hyponyms as
valid links for polarity propagation), we decided to unify this aspect
as well.  To this end, we established an edge between any two terms
that appeared in the same synset, and also linked all words whose
synsets were connected via \texttt{has\_participle},
\texttt{has\_pertainym}, \texttt{has\_hyponym}, \texttt{entails}, or
\texttt{is\_entailed\_by} relations. We intentionally ignored
relations \texttt{has\_hypernym} and \texttt{is\_related\_to}, because
hypernyms were not guaranteed to preserve the polarity of their
children (\eg{} ``bewertungsspezifisch'' [\emph{appraisal-specific}]
is a neutral term in contrast to its immediate hyponyms ``gut''
[\emph{good}] and ``schlecht'' [\emph{bad}]), and
\texttt{is\_related\_to} could connect both synonyms and antonyms of
the same term (\eg{} this relation holds between words ``Form''
[\emph{shape}] and ``unf\"ormig'' [\emph{misshapen}], but at the same
time, it also connects noun ``Dame'' [\emph{lady}] to its derived
adjective ``damenhaft'' [\emph{ladylike}]).

We fine-tuned the hyper-parameters of all approaches by using grid
search and optimizing the macro-averaged \F{}-score on the development
set.  In particular, instead of waiting for the full convergence of
the eigenvector in the approach of \citet{Blair-Goldensohn:08}, we
constrained the maximum number of multiplications to five.  Our
experiments showed that this limitation had a crucial impact on the
quality of the resulting polarity list (\eg{} after five
multiplications, the average precision of its positive terms amounted
to 0.499, reaching an average \F{}-score of 0.26 for this class; after
ten more iterations though, this precision decreased dramatically to
0.043, pulling the \F{}-score down to 0.078).  Furthermore, we limited
the maximum number of iterations in the label-propagation method of
\citet{Rao:09} to 300, although the effect of this setting was much
weaker than in the previous case (by comparison, the scores achieved
after 30 runs differed only by a few hundredths from the results
obtained after 300 iterations).  Finally, in the method of
\citet{Awadallah:10}, we allowed for seven simultaneous walkers with a
maximum number of 17 steps each, considering a word as polar if more
than a half of these walkers agreed on the same polarity class.

\begin{table}[h]
  \begin{center}
    \bgroup\setlength\tabcolsep{0.1\tabcolsep}\scriptsize
    \begin{tabular}{p{0.146\columnwidth} 
        >{\centering\arraybackslash}p{0.06\columnwidth} 
        *{9}{>{\centering\arraybackslash}p{0.072\columnwidth}} 
        *{2}{>{\centering\arraybackslash}p{0.058\columnwidth}}} 
      \toprule
      \multirow{2}*{\bfseries Lexicon} & %
      \multirow{2}{0.06\columnwidth}{\bfseries\centering \# of\newline{} Terms} & %
      \multicolumn{3}{c}{\bfseries Positive Expressions} & %
      \multicolumn{3}{c}{\bfseries Negative Expressions} & %
      \multicolumn{3}{c}{\bfseries Neutral Terms} & %
      \multirow{2}{0.068\columnwidth}{\bfseries\centering Macro\newline \F{}} & %
      \multirow{2}{0.068\columnwidth}{\bfseries\centering Micro\newline \F{}}\\
      \cmidrule(lr){3-5}\cmidrule(lr){6-8}\cmidrule(lr){9-11}

      & & Precision & Recall & \F{} & %
      Precision & Recall & \F{} & %
      Precision & Recall & \F{} & & \\\midrule


      \textsc{Seed Set} & 20 & \textbf{0.771} & 0.102 & 0.18 & %
      0.568 & 0.017 & 0.033 & %
      0.963 & \textbf{0.999} & \textbf{0.981} & %
      0.398 & \textbf{0.962}\\


      HL & 5,745 & 0.161 & 0.266 & 0.2 & %
      0.2 & 0.133 & 0.16 & %
      0.969 & 0.96 & 0.965 & %
      0.442 & 0.93\\


      BG & 1,895 & 0.503 & 0.232 & \textbf{0.318} & %
      0.285 & 0.093 & 0.14 & %
      0.968 & 0.991 & 0.979 & %
      \textbf{0.479} & 0.959\\


      KH & 356 & 0.716 & 0.159 & 0.261 & %
      0.269 & 0.044 & 0.076 & %
      0.965 & 0.997 & \textbf{0.981} & %
      0.439 & \textbf{0.962}\\


      ES & 39,181 & 0.042 & \textbf{0.564} & 0.078 & %
      0.033 & \textbf{0.255} & 0.059 & %
      \textbf{0.981} & 0.689 & 0.81 & %
      0.315 & 0.644\\


      RR$_{\textrm{mincut}}$ & 8,060 & 0.07 & 0.422 & 0.12 & %
      0.216 & 0.073 & 0.109 & %
      0.972 & 0.873 & 0.92 & %
      0.383 & 0.849\\


      RR$_{\textrm{lbl-prop}}$ & 1,105 & 0.567 & 0.176 & 0.269 & %
      \textbf{0.571} & 0.046 & 0.085 & %
      0.965 & 0.997 & \textbf{0.981} & %
      0.445 & \textbf{0.962}\\


      AR & 23 & 0.768 & 0.1 & 0.176 & %
      0.568 & 0.017 & 0.033 & %
      0.963 & \textbf{0.999} & \textbf{0.981} & %
      0.397 & \textbf{0.962}\\


      HL $\cap$ BG $\cap$ RR$_{\textrm{lbl}}$ & 752 & 0.601 & 0.165 & 0.259 & %
      0.567 & 0.045 & 0.084 & %
      0.965 & 0.997 & \textbf{0.981} & %
      0.441 & \textbf{0.962}\\


      HL $\cup$ BG $\cup$ RR$_{\textrm{lbl}}$ & 6,258 & 0.166 & 0.288 & 0.21 & %
      0.191 & 0.146 & \textbf{0.165} & %
      0.97 & 0.958 & 0.964 & %
      0.446 & 0.929\\\bottomrule
    \end{tabular}
    \egroup{}
    \caption[Results of dictionary-based approaches]{Results of
      dictionary-based approaches\\ {\small HL --- \citet{Hu:04}, BG
        --- \citet{Blair-Goldensohn:08}, KH --- \citet{Kim:04}, ES ---
        \citet{Esuli:06c}, RR --- \citet{Rao:09}, AR ---
        \citet{Awadallah:10}}}\label{snt-lex:tbl:lex-res}
  \end{center}
\end{table}

As we can see from the results in Table~\ref{snt-lex:tbl:lex-res}, the
scores of all automatic systems are significantly lower than the
values achieved by semi-automatic lexicons.  The best macro-averaged
\F{}-result for all three classes (0.479) is attained by the method of
\citet{Blair-Goldensohn:08}, which is still 0.11 points below the
highest score obtained by the intersection of GPC, SentiWS, and the
Zurich Polarity List.  Moreover, in general, the situation with
dictionary-based lexicons is more complicated than in the case of
manually curated polarity lists, as every system demonstrates a better
score on only one metric, but fails to convincingly outperform its
competitors on several (let alone all) aspects.  Nevertheless, we
still can notice at least the following main trends:
\begin{itemize}
\item the method of \citet{Esuli:06c} achieves the highest recall
  of positive and negative terms, but these entries have a very low
  precision;

\item simultaneously five approaches attain the same best \F{}-results
  for the neutral class, which, in turn, leads to the best
  micro-averaged \F{}-scores for these systems;

\item and, finally, the solution of \citet{Blair-Goldensohn:08}
  achieves the highest macro-averaged \F{} despite a rather low recall
  of negative expressions.
\end{itemize}

\subsection{Corpus-Based Methods}\label{subsec:snt-lex:corpus-based}

An alternative way of generating polarity lists is provided by
corpus-based approaches.  In contrast to dictionary-based methods,
these systems operate immediately on raw texts and are therefore
virtually independent of any manually annotated resources.

A pioneering work on these algorithms was done by
\citet{Hatzivassi:97}.  Assuming that coordinately conjoined
attributes would typically have the same semantic orientation, these
authors trained a supervised logistic classifier that predicted the
degree of dissimilarity between two co-occurring adjectives.
Afterwards, they constructed a word collocation graph, drawing a link
between any two adjectives that appeared in the same coordinate pair,
and using predicted dissimilarity score between these words as the
respective edge weight.  In the final stage,
\citet{Hatzivassi:97} partitioned this graph into two clusters
and assigned the positive label to the bigger part.

An attempt to unite dictionary- and corpus-based methods was made by
\citet{Takamura:05}, who adopted the Ising spin model from statistical
mechanics, considering words found in \textsc{WordNet}, the Wall
Street Journal, and the Brown corpus as electrons in a ferromagnetic
lattice.  The authors established a link between any two electrons
whose terms appeared in the same \textsc{WordNet} synset or
coordinately conjoined pair in the corpora.  In the final step, they
approximated the most probable orientation of all spins in this graph,
considering these orientations as polarity scores of the respective
terms.

Another way of creating a sentiment lexicon was proposed
by~\citet{Turney:03}, who induced a list of polar terms by computing
the difference between their point-wise mutual information (PMI) with
the positive and negative seeds.  In particular, the authors estimated
the polarity score of word $w$ as:
\begin{equation*}
  \textrm{SO-A}(w) = \sum_{w_p\in\mathcal{P}}PMI(w, w_p) - \sum_{w_n\in\mathcal{N}}PMI(w, w_n),
\end{equation*}
where $\mathcal{P}$ represents the set of all positive seeds;
$\mathcal{N}$ denotes the collection of known negative words; and
$PMI$ is computed as a log-ratio $PMI(w, w_x) = \log_2\frac{p(w,
  w_x)}{p(w)p(w_x)}$.  The joint probability $p(w, w_x)$ in the last
term was calculated as the number of hits returned by the AltaVista
search engine for the query ``$w\textrm{ NEAR }w_x$'' divided by the
total number of documents in the search index.

This method was later successfully adapted to Twitter by
\citet{Kiritchenko:14}, who harnessed the corpus of \citet{Go:09} and
an additional set of 775,000 tweets to create two sentiment lexicons,
Sentiment140 and Hashtag Sentiment Base, using frequent emoticons as
seeds for the first lexicons and taking common emotional hashtags such
as ``\#joy'', ``\#excitement'', ``\#fear'' as seed terms for the
second list.

Another Twitter-specific approach, which also relied on the corpus of
\citet{Go:09}, was presented by \citet{Severyn:15a}.  To derive their
lexicon, the authors trained an SVM classifier that used token n-grams
as features and then included n-grams with the greatest learned
feature weights into their final polarity list.

Graphical methods for corpus-based SLG were advocated by
\citet{Velikovich:10} and \citet{Feng:11}.  The former work adapted
the label-propagation algorithm of \citet{Rao:09} by replacing the
average of all incident scores for a potential subjective term with
their maximum value.  The latter approach induced a sentiment lexicon
using two popular techniques from information retrieval,
PageRank~\cite{Brin:98} and HITS~\cite{Kleinberg:99}.

For our experiments, we reimplemented the approaches
of~\citet{Takamura:05}, \citet{Velikovich:10}, \citet{Kiritchenko:14}
and~\citet{Severyn:15}, and applied these methods to the German
Twitter Snapshot~\cite{Scheffler:14}, a collection of 24~M German
microblogs, which we previously used for sampling one part of our
sentiment corpus.

We normalized all messages of this snapshot with the rule-based
normalization pipeline of~\citet{Sidarenka:13}, which will be
described in more detail in the next chapter, and lemmatized all
tokens with the \textsc{TreeTagger} of~\citet{Schmid:95}.  Afterwards,
we constructed a collocation graph from all normalized lemmas that
appeared at least four times in the snapshot.  For the method of
\citet{Takamura:05}, we additionally used \textsc{GermaNet} in order
to add more links between electrons.  As in the previous experiments,
all hyper-parameters (including the size of the lexicons) were
fine-tuned on the development set by maximizing the macro-averaged
\F{}-score on these data.

The results of this evaluation are presented in
Table~\ref{snt-lex:tbl:corp-meth}.



\begin{table}[h]
  \begin{center}
    \bgroup\setlength\tabcolsep{0.1\tabcolsep}\scriptsize
    \begin{tabular}{p{0.167\columnwidth} 
        >{\centering\arraybackslash}p{0.057\columnwidth} 
        *{9}{>{\centering\arraybackslash}p{0.07\columnwidth}} 
        *{2}{>{\centering\arraybackslash}p{0.055\columnwidth}}} 
      \toprule
      \multirow{2}*{\bfseries Lexicon} & %
      \multirow{2}{0.06\columnwidth}{\bfseries \# of Terms} & %
      \multicolumn{3}{c}{\bfseries Positive Expressions} & %
      \multicolumn{3}{c}{\bfseries Negative Expressions} & %
      \multicolumn{3}{c}{\bfseries Neutral Terms} & %
      \multirow{2}{0.068\columnwidth}{\bfseries\centering Macro\newline \F{}} & %
      \multirow{2}{0.068\columnwidth}{\bfseries\centering Micro\newline \F{}}\\
      \cmidrule(lr){3-5}\cmidrule(lr){6-8}\cmidrule(lr){9-11}

      & & Precision & Recall & \F{} & %
      Precision & Recall & \F{} & %
      Precision & Recall & \F{} & & \\\midrule


      \textsc{Seed Set} & 20 & \textbf{0.771} & 0.102 & 0.18 & %
      \textbf{0.568} & 0.017 & 0.033 & %
      0.963 & \textbf{0.999} & \textbf{0.981} & %
      0.398 & \textbf{0.962}\\


      TKM & 920 & 0.646 & \textbf{0.134} & \textbf{0.221} & %
      0.565 & \textbf{0.029} & \textbf{0.055} & %
      \textbf{0.964} & 0.998 & \textbf{0.981} & %
      \textbf{0.419} & \textbf{0.962}\\


      VEL & 60 & 0.764 & 0.102 & 0.18 & %
      \textbf{0.568} & 0.017 & 0.033 & %
      0.963 & 0.999 & 0.98 & %
      0.398 & \textbf{0.962}\\


      KIR & 320 & 0.386 & 0.106 & 0.166 & %
      \textbf{0.568} & 0.017 & 0.033 & %
      0.963 & 0.996 & 0.979 & %
      0.393 & 0.959\\


      SEV & 60 & 0.68 & 0.102 & 0.177 & %
      \textbf{0.568} & 0.017 & 0.033 & %
      0.963 & \textbf{0.999} & \textbf{0.981} & %
      0.397 & \textbf{0.962}\\

      TKM $\cap$ VEL $\cap$ SEV & 20 & \textbf{0.771} & 0.102 & 0.18 & %
      \textbf{0.568} & 0.017 & 0.033 & %
      0.963 & \textbf{0.999} & \textbf{0.981} & %
      0.398 & \textbf{0.962}\\


      TKM $\cup$ VEL $\cup$ SEV & 1,020 & 0.593 & \textbf{0.134} & 0.218  & %
      0.565 & \textbf{0.029} & \textbf{0.055} & %
      \textbf{0.964} & 0.998 & 0.98 & %
      0.418 & \textbf{0.962}\\\bottomrule
    \end{tabular}
    \egroup{}
    \caption[Results of corpus-based approaches]{Results of
      corpus-based approaches\\ {\small TKM --- \citet{Takamura:05},
        VEL --- \citet{Velikovich:10}, KIR --- \citet{Kiritchenko:14},
        SEV --- \citet{Severyn:15}}}\label{snt-lex:tbl:corp-meth}
  \end{center}
\end{table}

This time, we can observe a clear superiority of the system
of~\citet{Takamura:05}, which not only achieves the best recall and
\F{} for the positive and negative classes but also yields the highest
micro- and macro-averaged results for all three polarities.
The sizes and the scores of other lexicons, however, are much smaller
than the cardinalities and the results of the
\citeauthor{Takamura:05}'s polarity list.  Moreover, these lexicons
can hardly outperform the original seed set on the negative class.

Because the last result was somewhat unexpected, we decided to
investigate the reasons for potential problems in these systems.  A
closer look at their learning curves revealed that the macro-averaged
\F{}-values on the development data rapidly decreased from the very
beginning of their work.  Since we considered the lexicon size as one
of the parameters, we rapidly stopped populating these lists.  As a
consequence, only few highest ranked terms (all of which were
positive) were included into the final resource.  As it turned out the
main reason for this degradation was the ambiguity of the seed terms:
While adapting the original seed list of~\citet{Turney:03} to German,
we translated the English word ``correct'' as ``richtig.''  This
German word, however, also has another reading---\emph{real} (as in
``ein richtiges Spiel'' [\emph{a real game}] or ``ein richtiger
Rennwagen'' [\emph{a real sports car}]), which was much more frequent
in the analyzed snapshot and typically appeared in a negative context,
\eg{} ``ein richtiger Bombenanschlag'' (\emph{a real bomb attack}) or
``ein richtiger Terrorist'' (\emph{a real terrorist}).  As a
consequence, methods that relied on weak supervision had to deal with
extremely unbalanced training data (716,210 positive instances versus
92,592 negative ones) and got stuck in a local optimum from the very
beginning of their training.

\subsection{NWE-Based Methods}\label{subsec:snt:lex:nwe}

Finally, the last group of methods that we are going to explore in
this chapter are algorithms that operate on distributed vector
representations of words (neural word embeddings [NWEs]).  First
introduced by~\citet{Bengio:03} and significantly improved by
\citet{Collobert:11} and \citet{Mikolov:13}, NWEs had a great
``tsunami''-like effect on many downstream NLP
applications~\cite{Manning:15}.  Unfortunately, these advances have
largely bypassed the generation of sentiment lexicons, up to a few
exceptions introduced by the works of~\citet{Tang:14a} and
\citet{Vo:16}.  In the former approach, the authors used a large
collection of weakly labeled tweets in order to learn hybrid word
embeddings.  In contrast to standard word2vec
vectors~\cite{Mikolov:13} and purely task-specific
representations~\cite{Collobert:11}, such embeddings were optimized
with respect to both objectives---predicting the occurrence of nearby
words and classifying the overall polarity of a message.  Using these
hybrid vectors, \citeauthor{Tang:14a} trained a one-layer feed-forward
neural network that predicted the polarity of a microblog, and
subsequently applied this classifier separately to each word
embedding, considering the predicted value as polarity score for the
respective term.  In contrast to this approach, \citeauthor{Vo:16}
immediately optimized two-dimensional task-specific embeddings, and
regarded the two dimensions of these learned vectors as positive and
negative scores of corresponding words.

In order to evaluate these systems, we reimplemented both methods,
extending them to three-way classification (positive, negative, and
neutral), and applied them to weakly labeled snapshot tweets.

Apart from these solutions, we also came up with the following
alternative ways of generating polarity lists from neural word
embeddings:
\begin{itemize}
\item the method of the nearest centroids,
\item $k$-NN clustering;
\item principal component analysis (PCA),
\item and a new linear-projection algorithm.
\end{itemize}

\begin{algorithm}
  \begin{algorithmic}[1]
    \Function{ExpandPCA}{$\mathcal{P}, \mathcal{N}, \mathcal{O}, \vars{E}$}\Comment{$\mathcal{P}$~--~indices of positive terms,}
    \Statex\Comment{$\mathcal{N}$~--~indices of negative terms,}
    \Statex\Comment{$\mathcal{O}$~--~indices of objective terms, $\vars{E}$~--~embedding matrix}
    \State $\vars{U}, \Sigma, \vars{V}^\top\gets\func{SVD}(\vars{E})$;\Comment{obtain singular components of \vars{E}}
    \State $\vars{E}'\gets\left(\vars{E}^\top\cdot \vars{U}\right)^\top$;\Comment{project \vars{E} onto the eigenvectors of its row space}
    \State $\mathcal{S}\gets\mathcal{P}\cup\mathcal{N}$;\Comment{get the set of subjective terms}
    \State $\vars{u}_{subj}, \mu_{\mathcal{S}}, \mu_{\mathcal{O}}\gets\func{FindMeanAxis}(\vars{E}', \mathcal{S}, \mathcal{O})$;\Comment{find
      subjectivity axis}
    \State $\vars{u}_{pol}, \mu_{\mathcal{P}}, \mu_{\mathcal{N}}\gets\func{FindAxis}(\vars{E}', \mathcal{P}, \mathcal{N})$;\Comment{find polarity axis}
    \State\Return $\func{ComputePolScores}(\vars{E}', \mathcal{S}\cup\mathcal{O},%
    \vars{u}_{subj}, \vars{u}_{pol}, \mu_{\mathcal{S}}, \mu_{\mathcal{O}},%
    \mu_{\mathcal{P}}, \mu_{\mathcal{N}})$;
    \EndFunction
    \Statex
    \Function{FindMeanAxis}{$\vars{E}, \mathcal{S}_1, \mathcal{S}_2$}
    \State $\mu_1\gets 0$; $\mu_2\gets 0$; \vars{axis} $\gets 0$; \vars{max\_dist} $\gets 0$;
    \For{$\vars{i}\gets 1$; $\vars{i} <= \func{nrows}(\vars{E})$; $\vars{i}\gets\vars{i} + 1$}
    \State \vars{e} $\gets$\vars{M[i]};
    \State \vars{dist} $\gets\sum_{j_1\in\mathcal{S}_1,j_2\in\mathcal{S}_2}\left|\vars{e}[j_1] - \vars{e}[j_2]\right|$;
    \If {\vars{dist} $>$ \vars{max\_dist}}
    \State \vars{axis} $\gets$\vars{i}; \vars{max\_dist} $\gets$ \vars{dist};
    \State $\mu_1\gets\frac{\sum_{j_1\in\mathcal{S}_1}\vars{e}[j_1]}{|\mathcal{S}_1|}$; $\mu_2\gets\frac{\sum_{j_2\in\mathcal{S}_2}\vars{e}[j_2]}{|\mathcal{S}_2|}$;
    \EndIf
    \EndFor
    \State\Return\vars{axis}, $\mu_1$, $\mu_2$;
    \EndFunction
    \Statex
    \Function{ComputePolScores}{$\vars{E}, \mathcal{S}, \vars{u}_{subj}, \vars{u}_{pol},
      \mu_{\mathcal{S}}, \mu_{\mathcal{O}}, \mu_{\mathcal{P}}, \mu_{\mathcal{N}}$}
    \State\vars{scores} $\gets []$;

    \State\vars{O}$_{subj}\gets\mu_{\mathcal{O}} + \frac{\mu_{\mathcal{S}} -
    \mu_{\mathcal{O}}}{2}$;\Comment{Compute the origin of the
      subjectivity axis.}

    \State\vars{max\_score}$_{subj}\gets\max\left(\{|\vars{n}\lbrack\vars{u}_{subj}\rbrack%
      - \vars{O}_{subj}| |\forall\vars{n}\in\func{cols}(\vars{E})\}\right)$;

    \State\vars{O}$_{pol}\gets\mu_{\mathcal{N}} + \frac{\mu_{\mathcal{P}} -
    \mu_{\mathcal{N}}}{2}$;\Comment{Compute the origin of the polarity
      axis.}

    \State\vars{max\_score}$_{pol}\gets\max\left(\{|\vars{n}\lbrack\vars{u}_{pol}\rbrack%
      - \vars{O}_{pol}| |\forall\vars{n}\in\func{cols}(\vars{M})\}\right)$;

    \For{$\vars{i}\gets 1$; $\vars{i} <= \func{ncols}(\vars{M})$; $\vars{i}\gets\vars{i} + 1$}
    \If {$\vars{i} \in \mathcal{S}$}
    \State \textbf{continue};\Comment{known seeds will be added later by default}
    \EndIf
    \State \vars{n}$\gets$\vars{M[:, i]}$^\top$;\Comment{assign $i$-th column to \vars{n}}

    \If {$|\vars{n}\lbrack\vars{u}_{subj}\rbrack - \mu_{\mathcal{O}}| > %
      |\vars{n}\lbrack\vars{u}_{subj}\rbrack - \mu_{\mathcal{S}}|$}%
    \Comment{if the $i$-th word is subjective}

    \State\vars{score}$_{subj}\gets 1 + %
    \frac{|\vars{n}\lbrack\vars{u}_{subj}\rbrack - \vars{O}_{subj}|}{\vars{max\_score}_{subj}}$;
    \algstore{sentilex-pca}
  \end{algorithmic}
  \caption[Sentiment lexicon generation using PCA]{Sentiment lexicon
    generation with the PCA algorithm}\label{snt:lex:alg:pca}
\end{algorithm}

\begin{algorithm}
  \begin{algorithmic}[1]
    \algrestore{sentilex-pca}
    \Else
    \State\vars{score}$_{subj}\gets 1 - %
    \frac{|\vars{n}\lbrack\vars{u}_{subj}\rbrack - \vars{O}_{subj}|}{\vars{max\_score}_{subj}}$;
    \EndIf

    \If {$|\vars{n}\lbrack\vars{u}_{pol}\rbrack - \mu_{\mathcal{N}}| > %
      |\vars{n}\lbrack\vars{u}_{pol}\rbrack - \mu_{\mathcal{P}}|$}%
    \Comment{if the $i$-th word is positive}

    \State\vars{polarity}$\gets$ positive;
    \Else
    \State \vars{polarity}$\gets$ negative;
    \EndIf
    \State\vars{score}$_{pol}\gets%
    \frac{|\vars{n}\lbrack\vars{u}_{pol}\rbrack %
      - \vars{O}_{pol}|}{\vars{max\_score}_{pol}}$;
    \State\func{append}(\vars{scores},
    (\vars{i}, \vars{polarity}, %
    $\frac{1}{\vars{score}_{\textrm{subj}} + \vars{score}_{pol}}$));\Comment{The total score is inversed}
    \Statex\Comment{as we sort the resulting polarity list in the ascending order.}
    \EndFor
    \State\Return\vars{scores};
    \EndFunction
  \end{algorithmic}
\end{algorithm}

In the first method, we computed the centroids of the positive,
negative, and neutral clusters by taking the arithmetic mean of the
respective seed-term vectors and then assigned word $w$ to the
polarity group whose centroid was closest to the embedding of that
word.  We considered the distance to the cluster center as the
respective polarity score for that word and sorted the resulting
sentiment lexicon in ascending order of these values.

A similar procedure was used in $k$-NN, where we first determined $k$
seed vectors that were closest to the embedding of word $w$ and then
allocated this word to the polarity class whose seeds were nearest and
appeared most frequently in $w$'s neighborhood.

A different technique was used for the principal component analysis.
After normally decomposing the embedding
matrix~$E\in\mathbb{R}^{d\times|V|}$ (with $d = 300$ denoting the
dimension of word vectors, and $|V|$ representing the vocabulary size)
into singular components:
\begin{align*}
  E = U \Sigma V^T,
\end{align*}
we looked for a row axis $u_{\textrm{subj}}\in U$ that maximized the
distance between the embeddings of polar (positive or negative) and
neutral seeds projected on that line.  In the same way, we determined
a polarity axis $u_{\textrm{pol}}$ that maximized the distance between
projected positive and negative embeddings.  After finding both axes,
we projected all word vectors on these two lines, considering the
distances between these projections and lines' origins as the
respective subjectivity and polarity scores.  The pseudo-code of this
approach is shown in Algorithm~\ref{snt:lex:alg:pca}.

Since PCA, however, was not guaranteed to find the optimum projection
axes (the orthogonal bases $U$ and $V$ in SVD are typically computed
using the covariance matrix of $E$, and might therefore not reflect
semantic orientations of words, if terms with opposite polarities
occur in similar contexts), we devised our own \emph{linear projection
  method}, in which we explicitly encoded the above objective: Namely,
given two sets of vectors with opposite semantic orientations (let us
denote the set of positive vectors as~$\mathcal{P} =
\{\vec{p}_{+_1},\ldots,\vec{p}_{+_m}\}$ and the set of negative
embeddings as~$\mathcal{N} = \{\vec{p}_{-_1},\ldots,\vec{p}_{-_n}\}$),
we were looking for a line $\vec{b}$ that maximized the distance
between the projections of embeddings from these sets on that line, i.e.: {\small%
  \begin{align}
    \begin{split}
      \small
      \vec{b} &=\argmax\frac{1}{2}\sum_{\vec{p}_+}\sum_{\vec{p}_-}%
      \left\lVert\frac{\vec{b}\cdot\vec{p}_+}{\vec{b}^2}\vec{b}%
      - \frac{\vec{b}\cdot\vec{p}_-}{\vec{b}^2}\vec{b}\right\rVert^2\\
      &=\argmax\frac{1}{2}\sum_{\vec{p}_+}\sum_{\vec{p}_-}\left\lVert\frac{\vec{b}%
        \cdot\left(\vec{p}_{+}-\vec{p}_{-}\right)}{\vec{b}^2}\vec{b}\right\rVert^2,\label{eq:f}%
    \end{split}
  \end{align}\normalsize}%
where $\frac{\vec{b}\cdot\vec{p}_+}{\vec{b}^2}\vec{b}$ is the
projection of a word embedding with the positive polarity on line
$\vec{b}$, and $\frac{\vec{b}\cdot\vec{p}_-}{\vec{b}^2}\vec{b}$ is the
respective projection of a negative seed term.  Considering the
$\argmax$ argument in Expression~\ref{eq:f} as our objective
function~$f$, we computed the gradient of~$f$ with respect
to~$\vec{b}$ as follows: {\small%
  \begin{align}
    \nabla_{\vec{b}} f &= \sum_{\vec{p}_+}\sum_{\vec{p}_-}%
               \gamma\left(\Delta - \gamma\vec{b}\right),\label{eq:prj-line-grad}%
\end{align}\normalsize}%
where $\Delta$ stands for the difference between the positive and
negative vectors~$\vec{p}_{+}$ and~$\vec{p}_{-}$: $\Delta \defeq
\vec{p}_{+}-\vec{p}_{-}$; and $\gamma$ denotes the dot product of this
difference with vector~$\vec{b}$: $\gamma \defeq \Delta \cdot
\vec{b}$.\footnote{The details of this gradient computation are given
  in Appendix~\ref{chap:apdx:lex-grad}.}  With this gradient, we then
optimized $\vec{b}$ using gradient ascent until we reached a maximum
of function~$f$.\footnote{Since function $f$ is neither convex nor
  concave, we can only speak about \emph{a} maximum.  We hypothesize,
  however, based on our experiments and preliminary calculations, that
  this local maximum will simultaneously be the global one because the
  optimized projection vector will have two possible solutions, which
  will lie on the same line, but point to the opposite directions.}

\begin{figure*}[hbt!]
  \centering
  \begin{subfigure}{.45\textwidth}
    \centering
    \mbox{\includegraphics[height=15em]{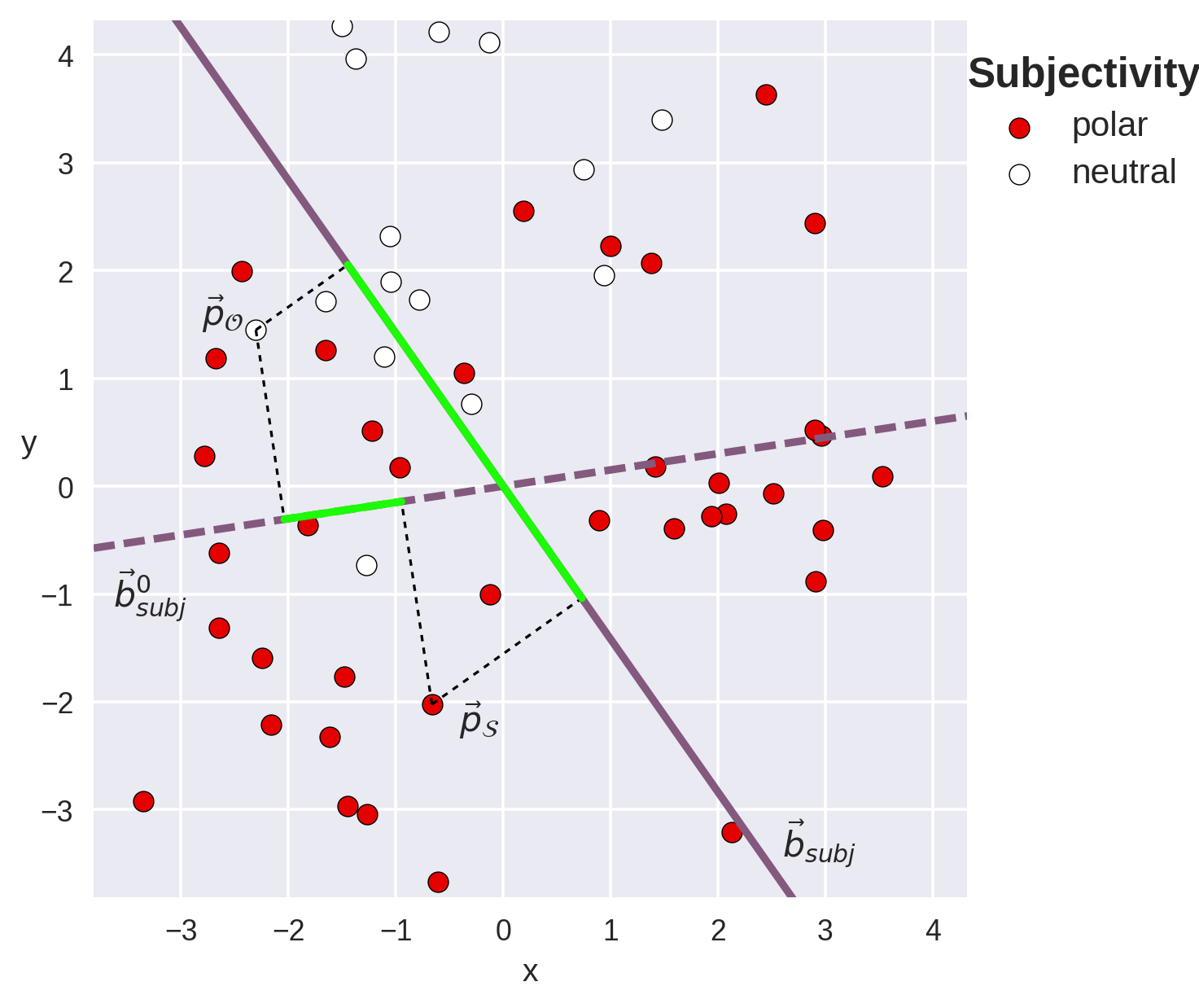}}
    \caption{Subjectivity line}
  \end{subfigure}
  \begin{subfigure}{.45\textwidth}
    \centering
    \mbox{\includegraphics[height=15em]{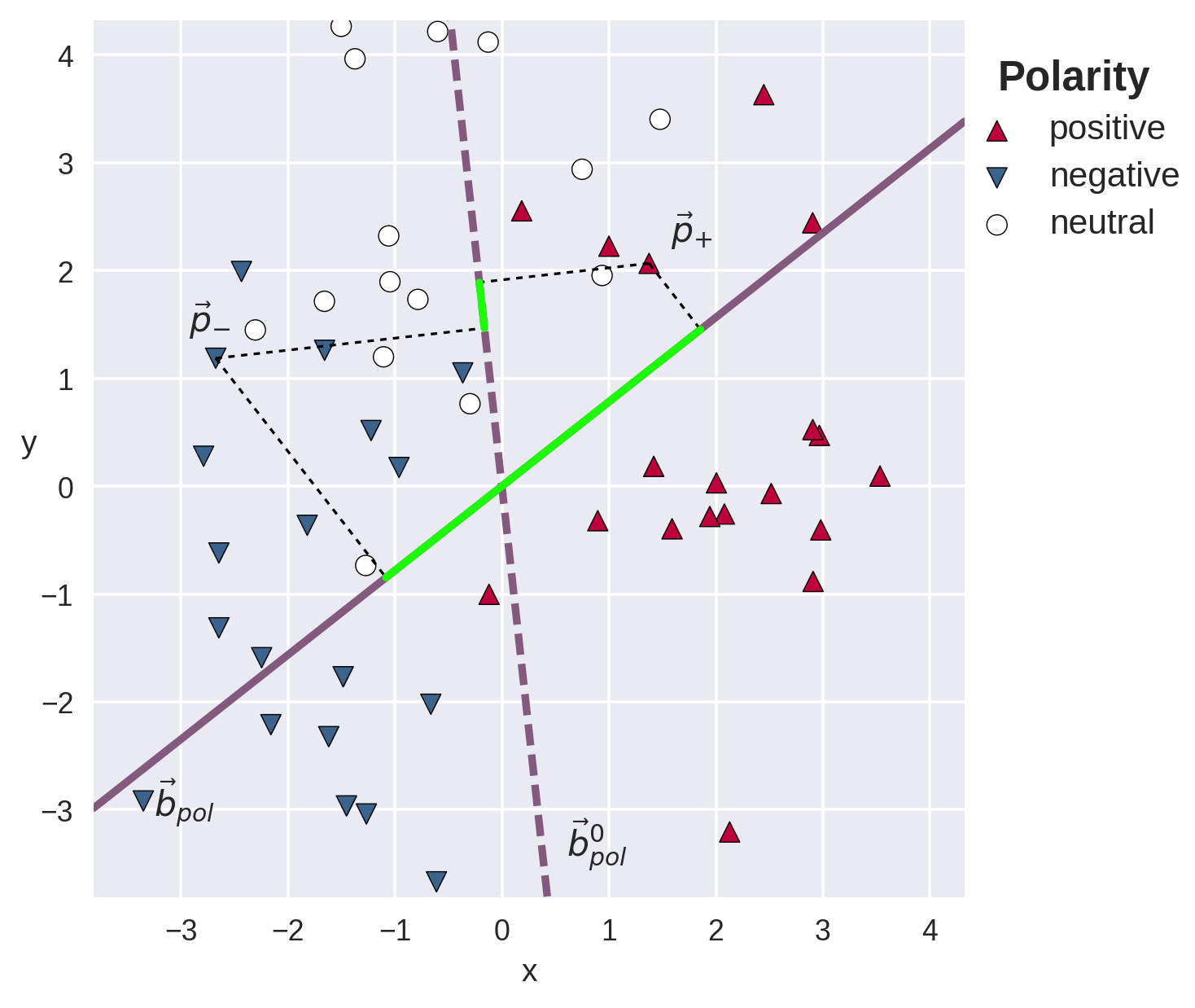}}
    \caption{Polarity line}
  \end{subfigure}
  \caption[Linear projection method in the
  two-dimensional vector space]{Visualization of the linear
    projection method in the
    two-dimensional vector space with unnormalized vectors\\
    (\small $\vec{b}^0_{subj}$ and $\vec{b}^0_{pol}$~--~initial
    guesses of the subjectivity and polarity lines; $\vec{b}_{subj}$
    and $\vec{b}_{pol}$~--~optimal projection vectors; the distances
    between the projections of sample seeds with opposite semantic
    orientations ($\vec{p}_{\mathcal{S}}$ vs. $\vec{p}_{\mathcal{O}}$
    and $\vec{p}_+$ vs. $\vec{p}_\_$ respectively) on these lines are
    highlighted in \textcolor{highlighter green}{green})}\label{fig:linproj}
\end{figure*}



As in PCA, we first used this approach to determine an optimal
subjectivity axis $\vec{b}_{subj}$, which maximized the distance
between the sets of polar and neutral embeddings.  After finding this
axis and projecting on it all remaining word vectors, we classified
all words into polar and neutral ones, depending on whether their
projections appeared closer to the mean of the former or latter set.
We then computed the subjectivity score for polar terms
as~$s^{pol}_{subj} = 1 +
\frac{\delta_{subj}^i}{\delta_{subj}^{\max}}$, where $\delta_{subj}^i$
stands for the distance between the projection of the $i$-th term and
the origin $O_{subj}$, and $\delta_{subj}^{\max}$ means the maximum
such distance observed in the data. Similarly, we estimated these
values for neutral items as $s^{neut}_{subj} = 1 -
\frac{\delta_{subj}^i}{\delta_{subj}^{\max}}$.

In the same way, we estimated the polarity score for the $i$-th word
by first finding a polarity line~$\vec{b}_{pol}$ and then computing
the respective score as~$s_{pol} = 1 +
\frac{\delta^i_{pol}}{{\delta}^{max}_{pol}}$.

In the final step, we united both values $s_{subj}$ and $s_{pol}$ into
a single score $s = \frac{1}{s_{subj} + s_{pol}}$ and sorted the
resulting polarity list in ascending order of these unified scores.
The pseudo-code of our approach is given in
Algorithm~\ref{snt:lex:alg:linproj}, and a visualization of the line
optimization step (in the two-dimensional vector space) is shown in
Figure~\ref{fig:linproj}.

\begin{algorithm}
  \begin{algorithmic}[1]
    \Function{ExpandLinProj}{$\mathcal{P}, \mathcal{N}, \mathcal{O}, \vars{E}$}\Comment{$\mathcal{P}$~--~indices of positive terms,}
    \Statex\Comment{$\mathcal{N}$~--~indices of negative terms,}
    \Statex\Comment{$\mathcal{O}$~--~indices of objective terms, $\vars{E}$~--~embedding matrix}
    \State $\mathcal{S}\gets\mathcal{P}\cup\mathcal{N}$;\Comment{get the set of subjective terms}
    \State $\vars{proj}_{subj}, \mu_{\mathcal{S}}, \mu_{\mathcal{O}}\gets%
    \func{Project}(\mathcal{S}, \mathcal{O}, \vars{E})$;
    \State $\vars{proj}_{pol}, \mu_{\mathcal{P}}, \mu_{\mathcal{N}}%
    \gets\func{Project}(\mathcal{P}, \mathcal{N}, \vars{E})$;
    \State\Return $\func{ComputePolScores}(\vars{E}, \mathcal{S}\cup\mathcal{O},%
    \vars{proj}_{subj}, \vars{proj}_{pol}, %
    \mu_{\mathcal{S}}, \mu_{\mathcal{O}}, \mu_{\mathcal{P}}, \mu_{\mathcal{N}})$;
    \EndFunction
    \Statex
    \Function{Project}{$\mathcal{S}_1, \mathcal{S}_2, \vars{E}$}
    \State $\vars{projs}\gets []$, $\mu_1\gets \vec{0}$; $\mu_2\gets \vec{0}$;
    \State $\vec{b}\gets\func{FindAxis}(\mathcal{S}_1, \mathcal{S}_2)$;\Comment{Determine the optimum polarity/subjectivity axis}
    \For{$\vars{i}\gets 1$; $\vars{i} <= \func{nrows}(\vars{E})$; $\vars{i}\gets\vars{i} + 1$}
    \State $\vec{p}\gets\vec{b}*\vars{E}$[$i$]$\cdot\vec{b}$;\Comment{Project $i$-th embedding onto the axis}
    \State $\func{append}(\vars{projs}, \vec{p})$;
    \If{$\vars{j}\in\mathcal{S}_1$}\Comment{Update means of the polarity/subjectivity classes}
    \State $\mu_1\gets\mu_1 + \vec{p}$;
    \Else
    \If{$\vars{k}\in\mathcal{S}_2$}
    \State $\mu_2\gets\mu_2 + \vec{p}$;
    \EndIf
    \EndIf
    \EndFor
    \State\Return $\vars{projs}$, $\frac{\mu_1}{|\mathcal{S}_1|}$; $\frac{\mu_2}{|\mathcal{S}_2|}$;
    \EndFunction
    \Statex
    \Function{FindAxis}{$\mathcal{S}_1, \mathcal{S}_2, \vars{E}$}
    \State $\vec{b}\gets\vec{1}$; $\vars{prev\_dist}\gets \infty$;
    \For{$\vars{i}\gets 1$; $\vars{i} \leqslant \vars{MAX\_ITERS}$; $\vars{i}\gets\vars{i} + 1$}
    \State $\vec{b}\gets\frac{\vec{b}}{\norm{\vec{b}}}$;\Comment{Ensure the projection line
      has a unit length}
    \State$\vars{dist}\gets 0$;
    \For{$\vars{j}\in\mathcal{S}_1$}
    \State $\vec{p}_1\gets\vec{b}*\vars{E}$[$j$]$\cdot\vec{b}$;\Comment{Project a seed term from the first set}
    \For{$\vars{k}\in\mathcal{S}_2$}
    \State $\vec{p}_2\gets\vec{b}*\vars{E}$[$k$]$\cdot\vec{b}$;\Comment{Project a seed term from the second set}
    \State$\vars{dist}\gets \vars{dist} + \norm{\vec{p}_1 - \vec{p}_2}$;\Comment{Accumulate the distance}
    \Statex\Comment{between the two projections}
    \EndFor
    \EndFor
    \algstore{sentilex-linproj-1}
  \end{algorithmic}
  \caption[Sentiment lexicon generation using linear
  projection]{Sentiment lexicon generation with the linear projection
    algorithm}\label{snt:lex:alg:linproj}
\end{algorithm}

\begin{algorithm}
  \begin{algorithmic}[1]
    \algrestore{sentilex-linproj-1}
    \If{$\vars{prev\_dist}\neq\infty$ \textbf{and} $\vars{dist} - \vars{prev\_dist} < \epsilon$}
    \State\textbf{break};\Comment{Break if the convergence criterion was reached}
    \EndIf
    \State $\vars{prev\_dist}\gets\vars{dist}$;
    \State $\vec{b}\gets\vec{b} + \alpha*\nabla\vec{b}$;\Comment{Optimize the projection line}
    \EndFor
    \State\Return $\vec{b}$;
    \EndFunction
    \Statex
    \Function{ComputePolScores}{$\vars{E}, \mathcal{S}, \vars{projections}_{subj}, %
      \vars{projections}_{pol}, \mu_{\mathcal{S}}, \mu_{\mathcal{O}}, %
      \mu_{\mathcal{P}}, \mu_{\mathcal{N}}$}
    \State\vars{scores} $\gets []$;

    \State\vars{O}$_{subj}\gets\mu_{\mathcal{O}} + \frac{\mu_{\mathcal{S}} -
    \mu_{\mathcal{O}}}{2}$;\Comment{Compute the origin of the
      subjectivity axis.}

    \State\vars{max\_score}$_{subj}\gets\max\left(\{|\vars{p}_{subj}%
      - \vars{O}_{subj}| |\forall\vars{p}_{subj}\in\vars{projections}_{subj}\}\right)$;

    \State\vars{O}$_{pol}\gets\mu_{\mathcal{N}} + \frac{\mu_{\mathcal{P}} -
    \mu_{\mathcal{N}}}{2}$;\Comment{Compute the origin of the polarity
      axis.}

    \State\vars{max\_score}$_{pol}\gets\max\left(\{|\vars{p}_{pol}%
      - \vars{O}_{pol}| |\forall\vars{p}_{pol}\in\vars{projections}_{pol}\}\right)$;

    \For{$\vars{i}\gets 1$; $\vars{i} <= \func{ncols}(\vars{M})$; $\vars{i}\gets\vars{i} + 1$}
    \If{$\vars{i} \in \mathcal{S}$}
    \State\textbf{continue};\Comment{known seeds will be added later by default}
    \EndIf{}
    \State{}$\vars{p}_{subj}$$\gets$\vars{projections}$_{subj}$[$i$];

    \If {$|\vars{p}_{subj}-\mu_{\mathcal{O}}| > |\vars{p}_{subj}-\mu_{\mathcal{S}}|$}%
    \Comment{if the $i$-th word is subjective}

    \State\vars{score}$_{subj}\gets{}1 + %
    \frac{|\vars{p}_{subj}-\vars{O}_{subj}|}{\vars{max\_score}_{subj}}$;
    \Else
    \State\vars{score}$_{subj}\gets 1 - \frac{|\vars{p}_{subj}-\vars{O}_{subj}|}{\vars{max\_score}_{subj}}$;
    \EndIf

    \State $\vars{p}_{pol}\gets\vars{projections}_{pol}$[$i$];
    \If {$|\vars{p}_{pol} - \mu_{\mathcal{N}}| > |\vars{p}_{pol} - \mu_{\mathcal{P}}|$}%
    \Comment{if the $i$-th word is positive}

    \State\vars{polarity}$\gets$ positive;
    \Else
    \State \vars{polarity}$\gets$ negative;
    \EndIf
    \State\vars{score}$_{pol}\gets%
    1 + \frac{|\vars{p}_{pol} - \vars{O}_{pol}|}{\vars{max\_score}_{pol}}$;
    \State\func{append}(\vars{scores},
    (\vars{i}, \vars{polarity}, %
    $\frac{1}{\vars{score}_{\textrm{subj}} + \vars{score}_{pol}}$));
    \EndFor
    \State\Return\vars{scores};
    \EndFunction
  \end{algorithmic}
\end{algorithm}

We applied all methods to word2vec embeddings, which had been
previously learned on the snapshot data, normalizing the length and
scaling the means of these vectors before passing them to our
algorithms.  The results of all systems are shown in
Table~\ref{snt-lex:tbl:nwe-meth}.

As we can see from the table, linear projection not only outperforms
all other NWE-based systems in terms of the micro-averaged \F-score
but also surpasses the results of dictionary-, corpus-based, and
semi-automatic lexicons, being only 0.1~percent below the overall best
\F-value achieved by the intersection of SentiWS, German Polarity
Clues, and Zurich Polarity List.  Our method also achieves the
second-best macro-averaged \F-result, being outmatched by $k$-NN.  The
third-best micro-averaged \F{} is attained by the approach
of~\citet{Tang:14a}, which, however, suffers from low precision of its
positive entries.  The results of the remaining systems are,
unfortunately, even lower and hardly improve on the scores of the
initial seed set.

\begin{table}[h]
  \begin{center}
    \bgroup \setlength\tabcolsep{0.1\tabcolsep}\scriptsize
    \begin{tabular}{p{0.142\columnwidth} 
        >{\centering\arraybackslash}p{0.06\columnwidth} 
        *{9}{>{\centering\arraybackslash}p{0.072\columnwidth}} 
        *{2}{>{\centering\arraybackslash}p{0.059\columnwidth}}} 
      \toprule
      \multirow{2}*{\bfseries Lexicon} & %
      \multirow{2}{0.06\columnwidth}{\bfseries \# of Terms} & %
      \multicolumn{3}{c}{\bfseries Positive Expressions} & %
      \multicolumn{3}{c}{\bfseries Negative Expressions} & %
      \multicolumn{3}{c}{\bfseries Neutral Terms} & %
      \multirow{2}{0.068\columnwidth}{\bfseries\centering Macro\newline \F{}} & %
      \multirow{2}{0.068\columnwidth}{\bfseries\centering Micro\newline \F{}}\\
      \cmidrule(lr){3-5}\cmidrule(lr){6-8}\cmidrule(lr){9-11}

      & & Precision & Recall & \F{} & %
      Precision & Recall & \F{} & %
      Precision & Recall & \F{} & & \\\midrule


      \textsc{Seed Set} & 20 & \textbf{0.771} & 0.102 & 0.18 & %
      0.568 & 0.017 & 0.033 & %
      0.963 & \textbf{0.999} & 0.981 & %
      0.398 & 0.962\\


      TNG & 1,600 & 0.088 & 0.153 & 0.112 & %
      0.193 & \textbf{0.155} & \textbf{0.172} & %
      \textbf{0.966} & 0.953 & 0.959 & %
      0.414 & 0.921\\


      VO & 40 & 0.117 & 0.115 & 0.116 & %
       0.541 & 0.017 & 0.033 & %
       0.963 & 0.98 & 0.971 & %
       0.374 & 0.944\\



      NC & 5,200 & \textbf{0.771} & 0.102 & 0.18 & %
       0.568 & 0.017 & 0.033 & %
       0.963 & \textbf{0.999} & 0.981 & %
       0.398 & 0.962\\

      $k$-NN & 420 & 0.486 & \textbf{0.182} & \textbf{0.265} & %
        \textbf{0.65} & 0.091 & 0.16 & %
        \textbf{0.966} & 0.995 & 0.98 & %
        \textbf{0.468} & 0.961\\


      PCA & 40 & 0.771 & 0.102 & 0.18 & %
         0.529 & 0.017 & 0.033 & %
         0.963 & \textbf{0.999} & 0.981 & %
         0.398 & 0.962\\


      LP & 6,340 & 0.741 & 0.156 & 0.257 & %
       0.436 & 0.088 & 0.147 & %
       \textbf{0.966} & 0.998 & \textbf{0.982} & %
       0.462 & \textbf{0.963}\\\bottomrule
    \end{tabular}
    \egroup
    \caption[Results of NWE-based approaches]{Results of
      NWE-based approaches\\ {\small TNG~--~\citet{Tang:14a}, %
        VO~--~\citet{Vo:16}, %
        NC~--~nearest centroids, %
        $k$-NN~--~$k$-nearest neighbors, %
        PCA~--~principal component analysis, %
        LP~--~linear projection}}%
    \label{snt-lex:tbl:nwe-meth}
  \end{center}
\end{table}

\subsubsection{Word Embeddings}\label{subsec:snt-lex:eowet}

An important aspect that could significantly affect the results of
NWE-algorithms was the type of word embeddings that we provided to
these systems as input.  As we already noted at the beginning of this
section, two most common kinds of such representations are standard
word2vec and task-specific vectors~\cite{Mikolov:13,Collobert:11}.
The former type seeks to find a word representation that maximizes the
probability of other tokens appearing in the nearby context, whereas
the latter type optimizes these representations with respect to a
specific custom task, such as polarity prediction of the whole text.

In order to see how these differences could affect the results of our
approaches, we have trained task-specific embeddings on snapshot
tweets, considering positive and negative emoticons that appeared in
these messages as their noisy polarity labels, and re-evaluated our
methods using these vectors.  Apart from that, we additionally
explored two in-between solutions:
\begin{itemize}
\item \emph{hybrid embeddings}, which were trained by simultaneously
  optimizing two objectives---predicting the surrounding context and
  classifying the polarity of the tweet;
\item and \emph{least-squares embeddings}, for which we first obtained
  both embedding types, word2vec ($V_{W2V}$) and task-specific ones
  ($V_{TS}$).  Since task-specific vectors, however, could only be
  learned on messages that contained emoticons or neutral seeds, many
  terms that had a word2vec representation did not have a
  task-specific counterpart.  To derive these missing embeddings, we
  computed a transformation matrix $W$ using the method of the
  ordinary least squares:
  \begin{equation}\label{eq:fgsa:least-sq}
    W = \argmin_{W}\lVert V_{TS} - W^T\cdot V^*_{W2V}\rVert_F,
  \end{equation}
  where $V^*_{W2V}$ represents a matrix of word2vec vectors whose
  words have both representations, and
  $\left\lVert\cdot\right\rVert_F$ means the Frobenius norm.
  Afterwards, we approximated task-specific representations for all
  terms that were missing in $V_{TS}$ by multiplying their word2vec
  vectors with matrix $W$.
\end{itemize}

As we can see from the results in Table~\ref{snt-lex:tbl:emb-eff},
$k$-NN and linear projection work best with the standard word2vec
embeddings (with the overall best score [0.468] achieved by $k$-NN),
but their performance degrades as the input vectors become more and
more aware of the polarity-prediction task.  An opposite situation is
observed for the nearest centroids and PCA, which show an improvement
in combination with task-specific and least-squares vectors.

\begin{table}[thb!]
  \begin{center}
    \bgroup\setlength\tabcolsep{0.1\tabcolsep}%
    \setlength{\belowrulesep}{0pt}\scriptsize
    \begin{tabular}{p{0.145\columnwidth} 
        *{4}{>{\centering\arraybackslash}p{0.21\columnwidth}}} 
      \toprule
      \multirow{2}*{\bfseries Lexicon} %
      & \multicolumn{4}{c}{\bfseries Embedding Type}\\
      & word2vec & task-specific + word2vec & task-specific + least squares %
                                            & task-specific\\\midrule
      NC & 0.398 & 0.398 & \textbf{0.401} & 0.399\\
      $k$-NN & \textbf{0.468} & 0.43 & 0.398 & 0.392\\
      PCA & 0.398 & 0.398 & 0.404 & \textbf{0.409}\\
      LP & \textbf{0.462} & 0.441 & 0.398 & 0.399\\\bottomrule

    \end{tabular}\egroup%
    {
      \captionsetup{justification=centering}
      \caption[Macro-averaged \F-scores of NWE-based methods with
      different embedding types]{Macro-averaged \F-scores of
        NWE-based methods with different embedding types%
      }\label{snt-lex:tbl:emb-eff}
    }
  \end{center}
\end{table}

\begin{figure}[bht]
  \centering
  \includegraphics[height=30em,width=\linewidth]{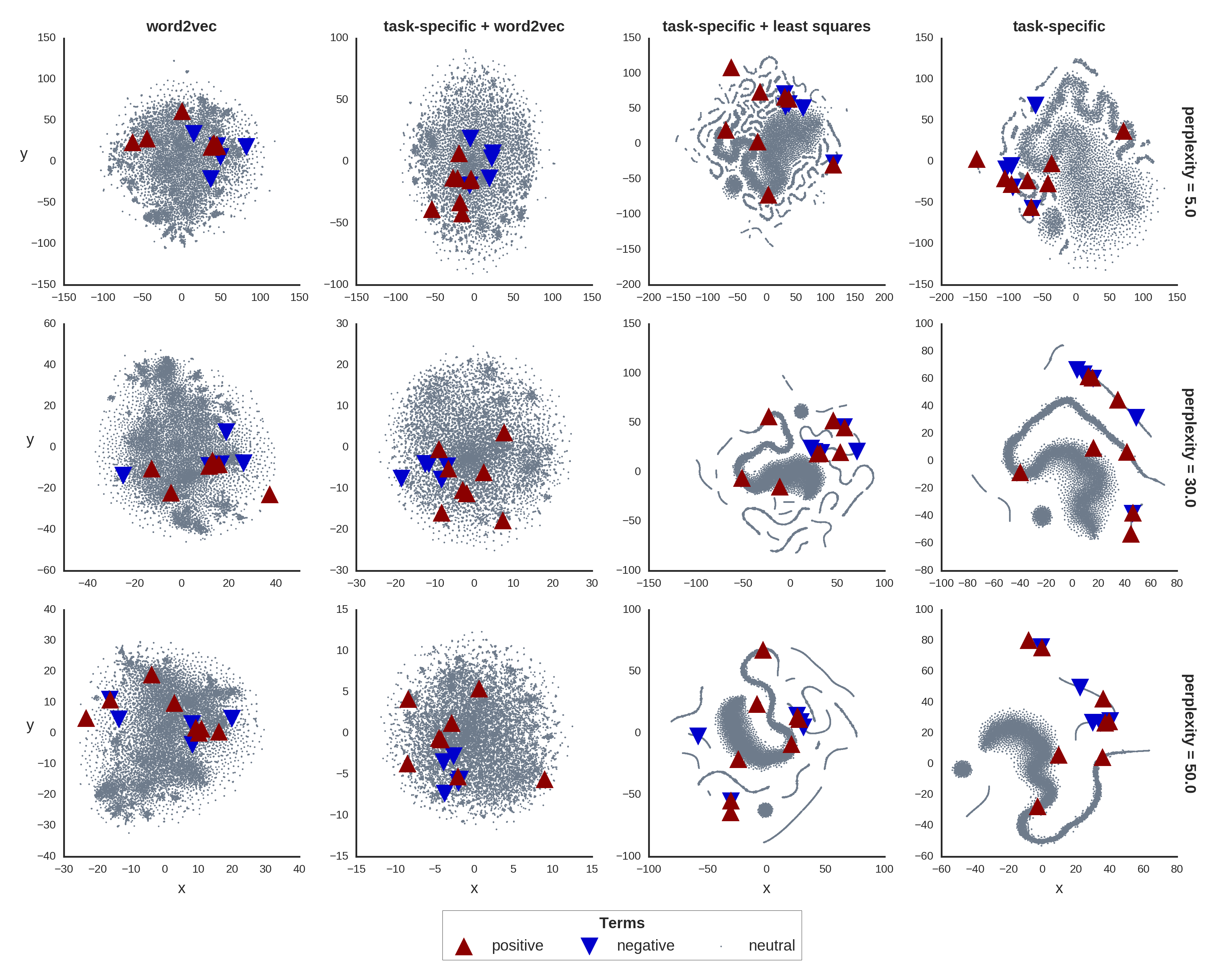}
  \caption[\citeauthor{Turney:03}'s seed
    set]{t-SNE visualization of PotTS' tokens and
    \citeauthor{Turney:03}'s seed set with different embedding
    types}\label{snt:fig:tsne-seeds}
\end{figure}

In order to understand the reasons for these differences, we projected
the embeddings of all tokens that appeared in our corpus onto the
two-dimensional vector space using the t-SNE method
of~\citet{Maaten:08}, and visualized these vectors, highlighting polar
terms from the \citeauthor{Turney:03}'s seed set.  Following the
recommended practices for analyzing t-SNE~\cite{Wattenberg:2016}, we
generated these lower-dimensional projections for different perplexity
values: $p\in\{5, 30, 50\}$; and present the results of this
visualization in Figure~\ref{snt:fig:tsne-seeds}.

As we can see from the right column of the figure, task-specific
representations of polar terms tend to appear close to each other, but
apart from the rest of the vectors.  Consequently, the centroids of
these terms will be far away from the center of neutral words, which
partially explains better results of the nearest centroids achieved
with this embedding type.  At the same time, because polar terms are
far away from the majority of embeddings, only few of them will appear
in the neighborhood of other words, which causes the $k$-NN classifier
to consider most terms as neutral.  Similarly, in the linear
projection method, the optimal polarity line will run parallel to both
polar and neutral lexemes, assigning high scores to both of these
classes.  Unfortunately, the subjectivity axis, which is supposed to
help distinguish between subjective and objective instances, is
apparently not strong enough to overcome this confusion.

\subsubsection{Vector Normalization}\label{subsec:snt-lex:eowet}

Another factor that influenced the quality of NWE-induced polarity
lists was length normalization and mean scaling of input vectors,
which we applied at the very beginning of the training.

To check the effect of this procedure, we reran our algorithms without
vector normalization, and present the results of our evaluation in
Table~\ref{snt-lex:tbl:emb-evn}.

As we can see from the scores, nearest centroids are almost
indifferent to this preprocessing, showing the same scores for all
settings.  But the remaining three approaches ($k$-NN, PCA, and linear
projection) achieve their best results when both normalization steps
are used.  We should, however, admit that this success is mostly due
to the length normalization rather than mean scaling.  We can
recognize this by comparing the scores in the first and third columns
of the table, where PCA shows identical results, and the scores of
$k$-NN and linear projection differ by only 0.001.  We also can
observe that mean-scaling alone has a strong negative effect on the
linear projection system, pulling its scores down by 0.026 in
comparison with unnormalized vectors.

\begin{table}[thb!]
  \begin{center}
    \bgroup\setlength\tabcolsep{0.1\tabcolsep}%
    \setlength{\belowrulesep}{0pt}\scriptsize
    \begin{tabular}{p{0.145\columnwidth} 
        *{4}{>{\centering\arraybackslash}p{0.21\columnwidth}}} 
      \toprule
      \multirow{2}*{\bfseries SLG Method} & \multicolumn{4}{c}{\bfseries Vector Normalization}\\
                                          & mean normalization + length normalization %
                                          & mean normalization %
                                          & length normalization %
                                          & no normalization\\\midrule
      NC & \textbf{0.398} & \textbf{0.398} & \textbf{0.398} & \textbf{0.398}\\
      $k$-NN & \textbf{0.468} & 0.418 & 0.467 & 0.417\\
      PCA & \textbf{0.398} & 0.396 & \textbf{0.398} & 0.396\\
      LP & \textbf{0.462} & 0.416 & 0.461 & 0.442\\\bottomrule

    \end{tabular}\egroup%
    {
      \captionsetup{justification=centering}
      \caption[Macro-averaged \F-scores of NWE-based methods with
        different vector normalizations]{Macro-averaged \F-scores of
        NWE-based methods with different vector
        normalizations}\label{snt-lex:tbl:emb-evn} }
  \end{center}
\end{table}

\section{Seed Sets}\label{subsec:snt-lex:eoss}

Finally, the presumably most important factor that significantly
affected the quality of all sentiment lexicons was the set of seed
terms that we used to initialize these polarity lists.  In order to
estimate the impact of this setting, we rerun our experiments using
the seed sets proposed by \citet{Hu:04}, \citet{Kim:04},
\citet{Esuli:06c}, and \citet{Remus:10}.  Since \citet{Hu:04},
however, only provided a few examples from their initial polarity
list, and \citet{Kim:04} did not specify any seeds at all, we filled
missing entries in these resources with common polar German words that
we came up with in order to match the reported cardinalities of these
sets.  Moreover, in the cases where the above seed lists were missing
the neutral category, we explicitly added a number of objective terms
proportional to the number of their polar entries.  Furthermore,
because the seed set of~\citet{Esuli:06c} had a total of~4,122 neutral
terms,\footnote{The authors considered as neutral all terms from the
  General Inquirer lexicon \cite{Stone:66} that were not marked there
  as either positive or negative.} which were difficult to translate
manually, we automatically translated these entries by using a
publicly available online
dictionary\footnote{\url{http://www.dict.cc}} and taking the first
suggested German translation for each neutral entry.\footnote{We also
  tried using all possible translations of original terms, but it
  considerably increased the number of neutral items (45,252 words)
  and lead to a substantial decrease of the final system scores.} A
short statistics on the cardinalities and compositions of the
resulting seed sets is presented in
Table~\ref{snt-lex:tbl:alt-seed-sets}.

\begin{table}[h]
  \begin{center}
    \bgroup \setlength\tabcolsep{0.1\tabcolsep}\scriptsize
    \begin{tabular}{ %
        >{\centering\arraybackslash}p{0.2\columnwidth} 
        >{\centering\arraybackslash}p{0.2\columnwidth} 
        >{\centering\arraybackslash}p{0.1\columnwidth} 
        *{2}{>{\centering\arraybackslash}p{0.243\columnwidth}}} 
      \toprule
      {\bfseries Seed Set} & %
      {\bfseries Cardinality} & %
      {\bfseries Part of Speech} & %
      {\bfseries Examples} & %
      {\bfseries Comments}\\
      \midrule
      \citet{Hu:04} & 14 positive, 15 negative, and 10 neutral terms & adjectives %
      & {{\itshape{}fantastisch, lieb, sympathisch, %
          b\"ose, dumm, schwierig}} & polar terms translated from the original paper~\cite{Hu:04}; neutral terms added by us;\\
      \citet{Kim:04} & 60 positive, 60 negative, and 60 neutral terms & any & %
      {\itshape{}fabelhaft, Hoffnung, lieben, h\"asslich, Missbrauch, t\"oten} %
      & devised by us to match the cardinality of the original set with %
      neutral terms added extra;\\
      \citet{Esuli:06c} & 16 positive, 35 negative, and 4,122 neutral terms & %
      any & {\itshape{}angenehm, ausgezeichnet, freundlich, %
        arm, bedauernswert, d\"urftig} & polar terms translated from the seed %
      set of \citet{Turney:03}; neutral terms automatically translated from %
      objective entries in the General Inquirer lexicon \cite{Stone:66};\\
      \citet{Remus:10} & 12 positive, 12 negative, and 10 neutral terms & %
      adjectives & {\itshape{}gut, sch\"on, richtig, %
        schlecht, unsch\"on, falsch} & %
      polar terms translated from the seed set of \citet{Turney:03}; %
      neutral terms added by us;\\
      \\\bottomrule
    \end{tabular}
    \egroup{}
    \caption[Overview of alternative seed sets]{ Overview of
      alternative seed sets\\ (all cardinalities are given with
      respect to the resulting German translations)}\label{snt-lex:tbl:alt-seed-sets}
  \end{center}
\end{table}

\begin{figure}[hbtp]
  \centering
  \includegraphics[height=12em,width=\linewidth]{%
    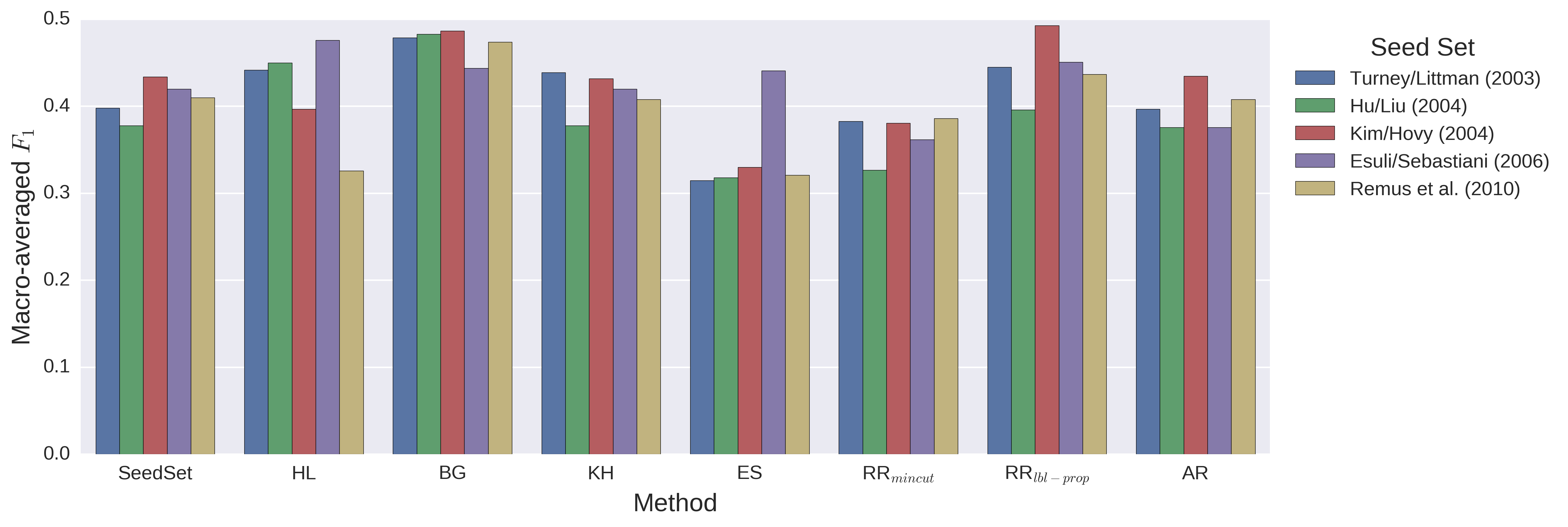}
  \caption{Macro-averaged \F{}-scores of dictionary-based approaches
    with different seed sets}\label{snt:fig:sent-dict-lex-alt-seeds}
\end{figure}

The results of dictionary-based approaches obtained with these seeds
are shown in Figure~\ref{snt:fig:sent-dict-lex-alt-seeds}.  This time,
we again can notice better scores achieved by the method of
\citet{Blair-Goldensohn:08}, which not only outperforms other systems
on average but is also less susceptible to the varying quality and
cardinalities of different sets.  The remaining methods typically
achieve their best macro-averaged \F{}-results with the polarity list
of \citet{Kim:04} or seed set of~\citet{Esuli:06c}.  The former option
works best for the label-propagation approach of \citet{Rao:09} and
the random walk algorithm of \citet{Awadallah:10}.  The latter seeds
yield best results for the approach of~\citet{Hu:04} and the
\textsc{SentiWordNet} system of~\citet{Esuli:06c}.

\begin{figure}[hbtp]
  \centering
  \includegraphics[height=12em,width=\linewidth]{%
    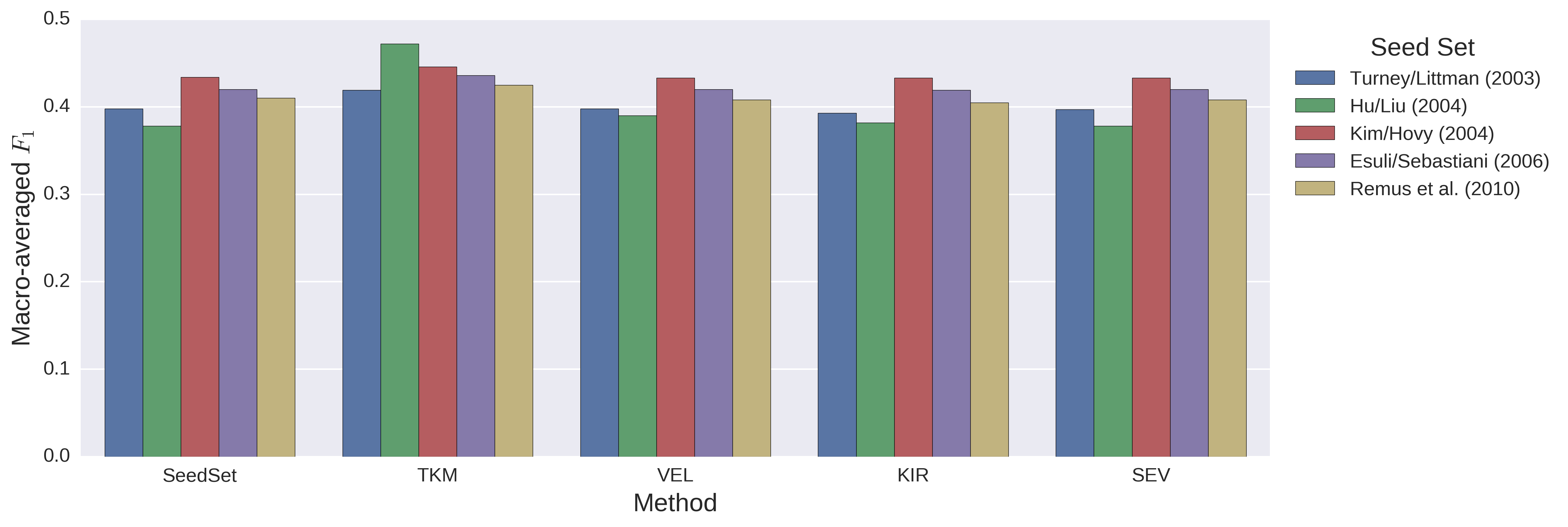}
  \caption{Macro-averaged \F{}-scores of corpus-based approaches with
    different seed sets}\label{snt:fig:sent-crp-lex-alt-seeds}
\end{figure}

A different situation is observed for corpus-based methods, whose
results are shown in Figure~\ref{snt:fig:sent-crp-lex-alt-seeds}.
Except for the approach of \citet{Takamura:05}, which achieves its
best score with the seed set of \citet{Hu:04}, all other systems (VEL,
KIR, and SEV) show very similar (though not identical) scores as the
ones reached with the seed set of \citet{Turney:03} in our initial
experiments.  The primary reason for this is again the ambiguity of
translated seeds, which leads to an early stopping of these
algorithms.

\begin{figure}[hbtp]
  \centering
  \includegraphics[height=12em,width=\linewidth]{%
    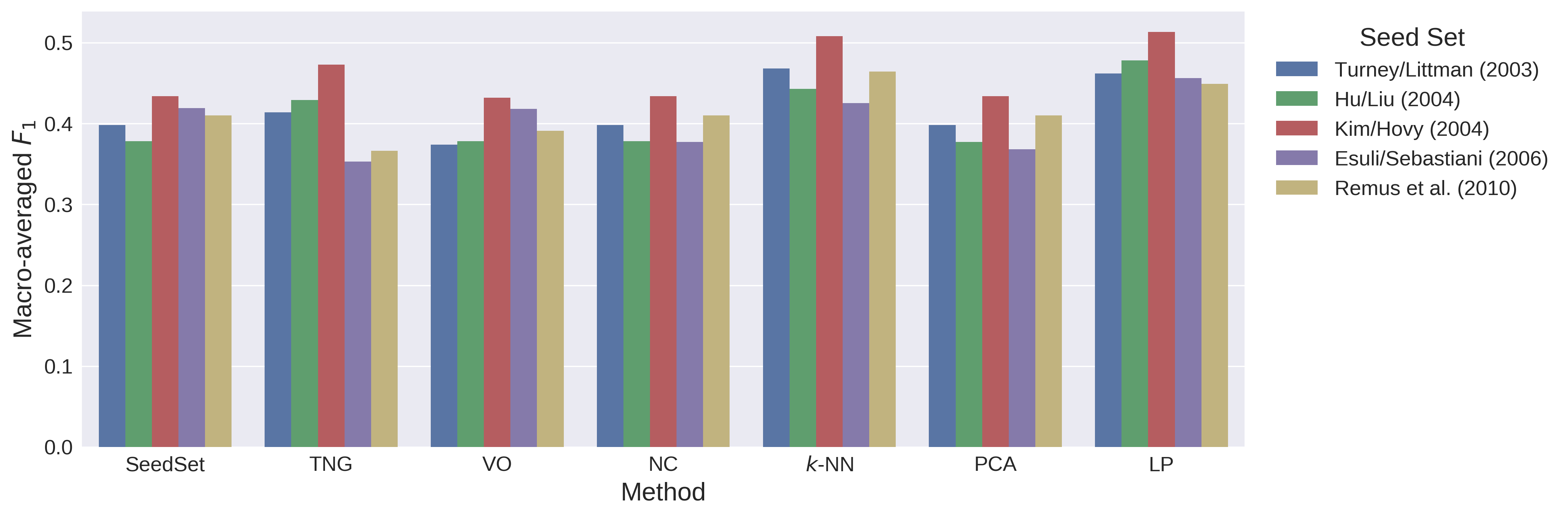}
  \caption{Macro-averaged \F{}-scores of NWE-based approaches with
    different seed sets}\label{snt:fig:sent-nwe-lex-alt-seeds}
\end{figure}

As to NWE-based methods, whose scores are presented in
Figure~\ref{snt:fig:sent-nwe-lex-alt-seeds}, we again can notice the
superior results of $k$-NN and linear projection, which both obtain
their best macro-averages (0.508 for $k$-NN and 0.513 for the linear
projection) with the seed set of~\citet{Kim:04}. Moreover, the
\F{}-score of the linear projection system achieved with these seeds
outperforms the results of all other SLG approaches, setting a new
state of the art on our corpus.  The remaining NWE-based systems also
attain their best scores with this seed list, which is not surprising
regarding the much bigger number of polar terms in this set.


\section{Analysis of Entries}\label{subsec:snt-lex:aoe}

Besides investigating the effects of different hyper-parameters and
seed sets, we also decided to have a closer look at the actual results
produced by the tested methods.  For this purpose, we extracted ten
highest-scored entries (not counting the seed terms) from each
dictionary-based automatic lexicon and present them in
Table~\ref{tbl:snt-lex:dict:top-10}.

\begin{table}[h]
  \begin{center}
    \bgroup \setlength\tabcolsep{0.03\tabcolsep}\scriptsize
    \begin{tabular}{%
        >{\centering\arraybackslash}p{0.065\columnwidth} 
        *{6}{>{\centering\arraybackslash}p{0.155\columnwidth}}} 
      \toprule
      \textbf{Rank} & %
      \textbf{HL} & \textbf{BG} & \textbf{KH} & %
      \textbf{ES} & \textbf{RR}$^{**}_{\textrm{mincut}}$ & %
      \textbf{RR}$_{\textrm{lbl-prop}}$\\\midrule
      1 & \ttranslate{perfekt}{perfect} & %
      \ttranslate{flei\ss{}ig}{diligent} &%
      \ttranslate{anr\"uchig}{indecent} &%
      \ttranslate{namenlos}{nameless} &%
      \ttranslate{planieren}{to plane} &%
      \ttranslate{prunkvoll}{splendid}\\

      2 & \ttranslate{musterg\"ultig}{immaculate} & %
      \ttranslate{b\"ose}{evil} &%
      \ttranslate{unecht}{artificial} &%
      \ttranslate{ruhelos}{restless} &%
      \ttranslate{Erdschicht}{stratum} &%
      \ttranslate{sinnlich}{sensual}\\

      3 & \ttranslate{vorbildlich}{commendable} & %
      \ttranslate{beispielhaft}{exemplary} &%
      \ttranslate{irregul\"ar}{irregular} &%
      \ttranslate{unbewaffnet}{unarmed} &%
      \ttranslate{gefallen}{please} &%
      \ttranslate{pomp\"os}{ostentatious}\\

      4 & \ttranslate{beispielhaft}{exemplary} & %
      \ttranslate{edel}{noble} &%
      \ttranslate{drittklassig}{third-class} &%
      \ttranslate{interesselos}{indifferent} &%
      \ttranslate{Zeiteinheit}{time unit} &%
      \ttranslate{unappetitlich}{unsavory}\\

      5 & \ttranslate{exzellent}{excellent} & %
      \ttranslate{t\"uchtig}{proficient} &%
      \ttranslate{sinnlich}{sensual} &%
      \ttranslate{reizlos}{unattractive} &%
      \ttranslate{Derivat}{derivate} &%
      \ttranslate{befehlsgem\"a\ss{}}{as ordered}\\

      6 & \ttranslate{exzeptionell}{exceptional} & %
      \ttranslate{emsig}{busy} &%
      \ttranslate{unprofessionell}{unprofessional} &%
      \ttranslate{w\"urdelos}{undignified} &%
      \ttranslate{Oberfl\"ache}{surface} &%
      \ttranslate{vierschr\"otig}{beefy}\\

      7 & \ttranslate{au\ss{}ergew\"ohnlich}{extraordinary} & %
      \ttranslate{eifrig}{eager} &%
      \ttranslate{abgeschlagen}{exhausted} &%
      \ttranslate{absichtslos}{unintentional} &%
      \ttranslate{Essbesteck}{cutlery} &%
      \ttranslate{regelgem\"a\ss}{regularly}\\

      8 & \ttranslate{au\ss{}erordentlich}{exceptionally} & %
      \ttranslate{arbeitsam}{hardworking} &%
      \ttranslate{gef\"allig}{pleasing} &%
      \ttranslate{ereignislos}{uneventful} &%
      \ttranslate{abl\"osen}{to displace} &%
      \ttranslate{wahrheitsgem\"a\ss}{true}\\

      9 & \ttranslate{viertklassig}{fourth-class} & %
      \ttranslate{musterg\"ultig}{exemplary} &%
      \ttranslate{musterg\"ultig}{exemplary} &%
      \ttranslate{regellos}{irregular} &%
      \ttranslate{Musikveranstaltung}{music event} &%
      \ttranslate{fettig}{greasy}\\

      10 & \ttranslate{sinnreich}{ingenious} & %
      \ttranslate{vorbildlich}{commendable} &%
      \ttranslate{unrecht}{wrong} &%
      \ttranslate{fehlerfrei}{accurate} &%
      \ttranslate{Gebrechen}{afflictions} &%
      \ttranslate{lumpig}{shabby}\\\bottomrule
    \end{tabular}
    \egroup
    \caption[Top-10 polar terms produced by dictionary-based
    methods]{%
      Top-10 polar terms produced by dictionary-based methods\\
      {\small ** -- the min-cut method of \citet{Rao:09} returns an
        unsorted set}}
    \label{tbl:snt-lex:dict:top-10}
  \end{center}
\end{table}

As we can see from the table, the approaches of \citet{Hu:04},
\citet{Blair-Goldensohn:08}, \citet{Kim:04}, and the label-propagation
algorithm of \citet{Rao:09} produce almost perfect polarity lists.
The \textsc{SentiWordNet} approach of \citet{Esuli:06c}, however,
already features some spurious terms among its top-scored entries
(\eg{} ``absichtslos'' [\emph{unintentional}]).  Finally, the min-cut
approach of \citet{Rao:09} returns mostly objective terms, which,
however, is due to the fact that this method performs a cluster-like
partitioning of the lexical graph without actually ranking the words
assigned to the clusters.

A different situation is observed with corpus-based systems as shown
in Table~\ref{tbl:snt-lex:crp:top-10}: The top-scoring polarity lists
returned by all of these approaches not only include many apparently
neutral terms but are also difficult to interpret in general, as they
contain a substantial number of slang and advertising expressions
(\eg{} ``BMKS65,'' ``\#gameinsight,'' ``\#androidgames'').

\begin{table}[hbt!]
  \begin{center}
    \bgroup \setlength\tabcolsep{0.03\tabcolsep}\scriptsize
    \begin{tabular}{%
        >{\centering\arraybackslash}p{0.06\columnwidth} 
        *{4}{>{\centering\arraybackslash}p{0.233\columnwidth}}} 
      \toprule
      \textbf{Rank} & %
      \textbf{TKM} & \textbf{VEL} & \textbf{KIR} & %
      \textbf{SEV} \\\midrule
      1 & \ttranslate{Stockfotos}{stock photos} &%
      \ttranslate{Wahl\-kampf\-ge\-schenk}{election gift} &%
      \ttranslate{Suchmaschinen}{search engines} &%
      \ttranslate{Scherwey}{Scherwey}\\

      2 & \ttranslate{BMKS65}{BMKS65} &%
      \ttranslate{Or\-dens\-ge\-schich\-te}{order history} &%
      \ttranslate{\#gameinsight}{\#gameinsight} &%
      \ttranslate{krebsen}{to crawl}\\

      3 & \ttranslate{Ziya}{Ziya} &%
      \ttranslate{Indologica}{Indologica} &%
      \ttranslate{\#androidgames}{\#androidgames} &%
      \ttranslate{kaschieren}{to conceal}\\

      4 & \ttranslate{Shoafoundation}{shoah found.} &%
      \ttranslate{Indologie}{Indology} &%
      \ttranslate{Selamat}{selamat} &%
      \ttranslate{Davis}{Davis}\\

      5 & \ttranslate{T1199}{T1199} &%
      \ttranslate{Energieverbrauch}{energy consumption} &%
      \ttranslate{Pagi}{Pagi} &%
      \ttranslate{\#Klassiker}{\#classics}\\

      6 & \ttranslate{Emilay55}{Emilay55} &%
      \ttranslate{Schimmelbildung}{mold formation} &%
      \ttranslate{\#Sparwelt}{\#savingsworld} &%
      \ttranslate{Nationalismus}{nationalism}\\

      7 & \ttranslate{Eneramo}{Eneramo} &%
      \ttranslate{Hygiene}{hygiene} &%
      \ttranslate{\#Seittest}{\#Seittest} &%
      \ttranslate{Kraftstoff}{fuel}\\

      8 & \ttranslate{GotzeID}{GotzeID} &%
      \ttranslate{wasserd}{waterp} &%
      \ttranslate{Gameinsight}{Gameinsight} &%
      \ttranslate{inaktiv}{idle}\\

      9 & \ttranslate{BSH65}{BSH65} &%
      \ttranslate{heizkostensparen}{saving heating costs} &%
      \ttranslate{\#ipadgames}{\#ipadgames} &%
      \ttranslate{8DD}{8DD}\\

      10 & \ttranslate{Saymak.}{Saymak.} &%
      \ttranslate{Re\-fe\-renz\-ar\-chi\-tek\-tu\-ren}{reference architectures} &%
      \ttranslate{Fitnesstraining}{fitness training} &%
      \ttranslate{Mailadresse}{mail address}\\\bottomrule
    \end{tabular}
    \egroup
    \caption[Top-10 polar terms produced by corpus-based
    methods]{Top-10 polar terms produced by corpus-based methods}
    \label{tbl:snt-lex:crp:top-10}
  \end{center}
\end{table}

We can also observe a similar trend for the most NWE-based methods,
whose results are presented in Table~\ref{tbl:snt-lex:NWE:top-10}.  As
we can see from the examples, many of these systems obviously overrate
Internet-specific terms (\eg{} ``\%user-playlist,'' ``\%user-video,''
``www.op''), and assign higher weights to foreign words (\eg{}
``nerelere,'' ``good,'' ``nativepride'') and interjections (\eg{}
``niedlichg\"ahn'' [\emph{cuteyawn}], ``vrrrum'').  Two notable
exceptions from this trend are {$k$-NN} and linear projection, whose
top-scoring entries contain exclusively polar terms.  At the same
time, we can notice a slight susceptibility of these approaches to the
negative polarity class as eight out of ten highest ranked words in
their results have negative semantic orientation.  One possible
explanation for this could be a more pronounced distribution of
negative expressions, which pushes the vectors of these terms to more
distinguishable regions than in the case of positive lexemes.

\begin{table}[hbt!]
  \begin{center}
    \bgroup \setlength\tabcolsep{0.03\tabcolsep}\scriptsize
    \begin{tabular}{%
        >{\centering\arraybackslash}p{0.07\columnwidth} 
        *{6}{>{\centering\arraybackslash}p{0.154\columnwidth}}} 
      \toprule
      \textbf{Rank} %
      & \textbf{TNG} & \textbf{VO} & \textbf{NC} %
      & \textbf{$k$-NN} & \textbf{PCA} & \textbf{LinProj} \\\midrule

      1 & \ttranslate{internetvorr\"ate}{Internet inventories} &%
      \ttranslate{guz}{guz} &%
      \ttranslate{paion}{paion} &%
      \ttranslate{eklig}{yukky} &%
      \ttranslate{gwiyomi.}{gwiyomi.} &%
      \ttranslate{dumm}{stupid}\\

      2 & \ttranslate{\%user-playlist}{\%user playlist} &%
      \ttranslate{nerelere}{nerelere} &%
      \ttranslate{auf$\lrcorner$}{on$\lrcorner$} &%
      \ttranslate{\"atzend}{lousy} &%
      \ttranslate{seitens}{on the part of} &%
      \ttranslate{eklig}{yukky}\\

      3 & \ttranslate{dumm}{stupid} &%
      \ttranslate{www.op}{www.op} &%
      \ttranslate{folgen!}{follow!} &%
      \ttranslate{l\"acherlich}{ridiculous} &%
      \ttranslate{kritisieren}{to criticize} &%
      \ttranslate{fies}{nasty}\\

      4 & \ttranslate{wundersch\"on}{gorgeous} &%
      \ttranslate{fernsehfestival}{TV festival} &%
      \ttranslate{teil8}{part8} &%
      \ttranslate{doof}{dumb} &%
      \ttranslate{nanda}{nanda} &%
      \ttranslate{doof}{dumb}\\

      5 & \ttranslate{\"olgem\"alde}{oil painting} &%
      \ttranslate{positip}{positip} &%
      \ttranslate{stanzmesser}{punch knife} &%
      \ttranslate{dumm}{stupid} &%
      \ttranslate{@deinskysport}{@deinskysport} &%
      \ttranslate{bl\"od}{stupid}\\

      6 & \ttranslate{\%user-video}{\%user video} &%
      \ttranslate{arn}{arn} &%
      \ttranslate{niedlichg\"ahn}{cuteyawn} &%
      \ttranslate{wunderbar}{winderful} &%
      \ttranslate{doubts}{doubts} &%
      \ttranslate{komisch}{funny}\\

      7 & \ttranslate{verlosen}{to raffle} & %
      \ttranslate{asri}{asri} &%
      \ttranslate{vrrrum}{vrrrum} &%
      \ttranslate{toll}{great} &%
      \ttranslate{temos}{temos} &%
      \ttranslate{traurig}{sad}\\

      8 & \ttranslate{w\"unschen}{to wish} &%
      \ttranslate{bewerten}{to rate} &%
      \ttranslate{good$\lrcorner$}{good$\lrcorner$} &%
      \ttranslate{widerlich}{disgusting} &%
      \ttranslate{temas}{temas} &%
      \ttranslate{d\"amlich}{silly}\\

      9 & \ttranslate{d\"amlich}{silly} &%
      \ttranslate{nacht}{night} &%
      \ttranslate{nativepride}{nativepride} &%
      \ttranslate{nervig}{annoying} &%
      \ttranslate{balas}{balas} &%
      \ttranslate{peinlich}{embarrassing}\\

      10 & \ttranslate{peinlich}{embarrassing} &%
      \ttranslate{morgen}{morning} &%
      \ttranslate{$\ulcorner$whistle$\lrcorner$}{$\ulcorner$whistle$\lrcorner$} &%
      \ttranslate{schrecklich}{awful} &%
      \ttranslate{hepi}{hepi} &%
      \ttranslate{schei\ss{}en}{to crap}\\\bottomrule
    \end{tabular}
    \egroup
    \caption{Top-10 polar terms produced by NWE-based methods}
    \label{tbl:snt-lex:NWE:top-10}
  \end{center}
\end{table}

\section{Summary and Conclusions}

Concluding this chapter, we would like to recapitulate that, in this
part, we have presented a thorough review of the most popular
sentiment lexicon generation methods.  For this purpose, we first
revised existing lexicon evaluation techniques and suggested our own
(stricter) metric, in which we explicitly counted all false positive,
false negative, and true positive occurrences of positive, negative,
and neutral terms on a real-life sentiment corpus, and also computed
the macro- and micro-averaged \F-results of these polarity classes.
Using our procedure, we first evaluated the most popular
semi-automatic German lexicons: German Polarity Clues
\cite{Waltinger:10}, SentiWS \cite{Remus:10}, and the Zurich Polarity
List \cite{Clematide:10}, finding the last resource working best in
terms of the macro-\F--score.  Afterwards, we estimated the quality of
automatic polarity lists that were created with dictionary- and
corpus-based methods, coming to the conclusion that the former group
generally produced better lexicons and was less susceptible to noisy
Twitter domain.  In the next step, we introduced several novel SLG
approaches that operate on neural embeddings of words, showing that at
least two of them ($k$-nearest neighbors and linear projection)
outperformed all other compared automatic SLG algorithms.  Last but
not least, we explored the effect of different hyper-parameters and
settings on the net results of these methods, rerunning them with
alternative sets of initial seed terms, checking their performance on
different kinds of embeddings, and estimating the impact of various
vector normalization techniques.

Based on these observations and experiments, we can formulate the main
conclusions of this chapter as follows:
\begin{itemize}
\item semi-automatic translations of common English polarity lists
  notably outperform purely automatic SLG methods, which are applied
  to German data directly;
\item despite their allegedly worse ability to accommodate new
  domains, dictionary-based approaches are still better than
  corpus-based systems (at least in terms of our intrinsic metric);
\item a potential weakness of these algorithms though is their
  dependence on various types of hyper-parameters and manually
  annotated linguistic resources, which might not necessarily be
  present for every language;
\item in this regard, a viable alternative to dictionary-based methods
  are SLG systems that induce polar lexicons from neural word
  embeddings, which not only avoid the above limitations but also
  yield competitive (or even better) results;
\item with at least two of such methods ($k$-NN and linear
  projection), we were able to establish a new state of the art for
  the macro- and micro-averaged \F-scores of automatically induced
  sentiment lexicons;
\item we also checked how different types of embeddings affected the
  performance of NWE-based SLG systems, noticing that the $k$-NN and
  linear projection methods worked best with standard word2vec
  vectors, while nearest centroids and PCA yielded better results when
  using task-specific representations;
\item furthermore, we saw that all NWE-based approaches benefited from
  mean-scaling and length normalization of input vectors, improving by
  up to~5\% on their macro-averaged \F-scores;
\item finally, an extensive evaluation of various sets of seed terms
  revealed that the results of almost all tested SLG algorithms
  crucially depend on the quality of their initial seeds, with larger
  and more balanced seed sets, \eg{} like the one proposed
  by~\citet{Kim:04}, typically leading to much higher scores.
\end{itemize}

Bearing this knowledge in mind, we will now move on to exploring
further opinion-mining fields: fine-grained and message-level
sentiment analysis, in which sentiment lexicons are traditionally
considered as one of the most valuable building blocks.



\chapter{Fine-Grained Sentiment Analysis}\label{chap:fgsa}

The task of fine-grained sentiment analysis (FGSA) is to automatically
recognize subjective evaluative opinions (\emph{sentiments}), holders
of these opinions (\emph{sources}), and their respective evaluated
entities (\emph{targets}) in text.  Since an accurate automatic
prediction of these elements would allow us to track public's attitude
towards literally any object (\eg{} a product, a service, or a
political decision), FGSA is traditionally considered as one of the
most attractive, necessary, but, unfortunately, also challenging
objectives in the opinion-mining field.

Researchers usually interpret this goal as a sequence labeling (SL)
objective, and address it with one of two most popular SL techniques:
conditional random fields (CRFs) or recurrent neural networks (RNNs).
The former approach represents a discriminative probabilistic
graphical model, which relies on an extensive set of hand-crafted
features, whereas the latter methods use a recursive computational
loop and learn their feature representations completely automatically.
In this chapter, we are going to evaluate each of these solutions in
detail in order to find out which of these algorithms is better suited
for the domain of German Twitter.  But before we proceed with our
experiments, we should first briefly discuss evaluation metrics that
we are going to use to estimate the quality of these systems.

\section{Evaluation Metrics}

Because fine-grained sentiment analysis operates on \emph{spans} of
sentiment labels, which typically consist of multiple contiguous tags,
we cannot straightforwardly apply metrics that are used for evaluation
of single independent instances to this objective, as it is unclear
which instances should be measured---single tokens or complete
spans---and how partial matches should be counted in the latter case.

One possibility to estimate the quality of FGSA prediction is to
compute precision, recall, and \F{}-scores of predicted spans by using
\emph{binary-overlap} or \emph{exact-match}
metrics~\cite[see][]{Choi:06,Breck:07}.  The first method considers an
automatically labeled span as correct if it has at least one token in
common with a labeled element from the gold annotation.  The second
metric only regards an automatic span as true positive if its
boundaries are absolutely identical with the span annotated by the
human expert.  Unfortunately, both of these approaches are problematic
to a certain extent: While binary overlap might be overly optimistic,
always assigning perfect scores to automatic spans that cover the
whole sentence; exact match might, vice versa, be too drastic,
considering the whole assignment as false if only one (possibly
irrelevant) token is classified incorrectly.

Instead of relying on these measures, we decided to use a ``golden
mean'' solution proposed by \citet{Johansson:10a}, in which they
penalize predicted spans proportionally to the number of tokens whose
labels are different from the gold annotation.  More precisely, given
two sets of manually and automatically tagged spans ($\mathcal{S}$ and
$\widehat{\mathcal{S}}$, respectively), \citeauthor{Johansson:10a}
estimate the precision of automatic assignment as:
\begin{equation}\label{eq:fgsa:jmmetric}
  P(\mathcal{S}, \widehat{\mathcal{S}}) = \frac{C(\mathcal{S},
    \widehat{\mathcal{S}})}{|\widehat{\mathcal{S}}|},
\end{equation}
where $C(\mathcal{S},\widehat{\mathcal{S}})$ stands for the proportion
of overlapping tokens across all pairs of manually ($s_i$) and
automatically ($s_j$) annotated spans:
\begin{equation*}
  C(\mathcal{S}, \widehat{\mathcal{S}}) = \sum_{s_i \in
    \mathcal{S}}\sum_{s_j \in \widehat{\mathcal{S}}}c(s_i, s_j),
\end{equation*}
and the $|\widehat{\mathcal{S}}|$ term denotes the total number of
spans automatically labeled with the given tag.

Similarly, the recall of this assignment is estimated as:
\begin{equation*}
  R(\mathcal{S}, \widehat{\mathcal{S}}) = \frac{C(\mathcal{S},
    \widehat{\mathcal{S}})}{|\mathcal{S}|}.
\end{equation*}

Using these two values, one can normally compute the \F{}-measure as:
\begin{equation*}
  F_1 = 2\times\frac{P \times R}{P + R}.
\end{equation*}

Because this estimation adequately accommodates both extrema of
automatic annotation (too long and too short spans) and also penalizes
erroneous labels, we will rely on this measure throughout our
subsequent experiments.

\section{Data Preparation}\label{snt:fgsa:subsec:data}

In order to evaluate CRFs and RNNs on our dataset, we split the
complete corpus annotated by the second annotator, which we will
henceforth consider as gold standard in all subsequent experiments,
into three parts, using 70\% of it for training, 10\% as development
data, and the remaining 20\% as a test set.  We tokenized all tweets
with the same adjusted version of Potts' tokenizer that we used
previously while creating the initial corpus files, and preprocessed
these microblogs with the rule-based normalization pipeline of
\citet{Sidarenka:13}.  In this procedure, we:
\begin{itemize}
\item \emph{unified Twitter-specific phenomena} such as @-mentions,
  hyperlinks, and e-mail addresses by replacing these entities with
  special tokens that represented their semantic classes (\eg{}
  ``\%Username'' for @-mentions, ``\%URI'' for hyperlinks).  We
  removed these elements from the input, if they were grammatically
  independent from the rest of the tweet and did not play a potential
  role for the expression of sentiments (\eg{} we stripped off all
  retweet mentions and hyperlinks appearing at the very end of the
  microblog if they were not preceded by a preposition).  Furthermore,
  we substituted all emoticons with special placeholders representing
  their semantic orientation (\eg{} \smiley{} $\rightarrow$
  ``\%PosSmiley,'' \frownie{} $\rightarrow$ ``\%NegSmiley,''
  \texttt{:-O} $\rightarrow$ ``\%Smiley''), and removed the hash sign
  (\#) from all hashtags (\eg{} ``\#gl\"ucklich'' $\rightarrow$
  ``gl\"ucklich'');

\item In addition to this, we \emph{restored frequent misspellings}
  (\eg{} ``zuguckn'' $\rightarrow$ ``zugucken'' [\emph{to watch}],
  ``Tach'' $\rightarrow$ ``Tag'' [\emph{day}]), using a set of
  manually-defined heuristic rules;

\item and, finally, \emph{replaced frequent slang terms and
  abbrebiations with their standard-language equivalents} (\eg{} ``n
  bissl'' $\rightarrow$ ``ein bisschen'' [\emph{a bit of}], ``iwie''
  $\rightarrow$ ``irgendwie'' [\emph{somehow}], ``nix'' $\rightarrow$
  ``nichts'' [\emph{nothing}]).
\end{itemize}

Afterwards, we labeled all normalized sentences with part-of-speech
tags using \textsc{TreeTagger}\footnote{In particular, we used
  \textsc{TreeTagger} Version~3.2 with the German parameter file
  UTF-8.}~\cite{Schmid:95}, and parsed them with the \textsc{Mate}
dependency parser\footnote{We used \textsc{Mate} Version \texttt{3.61}
  with the German parameter model 3.6.}
\cite{Bohnet:13}.\footnote{The choice of these tools was motivated by
  their better results in our evaluation study, which we conducted
  while working on the normalization module \cite{Sidarenka:13}.}
Finally, since \texttt{MMAX2} did not provide a straightforward
support for character offsets of annotated tokens and because
automatically tokenized data could disagree with the original corpus
tokenization, we aligned manual annotation with automatically split
words with the help of the Needleman-Wunsch
algorithm~\cite{Needleman:70}.

\section{Conditional Random Fields}

The first method that we evaluated using the obtained data was
conditional random fields.  First introduced by \citet{Lafferty:01},
CRFs have rapidly grown in popularity, turning into one of the most
widely used probabilistic frameworks, which was dominating the NLP
field for almost a decade.

The main reasons for the success of this model are:
\begin{enumerate}[1)]
\item the \emph{structural nature} of CRFs, which, in contrast to
  single-entity classifiers, such as logistic regression or SVM, make
  their predictions over structured input, trying to find the most
  likely label assignment to the whole structure (typically a chain or
  a tree) and not only its individual elements;
\item the \emph{discriminative power} of this framework, which, in
  contrast to generative probabilistic models such as HMMs
  \cite{Rabiner:86}, optimizes conditional probability
  $P(\boldsymbol{Y}|\boldsymbol{X})$ instead of joint distribution
  $P(\boldsymbol{X},\boldsymbol{Y})$ and consequently can efficiently
  deal with overlapping and correlated features;
\item and, finally, the \emph{avoidance of the label bias problem},
  which other discriminative classifiers, such as maximum entropy
  Markov networks~\cite{McCallum:00}, are known to be susceptible to.
  \begin{example}[Label Bias Problem]
    The label bias problem arises in the cases where a locally optimal
    decision outweighs globally superior solutions.  Consider, for
    example, the sentence ``Aber gerade Erwachsene haben damit
    Schwierigkeiten.'' (\textit{But especially adults have
      difficulties with it.}), for which we need to compute the most
    probable sequence of part-of-speech tags.

    \begin{center}
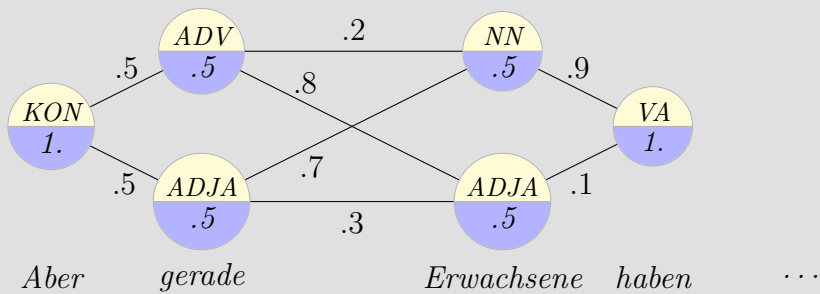

      \begin{tikzpicture}[node distance=5cm]
        \tikzstyle{tag}=[circle split,draw=gray!50,%
          minimum size=2.5em,inner ysep=2,inner xsep=0,%
          circle split part fill={yellow!20,blue!30}]
      \tikzstyle{word}=[draw=none,inner sep=10pt]

      \node[word] (A) at (1, 1) {Aber};
      \node[tag] (B) at (1, 3) {\footnotesize KON \nodepart{lower} 1.};
      \node[word] (D) at (3, 1) {gerade};
      \node[tag] (E) at (3, 2) {\footnotesize ADJA \nodepart{lower} .5};
      \node[tag] (F) at (3, 4) {\footnotesize ADV \nodepart{lower} .5} ;
      \node[word] (G) at (7, 1) {Erwachsene};
      \node[tag] (I) at (7,2) {\footnotesize ADJA \nodepart{lower} .5} ;
      \node[tag] (H) at (7,4) {\footnotesize NN \nodepart{lower} .5};
      \node[word] (J) at (9,1) {haben};
      \node[tag] (K) at (9,3) {\footnotesize VA \nodepart{lower}\small 1.};
      \node[word] (J) at (11,1) {\ldots};

      \path [-] (B) edge node[below] {$.5$} (E);
      \path [-] (B) edge node[above] {$.5$} (F);

      \path [-] (E) edge node[below] {$.3$} (I);
      \path [-] (E) edge node[below left=0.4] {$.7$} (H);
      \path [-] (F) edge node[above left=0.4] {$.8$} (I);
      \path [-] (F) edge node[above] {$.2$} (H);

      \path [-] (I) edge node[below] {$.1$} (K);
      \path [-] (H) edge node[above] {$.9$} (K);
    \end{tikzpicture}
    \captionof{figure}{Example of a CRF graph}\label{fig:snt:memm-crf}
    \end{center}
    Using features weights shown in Figure~\ref{fig:snt:memm-crf}, we
    will first estimate the probability of the correct label sequence
    for the initial part of this sentence using the Maximum Entropy
    Markov Model (MEMM)---the predecessor of the Conditional Random
    Fields.  According to the MEMM's definition, the probability of
    correct labeling ($KON-ADV-NN-VA$) is equal to:
    \begin{align*}
      P(KON, ADV, NN, VA) &= P(KON)\times P(ADV|KON)\\
      &\times P(NN|ADV)\times P(VA|NN)\\
      &=\frac{\exp(1)}{\exp(1)}\times\frac{\exp(0.5 + 0.5)}{\exp(0.5 + 0.5) + \exp(0.5 + 0.5)}\\%
      &\times\frac{\exp(0.2 + 0.5)}{\exp(0.2 + 0.5) + \exp(0.8 + 0.5)}\\
      &\times\frac{\exp(0.9 + 1.)}{\exp(0.9 + 1.)} \approx 0.177
    \end{align*}
    At the same time, the probability of the wrong variant
    ($KON-ADV-ADJA-VA$) amounts to $\approx$ 0.323 and will therefore
    be preferred by the automatic tagger.

    A different situation is observed with CRFs, where the normalizing
    factor in the denominator is computed over the whole input
    sequence without factorizing into individual terms for each
    transition as it is done in MEMM\@.  This way, the probability of
    correct labels will run up to:
    \begin{align*}
      P(KON, ADV, NN, VA) =& P(KON)\times
      P(ADV|KON)\times P(NN|ADV)\\
      &\times P(VA|NN)\\ =&\frac{\exp(1 + 0.5
        \times 3 + 0.2 + 0.9 + 1)}{Z} \approx 0.252,
    \end{align*}
    where $Z = \exp(1 + 0.5 \times 3 + 0.2 + 0.9 + 1) + \exp(1 + 0.5
    \times 3 + 0.8 + 0.1 + 1) + \exp(1 + 0.5 \times 3 + 0.7 + 0.9 + 1)
    + \exp(1 + 0.5 \times 3 + 0.3 + 0.1 + 1)$ is the total score of
    all possible label assignments; the incorrect alternative
    ($KON-ADV-ADJA-VA$), however, will get a probability score of
    $\approx$ 0.207, which is less than the score of the correct
    labeling.
  \end{example}
\end{enumerate}

\paragraph{Training.}

CRFs have these useful properties due to a neatly formulated objective
function in which they seek to optimize the global log-likelihood of
gold labels $\mathbf{Y}$ conditioned on training data $\mathbf{X}$.
In particular, given a set of training instances $\mathcal{D} =
\{(\mathbf{x}^{(n)}, \mathbf{y}^{(n)})\}_{n=1}^N$, where
$\mathbf{x}^{(n)}$ stands for the covariates of the $n$-th instance,
and $\mathbf{y}^{(n)}$ denotes its respective gold labels, CRFs try to
find feature coefficients $\mathbf{w}$ that maximize the
log-probabilities $\ell$ of $\mathbf{y}^{(i)}$ given
$\mathbf{x}^{(i)}$ over the whole corpus:
\begin{equation}\label{snt:fgsa:eq:crf-w}
  \mathbf{w} = \argmax_{\mathbf{w}}\sum_{n=1}^N\ell
  \left(\mathbf{y}^{(n)}|\mathbf{x}^{(n)}\right).
\end{equation}
The log-likelihood $\ell(\mathbf{y}^{(n)}|\mathbf{x}^{(n)})$ in this
equation is commonly estimated as the logarithm of globally (\ie{}
w.r.t\@. to the whole instance) normalized softmax function:
\begin{equation}\label{snt:fgsa:eq:crf-ell}
  \ell\left(\mathbf{y}^{(n)}|\mathbf{x}^{(n)}\right) =
  \ln\left(P(\mathbf{y}^{(n)}|\mathbf{x}^{(n)})\right) =
  \ln\left(\frac{ \exp\left(\sum_{m=1}^{M}\sum_{j}w_{j} \cdot f_j(x_{m},
    y_{m-1}, y_{m})\right)}{Z}\right),
\end{equation}
in which $M$ means the length of the $n$-th training example;
$f_j(x_{m}, y_{m-1}, y_{m})$ denotes the value of the $j$-th feature
function $f$ at position $m$; $w_j$ represents the corresponding
weight of this feature; and $Z$ is a normalization factor calculated
over all possible label assignments:
\begin{equation*}
  Z \defeq
  \sum_{y'\in\mathcal{Y},y''\in\mathcal{Y}}\exp\left(\sum_{m=1}^{M}\sum_{j}w_{j}
  \cdot f_j(x_{m}, y'_{m-1}, y''_{m})\right).
\end{equation*}
Since this normalizing term appears in the denominator and couples
together all feature weights that need to be optimized, it becomes
prohibitively expensive to find the best solution to
Equation~\ref{snt:fgsa:eq:crf-w} analytically, with a single shot.  A
possible remedy to this problem is to resort to other optimization
techniques, such as gradient descent, where feature weights are
successively changed in the direction of their gradient until they
reach the minimum of the loss function.

From Equation~\ref{snt:fgsa:eq:crf-ell}, we can see that the partial
derivative of log-likelihood w.r.t\@. a single feature weight $w_j$ is:
\begin{equation*}
  \frac{\partial}{\partial w_j}\ell =%
  \sum_{n=1}^N\sum_{m=1}^{M}f_j(x_{m}, y_{m-1}, y_{m}) -%
  \sum_{n=1}^N\sum_{m=1}^{M}\sum_{y'\in\mathcal{Y},y''\in\mathcal{Y}}f_j(x_{m},%
  y'_{m-1}, y''_{m})P(y',y''|\mathbf{x}^{(n)}),
\end{equation*}
which, after dividing both parts of the equation by the constant term
$N$ (the size of the corpus) can be transformed into:
\begin{equation*}
  \frac{1}{N}\frac{\partial}{\partial w_j}\ell = \E[f_j(\mathbf{x},
  \mathbf{y})] - \E_{\mathbf{w}}[f_j(\mathbf{x}, \mathbf{y})],
\end{equation*}
where the first term ($\E[f_j(\mathbf{x}, \mathbf{y})]$) is the
expectation of feature $f_j$ under empirical distribution, and the
second term ($\E_{\mathbf{w}}[f_j(\mathbf{x}, \mathbf{y})]$) is the
same expectation under model's parameters $\mathbf{w}$.  In other
words, the optimal solution to the log-likelihood objective in
Equation~\ref{snt:fgsa:eq:crf-ell} is achieved when model's
expectation of features matches their (true) empirical expectation on
the corpus.

The marginal probabilities of these features, which are required for
computing their expectations, can be estimated dynamically using the
forward-backward (FB) algorithm~\cite{Rabiner:90}, which is a
particular case of the more general belief-propagation
method~\cite[see][p.~81]{Barber:12}.

The only modification that one usually makes to
Equation~\ref{snt:fgsa:eq:crf-w} in practice, before applying it the
the provided training set, is the addition of so-called
\emph{regularization terms} (L1 and L2), which penalize excessively
high feature weights, thus preventing the model from overfitting the
training data, \ie{} we no longer seek feature weights that simply
maximize the probability of observed data, but we also want these
weights to be as small as possible:
\begin{equation}\label{snt:fgsa:eq:crf-w-regularization}
\mathbf{w} = \argmax_{\mathbf{w}}\sum_{n=1}^N\ell
\left(\mathbf{y}^{(n)}|\mathbf{x}^{(n)}\right) -
\lambda_1\lVert\mathbf{w}\rVert_1 - \lambda_2\lVert\mathbf{w}\rVert_2,
\end{equation}
where $\lambda_1$ and $\lambda_2$ are manually set hyper-parameters,
which control the amount of penalty that we want impose on the L1 and
L2 norms of the weights.

In our experiments, we also adopted this enhanced objective, picking
hyper-parameter values that yielded the best results on the held-out
development set.  Furthermore, in order to reduce the noise that is
typically introduced by rare, sporadic features, we only optimized the
weights of features that occurred two or more times in the training
corpus, ignoring all singleton attributes from these data.

\paragraph{Inference.}

Once optimal feature weights have been learned, one can
unproblematically compute the most likely label assignment for a new
instance by using the Viterbi algorithm~\cite{Viterbi:67}, which
effectively corresponds to the forward pass of the FB method with the
summation over the alternative preceding states replaced by the
maximum operator (hence the other name for this algorithm,
``max-product'').

\paragraph{Features.}

A crucial component that accounts for a huge part of the success (or
failure) of CRFs is features that are provided to this classifier as
input.

Traditionally, feature functions in CRFs are divided into transition-
and state-based ones.  Transition features represent real- or
binary-valued functions $f(\mathbf{x}, y'', y')\rightarrow\mathbb{R}$
associated with some data predicate
$\phi(\mathbf{x})\rightarrow\mathbb{R}$ and two labels $y''$
(typically the label of the previous token) and $y'$ (usually the
label of the current word).  The value of this function at position
$m$ in sequence $\mathbf{x}$ is then defined as:
\begin{equation*}
  f(\mathbf{x}_m, y'', y') = \begin{cases} \phi(\mathbf{x}_m), &
    \mbox{if } \mathbf{y}_{m-1} = y''\mbox{ and }\mathbf{y}_{m} =
    y'\\ 0, & \mbox{otherwise;}
  \end{cases}
\end{equation*}
where predicate~$\phi$ usually represents a simple unit function:
$\phi(\mathbf{x}_m)\mapsto 1$, $\forall\mathbf{x}_m$.

In contrast to ternary transition features, state attributes are
typically associated with binary predicates, whose output depends on
the input data at the given position and label $y'$ at the respective
state:
\begin{equation*}
  f(\mathbf{x}_m, y') = \begin{cases} \phi(\mathbf{x}_m), & \mbox{if }
    \mathbf{y}_{m} = y'\\ 0, & \mbox{otherwise.}
  \end{cases}
\end{equation*}
This time, predicate~$\phi$ is usually much more sophisticated and
reflects various properties of the input, such as whether the current
token is capitalized or whether it begins with a specific prefix or
ends with a certain suffix.  This type of features commonly accounts
for the overwhelming majority of all attributes in CRFs.

As state attributes in our experiments, we used the following
features, which, for simplicity, are listed in groups:
\begin{itemize}
\item\emph{formal}, which included the initial three characters of
  each token (\eg{} $\phi_{abc}(\mathbf{x}_m) = 1\mbox{ if
  }\mathbf{x}_m\sim\mbox{ /\textasciicircum{}abc/ else } 0$), its last
  three characters, and the spelling class of that word (\eg{}
  alphanumeric, digit, or punctuation);

\item\emph{morphological}, which encompassed part-of-speech tags of
  analyzed tokens, grammatical case and gender of inflectable PoS
  types, degree of comparison for adjectives, as well as mood, tense,
  and person forms for verbs;

\item\emph{lexical}, which comprised the actual lemma and form of the
  analyzed token (using one-hot encoding), its polarity class
  (positive, negative, or neutral), which we obtained from the Zurich
  Polarity Lexicon~\cite{Clematide:10};

\item and, finally, \emph{syntactic} features, which reflected the
  dependency relation via which token $x_m$ was connected to its
  parent.  In addition to this, we also used two binary attributes
  that showed whether the previous token in the sentence was the
  parent (first feature) or a child (second feature) of the current
  word.  Apart from that, we devised two more features, one of which
  encoded the dependency relation of the previous token in the
  sentence to its parent + the dependency relation of the current
  token to its ancestor; another feature reflected the dependency link
  of the next token + the dependency relation of the current token to
  its parent.
\end{itemize}

Besides the above attributes, we also introduced a set of complex
\emph{lexico-syntactic} features, which simultaneously reflected
several semantic and syntactic traits.  These were:
\begin{itemize}
\item the lemma of the syntactic parent;
\item the part-of-speech tag and polarity class of the grandparent in
  the syntactic tree;
\item the lemma of the child node + the dependency relation between
  the current token and its child;
\item the PoS tag of the child node + its dependency relation + the
  PoS tag of the current token;
\item the lemma of the child node + its dependency relation + the
  lemma of the current token;
\item the overall polarity of syntactic children, which was computed
  by summing up the polarity scores of all immediate dependents, and
  checking whether the resulting value was greater, less than, or
  equal to zero.\footnote{We again used the Zurich Polarity Lexicon
    of~\citet{Clematide:10} for computing these scores.}
\end{itemize}

\paragraph{Results.}

The results of our experiments are shown in
Table~\ref{snt-fgsa:tbl:crf-res}.  As we can see from the table, with
the given set of features, CRF can perfectly well fit the training
data, achieving a macro-averaged \F-score of~0.904.  This model,
however, can only partially generalize to unseen messages, where its
macro-\F{} reaches merely~0.287, despite the fact that the size of the
training corpus is almost 3.5 times bigger than the size of the test
set (5,616 versus 1,584 tweets).




\begin{table*}
  \begin{center}
    \bgroup\setlength\tabcolsep{0.1\tabcolsep}\scriptsize
    \begin{tabular}{p{0.162\columnwidth} 
        *{9}{>{\centering\arraybackslash}p{0.074\columnwidth}} 
        *{1}{>{\centering\arraybackslash}p{0.136\columnwidth}}} 
      \toprule
      \multirow{2}*{\bfseries Data Set} & \multicolumn{3}{c}{\bfseries Sentiment} & %
      \multicolumn{3}{c}{\bfseries Source} & %
      \multicolumn{3}{c}{\bfseries Target} & %
      \multirow{2}{0.136\columnwidth}{\bfseries\centering Macro\newline \F{}}\\\cmidrule(lr){2-4}\cmidrule(lr){5-7}\cmidrule(lr){8-10}

      & Precision & Recall & \F{} & %
      Precision & Recall & \F{} & %
      Precision & Recall & \F{} &\\\midrule

      Training Set & 0.949 & 0.908 & 0.928 & 0.903 & 0.87 & 0.886 & %
      0.933 & 0.865 & 0.898 & 0.904\\
      Test Set & 0.37 & 0.28 & 0.319 & 0.305 & 0.244 & 0.271 & 0.304 & %
      0.244 & 0.271 & 0.287\\\bottomrule
    \end{tabular}
    \egroup{}
    \caption{Results of fine-grained sentiment analysis with the
      first-order linear-chain CRFs}\label{snt-fgsa:tbl:crf-res}
  \end{center}
\end{table*}

\subsection{Feature Analysis}

To estimate the effect of different features on the net results of the
CRF system, we performed an ablation test, removing one group of state
attributes at a time and rechecking the performance of the model on
the development data.

\begin{table}[hbt]
  \begin{center}
    \bgroup\setlength\tabcolsep{0.47\tabcolsep}\scriptsize
    \begin{tabular}{p{0.14\columnwidth} 
        *{6}{>{\centering\arraybackslash}p{0.13\columnwidth}}} 
      \toprule
          \multirow{2}{0.2\columnwidth}{\bfseries Element} &
          \multirow{2}{0.1\columnwidth}{\bfseries Original\newline \F-Score} &
          \multicolumn{5}{c}{\bfseries \F-Score after Feature Removal}\\\cline{3-7}
          & & Formal & Morphological & Lexical & Syntactic & Complex\\\midrule

          Sentiment & 0.346 & 0.343\negdelta{0.003} & 0.344\negdelta{0.002} & 0.326\negdelta{0.02} & 0.345\negdelta{0.001} & 0.324\negdelta{0.022}\\
          Source & 0.309 & 0.321\posdelta{0.012} & 0.313\posdelta{0.004} & 0.265\negdelta{0.044} & 0.359\posdelta{0.05} & 0.271\negdelta{0.038}\\
          Target & 0.26 & 0.282\posdelta{0.022} & 0.252\negdelta{0.008} & 0.263\posdelta{0.003} & 0.233\negdelta{0.027} & 0.263\posdelta{0.003}\\\bottomrule
    \end{tabular}
    \egroup{}
    \caption[Results of the feature ablation tests for the CRF
      model]{Results of the feature ablation tests for the CRF
      model\\{\small\itshape (negative changes w.r.t\@. the original
        scores on the development set are shown in
        \textsuperscript{\textcolor{red3}{red}}; positive changes are
        depicted in \textsuperscript{\textcolor{seagreen}{green}}
        superscripts)\footnotemark}}\label{tbl:ablation}
  \end{center}
\end{table}

As we can see from the results in
Table~\ref{tbl:ablation},\footnotetext{Negative changes indicate good
  features in this context, since their removal leads to a degradation
  of the results.} all feature groups are useful for predicting
\markable{sentiment}s, as their removal leads to a degradation of its
scores.  This quality drop, however, is usually quite small,
suggesting that other features can easily make up for the removed
attributes.  A different situation is observed with \markable{source}s
and \markable{target}s though.  In the former case, removing formal,
morphological, and syntactic features shows a strong positive effect,
improving the \F{}-scores for \markable{source}s by up to five
percent.  Removing lexical and lexico-syntactic features, on the
contrary, worsens these results, tearing the \F-measure down by 4.4\%.
Except for the formal group, all these attributes behave completely
differently when applied to \markable{target}s, which benefit from
morphological and syntactic features, but apparently get confused by
lexical and complex attributes.

\begin{table}[hbt]
  \begin{center}
    \bgroup\setlength\tabcolsep{0.47\tabcolsep}\scriptsize
    \begin{tabular}{%
        >{\centering\arraybackslash}p{0.045\columnwidth} 
        >{\centering\arraybackslash}p{0.3\columnwidth} 
        >{\centering\arraybackslash}p{0.1\columnwidth} 
        >{\centering\arraybackslash}p{0.23\columnwidth} 
        >{\centering\arraybackslash}p{0.1\columnwidth}} 
      \toprule
          \multirow{2}{0.2\columnwidth}{Rank} &
          \multicolumn{2}{c}{\bfseries State Features} &
          \multicolumn{2}{c}{\bfseries Transition Features}\\\cmidrule(lr){2-3}\cmidrule(lr){4-5}
          & Feature & Score & Feature & Score\\\midrule

          1 & prntLemma=meiste $\rightarrow$ TRG & 18.68 & NON $\rightarrow$ TRG & -7.01\\
          2 & prntLemma=rettungsschirme $\rightarrow$ TRG & 18.3 & NON $\rightarrow$ SRC & -6.85\\
          3 & initChar=sty $\rightarrow$ NON & -16.04  & NON $\rightarrow$ SNT & -5.39\\
          4 & form=meisten $\rightarrow$ NON & 15.99 & TRG $\rightarrow$ SRC & -2.99\\
          5 & prntLemma=urlauberin $\rightarrow$ SNT & 14.74 & NON $\rightarrow$ NON & 2.69\\
          6 & lemma=anfechten  $\rightarrow$ SNT & 14.07 & SRC $\rightarrow$ NON & -2.59\\
          7 & form=thomasoppermann  $\rightarrow$ TRG & 13.44 & SNT $\rightarrow$ SNT & 2.54\\
          8 & form=bezeichnete $\rightarrow$ SNT & 13.25 & TRG $\rightarrow$ TRG & 2.31\\
          9 & deprel[0]|deprel[1]=NK|AMS $\rightarrow$ NON & 12.92 & SRC $\rightarrow$ SRC & 2.19\\
          10 & trailChar=te. $\rightarrow$ NON & 12.77 & SRC $\rightarrow$ TRG & -2.07\\\bottomrule
    \end{tabular}
    \egroup{}
    \caption[Top-10 state and transition features learned by the CRF
    model]{Top-10 state and transition features learned by the CRF
      model\\{\small (sorted by the absolute values of their
        weights)}}\label{fgsa:tbl:ablation}
  \end{center}
\end{table}

In order to get a better insight into the learned model's parameters,
we additionally extracted top-ten state and transition features,
ranked by the absolute values of their weights.  As we can see from
the statistics in Table~\ref{fgsa:tbl:ablation}, three of five
top-ranked state attributes (``meiste'' [\emph{most}],
``rettungsschirme'' [\emph{bailout}], and ``urlauberin'') are complex
features that reflect the lemma of the syntactic parent.  Another
common group of features is lemma and form of the current token: here,
we again encounter the word ``meisten'' (\emph{most}), which, however,
indicates the absence of any sentiments this time, and we also can see
two other attributes (``anfechten'' [\emph{doubt}] and ``bezeichnete''
[\emph{called}]) that represent the so-called \emph{direct speech
  events} and correlate with \markable{sentiment}s.  The remaining
feature (``thomasopperman'') is a person name, which frequently
appears as sentiment's \markable{target} in our corpus.

An interesting pattern can be observed with transition features: As we
can see from the results, top three of these attributes indicate a
strong belief in that an objective token is very unlikely to be
followed by a \markable{target}, \markable{source}, or
\markable{sentiment} tag (hence, the high negative weights of
transitions emanating from \textsc{NON}).  It is, however, quite
common that a \textsc{NON} tag will precede another \textsc{NON} (as
we can see from line 5 of the table).  Other transitions also mainly
reflect plausible regularities: It is, for instance, uncommon that a
\markable{target} of an opinion will appear immediately before a
source (\textsc{TRG}$\rightarrow$\textsc{SRC} $= -2.99$); in the same
vein, it is fairly improbable that an \textsc{SRC} tag will precede a
\markable{TRG} element (\textsc{SRC}$\rightarrow$\textsc{TRG} $=
-2.07$); nonetheless, is is perfectly acceptable that the same tag
will continue over multiple words (\eg{}
\textsc{SNT}$\rightarrow$\textsc{SNT} $= 2.54$,
\textsc{TRG}$\rightarrow$\textsc{TRG} $= 2.31$).

In order to better understand the reason for the observed overfitting
of the weights to the training data, we also compared all features
that appeared in the training set with the attributes that occurred in
the test part of the corpus.  As it turned out, more than two thirds
of all unique test features (34,186 out of 49,626) have never been
observed during the training and consequently had no meaningful model
weights.

\begin{figure*}[bht]
{
\centering
\begin{subfigure}{.5\textwidth}
  \centering
  \includegraphics[width=\linewidth]{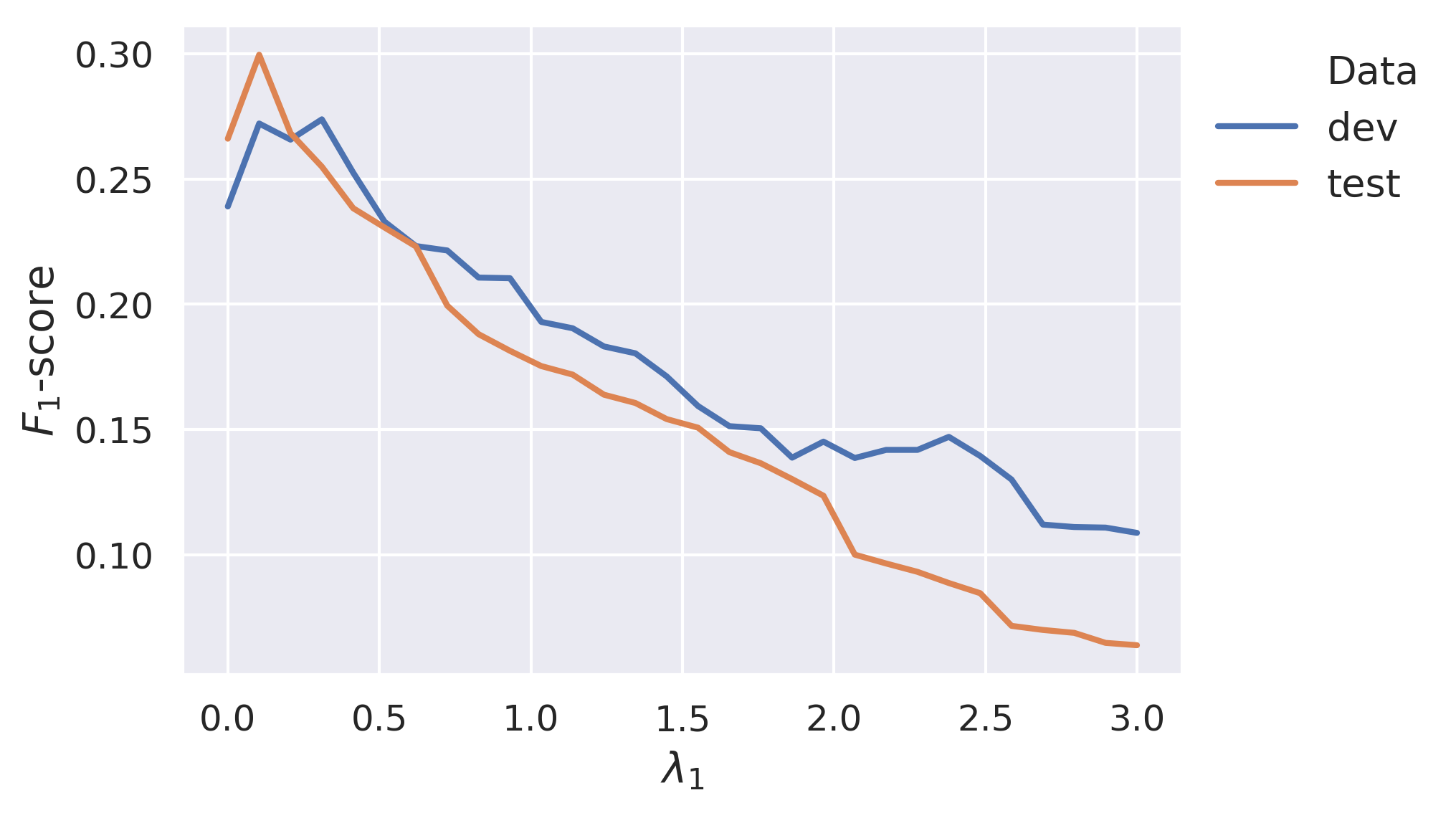}
  \caption{$\lambda_1$}\label{fgsa:fig:crf-lambda1}
\end{subfigure}%
\begin{subfigure}{.5\textwidth}
  \centering
  \includegraphics[width=\linewidth]{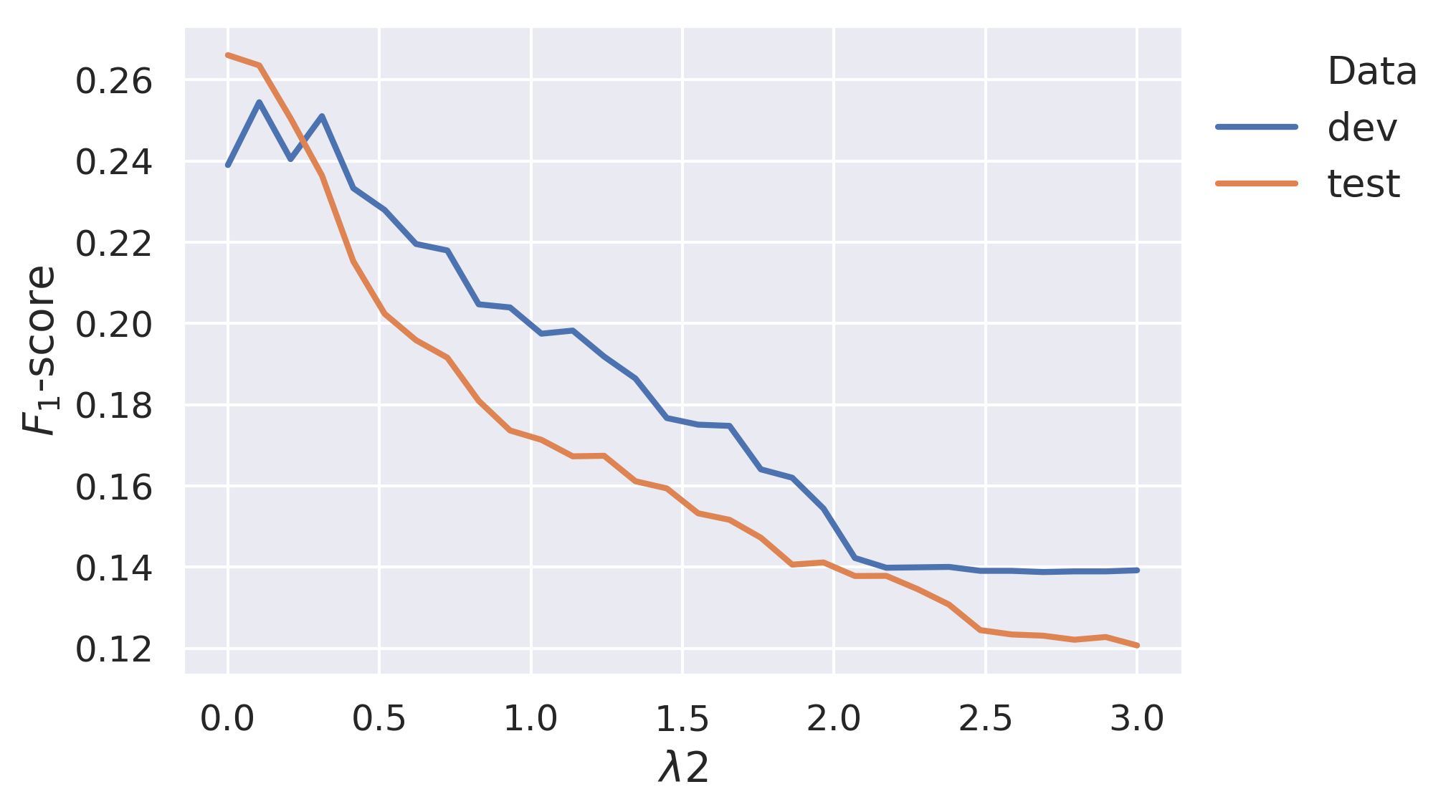}
  \caption{$\lambda_2$}\label{fgsa:fig:crf-lambda2}
\end{subfigure}
}
\caption[CRF results for different regularization values]{Results of
  the linear-chain CRFs with different values of regularization
  parameters}\label{fgsa:fig:crf-regularization}
\end{figure*}

Another factor that could significantly affect the generalization of
the CRF system was the regularization parameters $\lambda_1$ and
$\lambda_2$, which controlled the amount of penalty imposed on too big
learned feature weights (see
Equation~\ref{snt:fgsa:eq:crf-w-regularization}).  Because we chose
these parameters based on the model's results on the held-out
development data, a possible reason for rather low scores on the test
set could be a considerable difference between the distribution of
\markable{sentiment}s, \markable{source}s, and \markable{target}s in
the development and test parts of the corpus.  To see whether it
indeed was the case, we recomputed the \F{}-scores on the development
and test data, using different $\lambda$ values, and present the
results of this computation in
Figure~\ref{fgsa:fig:crf-regularization}.  As is evident from the
figure, model's \F-measure on the development set largely correlates
with its performance on the test corpus, and almost monotonically
decreases with larger $\lambda$s.

\subsection{Error Analysis}

Besides looking into model's parameters, we also decided to analyze
some errors made by the CRF system in order to understand the reasons
for its misclassifications.

\begin{example}[An Error Made by the CRF System]\label{snt:fgsa:exmp:crf-error-1}






  \noindent\textup{\bfseries\textcolor{darkred}{Gold Labels:}} {\upshape \"Uberall/TRG NPD/TRG Plakate/TRG \%NegSmiley/SNT}\\
  \noindent Everywhere/TRG NPD/TRG posters/TRG \%NegSmiley/SNT\\[\exampleSep]
  \noindent\textup{\bfseries\textcolor{darkred}{Predicted Labels:}} {\upshape \"Uberall/NON NPD/NON Plakate/NON \%NegSmiley/NON}\\
  \noindent Everywhere/NON NPD/NON posters/NON \%NegSmiley/NON\\[\exampleSep]
\end{example}

One such error is shown in Example~\ref{snt:fgsa:exmp:crf-error-1}.
In this case, the classifier has erroneously overlooked a negative
emoticon, which expresses author's attitude to election posters of the
National Democratic Party of Germany (NPD), and assigned the NON
(none) tags to all tokens of the tweet.  As it turns out, despite this
incorrect assignment, the state potentials of the smiley still achieve
their highest scores with the correct SNT (\markable{sentiment}) tag.
Moreover, the state scores of the word ``Plakate'' (\emph{posters})
also reach their maximum value (0.13 in the logarithmic domain) with
the correct TRG (\markable{target}) label.  Unfortunately, these good
guesses of single tags are overruled by the extremely high score of
the NON label (6.515) that is assigned to the first word of this
message (``\"uberall'' [\emph{everywhere}]) and is reinforced by the
transition features, which prefer contiguous runs of \textsc{NON}s.

This kind of mistakes is by far the most common type of errors that we
have observed on the development set, followed by spans with different
boundaries and invalid label sequences similar to the one shown
Example~\ref{snt:fgsa:exmp:crf-error-2}, where the classifier assigned
only SNT tags to all input tokens, although a \markable{sentiment} in
our original corpus annotation could only appear in the presence of a
\markable{target} element.

\begin{example}[An Error Made by the CRF System]\label{snt:fgsa:exmp:crf-error-2}








  \noindent\textup{\bfseries\textcolor{darkred}{Gold Labels:}}
                  {\upshape So/SNT muss/SNT das/SNT sein/SNT
                    \%PosSmiley/SNT piraten+/TRG}\\
  \noindent That/SNT 's/SNT the/SNT way/SNT how/SNT it/SNT 's/SNT\\
  supposed/SNT to/SNT be/SNT \%PosSmiley/SNT
  piraten+/TRG\\[\exampleSep]
  \noindent\textup{\bfseries\textcolor{darkred}{Predicted Labels:}}
                  {\upshape So/SNT muss/SNT das/SNT sein/SNT
                    \%PosSmiley/SNT piraten+/SNT}\\
  \noindent That/SNT 's/SNT the/SNT way/SNT how/SNT it/SNT 's/SNT\\
  supposed/SNT to/SNT be/SNT \%PosSmiley/SNT
  piraten+/TRG\\[\exampleSep]
\end{example}

\section{Recurrent Neural Networks}

A competitive alternative to CRFs is deep recurrent neural networks
(RNNs).  Introduced in the mid-nineties~\cite{Hochreiter:97}, RNNs
have become one of the most popular trends in the raging tsunami of
deep learning applications, demonstrating superior results on many
important NLP tasks including part-of-speech
tagging~\cite{Wang:15:pos}, dependency parsing~\cite{Kiperwasser:16a},
and machine
translation~\cite{Kalchbrenner:13,Bahdanau:14,Sutskever:14}.  Key
factors that account for this success are
\begin{enumerate}[1)]
\item \emph{the ability of RNN systems to learn optimal feature
  representations automatically}, which favorably sets them apart from
  traditional supervised machine-learning frameworks, such as SVMs or
  CRFs, where all features need to be defined by the user; and
\item \emph{the ability to deal with arbitrary sequence lengths},
  which advantageously distinguishes these methods from other NN
  architectures, such as plain feed-forward networks or convolutional
  systems without pooling, where the size of the input layer has to be
  constant.
\end{enumerate}

The main component that underlies any modern RNN approach is a
fixed-size hidden vector $\vec{h}$, which is recurrently updated
during the analysis of an input sequence $\mathbf{x}$ and is meant to
encode the meaning of that sequence.  The general form of this vector
at input state $t$ is usually defined as:
\begin{align*}
  \vec{h}^{(t)} = f(\vec{h}^{(t-1)}, \mathbf{x}^{(t)});
\end{align*}
where $f$ represents some non-linear transformation function,
$\vec{h}^{(t-1)}$ denotes the state of the hidden vector at the
previous time step, and $\mathbf{x}^{(t)}$ is the input vector at
position $t$.

\paragraph{LSTM.}

A fundamental problem that arises from the above definition is that
the gradients of model's parameters rapidly vanish to zero or explode
to infinity (depending on whether the absolute values of $\vec{h}$ are
less or greater than one) as the length of the input sequence
increases.  In order to solve this issue, \citet{Hochreiter:97}
proposed the long short-term memory mechanism (LSTM), in which they
explicitly incorporated the goal of keeping the gradients within an
appropriate range.  In particular, given an input sequence
$\mathbf{x}$, they introduced a special \emph{activation unit}
$\vec{i}^{(t)}$:
\begin{align*}
  \vec{i}^{(t)} &= \sigma\left(W_i\cdot \mathbf{x}^{(t)} + U_i \cdot \vec{h}^{(t-1)} + \vec{b}_i\right);
\end{align*}
where $\sigma$ denotes the sigmoid function; $W_i$, $U_i$, and
$\vec{b_i}$ represent model's parameters; $\mathbf{x}^{(t)}$ stands
for the input state; and $\vec{h}^{(t-1)}$ means the previous hidden
state.  In addition to the activation unit, the authors also estimated
a dedicated \emph{forget gate}~$\vec{f}^{(t)}$:
\begin{align*}
  \vec{f}^{(t)} &= \sigma\left(W_f\cdot \mathbf{x}^{(t)} + U_f
  \cdot \vec{h}^{(t-1)} + \vec{b}_f\right),
\end{align*}
which is used to erase parts of the previous input that appear to be
irrelevant.

After computing an \emph{intermediate update state}
$\widetilde{c}^{(t)}$ for the current time step~$t$:
\begin{align*}
  \widetilde{c}^{(t)} &= \tanh\left(W_c\cdot \mathbf{x}^{(t)} + U_c
  \cdot \vec{h}^{(t-1)} + \vec{b}_c\right),
\end{align*}
they estimated the \emph{final update}~$\vec{c}^{(t)}$ by taking a
weighted sum of the candidate update vector~$\widetilde{c}^{(t)}$ and
the previous update value~$\vec{c}^{(t-1)}$:
\begin{align*}
  \vec{c}^{(t)} &= \vec{i}^{(t)} \odot \widetilde{c}^{(t)} +
  \vec{f}^{(t)} \odot \vec{c}^{(t-1)};
\end{align*}
from which, they finally computed the output vector $\vec{o}^{(t)}$
and the new value of the hidden state $\vec{h}^{(t)}$:
\begin{align*}
  \vec{o}^{(t)} &= \sigma\left(W_o\cdot \mathbf{x}^{(t)} + U_o \cdot \vec{h}^{(t-1)} + V_o \cdot \vec{c}^{(t)} + \vec{b}_o\right),\\
  \vec{h}^{(t)} &= \vec{o}^{(t)} \odot \tanh(\vec{c}^{(t)}).
\end{align*}

\paragraph{GRU.}

Despite their enormous
popularity~\cite[\eg{}][]{Filippova:15,Ghosh:16,Rao:16}, LSTMs have
been criticized for the high complexity of their recurrent unit.  In
order to overcome this deficiency, while still keeping the gradients
within an acceptable range, \citet{Cho:14a} proposed an alternative
architecture called Gated Recurrent Units (GRU).  In this framework,
the authors also used activation and forget gates ($\vec{i}^{(t)}$ and
$\vec{f}^{(t)}$) similar to the ones defined by~\citet{Hochreiter:97}:
\begin{align*}
  \vec{i}^{(t)} &= \sigma\left(W_i\cdot \mathbf{x}^{(t)} + U_i \cdot \vec{h}^{(t-1)} + \vec{b}_i\right),\\
  \vec{f}^{(t)} &= \sigma\left(W_f\cdot \mathbf{x}^{(t)} + U_f \cdot \vec{h}^{(t-1)} + \vec{b}_f\right).
\end{align*}
With the help of these gates, they estimated the candidate activation
$\widetilde{c}^{(t)}$ as:
\begin{align*}
  \widetilde{c}^{(t)} &= tanh\left(W_c\cdot \mathbf{x}^{(t)} + U_c
  \cdot \left(\vec{f}^{(t)} \odot \vec{h}^{(t-1)}\right)  + \vec{b}_c\right),
\end{align*}
and computed the hidden state $\vec{h}^{(t)}$ as:
\begin{align*}
  \vec{h}^{(t)} &= \vec{i}^{(t)} \odot \vec{h}^{(t-1)} + \left(\vec{1} -
  \vec{i}^{(t)}\right) \odot \widetilde{c}^{(t)}.
\end{align*}

\paragraph{Final Layer.}

Because the output vectors of these recurrences ($\vec{o}^{(t)}$ in
the LSTM case, and $\vec{h}^{(t)}$ in the case of GRU) do not strictly
represent label probabilities (since elements of these vectors can
also be negative and typically do not sum to one), and, moreover,
because the size of our tagset (four tags: \textsc{SNT}, \textsc{SRC},
\textsc{TRG}, and \textsc{NON}) was obviously too small for the size
of the hidden unit, we set the dimensionality of the intermediate RNN
vectors to 100, and apply a linear transformation matrix $O \in
\mathbb{R}^{4 \times 100}$ to the final output of the recursion loop,
computing the softmax of their dot product:
\begin{align*}
  \vec{p}^{(t)} &= softmax\left(O\cdot\vec{o}^{(t)}\right).
\end{align*}
and considering the greatest value in the resulting vector as the
probability of the most likely tag.

\paragraph{Training.}

A neat property of LSTM and GRU is that the final equation, which is
obtained after unrolling the recurrence loop, is differentiable with
respect to all of its parameters, and can therefore be optimized with
standard gradient update techniques.  Since most of these parameters,
however, represent high-dimensional matrices or vectors, finding an
optimal learning rate (\ie{} the size of the update step taken in the
direction of the gradient) might pose considerable difficulties,
leading either to prohibitively large training times (if the steps are
too small) or complete divergence of the trained model (if the steps
are too large).

Several algorithms have been proposed for solving this problem,
including the method of momentum~\cite{Rumelhart:88},
AdaGrad~\cite{Duchi:11}, AdaDelta~\cite{Zeiler:12},
RMSProp~\cite{Tieleman:12}, etc.  In our experiments, we used the last
of these options---RMSProp~\cite{Tieleman:12}---as this algorithm
showed both a faster convergence and better classification results.

Another important factor that could significantly influence the
results were initial values of models' parameters.  As shown
by~\citet{He:15}, an inappropriate initialization of neural network
might lead to a complete stalling of the whole learning process.
Following recommended practices~\cite{Saxe:13}, we used orthogonal
initialization for all linear transformation matrices, and applied
uniform He sampling~\cite{He:15} for setting the initial values of
bias vectors.

Finally, due to a high imbalance of the target classes in the training
set (where most of the instances represent objective statements
without any sentiment tags), we ``upsampled'' sentiment tweets (\ie{}
we randomly repeated microblogs containing \markable{sentiment}s until
we reached an equal proportion of subjective and objective messages),
and chose the \emph{hinge-loss} as the optimized objective function
$L$:\footnote{Since most of the tokens in the over-sampled training
  set still have the \textsc{NON} tag, the easiest way for a
  classifier to minimize the objective function is to always predict
  this tag with a very high confidence.  We hoped to mitigate this
  effect by using the hinge-loss, since this function only penalizes
  incorrectly predicted labels or correct tags whose probability is
  insufficiently high (less than $c$), but does not reward any
  over-confident decisions.}
\begin{align}
  L &= \sum_{i}^{N}\sum_{t=0}^{\lvert\mathbf{x}_i\rvert}\max\left(0, %
  c + \max\limits_{y'\neq y}\vec{p}_{t,y'} - \vec{p}_{t,y}\right) + \alpha \norm{O}^2_2,
\end{align}
where $\vec{p}_{t,y'}$ stands for the probability of the most likely
wrong tag $y'$ at position $t$ in the training instance
$\mathbf{x}_i$, $\vec{p}_{t,y}$ represents the probability of the gold
label, and $\norm{O}^2_2$ stands for the $L2$-norm of the $O$ matrix.

We optimized the scalar hyper-parameters $c$ and $\alpha$ on the
development set, and trained the final model for 256 epochs, choosing
parameter values that maximized the macro-averaged \F-score on the
development set.

\paragraph{Inference.}

Since each of the above approaches (LSTM and GRU) explicitly defines
an output unit, the inference of the most likely label assignment for
an input instance $\mathbf{x}$ is straightforward and amounts to
finding the $\argmax$ value of the output vector at each time step of
the recurrence:
\begin{equation*}
  \mathbf{\hat{y}} =
  \argmax{\vec{p}^{(1)}},\argmax{\vec{p}^{(2)}},\ldots,\argmax{\vec{p}^{(|\mathbf{x}|)}}.
\end{equation*}

\paragraph{Results.}

To account for the random factors in the initialization, we repeated
each training experiment three times, and show the mean and the
standard deviation of these results in
Table~\ref{snt-fgsa:tbl:rnn-res}.

As we can see from the table, the LSTM model shows generally better
scores than the GRU system on both training and test sets.  The only
aspect at which it yields slightly worse results than the latter
approach is precision of \markable{sentiment}s, which, however, is
more than compensated for by a much higher recall.  Moreover, the
overfitting effect is significantly less pronounced than in the CRF
case (where the \F{}-scores on the training and test data differed by
a factor of three).  Nonetheless, both RNN systems achieve lower
results than the linear-chain CRFs, which indicates the fact that the
learned features still cannot capture the full extent of information
that a human expert can encode with manually defined attributes.

\begin{table*}
  \begin{center}
    \bgroup \setlength\tabcolsep{0.1\tabcolsep}\scriptsize
    \begin{tabular}{p{0.162\columnwidth} 
        *{9}{>{\centering\arraybackslash}p{0.074\columnwidth}} 
        *{1}{>{\centering\arraybackslash}p{0.136\columnwidth}}} 
      \toprule
      \multirow{2}*{\bfseries Data Set} & \multicolumn{3}{c}{\bfseries Sentiment} & %
      \multicolumn{3}{c}{\bfseries Source} & %
      \multicolumn{3}{c}{\bfseries Target} & %
      \multirow{2}{0.136\columnwidth}{\bfseries\centering Macro\newline \F{}}\\\cmidrule(lr){2-4}\cmidrule(lr){5-7}\cmidrule(lr){8-10}
      & Precision & Recall & \F{} & %
      Precision & Recall & \F{} & %
      Precision & Recall & \F{} &\\\midrule

      \multicolumn{11}{c}{\cellcolor{cellcolor}LSTM}\\



      Training Set & 0.49\stddev{0.16} & 0.75\stddev{0.01} & 0.58\stddev{0.13} & %
      0.45\stddev{0.05} & 0.63\stddev{0.12} & 0.52\stddev{0.08} %
      & 0.41\stddev{0.11} & 0.73\stddev{0.06} & 0.52\stddev{0.11} %
      & 0.54\stddev{0.11}\\




      Test Set & 0.29\stddev{0.03} & \textbf{0.31}\stddev{0.11} & \textbf{0.29}\stddev{0.03} &%
      \textbf{0.25}\stddev{0.02} & \textbf{0.31}\stddev{0.0} & \textbf{0.27}\stddev{0.01} & %
      \textbf{0.23}\stddev{0.02} & \textbf{0.25}\stddev{0.05} & \textbf{0.24}\stddev{0.01} & %
      \textbf{0.27}\stddev{0.02}\\

      \multicolumn{11}{c}{\cellcolor{cellcolor}GRU}\\




      Training Set & 0.51\stddev{0.08} & 0.66\stddev{0.05} & 0.57\stddev{0.03} & %
      0.42\stddev{0.03} & 0.62\stddev{0.05} & 0.5\stddev{0.03} & %
      0.47\stddev{0.11} & 0.63\stddev{0.11} & 0.52\stddev{0.04} & 0.53\stddev{0.03}\\




      Test Set & \textbf{0.3}\stddev{0.01} & 0.26\stddev{0.06} & 0.28\stddev{0.03} & %
      0.22\stddev{0.03} & 0.28\stddev{0.02} & 0.24\stddev{0.02} & %
      0.24\stddev{0.03} & 0.21\stddev{0.07} & 0.22\stddev{0.03} & 0.25\stddev{0.01}\\\bottomrule
    \end{tabular}
    \egroup
    \caption{Results of fine-grained sentiment analysis with recurrent
      neural networks}
    \label{snt-fgsa:tbl:rnn-res}
  \end{center}
\end{table*}

\subsection{Word Embeddings}

To see whether using different embeddings would improve the results of
the tested methods, we reran our experiments with two alternative
embedding types:
\begin{itemize}
\item \emph{word2vec vectors}~\cite{Mikolov:13}, which had been
  pretrained on the German Twitter snapshot~\cite{Scheffler:14} and
  were kept fixed during the RNN optimization;
\item and \emph{least-squares embeddings}, which were previously
  described in Chapter~\ref{chap:snt:lex};
\end{itemize}
subsequently evaluating all systems on the development set.

The results of this evaluation are shown in
Table~\ref{snt-fgsa:tbl:embeddings}.  As we can see from the scores,
least-squares representations significantly improve the recall of all
classes, which, in turn, leads to much higher macro-averaged
\F-measures in comparison with other embeddings.  The task-specific
variant shows second-best results, mainly due to a higher precision of
\markable{target}s and \markable{source}s.  Finally, word2vec vectors
also improve the prediction of \markable{sentiment} spans, but
otherwise cause a notable degradation of literally every other aspect.

\begin{table*}
  \begin{center}
    \bgroup \setlength\tabcolsep{0.1\tabcolsep}\scriptsize
    \begin{tabular}{p{0.162\columnwidth} 
        *{9}{>{\centering\arraybackslash}p{0.074\columnwidth}} 
        *{1}{>{\centering\arraybackslash}p{0.136\columnwidth}}} 
      \toprule
      \multirow{2}*{\bfseries RNN} & \multicolumn{3}{c}{\bfseries Sentiment} & %
      \multicolumn{3}{c}{\bfseries Source} & %
      \multicolumn{3}{c}{\bfseries Target} & %
      \multirow{2}{0.136\columnwidth}{\bfseries\centering Macro\newline \F{}}\\\cmidrule(lr){2-4}\cmidrule(lr){5-7}\cmidrule(lr){8-10}
      & Precision & Recall & \F{} & %
      Precision & Recall & \F{} & %
      Precision & Recall & \F{} &\\\midrule

      \multicolumn{11}{c}{\cellcolor{cellcolor}Task-Specific Embeddings}\\

      LSTM & 0.283 & 0.288 & 0.278 & %
       \textbf{0.293} & 0.372 & 0.328 & %
       \textbf{0.254} & 0.27 & \textbf{0.259} & 0.288\\

      GRU & 0.287 & 0.246 & 0.263 & %
       0.287 & 0.405 & \textbf{0.335} & %
       0.252 & 0.205 & 0.216 & 0.271\\

      \multicolumn{11}{c}{\cellcolor{cellcolor}Least-Squares Embeddings}\\

      LSTM & 0.268 & \textbf{0.37} & 0.307 & %
      0.261 & \textbf{0.414} & 0.314 & %
      0.223 & \textbf{0.275} & 0.245 & \textbf{0.289}\\

      GRU & 0.256 & 0.341 & 0.291 & %
       0.267 & 0.395 & 0.318 & %
       0.229 & 0.262 & 0.245 & 0.285\\

      \multicolumn{11}{c}{\cellcolor{cellcolor}word2vec Embeddings}\\

      LSTM & \textbf{0.291} & 0.329 & \textbf{0.309} & %
       0.2 & 0.311 & 0.244 & %
       0.221 & 0.219 & 0.22 & 0.257\\

      GRU & 0.273 & 0.355 & 0.301 & %
       0.207 & 0.353 & 0.257 & %
       0.213 & 0.26 & 0.233 & 0.264\\\bottomrule
    \end{tabular}
    \egroup
    \caption{Results of fine-grained sentiment analysis with different
      word embeddings}
    \label{snt-fgsa:tbl:embeddings}
  \end{center}
\end{table*}

\subsection{Error Analysis}

As in the previous case, we also decided to have a closer look at some
sample errors, which were committed by the tested systems.  As it
turned out, the most common type of mistakes made by both classifiers
was confusion of NON labels with other tags, which we also can see in
Examples~\ref{snt:fgsa:exmp:lstm-error} and
\ref{snt:fgsa:exmp:gru-error}.

\begin{example}[An Error Made by the LSTM System]\label{snt:fgsa:exmp:lstm-error}
  \noindent\textup{\bfseries\textcolor{darkred}{Gold Labels:}}
                  {\upshape Meine/NON Mama/NON liest/NON bei/NON
                    Twitter/NON mit/NON}\\
  \noindent My/NON mom/NON is/NON reading/NON Twitter/NON together/NON
  with/NON me/NON\\[\exampleSep]
  \noindent\textup{\bfseries\textcolor{darkred}{Predicted Labels:}}
                  {\upshape Meine/TRG Mama/NON liest/NON bei/NON
                    Twitter/NON mit/NON}\\
  \noindent My/TRG mom/NON is/NON reading/NON Twitter/NON together/NON
  with/NON me/NON\\[\exampleSep]
\end{example}

The obvious reason for these wrong predictions was the upsampling of
sentiment tweets that we used to balance the class distribution in the
training data.  Unfortunately, switching this component off caused all
classifiers to always predict only the NON tag and significantly
worsened the scores of these approaches in comparison with our initial
experiments.

\begin{example}[An Error Made by the GRU System]\label{snt:fgsa:exmp:gru-error}
  \noindent\textup{\bfseries\textcolor{darkred}{Gold Labels:}}
                  {\upshape Ich/NON habe/NON das/NON ``noch''/NON\\
                    vergessen/NON}\\
  \noindent I/NON have/NON forgotten/NON the/NON ``still''/NON\\[\exampleSep]
  \noindent\textup{\bfseries\textcolor{darkred}{Predicted Labels:}}
                  {\upshape Ich/SRC habe/NON das/TRG ``noch''/NON\\
                    vergessen/NON}\\
  \noindent I/SRC have/NON forgotten/NON the/TRG ``still''/NON\\[\exampleSep]
\end{example}

\section{Evaluation}

After estimating the results of popular FGSA approaches with their
(mostly) standard settings, evaluating their specific components
(features and word embeddings), and looking at their sample errors, we
also decided to investigate the impact of common factors, such as
annotation scheme, graph structure, and text normalization on the net
results of these methods.  For this purpose, we reran the evaluation,
changing one aspect of the training procedure at a time, and
re-estimated the scores of these systems on the development set.  The
results of these experiments are presented below.

\subsection{Annotation Scheme}

As the first factor that could affect the quality of automatic FGSA
methods, we considered the annotation scheme that we used to create
the corpus.  As described in Section~\ref{subsec:snt:ascheme}, we
initially asked our experts to assign the \markable{sentiment} label
to complete syntactic or discourse-level units that included both the
target of an opinion and its immediate evaluative expression.  Even
though this decision was linguistically plausible and helpful for
determining the boundaries of \markable{sentiment}s and their relevant
components, it also posed considerable difficulties for sequence
labeling techniques, since \markable{sentiment} tags were assigned not
only to the immediate polar terms but also to neutral words that
occurred within the same syntactic constituent as the polar item and
its target.  Since none of the tested methods could explicitly
incorporate this logic, we decided to check whether an alternative
interpretation of the annotation scheme could alleviate their
inference.

In particular, instead of unconditionally labeling all words belonging
to a \markable{sentiment} span in the original annotation with the
\textsc{SNT} tag as we did previously (which we call a \emph{broad}
interpretation of the annotation scheme), we only assigned this label
to the polar terms found in the corpus (which we call a \emph{narrow}
interpretation).  The difference between these two takes is shown in
Examples~\ref{snt:fgsa:exmp:wide} and~\ref{snt:fgsa:exmp:narrow}.
\begin{example}[Broad Sentiment
  Interpretation]\label{snt:fgsa:exmp:wide}
  \noindent\sentiment{\target{Francis} makes a \intensifier{very}
    \emoexpression{good} impression on\\ \source{me}!
    \emoexpression{:)}}

  $\rightarrow$

  \noindent Francis/TRG makes/SNT a/SNT very/SNT good/SNT
  impression/SNT on/SNT\\ me/SRC !/SNT :)/SNT
\end{example}

\begin{example}[Narrow Sentiment Interpretation]\label{snt:fgsa:exmp:narrow}
  \noindent\sentiment{\target{Francis} makes a \intensifier{very}
    \emoexpression{good} impression on\\ \source{me}!
    \emoexpression{:)}}

  $\rightarrow$

  \noindent Francis/TRG makes/NON a/NON very/NON good/SNT
  impression/NON on/NON\\ me/SRC !/NON :)/SNT
\end{example}
\noindent In the former (broad) case, we labeled the whole subjective
sentence with the \textsc{SNT} tag except for the words that denoted
the target and source of the opinion.  In the latter (narrow) case, we
only assigned the \textsc{SNT} tag to the polar term ``good'' and the
emoticon ``:),'' which, however, were expressive enough to convey the
main evaluative sense of the whole subjective statement.

\begin{table*}[hbt!]
  \begin{center}
    \bgroup \setlength\tabcolsep{0.1\tabcolsep}\scriptsize
    \begin{tabular}{p{0.162\columnwidth} 
        *{9}{>{\centering\arraybackslash}p{0.074\columnwidth}} 
        *{1}{>{\centering\arraybackslash}p{0.136\columnwidth}}} 
      \toprule
      \multirow{2}*{\bfseries Method} & \multicolumn{3}{c}{\bfseries Sentiment} & %
      \multicolumn{3}{c}{\bfseries Source} & %
      \multicolumn{3}{c}{\bfseries Target} & %
      \multirow{2}{0.136\columnwidth}{\bfseries\centering Macro\newline \F{}}\\\cmidrule(lr){2-4}\cmidrule(lr){5-7}\cmidrule(lr){8-10}
      & Precision & Recall & \F{} & %
      Precision & Recall & \F{} & %
      Precision & Recall & \F{} &\\\midrule

      \multicolumn{11}{c}{\cellcolor{cellcolor}Broad Interpretation}\\


      CRF & 0.38 & 0.32 & 0.34 & %
      \textbf{0.3} & 0.33 & 0.31 & %
      \textbf{0.29} & 0.23 & \textbf{0.26} & 0.31\\





      LSTM & 0.28 & 0.29 & 0.28 & %
      0.29 & \textbf{0.37} & \textbf{0.33} & %
      0.25 & \textbf{0.27} & \textbf{0.26} & 0.29\\





      GRU & 0.29 & 0.25 & 0.26 & %
       0.29 & 0.4 & 0.34 & %
       0.25 & 0.21 & 0.22 & 0.27\\

      \multicolumn{11}{c}{\cellcolor{cellcolor}Narrow Interpretation}\\


      CRF & 0.59 & 0.64 & 0.62 & %
      0.26 & 0.23 & 0.24 & %
      0.22 & 0.20 & 0.21 & 0.36\\




      LSTM & \textbf{0.62} & \textbf{0.65} & \textbf{0.63} & %
      \textbf{0.3} & 0.35 & 0.32 & %
      0.26 & 0.14 & 0.18 & \textbf{0.38}\\




      GRU & \textbf{0.62} & 0.63 & 0.62 & %
      0.28 & 0.33 & 0.3 & %
      0.23 & 0.24 & 0.23 & \textbf{0.38}\\\bottomrule

    \end{tabular}
    \egroup
    \caption{Results of fine-grained analysis with broad and narrow
      sentiment interpretations}
    \label{snt-fgsa:tbl:broad-narrow}
  \end{center}
\end{table*}

The results of the automatic systems with these two approaches are
given in Table~\ref{snt-fgsa:tbl:broad-narrow}.  As we can see from
the table, the broad interpretation generally leads to notably lower
scores for \markable{sentiment} spans, but yields much better results
for their \markable{source}s and \markable{target}s.  An opposite
situation is observed with the narrow scheme: even though the
\F-values for \markable{sentiment}s are twice as high as in the broad
case, the scores for the remaining elements are up to seven percent
lower.

An obvious explanation for these results is the expected better
amenability of the narrow scheme to the prediction of
\markable{sentiment} labels: since \markable{sentiment} tags are only
assigned to obvious polar terms, it becomes easier for the models to
infer this class using their state features, especially morphological
or lexical ones, or word embeddings.  But, on the other hand, such
short spans lead to disrupted label chains for other opinion-related
elements, setting \markable{sentiment} tags far apart from the spans
of their respective \markable{source}s and \markable{target}s.  As a
consequence, these classes suffer from the lack of context and become
heavily dependent on the state attributes as well.  But, this time,
the effect of state features is rather negative, because in contrast
to \markable{polar term}s, being a source or a target of an opinion is
not an inherent property of the lexical term, but arises solely from
the context which this term appears in.

Consider, for instance, the name ``Silvio Berlusconi'' in
Example~\ref{snt:fgsa:trg-ctxt}, where it appears as the target of a
sentiment in the first sentence (which expresses author's hope that
Silvio Berlusconi will not be the new Pope), but serves as a normal
subject of an objective clause in the second case.  The decision about
the role of this name depends primarily on the sense of the whole
statement rather than the name itself.  Consequently, state attributes
might only increase our prior belief that certain words would rather
appear in a subjective context, but cannot tell for sure whether they
actually do so or not.\footnote{The negative effect of state features
  on prediction of \markable{source}s and \markable{target}s was
  actually observed in our corpus, where one of the most frequently
  made mistakes was the unconditional assignment of the TRG tag to the
  word ``Nordkorea'' (\emph{North Korea}) regardless of its
  surrounding context.}  As a consequence, prediction of
\markable{source}s and \markable{target}s becomes much harder when
they do not have enough context information.

\begin{example}[Contextual Dependence of Target
  Elements]\label{snt:fgsa:trg-ctxt}
  Hoffentlich ist es nicht \target{Silvio Berlusconi}. \#Papst\\[0.5em]
  \noindent Hopefully, this won't be \target{Silvio Berlusconi}. \#Pope\\[1em]
  Silvio Berlusconi ist ein italienischer Medienmagnat und Politiker.\\[0.5em]
  \noindent Silvio Berlusconi is an Italian media tycoon and politician.\\
\end{example}

\subsection{Graph Structure}

Since the lack of contextual links played an important role for
prediction of \markable{source}s and \markable{target}s, we decided to
investigate whether redefining the way these links were established in
the models would improve the results.  For this purpose, we
implemented three possible extensions to the traditional first-order
linear-chain CRFs, which are shown in Figure~\ref{fig:snt:ho-crf}:
\begin{itemize}
  \item higher-order linear-chain CRFs,
  \item first- and higher-order semi-Markov models, and
  \item tree-structured CRFs.\footnote{The training and inference
    algorithms of these CRF variants are described in
    Appendix~\ref{chap:apdx:crf-inference}.}
\end{itemize}

\begin{figure*}[thb]
  \centering
  \begin{subfigure}[t]{0.4\textwidth}
    \centering
      {\scriptsize
  \begin{tikzpicture}
    \tikzstyle{xnode}=[rectangle,draw,fill=gray76,minimum size=2em] %
    \tikzstyle{ynode}=[circle,draw,fill=gray76,inner sep=1pt,%
    minimum width=width("MMM"),minimum height=height("MMM")] %
    \tikzstyle{factor}=[rectangle,fill=black,midway,inner sep=0pt,%
    minimum size=0.4em] %
    \tikzstyle{ctxt}=[] %

    \node[ynode] (NON0) at (1, 4) {NON};
    \node[ynode] (SRC0) at (1, 3) {SRC};
    \node[ynode] (TRG0) at (1, 2) {TRG};
    \node[ynode] (SNT0) at (1, 1) {SNT};
    \hyperNodeX{NON0}{SRC0}{TRG0}{SNT0}{w$_1$};

    \node[ynode] (NON1) at (3, 4) {NON};
    \node[ynode] (SRC1) at (3, 3) {SRC};
    \node[ynode] (TRG1) at (3, 2) {TRG};
    \node[ynode] (SNT1) at (3, 1) {SNT};
    \hyperNodeX{NON1}{SRC1}{TRG1}{SNT1}{w$_2$};

    \crfFeaturesX{0/2, 1/3, 2/4}{1}{0};


    \crfEdgesX{1}{0};
\end{tikzpicture}
}
    \caption{First-order linear-chain CRF}
  \end{subfigure}
  ~
  \begin{subfigure}[t]{0.4\textwidth}
    \centering
      {\scriptsize
  \begin{tikzpicture}
    \tikzstyle{xnode}=[rectangle,draw,fill=gray76,minimum size=2em] %
    \tikzstyle{ynode}=[rounded rectangle,draw,fill=gray76,inner sep=1pt,%
    minimum size=2.3em,minimum width=width("MMM|MMM")] %
    \tikzstyle{znode}=[rounded rectangle,fill=none,inner sep=1pt] %
    \tikzstyle{factor}=[rectangle,fill=black,midway,inner sep=0pt,%
    minimum size=0.4em] %

    \node[ynode] (NON0) at (1, 5) {NON|NON};
    \node[ynode] (DOTS0) at (1, 4) {$\ldots$};
    \node[ynode] (SRC0) at (1, 3) {SRC|SNT};
    \node[ynode] (TRG0) at (1, 2) {TRG|SNT};
    \node[ynode] (SNT0) at (1, 1) {SNT|SNT};
    \hyperNodeXX{NON0}{DOTS0}{SRC0}{TRG0}{SNT0}{w$_1$};

    \node[ynode] (NON1) at (3, 5) {NON|NON};
    \node[ynode] (DOTS1) at (3, 4) {$\ldots$};
    \node[ynode] (SRC1) at (3, 3) {SRC|SNT};
    \node[ynode] (TRG1) at (3, 2) {TRG|SNT};
    \node[ynode] (SNT1) at (3, 1) {SNT|SNT};
    \hyperNodeXX{NON1}{DOTS1}{SRC1}{TRG1}{SNT1}{w$_2$};

    \crfFeaturesXX{0/2, 1/3, 2/4}{1}{0};


    \begin{scope}[on background layer]
      \path [-] (NON0.10) edge[] node [factor] {} (NON1.177);
      \path [-] (NON0.350) edge[] node [factor] {} (NON1.185);
      \path [-] (SRC0.340) edge[] node [factor] {} (SNT1.155);
      \path [-] (SRC0.320) edge[] node [factor] {} (SNT1.160);
      \path [-] (TRG0.350) edge[] node [factor] {} (SNT1.165);
      \path [-] (TRG0.335) edge[] node [factor] {} (SNT1.170);
      \path [-] (SNT0.10) edge[] node [factor] {} (SNT1.177);
      \path [-] (SNT0.350) edge[] node [factor] {} (SNT1.185);
    \end{scope}
  \end{tikzpicture}
}
    \caption{Second-order linear-chain CRF}
  \end{subfigure}\\[1em]
  \begin{subfigure}[t]{0.4\textwidth}
    \centering
      {\scriptsize
  \begin{tikzpicture}
    \tikzstyle{xnode}=[rectangle,draw,fill=gray76,minimum size=2em] %
    \tikzstyle{ynode-el}=[rounded rectangle,draw,fill=gray76,inner sep=1pt,%
        minimum size=2.3em,minimum width={13em}] %
    \tikzstyle{ynode}=[rounded rectangle,draw,fill=gray76,inner sep=1pt,%
    minimum size=2.3em,minimum width=width("MMM")] %
    \tikzstyle{znode}=[rounded rectangle,draw=none,inner sep=1pt,minimum size=2.3em] %
    \tikzstyle{factor}=[rectangle,fill=black,midway,inner sep=0pt,%
    minimum size=0.4em] %

    \node[ynode] (NON0) at (1, 4) {NON};
    \node[ynode] (SRC0) at (1, 3) {SRC};
    \node[ynode] (TRG0) at (1, 2) {TRG};
    \node[ynode] (SNT0) at (1, 1) {SNT};
    \hyperNodeX{NON0}{SRC0}{TRG0}{SNT0}{w$_1$};

    \node[znode] (NON2) at (2.7, 4) {};
    \node[znode] (SRC2) at (2.7, 3) {};
    \node[znode] (TRG2) at (2.7, 2) {};
    \node[znode] (SNT2) at (2.7, 1) {};
    \hyperNodeX{NON2}{SRC2}{TRG2}{SNT2}{w$_2$};

    \node[znode] (NON4) at (5.3, 4) {};
    \node[znode] (SRC4) at (5.3, 3) {};
    \node[znode] (TRG4) at (5.3, 2) {};
    \node[znode] (SNT4) at (5.3, 1) {};
    \hyperNodeX{NON4}{SRC4}{TRG4}{SNT4}{w$_3$};

    \node[ynode-el] (NON1) at (4, 4) {NON};
    \node[ynode-el] (SRC1) at (4, 3) {SRC};
    \node[ynode-el] (TRG1) at (4, 2) {TRG};
    \node[ynode-el] (SNT1) at (4, 1) {SNT};

    \crfFeaturesSemiMarkov{1/2/1.193, 2/2.7/1.195, 3/3.4/1.197}{%
    1/4.6/1.343, 2/5.3/1.345, 3/6/1.347}{0};



    \begin{scope}[on background layer]

      \path [-] (NON0) edge[] node [factor] {} (NON1);
      \path [-] (SRC0) edge[] node [factor] {} (NON1.184);
      \path [-] (TRG0) edge[] node [factor] {} (NON1.186);
      \path [-] (SNT0) edge[] node [factor] {} (NON1.188);

      \path [-] (NON0) edge[] node [factor] {} (SRC1.178);
      \path [-] (SRC0) edge[] node [factor] {} (SRC1.180);
      \path [-] (TRG0) edge[] node [factor] {} (SRC1.182);
      \path [-] (SNT0) edge[] node [factor] {} (SRC1.185);

      \path [-] (NON0) edge[] node [factor] {} (TRG1.175);
      \path [-] (SRC0) edge[] node [factor] {} (TRG1.178);
      \path [-] (TRG0) edge[] node [factor] {} (TRG1.180);
      \path [-] (SNT0) edge[] node [factor] {} (TRG1.182);

      \path [-] (NON0) edge[] node [factor] {} (SNT1.172);
      \path [-] (SRC0) edge[] node [factor] {} (SNT1.174);
      \path [-] (TRG0) edge[] node [factor] {} (SNT1.176);
      \path [-] (SNT0) edge[] node [factor] {} (SNT1);
    \end{scope}
  \end{tikzpicture}
}
    \caption{Semi-Markov CRF}
  \end{subfigure}
  ~
  \begin{subfigure}[t]{0.4\textwidth}
    \centering
      {\scriptsize
  \begin{tikzpicture}
    \tikzstyle{xnode}=[rectangle,draw,fill=gray76,minimum size=2em] %
    \tikzstyle{ynode}=[circle,draw,fill=gray76,inner sep=1pt,%
    minimum width=width("MMM"),minimum height=height("MMM")] %
    \tikzstyle{factor}=[rectangle,fill=black,midway,inner sep=0pt,%
    minimum size=0.4em] %

    \node[ynode] (SNT1) at (1, 4) {SNT};

    \node[ynode] (TRG1) [right=0.4em of SNT1] {TRG};
    \node[ynode] (SRC1) [right=0.4em of TRG1] {SRC};
    \node[ynode] (NON1) [right=0.4em of SRC1] {NON};
    \hyperNodeTreeCRF{SNT1}{TRG1}{SRC1}{NON1}{w$_2$};

    \node[ynode] (SNT0) [below left=8em and 4em of SNT1] {SNT};
    \node[ynode] (TRG0) [right=0.4em of SNT0] {TRG};
    \node[ynode] (SRC0) [right=0.4em of TRG0] {SRC};
    \node[ynode] (NON0) [right=0.4em of SRC0] {NON};
    \hyperNodeTreeCRF{SNT0}{TRG0}{SRC0}{NON0}{w$_1$};

    \node[ynode] (SNT2) [right=1.2em of NON0] {SNT};
    \node[ynode] (TRG2) [right=0.4em of SNT2] {TRG};
    \node[ynode] (SRC2) [right=0.4em of TRG2] {SRC};
    \node[ynode] (NON2) [right=0.4em of SRC2] {NON};
    \hyperNodeTreeCRF{SNT2}{TRG2}{SRC2}{NON2}{w$_3$};

    \crfFeaturesX{0/1.5, 1/2.4, 2/3.3}{1}{2.85};

    \crfEdgesX{1}{0};
    \crfEdgesX{1}{2};

\end{tikzpicture}
}
    \caption{Tree-structured CRF}
  \end{subfigure}
  \caption[Factor graphs of different CRF structures]{Factor graphs of
    different CRF structures\\{\small (circles represent random
      variables; gray boxes denote observed input; factors [\ie{}
        feature functions] are shown as tiny black
      squares)}\label{fig:snt:ho-crf}}
\end{figure*}

The results of these systems on the training and development sets are
shown in Table~\ref{fgsa:tbl:crf-topologies}.
\begin{table*}[hbt!]
  \begin{center}
    \bgroup \setlength\tabcolsep{0.1\tabcolsep}\scriptsize
    \begin{tabular}{p{0.12\columnwidth} 
        *{9}{>{\centering\arraybackslash}p{0.094\columnwidth}}} 
      \toprule
      \multirow{2}*{\bfseries Element} & %
      \multicolumn{9}{c}{\bfseries Structure}\\\cline{2-10}
      & lcCRF$^1$ & lcCRF$^2$ & lcCRF$^3$ & lcCRF$^4$ & %
      smCRF$^1$ & smCRF$^2$ & smCRF$^3$ & smCRF$^4$ & trCRF$^1$\\\midrule

      \multicolumn{10}{c}{\cellcolor{cellcolor}Training Set}\\

      Sentiment & 0.928 & 0.919 & 0.922 & 0.925 & 0.931 & 0.931 & 0.933 & 0.931 & 0.906\\
      Source & 0.887 & 0.876 & 0.89  & 0.901 & 0.869 & 0.886 & 0.874 & 0.878 & 0.881\\
      Target & 0.898 & 0.811 & 0.816 & 0.827 & 0.813 & 0.827 & 0.815 & 0.817 & 0.876\\

      \multicolumn{10}{c}{\cellcolor{cellcolor}Development Set}\\

      Sentiment & 0.345 & 0.334 & 0.332 & 0.335 & \textbf{0.395} & 0.385 & 0.389 & 0.378 & 0.331\\
      Source & 0.313 & \textbf{0.32} & 0.272 & 0.304 & 0.298 & 0.282 & 0.287 & 0.291 & 0.223\\
      Target & 0.258 & 0.235 & 0.24 & 0.229 & 0.287 & \textbf{0.309} & 0.301 & 0.292 & 0.243\\\bottomrule
    \end{tabular}
    \egroup{}
    \caption[Results of fine-grained sentiment analysis with different
      CRF topologies]{Results of fine-grained sentiment analysis with
      different CRF topologies\\ {\small lcCRF---linear-chain CRFs,
        smCRF---semi-Markov CRFs, trCRF---tree-structured CRFs;\\1, 2,
        3, and 4 in the superscripts denote the order}}\label{fgsa:tbl:crf-topologies}
  \end{center}
\end{table*}

As we can see from the scores, semi-Markov CRFs achieve better results
at predicting \markable{sentiment}s and \markable{target}s, but show a
degradation when classifying \markable{source}s of sentiments.
Furthermore, second-order semi-Markov and linear-chain structures
outperform the first-order models at classifying \markable{target}s
and \markable{source}s, but further increasing the order of these
structures does not bring about any improvements.  Somewhat
surprisingly, tree-structured CRFs show even worse scores than their
linear counterparts.

In order to see whether the same tendencies would hold for
deep-learning methods, we also implemented higher-order and
tree-structured extensions of LSTM and GRU.  In the former case, we
passed a concatenation of $n$ preceding $\vec{h}$ vectors (where $n$
is the order of the model) as input to the recurrence loop.  In the
tree-structure modification, we followed the approach
of~\citet{Tai:15} and defined the LSTM unit as follows:
\begin{align*}
  \tilde{h}^{(t)} &= \sum_{k \in C\left(t\right)}\vec{h}^{(k)},\\
  \vec{i}^{(t)} &= \sigma\left(W_i\cdot\vec{x}^{(t)} + U_i\cdot\tilde{h}^{(t)} + \vec{b}_i\right),\\
  \vec{o}^{(t)} &= \sigma\left(W_o\cdot\vec{x}^{(t)} + U_o\cdot\tilde{h}^{(t)} + \vec{b}_o\right),\\
  \vec{u}^{(t)} &= \sigma\left(W_o\cdot\vec{x}^{(t)} + U_o\cdot\tilde{h}^{(t)} + \vec{b}_u\right),\\
  \vec{f}^{(t,k)} &= \sigma\left(W_f\cdot\vec{x}^{(t)} + U_f\cdot\vec{h}^{(k)} + \vec{b}_f\right),\\
  \vec{c}^{(t)} &= \vec{i}^{(t)}\odot\vec{u}^{(t)} + \sum_{k \in C(t)}f^{(t,k)}\odot c^{(k)},\\
  \vec{h}^{(t)} &= \vec{o}^{(t)}\odot tanh\left(\vec{c}^{(t)}\right);
\end{align*}
where $C\left(t\right)$ stands for the indices of all child nodes of
the token $t$.

In a similar way, we also redefined the GRU unit to the following
solutions:
\begin{align*}
  \tilde{h}^{(t)} &= \sum_{k \in C\left(t\right)}\vec{h}^{(k)},\\
  \vec{i}^{(t)} &= \sigma\left(W_i\cdot \mathbf{x}^{(t)} + U_i \cdot \tilde{h}^{(t)}\right),\\
  \vec{f}^{(t,k)} &= \sigma\left(W_f\cdot \mathbf{x}^{(t)} + U_f \cdot \vec{h}^{(t,k)}\right),\\
  \widetilde{c}^{(t)} &= tanh\left(W_c\cdot \mathbf{x}^{(t)} + U_c
  \cdot \sum_{k\in C(t)}\left(\vec{f}^{(t,k)} \odot \vec{h}^{(k)}\right)\right),\\
  \vec{h}^{(t)} &= \vec{i}^{(t)} \odot \tilde{h}^{(t)} + \left(\vec{1} -
  \vec{i}^{(t)}\right) \odot \widetilde{c}^{(t)}.
\end{align*}

The results of these modifications are shown in
Table~\ref{fgsa:tbl:nn-topologies}, from which we can see that
first-order LSTM still outperforms all higher-order LSTM and GRU
variants at predicting \markable{target}s and \markable{source}s of
opinions.  Furthermore, first-order GRU also achieves the best scores
on predicting \markable{sentiment} spans among all compared models.
This time, again, none of the tree-structured extensions can
outperform the linear-chain systems, which might be partially
explained by the errors produced by the parser, whose original target
domain is standard-language news texts.

\begin{table*}
  \begin{center}
    \bgroup \setlength\tabcolsep{0.1\tabcolsep}\scriptsize
    \begin{tabular}{p{0.12\columnwidth} 
        *{8}{>{\centering\arraybackslash}p{0.10575\columnwidth}}} 
      \toprule
      \multirow{2}*{\bfseries Element} & %
      \multicolumn{8}{c}{\bfseries Structure}\\\cline{2-9}
      & lcLSTM$^1$ & lcLSTM$^2$ & lcLSTM$^3$ & lcGRU$^1$ & %
      lcGRU$^2$ & lcGRU$^3$ & trLSTM$^1$ & trGRU$^1$\\\midrule

      \multicolumn{9}{c}{\cellcolor{cellcolor}Training Set}\\

      Sentiment & 0.584 & 0.559 & 0.54 & 0.57 & 0.587 & 0.606 & 0.43 & 0.518\\
      Source & 0.525 & 0.458 & 0.424  & 0.503 & 0.546 & 0.548 & 0.317 & 0.372 \\
      Target & 0.521 & 0.513 & 0.501 & 0.519 & 0.544 & 0.605 & 0.305 & 0.425\\

      \multicolumn{9}{c}{\cellcolor{cellcolor}Development Set}\\

      Sentiment & 0.278 & 0.285 & 0.281 & \textbf{0.335} & 0.252 & 0.253 & 0.314 & 0.292\\
      Source & \textbf{0.328} & 0.314 & 0.303  & 0.263 & 0.298 & 0.306 & 0.256 & 0.262\\
      Target & \textbf{0.259} & 0.218 & 0.222 & 0.216 & 0.219 & 0.188 & 0.205 & 0.193\\\bottomrule
    \end{tabular}
    \egroup{}
    \caption[Results of fine-grained sentiment analysis with different
    neural network topologies]{Results of fine-grained sentiment
      analysis with different neural network topologies\\ {\small
        lcLSTM---linear-chain LSTM, lcGRU---linear-chain GRU,
        trLSTM---tree-structured LSTM, trGRU---tree-structured
        GRU;\\1, 2, and 3 in the superscripts denote the order}}\label{fgsa:tbl:nn-topologies}
  \end{center}
\end{table*}

\subsection{Text Normalization}

Another question that remained open in the previous experiments was
whether the input passed to the models actually had to be normalized
or not.  As mentioned in Section~\ref{snt:fgsa:subsec:data}, when
preparing the data, we preprocessed all corpus tweets using the
rule-based normalization procedure of~\citet{Sidarenka:13}.

Even though these transformations were supposed to improve the
grammaticality of sentences, an opposite consequence of this
normalization was the loss of (potentially valuable) surface features.
In order to check which of these effects had a stronger influence on
the FGSA results, we repeated the evaluation once again, turning the
preprocessing pipeline off this time.

\begin{table*}[bht!]
  \begin{center}
    \bgroup \setlength\tabcolsep{0.1\tabcolsep}\scriptsize
    \begin{tabular}{p{0.162\columnwidth} 
        *{9}{>{\centering\arraybackslash}p{0.074\columnwidth}} 
        *{1}{>{\centering\arraybackslash}p{0.136\columnwidth}}} 
      \toprule
      \multirow{2}*{\bfseries Data Set} & \multicolumn{3}{c}{\bfseries Sentiment} & %
      \multicolumn{3}{c}{\bfseries Source} & %
      \multicolumn{3}{c}{\bfseries Target} & %
      \multirow{2}{0.136\columnwidth}{\bfseries\centering Macro\newline \F{}}\\\cmidrule(lr){2-4}\cmidrule(lr){5-7}\cmidrule(lr){8-10}
      & Precision & Recall & \F{} & %
      Precision & Recall & \F{} & %
      Precision & Recall & \F{} &\\\midrule

      \multicolumn{11}{c}{\cellcolor{cellcolor}w Normalization}\\

      CRF & \textbf{0.376} & \textbf{0.319} & \textbf{0.345} & %
       \textbf{0.298} & 0.33 & 0.313 & %
       \textbf{0.293} & 0.231 & 0.258 & \textbf{0.305}\\

      LSTM & 0.283 & 0.288 & 0.278 & %
       0.293 & 0.372 & 0.328 & %
       0.254 & \textbf{0.27} & \textbf{0.259} & 0.288\\

      GRU & 0.287 & 0.246 & 0.263 & %
       0.287 & \textbf{0.405} & \textbf{0.335} & %
       0.252 & 0.205 & 0.216 & 0.271\\

      \multicolumn{11}{c}{\cellcolor{cellcolor}w/o Normalization}\\

      CRF & 0.301 & 0.278 & 0.289 & %
       0.276 & 0.3  & 0.287 & %
       0.255 & 0.23 & 0.242 & 0.273\\

      LSTM & 0.274 & 0.252 & 0.261 & %
       0.284 & 0.367 & 0.32 & %
       0.237 & 0.241 & 0.237 & 0.273\\

      GRU & 0.266 & 0.245 & 0.252 & %
       0.296 & 0.369 & 0.328 & %
       0.232 & 0.268 & 0.245 & 0.275\\\bottomrule

    \end{tabular}\egroup{}
    \caption{Results of fine-grained sentiment analysis with (w) and
      without (w/o) text normalization}\label{snt-fgsa:tbl:normalization}
  \end{center}
\end{table*}

As we can see from the results in
Table~\ref{snt-fgsa:tbl:normalization}, text preprocessing clearly
helps sentiment classification, as all of the best observed results
are achieved exclusively with normalized text.  The only aspect that
benefits from keeping the input unchanged is precision of
\markable{target} classification with GRU, which, in turn, leads to a
slightly higher ($+0.004$) macro-averaged \F-score for this system.
Apart from that, all other aspects and classifiers show a notable
degradation when the preprocessing module is switched off.

\section{Summary and Conclusions}\label{fgsa:subsec:conclusions}

Summarizing these findings, we would like to remind the reader that in
this chapter we have evaluated two most common approaches to
fine-grained sentiment analysis: conditional random fields and
recurrent neural networks.  Our experiments showed that CRFs with
manually defined features outperform both recurrent neural networks
(LSTM and GRU), reaching a macro-averaged \F-score of~0.287 on
predicting \markable{sentiment}s, \markable{source}s, and
\markable{target}s.

Furthermore, a closer look at these systems revealed that:
\begin{itemize}
\item CRFs can learn meaningful weights for state- and
  transition-features, although different features types might have
  different effects on classification of opinion elements: whereas
  \markable{sentiment}s benefited from all features used in our
  experiments, \markable{source}s profited most from lexical and
  complex attributes, and \markable{target}s were positively
  influenced by morphological and syntactic features only;
\item Apart from that, we analyzed the effect of different embedding
  types on the net results of RNN systems, finding that least-squares
  embeddings yield the best overall scores for these methods;
\item Furthermore, even higher prediction scores for
  \markable{sentiment}s can be achieved by narrowing the spans of
  these elements to polar terms.  This, however, might negatively
  affect the classification of \markable{source}s and
  \markable{target}s;
\item Even though context seems to play an important role, redefining
  models' structures by increasing the order of their dependencies or
  performing inference over trees instead of linear chains does not
  bring much improvement.  We could, however, still outperform the
  results of traditional first-order linear-chain CRFs with their
  first- and second-order semi-Markov modifications;
\item In the final step, we estimated the effect of text normalization
  by rerunning all experiments with original (unnormalized) tweets.
  This test showed that preprocessing is an extremely helpful
  procedure, which might improve the results of FGSA methods by up to
  3\%.
\end{itemize}


\chapter{Message-Level Sentiment Analysis}\label{chap:cgsa}

Having familiarized ourselves with the peculiarities of the creation
of a sentiment corpus, the different ways to automatically induce new
polarity lists, and the difficulties of fine-grained opinion mining,
we now move on to the presumably most popular sentiment analysis
task---message-level sentiment analysis or MLSA, in which we need to
determine the overall polarity of a message.

Traditionally, this objective is addressed with either of the three
popular method groups:
\begin{itemize}
  \item lexicon-based approaches,
  \item machine-learning--based (ML) techniques,
  \item and deep-learning--based (DL) systems.
\end{itemize}
In this chapter, we are going to scrutinize the most successful
representatives of each of these paradigms, propose our own solution,
and also analyze errors, the utility of single components, and the
effect of additional training factors on the net results of these
methods.


We begin our comparison by first presenting two metrics that we will
use in our subsequent evaluation.  After briefly describing the data
preparation step, we proceed to the actual estimation of popular
lexicon-, ML-, and DL-based approaches, explaining and evaluating them
in Sections~\ref{sec:cgsa:lexicon-based},~\ref{sec:cgsa:ml-based},
and~\ref{sec:cgsa:dl-based}.  Finally, we conclude with an extensive
evaluation of different hyperparameters and settings (including the
impact of additional noisily labeled training data, various types of
sentiment lexicons, and text normalization), summarizing our results
and recapping our findings at the end of this part.

\section{Evaluation Metrics}\label{sec:cgsa:eval-metrics}

To estimate the quality of compared systems, we will rely on two
established evaluation metrics that are commonly used to measure MLSA
results: The first of these metrics is the \emph{macro-averaged
  \F-score} over two main polarity classes (positive and negative):
\begin{equation*}
  F_1 = \frac{F_{pos} + F_{neg}}{2}.
\end{equation*}
This measure was first introduced by the organizers of the SemEval
competition~\cite{Nakov:13,Rosenthal:14,Rosenthal:15} and has become a
de facto standard not only for the SemEval dataset but virtually for
all related message-level sentiment corpora and tasks.  This score is
supposed to emphasize the ability of a classifier to distinguish
between opposite semantic orientations.  Although it seemingly ignores
the neutral class, this type of misclassifications is indirectly taken
into account as well, because confusing the neutral label with another
polarity will automatically pull down the values of $F_{pos}$ or
$F_{neg}$.

The second metric, \emph{micro-averaged \F-score}, explicitly
considers all three semantic orientations (positive, negative, and
neutral) and essentially corresponds to the prediction accuracy on the
complete dataset~\cite[see][p.~577]{Manning:99}.  This measure both
predates and supersedes the SemEval evaluation as it had already been
used in the very first works on sentence-level opinion
mining~\cite{Wiebe:99,Das:01,Read:05,Kennedy:06,Go:09} and was
reintroduced again at the GermEval shared task
in~2017~\cite{Wojatzki:17}.

Besides these two metrics, we will also give a detailed information
about precision, recall, and \F-scores for each particular polarity
class.

\section{Data Preparation}\label{sec:cgsa:data}

As in the previous experiments, we preprocessed all tweets labeled by
the second annotator with the text normalization system
of~\citet{Sidarenka:13}, tokenized them using the same adjusted
version of Potts'
tokenizer,\footnote{\url{http://sentiment.christopherpotts.net/code-data/happyfuntokenizing.py}}
lemmatized and assigned part-of-speech tags to these tokens with the
\textsc{TreeTagger} of \citet{Schmid:95}, and obtained morphological
features and syntactic analyses with the \texttt{Mate} dependency
parser~\cite{Bohnet:13}.

We again divided our corpus into training, development, and test sets,
using 70\% of the tweets for learning, 10\% for tuning and picking
optimal model parameters, and the remaining 20\% for evaluating the
results.  Drawing on the work of~\citet{Wiebe:05a}, we inferred the
polarity of these microblogs, which we will consider as gold labels in
our experiments, using a simple heuristic rule in which we assigned
the positive (negative) class to the messages that had exclusively
positive (negative) annotated \markable{sentiment}s, skipping all
microblogs that simultaneously contained multiple labeled opinions
with different semantic orientations (178 tweets). In the cases when
there was no \markable{sentiment}, we recoursed to a fallback strategy
by considering all tweets that contained exclusively positive
(negative) annotated \markable{polar term}s as positive (negative),
and ignoring all messages that featured polar elements from both
polarity classes (335 messages).\footnote{Note that we inferred all
  message-level labels based on \emph{annotated} \markable{sentiment}s
  and \markable{polar term}s and did not rely on the mere occurrence
  of positive or negative smileys, which not necessarily implied an
  expression of polarity.}  Finally, all microblogs without any
\markable{sentiment}s or \markable{polar term}s were regarded as
neutral.

A few examples of such heuristically inferred labels are provided
below:
\begin{example}[Message-Level Sentiment Annotations]\label{snt:cgsa:exmp:anno1}
  \noindent\textup{\bfseries\textcolor{darkred}{Tweet:}} {\upshape
    \sentiment[polarity=positive]{Ich finde den Papst
      \emoexpression[polarity=positive]{putzig}\\
      \emoexpression[polarity=positive]{\smiley{}}}}\\
  \noindent \sentiment[polarity=positive]{I find the Pope \emoexpression[polarity=positive]{cute}\\
    \emoexpression[polarity=positive]{\smiley{}}.}\\
  \noindent\textup{\bfseries\textcolor{darkred}{Label:}}\hspace*{2em}\textbf{%
    \upshape\textcolor{green3}{positive}}\\[1.5em]
  \noindent\textup{\bfseries\textcolor{darkred}{Tweet:}} {\upshape
    \emoexpression[polarity=negative]{typisch} Bayern kaum ist der
    neue Papst da und schon haben sie ihn
    \emoexpression[polarity=negative]{in der Tasche} \ldots}\\
  \noindent \emoexpression[polarity=negative]{Typical} Bavaria The new
  Pope is hardly there, as they already have him
  \emoexpression[polarity=negative]{in their pocket}\\
  \noindent\textup{\bfseries\textcolor{darkred}{Label:}}\hspace*{2em}\textbf{%
    \upshape\textcolor{midnightblue}{negative}}
\end{example}
As we can see from the examples, our simple rule makes fairly
reasonable decisions, assigning the positive class to the first tweet,
which also expresses a positive sentiment, and labeling the second
message as negative, since it contains two negative polar terms
(``typisch'' [\emph{typical}] and ``in der Tasche haben'' [\emph{to
    have sb.\ in one's pocket}]).

But because our approach is still an approximation and consequently
prone to errors (especially in the cases where the polarity of the
whole microblog differs from the semantic orientation of its polar
terms, as in the first tweet in Example~\ref{snt:cgsa:exmp:anno2}, or
when it is expressed without any explicit polar terms at all, as in
the second microblog of this example), we decided to evaluate all MLSA
methods also on another German Twitter corpus,
SB10k~\cite{Cieliebak:17}, which was introduced when we already
started working on this chapter and which had been explicitly
annotated with message-level polarities of the tweets.

\begin{example}[Erroneous Sentiment
  Annotations]\label{snt:cgsa:exmp:anno2}
  \noindent\textup{\bfseries\textcolor{darkred}{Tweet:}} {\upshape
    Unser Park, unser Geld, unsere Stadt! -NICHT unser Finanzminister!
    \emoexpression[polarity=positive]{\smiley{}} \#schmid \#spd \#s21
    \#btw13}\\
  \noindent Our park, our money, our city! -NOT our Finance Minister!\\
  \emoexpression[polarity=positive]{\smiley{}} \#schmid \#spd \#s21
  \#btw13\\
  \noindent\textup{\bfseries\textcolor{darkred}{Label:}}\hspace*{2em}\textbf{%
    \upshape\textcolor{green3}{positive*}}\\[1.5em]
  \noindent\textup{\bfseries\textcolor{darkred}{Tweet:}} {\upshape Auf
    die Lobby-FDP von heute kann Deutschland verzichten \ldots}\\
  \noindent Germany can go without today's lobby FDP\\
  \noindent\textup{\bfseries\textcolor{darkred}{Label:}}\hspace*{2em}\textbf{%
    \upshape\textcolor{black}{neutral*}}
\end{example}

The SB10k dataset comprises a total of 9,738 microblogs, which were
sampled from a larger snapshot of 5M German tweets gathered between
August and November~2013.  To ensure lexical diversity and
proportional polarity distribution in this corpus, the authors first
grouped all posts of this snapshot into 2,500 clusters using the
$k$-means algorithm with unigram features.  Afterwards, from each of
these groups, they selected tweets that contained at least one
positive or one negative term from the German Polarity Clues
lexicon~\cite{Waltinger:10}.  Each message was subsequently annotated
by at least three human experts from a pool of 34 different
annotators.  The resulting inter-rater reliability (IRR) of this
annotation run up to 0.39 Krippendorff's
$\alpha$~\cite{Krippendorff:07}.  Unfortunately, due to the
restrictions of Twitter's terms of use, which only allow to distribute
the ids of the microblogs and their labels, we could only retrieve
7,476 tweets of this collection, which, however, still represents a
substantial part of the original dataset.

In addition to the aforementioned two corpora (PotTS and SB10k), we
also automatically annotated all microblogs of the German Twitter
Snapshot~\cite{Scheffler:14} by following the procedure
of~\citet{Read:05} and~\citet{Go:09} and assigning the positive
(negative) class to the tweets that contained respective emoticons,
regarding the rest of the microblogs as neutral.  In contrast to the
previous two datasets, whose labels were inferred or directly obtained
from manual annotations, we will not use this automatically tagged
corpus for evaluation, but will only harness it for training in our
later weak-supervision experiments.

The resulting statistics on the number of messages and polarity class
distribution in these data are shown in
Table~\ref{snt-cgsa:tbl:corp-dist}.
\begin{table}[h]
  \begin{center}
    \bgroup\setlength\tabcolsep{0.1\tabcolsep}\scriptsize
    \begin{tabular}{p{0.163\columnwidth} 
        *{6}{>{\centering\arraybackslash}p{0.135\columnwidth}}} 
      \toprule
      \textbf{Dataset} & \multicolumn{4}{c}{\bfseries Polarity Class}%
      & \multicolumn{2}{c}{\bfseries Label Agreement}\\\cmidrule(lr){2-5}\cmidrule(lr){6-7}
                       & \textbf{Positive} & \textbf{Negative} %
                                           & \textbf{Neutral} & \textbf{Mixed*} %
                                                              & $\alpha$ & $\kappa$\\\midrule

      \textbf{PotTS} & 3,380 & 1,541 & 2,558 & 513 & 0.66 & 0.4\\
      \textbf{SB10k} & 1,717 & 1,130 & 4,629 & 0 & 0.39 & \NA{}\\
      \textbf{GTS} & 3,326,829 & 350,775 & 19,453,669 & 73,776 & \NA{} & \NA{}\\\bottomrule
    \end{tabular}
    \egroup{}
    \caption[Polarity class distribution in PotTS, SB10k, and the
    German Twitter Snapshot]{Polarity class distribution in PotTS,
      SB10k, and the German
      Twitter Snapshot (GTS)\\
      \emph{(* --- the \emph{mixed} polarity was excluded from our
        experiments)}}\label{snt-cgsa:tbl:corp-dist}
  \end{center}
\end{table}

As we can see, each dataset has its own unique composition of polar
tweets: The PotTS corpus, for example, shows a conspicuous bias
towards the positive class, with 42\% of its microblogs belonging to
this polarity.  We can partially explain this skewness by the
selection criteria that we used to compile the initial data for this
collection: Because a big part of this dataset was composed from
tweets that contained smileys, and most of these emoticons were
positive, which is evident from the statistics of the German Twitter
snapshot, the selected microblogs also got biased towards this
semantic orientation.

The second most frequent group in the PotTS corpus are neutral tweets,
which account for 32\% of the data.  Negative messages, vice versa,
represent a clear minority in this collection (only 19\%), which,
however, is less surprising as the same tendency can be observed for
SB10k and the German Twitter Snapshot too.

Regarding the last two corpora, we can observe a more uniform (though
not identical) behavior, where both datasets are dominated by neutral
posts, which constitute 62\% of SB10k and 84\% of all snapshot tweets.
The positive class, again, makes up a big part of these data (23\% of
the former corpus and 14\% of the latter dataset), but its influence
this time is much less pronounced than in the PotTS case.  Finally,
negative tweets are again the least represented semantic orientation.
The only group that has even less instances than this class is the
\textsc{Mixed} polarity.  We, however, will skip the mixed orientation
in our experiments for the sake of simplicity and uniformity of
evaluation.



\section{Lexicon-Based Methods}\label{sec:cgsa:lexicon-based}

The first group of approaches that we are going to explore in this
chapter using the aforementioned data are lexicon-based (LB) systems.
Just like sentiment lexicons themselves, LB methods for message-level
opinion mining have attracted a lot of attention from the very
inception of the sentiment analysis field.  Starting from the work
of~\citet{Hatzivassi:00}, who gave a statistical proof that the mere
occurrence of a subjective adjective from an automatically compiled
polarity list was a sufficiently reliable indicator that the whole
sentence was subjective, more and more researchers started using
lexicons in order to estimate the overall polarity of a text.

One of the first notable steps in this direction was made
by~\citet{Das:01}, who proposed an ensemble of five classifiers (two
of which were purely lexicon-based and the other three heavily relied
on lexicon features) to predict the polarity of stock messages,
achieving an accuracy of 62\% on a corpus of several hundreds stock
board messages.  A much simpler method for a related task was
suggested by~\citet{Turney:02}, who determined the \emph{semantic
  orientation} (SO) of reviews by averaging the PMI scores of their
terms, getting these scores from an automatically generated sentiment
lexicon.  With this approach, the author could reach an accuracy of
74\% on a corpus of 410 manually labeled Epinions comments.  In the
same vein, \citet{Hu:04} computed the overall polarity of a sentence
by comparing the numbers of its positive and negative terms, reversing
their orientation if they appeared in a negated context.




In~\citeyear{Polanyi:06}, \citeauthor{Polanyi:06} presented an
extensive overview and analysis of common lexicon-based sentiment
methods that existed at that time, arguing that besides considering
the lexical valence (\ie{} semantic orientation) of polar expressions,
it was also important to incorporate syntactic, discourse-level, and
extra-linguistic factors such as negations, intensifiers, modal
operators (\eg{} \emph{could} or \emph{might}), presuppositional items
(\eg{} \emph{barely} or \emph{failure}), irony, reported speech,
discourse connectors, genre, attitude assessment, reported speech, and
multi-entity evaluation.  This theoretical hypothesis was also proven
empirically by \citet{Kennedy:06}, who investigated two ways to
determine the polarity of a customer review: In the first approach,
the authors simply compared the numbers of positive and negative terms
in the text, assigning the review to the class with the greater number
of items.  In the second attempt, they enhanced the original system
with an additional information about contextual valence shifters,
increasing or decreasing the sentiment score of a term if it was
preceded by an intensifier or downtoner, and changing the polarity
sign of this score to the opposite in case of a negation.  

Finally, a seminal work on lexicon-based techniques was presented
by~\citet{Taboada:11}, who introduced a manually compiled polarity
list\footnote{The authors hand-annotated all occurrences of
  adjectives, nouns, and verbs found in a corpus of 400 Epinions
  reviews with ordinal categories ranging from -5 to 5 that reflected
  the semantic orientation of a term (positive vs.\ negative) and its
  polar strength (weak vs.\ strong).} and used this resource to
estimate the overall semantic orientation of texts.  Drawing on the
ideas of~\citet{Polanyi:06}, the authors incorporated a set of
additional heuristic rules into their computation by changing the
prior SO values of negated, intensified, and downtoned terms, ignoring
irrealis and interrogative sentences, and adjusting the weights of
specific document sections.  An extensive evaluation of this approach
showed that the manual lexicon performed much better than
automatically generated polarity lists, such as Subjectivity
Dictionary~\cite{Wilson:05}, Maryland Polarity Set~\cite{Mohammad:09},
and \textsc{SentiWordNet} of~\citet{Esuli:06c}.  Moreover, the authors
also demonstrated that their method could be successfully applied to
other topics and genres, hypothesizing that lexicon-based approaches
were in general more amenable to domain shifts than traditional
supervised machine-learning techniques.


It is therefore not surprising that lexicon-based systems have also
quickly found their way into the sentiment analysis of social media:
For example, one such approach, explicitly tailored to Twitter
specifics, was proposed by~\citet{Musto:14}, who examined four
different ways to compute the overall polarity scores of microblogs:
\emph{basic}, \emph{normalized}, \emph{emphasized}, and
\emph{normalized-emphasized}.  
In each of these methods, the authors first split the input message
into a list of \emph{micro-phrases} based on the occurrence of
punctuation marks and conjunctions.  Afterwards, they calculated the
polarity score for each of these segments and finally estimated the
overall polarity of the whole tweet by uniting the scores of its
micro-phrases.  \citeauthor{Musto:14} obtained their best results
(58.99\% accuracy on the SemEval-2013 dataset) with the
normalized-emphasized approach, in which they averaged the polarity
scores of segments' tokens, boosting these values by 50\% for
adjectives, adverbs, nouns, and verbs; and computed the final overall
polarity of the microblog by taking the sum of all micro-phrase
scores.


Another Twitter-aware system was presented by~\citet{Jurek:15}, who
computed the negative and positive polarity of a message ($F_p$ and
$F_n$ respectively) as:
\begin{align}
  \small
  \begin{split}
  F_P &= \min\left(\frac{A_P}{2 - \log(3.5\times W_P + I_P)}, 100\right),\\
  F_N &= \max\left(\frac{A_N}{2 - \log(3.5\times W_N + I_N)}, -100\right);\label{cgsa:eq:jurek}
  \end{split}
\end{align}%
where $A_P$ and $A_N$ represent the average scores of positive and
negative lexicon terms found in the tweet; $W_P$ and $W_N$ stand for
the raw counts of polar tokens; and $I_P$ and $I_N$ denote the number
of intensifiers preceding these words.  In addition to that, before
estimating the average values, the authors modified the polarity
scores $s_w$ of all negated words $w$ using the following rule:
\begin{align*}
  \small%
neg(s_w) =
    \begin{cases}
        \min\left(\frac{s_w - 100}{2}, -10\right) & \text{if } s_w > 0,\\
        \max\left(\frac{s_w + 100}{2}, 10\right), & \text{if } s_w < 0.
    \end{cases}
\end{align*}%
Furthermore, besides computing the polarity scores $F_p$ and $F_n$,
\citeauthor{Jurek:15} also determined the subjectivity degree of the
message by replacing the $A_P$ and $A_N$ terms in
Equation~\ref{cgsa:eq:jurek} with the average of conditional
probabilities of the tweet being subjective given the occurrences of
the respective polar terms.\footnote{These probabilities were
  calculated automatically on the noisily labeled data set
  of~\citet{Go:09}.}  The authors considered a microblog as neutral if
its absolute polarity was less than 25, and the subjectivity value was
not greater than 0.5.  Otherwise, they assigned a positive or negative
label to this message depending on the sign of the polarity score.
With this approach, \citeauthor{Jurek:15} achieved an accuracy
of~77.3\% on the manually annotated subset of the \citeauthor{Go:09}'s
corpus and reached 74.2\% on the IMDB review dataset~\cite{Maas:11}.

Finally, \citet{Kolchyna:15} also explored two different ways of
computing the overall polarity of a microblog:
\begin{inparaenum}[(i)]
\item by simply averaging the scores of all lexicon terms found in the
  message and
\item by taking a signed logarithm of this average:
\end{inparaenum}
\begin{equation*}
  \text{Score}_{\log} =
  \begin{cases}
    \text{sign}(\text{Score}_{\text{AVG}})\log_{10}(|\text{Score}_{\text{AVG}}|) & %
    \text{if |Score}_{\text{AVG}}| > 0.1,\\
    0, & \text{otherwise};
  \end{cases}
\end{equation*}%
The authors determined the final polarity of a tweet by using
$k$-means clustering, which utilized both of the above polarity values
as features.  They showed that the logarithmic strategy performed
better than the simple average solution, yielding an accuracy of
61.74\% on the SemEval-2013 corpus~\cite{Nakov:13}.


As it was unclear how each of these methods would perform on PotTS and
SB10k, we reimplemented the approaches of~\citet{Hu:04} (as a
relatively simple baseline), \citet{Taboada:11}, \citet{Musto:14},
\citet{Jurek:15}, and \citet{Kolchyna:15}, and applied these systems
to the test sets of these corpora.

Based on our comparison in Chapter~\ref{chap:snt:lex}, we chose the
Zurich Polarity List~\cite{Clematide:10} as the primary sentiment
lexicon for the tested methods.  However, a significant drawback of
this resource is that most of its entries have uniform weights, with
their polarity scores being either 0.7 or 1.  We decided to keep the
original values as is, and only multiplied the scores of negative
terms by -1, since all of the tested approaches presupposed different
signs for the terms with opposite semantic orientations.\footnote{We
  will investigate the impact of other lexicons with presumably better
  scoring later in Section~\ref{cgsa:subsec:eval:lexicons}.}
Moreover, because some analyzers (\eg{} \citeauthor{Taboada:11}
[\citeyear{Taboada:11}] and \citeauthor{Musto:14}
[\citeyear{Musto:14}]) required part-of-speech tags of lexicon
entries, we automatically tagged all terms in this polarity list with
the \textsc{TreeTagger}~\cite{Schmid:95}, choosing the most probable
part-of-speech tag for each entry and also using the tag sequences
whose probabilities were at least two times lower than the likelihood
of the best assignment, duplicating the lexicon entries in the second
case.

Furthermore, since all of the systems except for that
of~\citet{Kolchyna:15} by default returned continuous real values, but
our evaluation required discrete polarity labels (\emph{positive},
\emph{negative}, or \emph{neutral}), we discretized the results of
these approaches using the following simple procedure: We first
determined the optimal threshold values for each particular polar
class on the training and development sets,\footnote{Since none of the
  methods required training or involved any sophisticated
  hyper-parameters, we used both training and development data to
  optimize the threshold scores.} and then derived polarity labels for
the test messages by comparing their predicted SO scores with these
thresholds.  To achieve the former goal (\ie{} to find the optimal
thresholds), we exhaustively searched through all unique polarity
values assigned to the training and development instances and checked
whether using these values as a boundary between two adjacent polarity
classes (sorted in ascending order of their positivity) would increase
the overall macro-\F{} on the training and development sets.

The final results of this evaluation are shown in
Table~\ref{snt-cgsa:tbl:lex-res}.

\begin{table}[h]
  \begin{center}
    \bgroup\setlength\tabcolsep{0.1\tabcolsep}\scriptsize
    \begin{tabular}{p{0.162\columnwidth} 
        *{9}{>{\centering\arraybackslash}p{0.074\columnwidth}} 
        *{2}{>{\centering\arraybackslash}p{0.068\columnwidth}}} 
      \toprule
      \multirow{2}*{\bfseries Method} & %
      \multicolumn{3}{c}{\bfseries Positive} & %
      \multicolumn{3}{c}{\bfseries Negative} & %
      \multicolumn{3}{c}{\bfseries Neutral} & %
      \multirow{2}{0.068\columnwidth}{\bfseries\centering Macro\newline \F{}$^{+/-}$} & %
      \multirow{2}{0.068\columnwidth}{\bfseries\centering Micro\newline \F{}}\\
      \cmidrule(lr){2-4}\cmidrule(lr){5-7}\cmidrule(lr){8-10}

      & Precision & Recall & \F{} & %
      Precision & Recall & \F{} & %
      Precision & Recall & \F{} & & \\\midrule

      \multicolumn{12}{c}{\cellcolor{cellcolor}PotTS}\\

      %
      %
      %

      HL & 0.75 & \textbf{0.76} & \textbf{0.76} & %
       0.53 & 0.43 & 0.47 & %
       0.67 & 0.73 & 0.69 & %
       \textbf{0.615} & \textbf{0.685}\\

       %
       %
       %
      TBD & \textbf{0.77} & 0.71 & 0.74 & %
        \textbf{0.54} & 0.39 & 0.45 & %
        0.63 & 0.77 & 0.69 & %
        0.597 & 0.674\\

       %
       %
       %

       MST & 0.75 & 0.72 & 0.74 & %
        0.48 & \textbf{0.47} & \textbf{0.48} & %
        \textbf{0.68} & 0.72 & 0.7 & %
        0.606 & 0.675\\

       %
       %
       %

      JRK & 0.6 & 0.31 & 0.41 & %
       0.42 & 0.2 & 0.27 & %
       0.43 & 0.8 & 0.56 & %
       0.339 & 0.467\\

       %
       %
       %

      KLCH & 0.71 & 0.72 & 0.71 & %
       0.34 & 0.17 & 0.22 & %
       0.66 & \textbf{0.82} & \textbf{0.73} & %
       0.468 & 0.651\\

      \multicolumn{12}{c}{\cellcolor{cellcolor}SB10k}\\

      HL & \textbf{0.49} & \textbf{0.62} & \textbf{0.55} & %
        0.27 & 0.33 & 0.3 & %
        \textbf{0.73} & 0.62 & 0.67 & %
        \textbf{0.421} & 0.577\\


      TBD & 0.48 & 0.6 & 0.53 & %
        0.24 & 0.27 & 0.25 & %
        0.72 & 0.63 & 0.67 & %
        0.393 & 0.57\\

      MST & 0.45 & 0.49 & 0.47 & %
        0.29 & \textbf{0.35} & \textbf{0.32} & %
        0.7 & 0.64 & 0.67 & %
        0.395 & 0.568\\

      JRK & 0.41 & 0.39 & 0.4 & %
        \textbf{0.36} & 0.26 & 0.3 & %
        0.69 & 0.75 & 0.72 & %
        0.351 & 0.592\\

      KLCH & 0.39 & 0.22 & 0.28 & %
        0.34 & 0.13 & 0.19 & %
        0.66 & \textbf{0.86} & \textbf{0.75} & %
        0.235 & \textbf{0.606}\\\bottomrule
    \end{tabular}
    \egroup{}
    \caption[Results of lexicon-based MLSA methods]{
      Results of lexicon-based MLSA methods\\
      {\small HL~--~\citet{Hu:04}, TBD~--~\citet{Taboada:11}, MST~--~\citet{Musto:14},
        JRK~--~\citet{Jurek:15}, KLCH~--~\citet{Kolchyna:15}}}\label{snt-cgsa:tbl:lex-res}
  \end{center}
\end{table}

As we can see, the performance of the tested methods significantly
varies across different polarity classes, but follows more or less the
same pattern on both datasets: For example, the most simple approach
of~\citet{Hu:04} achieves surprisingly good quality at predicting
positive tweets, showing the highest recall and \F{}-measure on the
PotTS corpus and yielding the best overall scores for this polarity
class on the SB10k set.  Moreover, on the latter data, it also
outperforms all other systems in terms of the precision of neutral
microblogs.  Combined with its generally good results on other
metrics, this classifier attains the highest macro-averaged
\F{}-result for all classes and sets up a new benchmark for the
micro-\F{} on the PotTS test set.

The approach of~\citet{Taboada:11}, which can be viewed as an
extension of the previous method, only surpasses the HL classifier
w.r.t.\ the precision of positive and negative messages, but still
loses more than 0.02 macro-\F{} due to a lower recall of the neutral
class.  A better performance in this regard is shown by the analyzer
of~\citet{Musto:14}, which shows a fairly strong recall of negative
tweets, which in turn leads to the best \F{}-score for this polarity.
Unfortunately, since this semantic orientation is the most
underrepresented one in both corpora, this success is not reflected in
the overall statistics: Although this methods ranks second in terms of
the macro-averaged \F{}, it lags behind its competitors with regard to
the micro-averaged value on the SB10k corpus.

Finally, the system of~\citet{Kolchyna:15} shows very strong recall
and \F{}-scores for the neutral class on both sets and also achieves
the best accuracy (0.606) on the SB10k data, but its quality for the
remaining two polarities is fairly suboptimal, with the \F{}-scores
for these semantic orientations ranking last or second to last in both
cases.

\subsection{Polarity-Changing Factors}\label{subsec:cgsa:lex-methods:pol-change}

Since the analysis of context factors is commonly considered to be one
of the most important components of any lexicon-based MLSA system, and
because the method with the simplest approach to this task achieved
surprisingly good results, outperforming other more sophisticated
competitors, we decided to recheck the utility of this module for all
classifiers.  In order to do so, we successively deactivated, one by
one, parts of the classifiers that analyzed the surrounding context of
polar terms and recomputed the \F{}-scores of all systems after these
changes.

\begin{table}[h]
  \begin{center}
    \bgroup\setlength\tabcolsep{0.1\tabcolsep}\scriptsize
    \begin{tabular}{p{0.16\columnwidth} 
        *{10}{>{\centering\arraybackslash}p{0.082\columnwidth}}}
      \toprule
      \multirow{2}{0.15\columnwidth}{%
      \bfseries Polarity-Changing\newline Factors} & %
      \multicolumn{10}{c}{\bfseries System Scores}\\
      & \multicolumn{2}{c}{\bfseries HL} & \multicolumn{2}{c}{\bfseries TBD} %
      & \multicolumn{2}{c}{\bfseries MST} %
      & \multicolumn{2}{c}{\bfseries JRK} & \multicolumn{2}{c}{\bfseries KLCH}\\%
      \cmidrule(lr){2-3}\cmidrule(lr){4-5}\cmidrule(lr){6-7} %
      \cmidrule(lr){8-9}\cmidrule(lr){10-11}

      & Macro\newline \F{}$^{+/-}$ & Micro\newline \F{} %
      & Macro\newline \F{}$^{+/-}$ & Micro\newline \F{} %
      & Macro\newline \F{}$^{+/-}$ & Micro\newline \F{} %
      & Macro\newline \F{}$^{+/-}$ & Micro\newline \F{} %
      & Macro\newline \F{}$^{+/-}$ & Micro\newline \F{}\\\midrule

      \multicolumn{11}{c}{\cellcolor{cellcolor}PotTS}\\
      All & 0.615 & 0.685 & 0.593 & 0.671 & 0.606 & 0.675 %
      & 0.339 & 0.467 & 0.468 & 0.651\\

      --Negation & 0.622 & \textbf{0.691} & 0.596 & 0.672 & \textbf{0.641} & %
       0.7 & 0.357 & 0.473 & 0.298 & 0.463\\


      --Intensification & \NA{} & \NA{} & 0.595 & 0.672 & \NA{} &  %
      \NA{} & 0.339 & 0.467 & \NA{} & \NA{}\\

      --Other Modifiers & \NA{} & \NA{} & 0.613 & 0.684 & \NA{} &  %
      \NA{} & \NA{} & \NA{} & \NA{} & \NA{}\\

      \multicolumn{11}{c}{\cellcolor{cellcolor}SB10k}\\

      All & \textbf{0.421} & 0.577 & 0.392 & 0.569 & 0.395 & 0.568 %
      & 0.351 & 0.592 & 0.235 & 0.606\\

      --Negation & 0.415 & 0.576 & 0.395 & 0.572 & 0.381 & %
        0.559 & 0.316 & 0.586 & 0.218 & \textbf{0.609}\\


      --Intensification & \NA{} & \NA{} & 0.4 & 0.576 & \NA{} &  %
      \NA{} & 0.352 & 0.59 & \NA{} & \NA{}\\

      --Other Modifiers & \NA{} & \NA{} & 0.406 & 0.566 & \NA{} &  %
      \NA{} & \NA{} & \NA{} & \NA{} & \NA{}\\\bottomrule
    \end{tabular}
    \egroup{}
    \caption{Effect of polarity-changing factors on lexicon-based MLSA
      methods}\label{snt-cgsa:tbl:lex-res-ablation}
  \end{center}
\end{table}

As we can see from the results in
Table~\ref{snt-cgsa:tbl:lex-res-ablation}, various methods respond in
different ways to this ablation: For example, the scores of the
\citeauthor{Hu:04} system improve on the PotTS corpus, but degrade on
the SB10k dataset after switching off the negation handling.  The same
situation can also be observed with the analyzers of \citet{Musto:14}
and~\citet{Jurek:15}.  The classifier of~\citet{Taboada:11}, however,
benefits from this deactivation in both cases, and the approach
of~\citet{Kolchyna:15} vice versa shows a performance drop on either
dataset with the only exception being the micro-averaged \F{} on the
SB10k data, which unexpectedly improves from 0.606 to 0.609.

As to the intensification handling, we can see that only two
approaches (TBD and JRK) have this component at all.  As in the
previous case, the Taboada system profits from its deactivation, with
the macro- and micro-averaged \F{}-scores going up by 0.002 on PotTS
and 0.008 on the SB10k corpus.  A more varied situation is observed
with the analyzer of~\citet{Jurek:15}, whose PotTS results are
virtually unaffected by these changes, but the macro-averaged \F{}
slightly increases and the micro-averaged score slightly decreases on
the \citeauthor{Cieliebak:17}'s dataset.

Finally, ``other modifiers'' (such as irrealis and interrogative
clauses) only play a role as a polarity-changing factor in the system
of~\citet{Taboada:11} and, as we can see from the figures, do there
rather more harm than good: deactivating this part boosts the
macro-averaged \F{}-scores on PotTS and SB10k by~0.02 and~0.014
respectively.  At the same time, the micro-averaged result of this
system climbs up from 0.671 to 0.684 on the former dataset, but drops
from~0.569 to~0.566 on the latter corpus.

\subsection{Error Analysis}\label{subsec:cgsa:lex-methods:err-analysis}

In order to get a better intuition about the strengths and weaknesses
of each particular classifier, we additionally collected a set of
errors that were specific to only one of above the systems and will
discuss some of these cases here in detail.

The first such error, which was made by the system
of~\citet{Taboada:11}, is shown in
Example~\ref{snt:cgsa:exmp:taboada-error-0}.  Here, a strongly
positive tweet describing one's excitement about a technical report
was erroneously classified as neutral despite the presence of the
prototypical positive term ``gut'' (\emph{good}) in its superlative
form ``beste'' (\emph{best}).  Unfortunately, it is the degree of
comparison which becomes fatal in this case: According to the
implementation of~\citet{Taboada:11}, any superlative adjective has to
be preceded by the definite article and a verb in order to be
considered as a polar term for the final SO computation. Although the
adjective ``beste'' (\emph{best}) can fulfill the first criterion (it
immediately follows the determiner ``der'' [\emph{the}]), the lack of
the preceding verb nullifies its effect.

\begin{example}[An Error Made by the System of~\citeauthor{Taboada:11}]\label{snt:cgsa:exmp:taboada-error-0}
  \noindent\textup{\bfseries\textcolor{darkred}{Tweet:}} {\upshape Der beste Microsoft Knowledgebase-Artikel, den ich je gelesen habe.}\\
  \noindent The best Microsoft-Knowledgebase article I've ever read.\\[\exampleSep]
  \noindent\textup{\bfseries\textcolor{darkred}{Gold Label:}}\hspace*{4.3em}\textbf{%
    \upshape\textcolor{green3}{positive}}\\
 \noindent\textup{\bfseries\textcolor{darkred}{Predicted Label:}}\hspace*{2em}\textbf{%
    \upshape\textcolor{black}{neutral*}}
\end{example}

\noindent Another error of this method is shown in
Example~\ref{snt:cgsa:exmp:taboada-error-1}.  This time, the presence
of the colloquial term ``verarschen'' (\emph{to hoax}) suggests that
the tweet at hand is negative.  Alas, the occurrence of another verb
(``wollt'' [\emph{wanna}]) is interpreted as an irrealis clue, which
prevents further SO computation and leads to a zero score to the whole
message.\footnote{Please note that the occurrence of the question mark
  does not affect the sentiment score because ``?!'' is not included
  in the list of valid punctuation marks in the original
  implementation.}

\begin{example}[An Error Made by the System of~\citeauthor{Taboada:11}]\label{snt:cgsa:exmp:taboada-error-1}
  \noindent\textup{\bfseries\textcolor{darkred}{Tweet:}} {\upshape Die Konklave w\"ahlt den Papst und dann sagen sie Gott war es --- Wollt ihr mich verarschen ?!}\\
  \noindent The conclave elects the Pope and then they say it was God
  --- do you wanna hoax me ?!\\[\exampleSep]
  \noindent\textup{\bfseries\textcolor{darkred}{Gold Label:}}\hspace*{4.3em}\textbf{%
    \upshape\textcolor{midnightblue}{negative}}\\
 \noindent\textup{\bfseries\textcolor{darkred}{Predicted Label:}}\hspace*{2em}\textbf{%
    \upshape\textcolor{black}{neutral*}}
\end{example}

At this point, we already can see that the main flaws of the TBD
approach apparently stem from its overly coarse rules, which, in
addition, are not always valid in German, whose word order is
significantly laxer than English syntax.\footnote{In order to check
  this claim, we tried to temporarily deactivate the above two
  heuristics (predicate check for superlative adjectives and irrealis
  blocking by model verbs) and recomputed the scores of this system,
  getting in both cases an improvement by almost one percent on either
  corpus.}

Returning back to our error analysis, let us look at another erroneous
case shown in Example~\ref{snt:cgsa:exmp:musto-error-0}. This time,
the system of~\citet{Musto:14} incorrectly assigned the neutral label
to a positive tweet even though the positive term ``gut''
(\emph{good}) again appears in this message.  As it turns out, the
occurrence of this word is still insufficient for the classifier to
predict the positive class although this term has the highest possible
positive score in the lexicon (1.0), which is additionally boosted by
a factor of 1.5, since this word is an adjective.  But the crushing
factor in this case is the length of the tweet: since this approach
relies on the average SO-score for all words in a sentence, the value
1.5 of the only positive term is divided by 7 (the length of the
sentence) and drops down to 0.214, which is below the threshold for
the positive class (0.267).

\begin{example}[An Error Made by the System of~\citeauthor{Musto:14}]\label{snt:cgsa:exmp:musto-error-0}
  \noindent\textup{\bfseries\textcolor{darkred}{Tweet:}} {\upshape
    Mensch Meier, Mensch Meier! Das sieht gut aus f\"ur
    die \%User:}\\
  \noindent Gosh Meier, Gosh Meier! It looks good for
  the \%User:\\[\exampleSep]
  \noindent\textup{\bfseries\textcolor{darkred}{Gold Label:}}\hspace*{4.3em}\textbf{%
    \upshape\textcolor{green3}{positive}}\\
 \noindent\textup{\bfseries\textcolor{darkred}{Predicted Label:}}\hspace*{2em}\textbf{%
    \upshape\textcolor{black}{neutral*}}
\end{example}

\noindent As it turns out, this kind of mistakes is by far the most
common type of errors characteristic to the MST system.  Further
examples of such incorrect decisions are provided in
Example~\ref{snt:cgsa:exmp:musto-error-1}:
\begin{example}[Errors Made by the System of~\citeauthor{Musto:14}]\label{snt:cgsa:exmp:musto-error-1}
  \noindent\textup{\bfseries\textcolor{darkred}{Tweet:}} {\upshape
    Der \%User tut echt geile musik machen. Nichts mit Boyband hier.}\\
  \noindent The \%User is making really great music. Nothing with Boyband here.\\[\exampleSep]
  \noindent\textup{\bfseries\textcolor{darkred}{Gold Label:}}\hspace*{4.3em}\textbf{%
    \upshape\textcolor{green3}{positive}}\\
 \noindent\textup{\bfseries\textcolor{darkred}{Predicted Label:}}\hspace*{2em}\textbf{%
    \upshape\textcolor{black}{neutral*}}\\[2\exampleSep]
  \noindent\textup{\bfseries\textcolor{darkred}{Tweet:}} {\upshape
    Diese S5E5 Episode mit den Zug\"uberfall war wieder genial! BreakingBad}\\
  \noindent This S5E5 episode with train robbery was brilliant again! BreakingBad\\[\exampleSep]
  \noindent\textup{\bfseries\textcolor{darkred}{Gold Label:}}\hspace*{4.3em}\textbf{%
    \upshape\textcolor{green3}{positive}}\\
 \noindent\textup{\bfseries\textcolor{darkred}{Predicted Label:}}\hspace*{2em}\textbf{%
    \upshape\textcolor{black}{neutral*}}
\end{example}

A different kind of problems is experienced by the approach
of~\citet{Jurek:15}, which apparently has difficulties with correctly
predicting the positive class.  A deeper analysis of its
misclassifications revealed that the reason for it is relatively
simple: Because this classifier uses conditional probabilities of
polar terms instead of their original lexicon scores, and we have
estimated these probabilities on the noisily labeled German Twitter
Snapshot, which was extremely biased towards the positive class (see
Table~\ref{snt-cgsa:tbl:corp-dist}), all positive lexicon entries
received extremely high scores.  As a consequence, even a single
occurrence of a positive term in a message outweighed the effect of
any negative expressions, even if they were more frequent in that
tweet.  This is, for instance, the case in
Example~\ref{snt:cgsa:exmp:jurek-error-0} where the score of the
(questionable) positive expression ``Normal'' (\emph{normally}) is
greater than the absolute sum of two negative values for the terms
``sich beschweren'' (\emph{to complain}) and ``ekelhaft''
(\emph{disgusting}).

\begin{example}[An Error Made by the System of~\citeauthor{Jurek:15}]\label{snt:cgsa:exmp:jurek-error-0}
  \noindent\textup{\bfseries\textcolor{darkred}{Tweet:}} {\upshape
    Normal bin ich ja nicht der mensch dwer sich beschwert wegen dem
    essen aber diese Pizza von Joeys\ldots{} boah wie ekelhaft}\\
  \noindent Normally I'm not a person who complains about food but
  this pizza from Joeys\ldots{} Boah it's so disgusting\\[0.65em]
  \noindent\textup{\bfseries\textcolor{darkred}{Gold Label:}}\hspace*{4.3em}\textbf{%
    \upshape\textcolor{midnightblue}{negative}}\\
 \noindent\textup{\bfseries\textcolor{darkred}{Predicted Label:}}\hspace*{2em}\textbf{%
    \upshape\textcolor{green3}{positive*}}\\
\end{example}

The same problem also afflicts the system of~\citeauthor{Kolchyna:15},
whose error example is given in~\ref{snt:cgsa:exmp:kolchyna-error-0}.
In contrast to the previous approaches, which mainly rely on manually
designed heuristic rules, this method makes its decisions using a
trained $k$-NN classifier.  Nevertheless, its prediction in the
provided case is still incorrect as it evidently confuses the positive
class with the neutral polarity.

\begin{example}[An Error Made by the System of~\citeauthor{Kolchyna:15}]\label{snt:cgsa:exmp:kolchyna-error-0}
  \noindent\textup{\bfseries\textcolor{darkred}{Tweet:}} {\upshape das
    H\"ort sich echt Super an! \%PosSmiley macht sami nicht auch so
    ein Video? Noah s\"usse beste Freunde! \heart{} \%User isilie
    saminator}\\
  \noindent It sounds really fantastic! \%PosSmiley won't sami also
  make such a video? Noah's sweet best friends! \heart{} \%User isilie
  saminator\\[0.65em]
  \noindent\textup{\bfseries\textcolor{darkred}{Gold Label:}}\hspace*{4.3em}\textbf{%
    \upshape\textcolor{green3}{positive}}\\
 \noindent\textup{\bfseries\textcolor{darkred}{Predicted Label:}}\hspace*{2em}\textbf{%
    \upshape\textcolor{black}{neutral*}}\\
\end{example}

\noindent In order to understand the reason for this
misclassification, we first looked at the initial SO scores computed
by the Kolchyna analyzer.  As it turned out, both values that were
used by the internal $k$-NN predictor of this system as features (the
average SO score of all polar terms found in the message and the
logarithm of this average) were relatively high, amounting to 33.42
and 2.52 respectively.  But a closer look at the selected nearest
neighbors revealed that even despite such high SO values, top three of
the closest neighbors of this microblog were indeed neutral, as we can
see from the list below:

\begin{enumerate}
\item \textcolor{darkred}{\bfseries Tweet:} ``Not in my backyard''
  -Mentalit\"at dt. Politik: ``N\"achster Castor geht wohl doch nach
  Gorleben\ldots{} -\%Link antiatom''\\
  \textit{``Not in my backyard'' -Mentality of German politics: ``Next Castor will probably still got to Gorleben\ldots{} -\%Link antiatom''}\\
  \textcolor{darkred}{\bfseries Label:}~\textcolor{black}{neutral}\\
  \textcolor{darkred}{\bfseries Distance:}~$\expnumber{6.83}{-03}$;

\item \textcolor{darkred}{\bfseries Tweet:} Kanzlerin im Google-Hangout: ``Die Technik soll sich mal bem\"uhen''\\
  \textit{Chancellor in Google-Hangout: ``The technology should make an effort''}\\
  \textcolor{darkred}{\bfseries Label:}~\textcolor{black}{neutral}\\
  \textcolor{darkred}{\bfseries Distance:}~$\expnumber{1.6}{-02}$;\label{snt:cgsa:exmp:kolchyna-error-0.1}

\item \textcolor{darkred}{\bfseries Tweet:} Kanzlerin im
  Google-Hangout: ``Die Technik soll sich mal
  bem\"uhen''\\ \textit{Chancellor in Google-Hangout: ``The technology
    should make an effort''}\\ \textcolor{darkred}{\bfseries
    Label:}~\textcolor{black}{neutral}\\ \textcolor{darkred}{\bfseries
    Distance:}~$\expnumber{1.6}{-02}$;\footnote{Please note that this
    tweet is not a duplicate of the previous microblog, but a
    different message (with its distinct message id), which, however,
    has the same wording.}\label{snt:cgsa:exmp:kolchyna-error-0.2}

\item \textcolor{darkred}{\bfseries Tweet:} W\"unsche mir ein Format wie zdflogin auch f\"ur das \%User. Viele Themen, klare Aussagen. Sch\"ones Special \%User zur Landtagswahl! \%PosSmiley\\
  \textit{Wish \%User had a format like zdflogin. Many topics, clear statements. Nice Special \%User zur Landtagswahl! \%PosSmiley}\\
  \textcolor{darkred}{\bfseries Label:}~\textcolor{green3}{positive}\\
  \textcolor{darkred}{\bfseries Distance:}~$\expnumber{2.1}{-02}$;

\item \textcolor{darkred}{\bfseries Tweet:} Ich bin ja so gespannt ob
  die FDP im September erst den Zahn\"arzten und dann den Apothekern
  mit Geschenken dankt, oder anders rum\ldots\\ \textit{I'm so curious
    whether FDP will first give gifts to dentists and then to
    pharmacists in September, or whether it'll be vice
    versa}\\ \textcolor{darkred}{\bfseries
    Label:}~\textcolor{green3}{positive}\\ \textcolor{darkred}{\bfseries
    Distance:}~$\expnumber{4.12}{-02}$;
\end{enumerate}

\noindent Even more surprisingly, the SO scores of the neighboring
neutral instances were indeed also relatively high: In the first
microblog, for example, the system recognized two polar terms: the
English word ``Not'', which was confused with the German term ``Not''
(\emph{distress}), and ``n\"achster'' (\emph{next}).  Another polar
expression (``sich bem\"uhen'' [\emph{to make an effort}]) was found
in messages~\ref{snt:cgsa:exmp:kolchyna-error-0.1}
and~\ref{snt:cgsa:exmp:kolchyna-error-0.2}.  Although two of these
terms (``Not'' and ``sich bem\"uhen'') had a negative label in the
sentiment lexicon, their conditional probability of being associated
with the positive class was more than ten times bigger than the chance
to appear in a negative microblog (according to the computed
statistics). As a consequence of this positive probability bias, many
neutral tweets from the training set ended up in close vicinity to
actual positive examples.

As we can see, lexicon-based methods experience various kinds of
problems with predicting the polarity of short casually written
microblogs: Some of these systems apply rules that are too specific to
a particular language and domain, so that they do not generalize well
to German tweets; others rely on noisy statistics, which might be
extraordinarily skewed towards just one polarity.  Now, we should
check whether other approaches to the message-level sentiment analysis
(which rely on completely different principles and paradigms) will
also be susceptible to these kinds of errors.

\section{Machine-Learning Methods}\label{sec:cgsa:ml-based}

Despite their immense popularity, linguistic plausibility, and
simplicity to implement, lexicon-based approaches often have been
criticized for the rigidness of their classification\footnote{Since
  these systems only rely on the precomputed weights of lexicon
  entries, considering these coefficients as constant, their decision
  boundaries frequently appear to be suboptimal as many terms might
  have different polarity and intensity values depending on the
  domain~\cite[see][]{Eisenstein:17,Yang:17}.} and the inability to
incorporate additional, non-lexical attributes into their final
decisions.  Moreover, as noted by~\citet{Pang:02} and also confirmed
empirically by~\citet{Riloff:03} and \citet{Gamon:04}, many linguistic
expressions that actually correlate with the subjectivity and polarity
of a sentence (\eg{} exclamation marks or spelling variations) are
very unlikely to be included into a sentiment lexicon even by a human
expert.  As a consequence of this, with the emergence of manually
annotated corpora, lexicon-based systems have been gradually
superseded by supervised machine-learning techniques.

One of the first steps in this direction was taken
by~\citet{Wiebe:99}, who used a Na\"{\i}ve Bayes classifier to
differentiate between subjective and objective statements.  Using
binary features that reflected the presence of a pronoun, an
adjective, a cardinal number, or a modal verb in the analyzed
sentence, the authors achieved an accuracy of~72.17\% on the two-class
prediction task (differentiating between positive and negative
classes), outperforming the majority class baseline by more than 20\%.
An even better result (81.5\%) could be reached when the dataset was
restricted only to the examples with the most confident annotation.

Inspired by this success,~\citet{Yu:03} presented a more elaborated
system in which they first distinguished between subjective and
objective documents, then differentiated between polar and neutral
sentences, and, finally, determined the polarity of that clauses.  As
in the previous case, the authors used a Na\"{\i}ve Bayes predictor
for the document-level task, reaching a remarkable \F-score of~0.96 on
this objective; and applied an ensemble of NB systems to predict the
subjectivity of single sentences.  To determine the semantic
orientation of subjective clauses, \citeauthor{Yu:03} averaged the
polarity scores of their tokens, obtaining these scores from an
automatically constructed sentiment lexicon~\cite{Hatzivassi:97}.
This way, they attained an accuracy of~91\% on a set of 38 sentences
that had a perfect inter-annotator agreement.


In order to check the effectiveness of the Na\"{\i}ve Bayes approach,
\citet{Pang:02} compared the results of NB, MaxEnt, and SVM systems on
the movie review classification task, trying to predict whether a
review was perceived as thumbs up or thumbs down.  In contrast to the
previous works, they found the SVM classifier working best for this
objective, yielding 82.9\% accuracy when used with unigram features
only.  This conclusion paved the way for the following triumph of the
support-vector approach, which was dominating the whole sentiment
research field for almost a decade ever since.  For example,
\citet{Gamon:04} also trained an SVM predictor using a set of
linguistic and surface-level features (including part-of-speech
trigrams, context-free phrase-structure patterns, and part-of-speech
information coupled with syntactic relations) to distinguish between
positive and negative customer feedback, achieving 77.5\% accuracy and
$\approx$0.77~\F{} by using only top 2,000 attributes that had the
highest log-likelihood ratio with the target class.
Furthermore, \citet{Pang:05} addressed the problem of multi-class
rating, attempting to predict the number of stars assigned to a
review.  For this purpose, they compared three different SVM types:
\begin{inparaenum}[(i)]
\item one-versus-all SVM (OVA-SVM),
\item SVM regression,
\item and OVA-SVM with metric labeling;
\end{inparaenum}
getting their best results ($\approx$52\%~accuracy) with the last
option.
Finally, \citet{Ng:06} proposed a multi-stage SVM system, in which
they first classified whether the given text was a review or not and
then tried to predict its polarity.  Due to a better usage of
higher-order $n$-grams (where, instead of na\"{\i}vely considering all
token sequences up to length $n$ as new features, the authors only
took 5,000 most useful ones), \citet{Ng:06} even improved the state of
the art on the \citeauthor{Pang:04}'s corpus, boosting the
classification accuracy from~87.1 to~90.5\%.

But a real game change in the MLSA research field happened with the
introduction of the SemEval shared task on sentiment analysis in
Twitter~\cite{Nakov:13}.  Starting from its inaugural run
in~\citeyear{Nakov:13}, this competition has rapidly caught the
attention of the broader NLP community and has been rerun five times,
attracting more than 40 active participants every year.

It is not surprising that the first winning systems in this task
closely followed in the footsteps of the advances in the general
opinion mining at that time.  For example, the two top-scoring
submissions in the initial iteration~\cite{Mohammad:13,Guenther:13}
both relied on the SVM algorithm: The first of these approaches, an
analyzer developed by~\citet{Mohammad:13}, was the absolute winner of
SemEval~2013, scoring impressive 0.69 macro-averaged two-class \F{} on
the provided Twitter corpus.  The key to the success of this method
was an extensive set of linguistic features devised by the authors,
which included character and token $n$-grams, Brown
clusters~\cite{Brown:92}, statistics on part-of-speech tags,
punctuation marks, elongated words etc.  But the most useful type of
attributes according to the feature ablation test turned out to be the
features that reflected information from various sentiment
lexicons. In particular, depending on the type of the polarity list
from which such information was extracted, \citeauthor{Mohammad:13}
introduced two types of lexicon attributes: \emph{manual} and
\emph{automatic} ones.  The former group was computed with the help of
the NRC emotion lexicon~\cite{Mohammad:13a}, MPQA polarity
list~\cite{Wilson:05}, and Bing Liu's manually compiled polarity
set~\cite{Hu:04}.  For each of these resources and for each of the
non-neutral polarity classes (positive and negative), the authors
estimated the total sum of the lexicon scores for all message tokens
and also separately calculated these statistics for each particular
part-of-speech tag, considering them as additional attributes.
Automatic features were obtained using the Sentiment140 and Hashtag
Sentiment Base polarity lists~\cite{Kiritchenko:14}.  Again, for each
of these lexicons, for each of the two polarity classes, the authors
produced four features representing the number of tokens with non-zero
scores, the sum and the maximum of all respective lexicon values for
all words, and the score of the last term in the tweet.  These two
feature groups (manual and automatic lexicon attributes) improved the
macro-averaged \F{}$^{+/-}$-score by almost five percent,
outperforming in this regard all other traits.

Another notable submission, the system of~\citet{Guenther:13}, also
relied on a linear SVM predictor with a rich set of features.
Like~\citet{Mohammad:13}, the authors used original and lemmatized
unigrams, word clusters, and lexicon features.  But in contrast to the
previous approach, this application utilized only one polarity
list---that of~\citet{Esuli:05}.  Partially due to this fact,
\citeauthor{Guenther:13} found the word clusters working best among
all features.  This method also yielded competitive results
(0.653~\F{}$^{+/-}$) on the message-level polarity task, attaining
second place in that year.

Later on, \citet{Guenther:14} further improved their results (from
0.653 to 0.691 two-class \F) by extending the original system with a
Twitter-aware tokenizer~\cite{Owoputi:13}, spelling normalization
module, and a significantly increased set of lexicon-based features.
In particular, instead of simply relying on
\textsc{SentiWordNet}~\cite{Esuli:05}, \citeauthor{Guenther:14}
applied a whole ensemble of various polarity lists including Liu's
opinion lexicon, MPQA subjectivity list, and TwittrAttr polarity
resource.  As mentioned by the authors, the last change was of
particular use to the classification accuracy, improving the
macro-\F{}$^{+/-}$ by almost four percent.

An even better score on this task, could be attained with the approach
of~\citet{Miura:14}, who also utilized a supervised ML classifier with
character and word $n$-grams, word clusters, disambiguated senses, and
lexicon scores of message tokens as features.  Similarly to the
systems of~\citet{Mohammad:13} and \citet{Guenther:14}, the authors
made heavy use of various kinds of polarity lists including
AFINN-111~\cite{Nielsen:11}, Liu's Opinion Lexicon~\cite{Hu:04},
General Inquirer~\cite{Stone:66}, MPQA Polarity List~\cite{Wiebe:05a},
NRC Hashtag and Sentiment140 Lexicon~\cite{Mohammad:13}, as well as
\textsc{SentiWordNet}~\cite{Esuli:06a}, additionally applying a whole
set of preprocessing steps such as spelling correction, part-of-speech
tagging with lemmatization, and a special weighting scheme for
underrepresented classes.  Due to these enhancements, combined with a
carefully tuned LogLinear classifier, \citet{Miura:14} were able to
boost the sentiment classification results on the SemEval~2014 test
set to~0.71~\F{}$^{+/-}$.

In order to see how this family of methods would perform on our
Twitter corpora, we have reimplemented the approaches of
\citet{Gamon:04}, \citet{Mohammad:13}, and \citet{Guenther:14} with
the following modifications: In the system of~\citet{Gamon:04}, we
used the available dependency analyses from the
\texttt{MateParser}~\cite{Bohnet:09} instead of constituency trees,
considering each node of the dependency tree as a syntactic
constituent and regarding the two-tuple
(\texttt{dependency-link-to-the-parent}, \texttt{node's-PoS-\-tag}) as
the name of that constituent (for example, a finite verb at the root
of the tree was mapped to the constituent (\texttt{--},
\texttt{VVFIN}), where \texttt{--} is the name of the root relation).
Furthermore, because the Brown clusters were not available for German,
we had to remove this attribute altogether from the feature sets of
\citeauthor{Mohammad:13}'s and \citeauthor{Guenther:14}'s methods.
Moreover, because the former system relied on two types of lexicon
attributes---manual and automatic ones, we used two polarity lists for
these approaches: the Zurich Sentiment Lexicon of~\citet{Clematide:10}
as a manual resource and our Linear Projection Lexicon, which was
introduced in Chapter~\ref{chap:snt:lex}, as an automatically
generated polarity list.  All remaining attributes and training
specifics were kept maximally close to their original descriptions.

\begin{table}[h]
  \begin{center}
    \bgroup\setlength\tabcolsep{0.1\tabcolsep}\scriptsize
    \begin{tabular}{p{0.162\columnwidth} 
        *{9}{>{\centering\arraybackslash}p{0.074\columnwidth}} 
        *{2}{>{\centering\arraybackslash}p{0.068\columnwidth}}} 
      \toprule
      \multirow{2}*{\bfseries Method} & %
      \multicolumn{3}{c}{\bfseries Positive} & %
      \multicolumn{3}{c}{\bfseries Negative} & %
      \multicolumn{3}{c}{\bfseries Neutral} & %
      \multirow{2}{0.068\columnwidth}{\bfseries\centering Macro\newline \F{}$^{+/-}$} & %
      \multirow{2}{0.068\columnwidth}{\bfseries\centering Micro\newline \F{}}\\
      \cmidrule(lr){2-4}\cmidrule(lr){5-7}\cmidrule(lr){8-10}

      & Precision & Recall & \F{} & %
      Precision & Recall & \F{} & %
      Precision & Recall & \F{} & & \\\midrule

      \multicolumn{12}{c}{\cellcolor{cellcolor}PotTS}\\

       %
       %
       %
       %

       GMN & 0.67 & 0.73 & 0.7 & %
       0.35 & 0.15 & 0.21 & %
       0.6 & 0.72 & 0.66 & %
       0.453 & 0.617\\

       %
       %
       %
       %
       MHM & \textbf{0.79} & 0.77 & \textbf{0.78} & %
       \textbf{0.58} & \textbf{0.56} & \textbf{0.57} & %
       \textbf{0.73} & \textbf{0.76} & \textbf{0.74} & %
       \textbf{0.674} & \textbf{0.727}\\

       %
       %
       %
       %

       GNT & 0.71 & \textbf{0.8} & 0.75 & %
       0.55 & 0.45 & 0.5 & %
       0.68 & 0.63 & 0.65 & %
       0.624 & 0.673\\

      \multicolumn{12}{c}{\cellcolor{cellcolor}SB10k}\\

       %
       %
       %
       %

      GMN & 0.65 & 0.45 & 0.53 & %
       0.38 & 0.08 & 0.13 & %
       0.72 & \textbf{0.93} & 0.81 & %
       0.329 & 0.699\\

       %
       %
       %
       %

       MHM & \textbf{0.71} & \textbf{0.65} & \textbf{0.68} & %
        \textbf{0.51} & \textbf{0.4} & \textbf{0.45} & %
        \textbf{0.8} & 0.87 & \textbf{0.84} & %
        \textbf{0.564} & \textbf{0.752}\\

       %
       %
       %
       %

       GNT & 0.67 & 0.62 & 0.64 & %
       0.44 & 0.28 & 0.34 & %
       0.78 & 0.87 & 0.82 & %
       0.491 & 0.724\\\bottomrule
    \end{tabular}
    \egroup{}
    \caption[Results of ML-based MLSA methods]{ Results of
      machine-learning--based MLSA methods\\ {\small
        GMN~--~\citet{Gamon:04}, MHM~--~\citet{Mohammad:13},
        GNT~--~\citet{Guenther:14}}}\label{snt-cgsa:tbl:ml-res}
  \end{center}
\end{table}

The results of our reimplementations are shown in
Table~\ref{snt-cgsa:tbl:ml-res}.  As we can see from the scores, the
system of~\citet{Mohammad:13} clearly dominates its competitors on
both corpora. This holds for all presented metrics except for the
recall of positive tweets on the PotTS dataset and neutral messages on
the SB10k data, where it is outperformed by the analyzers of
\citet{Guenther:14} and \citet{Gamon:04} respectively.  In any other
respect, however, the results of the MHM classifier are notably higher
than those of the GNT method, sometimes surpassing it by up to 12\%
(this is, for instance, the case for the recall of negative microblogs
on the SB10k corpus).  This margin becomes even larger if we compare
the scores of Mohammad's system with the performance of
\citeauthor{Gamon:04}'s predictor, which is by far the weakest ML
method in this survey.  This weakness, however, is less surprising
regarding the fact that \citeauthor{Gamon:04}'s approach is purely
grammar-based and relies only on information about part-of-speech tags
and constituency parses without any lexicon traits or even plain
$n$-gram features.  Partially due to these limited input attributes,
the results of this analyzer are even worse than the average scores of
lexicon-based methods.

\subsection{Feature Analysis}\label{subsec:cgsa:ml-methods:feature-analysis}

Because input features appeared to play a crucial role for the success
of ML-based systems, we decided to investigate the impact of this
factor in more detail and performed an ablation test for each of the
tested classifiers, removing one of their feature groups at a time and
recomputing their scores.

\begin{table}[h]
  \begin{center}
    \bgroup\setlength\tabcolsep{0.1\tabcolsep}\scriptsize
    \begin{tabular}{p{0.2\columnwidth} 
        *{6}{>{\centering\arraybackslash}p{0.13097\columnwidth}}}
      \toprule
      \multirow{2}{0.145\columnwidth}{%
      \bfseries Features} & %
      \multicolumn{6}{c}{\bfseries System Scores}\\
      & \multicolumn{2}{c}{\bfseries GMN} & \multicolumn{2}{c}{\bfseries MHM} %
      & \multicolumn{2}{c}{\bfseries GNT}\\%
      \cmidrule(lr){2-3}\cmidrule(lr){4-5}\cmidrule(lr){6-7}

      & Macro\newline \F{}$^{+/-}$ & Micro\newline \F{} %
      & Macro\newline \F{}$^{+/-}$ & Micro\newline \F{} %
      & Macro\newline \F{}$^{+/-}$ & Micro\newline \F{}\\\midrule

      \multicolumn{7}{c}{\cellcolor{cellcolor}PotTS}\\
      All & \textbf{0.453} & \textbf{0.617} & \textbf{0.674} & 0.727 & \textbf{0.624} & 0.673\\
      --Constituents & 0.388 & 0.545 & \NA{} & \NA{} & \NA{} & \NA{}\\
      --PoS Tags & 0.417 & 0.607 & 0.669 & 0.721 & \NA{} & \NA{}\\
      --Character Features & \NA{} & \NA{} & 0.671 & \textbf{0.734} & \NA{} & \NA{}\\
      --Token Features & \NA{} & \NA{} & 0.659 & 0.704 & 0.0 & 0.366\\
      --Automatic Lexicons & \NA{} & \NA{} & 0.667 & 0.717 & 0.613 & 0.666\\
      --Manual Lexicons & \NA{} & \NA{} & 0.665 & 0.715 & 0.617 & \textbf{0.675}\\

      \multicolumn{7}{c}{\cellcolor{cellcolor}SB10k}\\
      All & \textbf{0.329} & 0.699 & 0.564 & 0.752 & 0.491 & 0.724\\
      --Constituents & 0.127 & 0.646 & \NA{} & \NA{} & \NA{} & \NA{}\\
      --PoS Tags & 0.301 & \textbf{0.7} & \textbf{0.57} & \textbf{0.757} & \NA{} & \NA{}\\
      --Character Features & \NA{} & \NA{} & 0.546 & 0.753 & \NA{} & \NA{}\\
      --Token Features & \NA{} & \NA{} & 0.559 & 0.741 & 0.046 & 0.62\\
      --Automatic Lexicons & \NA{} & \NA{} & 0.54 & 0.753 & \textbf{0.517} & 0.735\\
      --Manual Lexicons & \NA{} & \NA{} & 0.553 & 0.751 & 0.51 & \textbf{0.739}\\\bottomrule
    \end{tabular}
    \egroup{}
    \caption[Feature-ablation test of ML-based MLSA methods]{
      Results of the feature-ablation test for ML-based MLSA methods}\label{snt-cgsa:tbl:ml-res-ablation}
  \end{center}
\end{table}

As we can see from the results in
Table~\ref{snt-cgsa:tbl:ml-res-ablation}, the approach
of~\citet{Gamon:04} typically achieves its best performance when all
of the input attributes (PoS tags and syntactic constituents) are
active.  This is for example the case for the micro- and
macro-averaged \F{} on the PotTS corpus, and also holds for the
two-class macro-\F{} on the SB10k data.  The only exception to this
tendency is the micro-averaged \F{}-score on the latter dataset, which
shows a slight improvement (from 0.699 to 0.7) after the removal of
part-of-speech features.

Similarly, the analyzer of~\citet{Mohammad:13} seems to rather suffer
than benefit from the part-of-speech attributes, which decrease its
micro-averaged scores by almost 0.07 points on PotTS and 0.05~\F{} on
SB10k.  One possible explanation for this degradation could be the
differences in the utilized PoS taggers and tagsets: Whereas the
original \citeauthor{Mohammad:13}'s classifier relied on a special
Twitter-aware tagger~\cite{Owoputi:13}, whose tags were explicitly
adjusted to the peculiarities of social media texts (including special
labels for the @-mentions and \#hashtags), we instead used the output
of the standard \textsc{TreeTagger}~\cite{Schmid:95}, which, apart
from lacking any Twitter-specific information, was also trained on a
completely different text genre (newspaper articles) and therefore a
priori produced unreliable output.  As a consequence, the effect of
part-of-speech information is rather harmful, and the only aspect
where it comes in handy is the macro-averaged \F{} on the PotTS
corpus, which improves by 0.003 when these features are used.  A
better alternative in this regard could be the Twitter-specific tagger
for German developed by~\citet{Rehbein:13}, we could not, however,
find this tagger in the public domain, and, moreover, its usage would
preclude the following \textsc{Mate} analysis due to the difference in
the tagsets.

An even more controversial situation is observed with the classifier
of~\citet{Guenther:14}. Although this system lacks any part-of-speech
attributes, its reaction to the deletion of other features (first of
all token and lexicon traits) is quite unexpected.  For example, the
macro-averaged \F{}-scores on both corpora drop almost to zero when
the information about tokens is excluded.  On the other hand, the
deactivation of manual lexicons surprisingly improves the
micro-averaged results on both datasets and also increases the
macro-\F{}$^{+/-}$ on the SB10k data.  We also notice a similar
(though less pronounced) trend with automatic lexicons: the ablation
of these features lowers the scores on PotTS, but improves both
results on SB10k.  We can partially explain this negative effect of
polarity lists by the coarseness of lexicon features: This classifier
uses only binary attributes, which reflect whether the given tweet has
more positive or more negative lexicon items, but it does not
distinguish between the scores or intensities of these terms.

Besides analyzing the utility of each particular feature group, we
also decided to have a look at the top-10 most relevant attributes
learned by each system.  The summarized overview in
Table~\ref{fgsa:tbl:ml:to10-features} partially confirms our previous
findings: For example, the most useful traits for the analyzer
of~\citet{Gamon:04} are attributes reflecting the information about
both constituents and part-of-speech tags, with five of its ten
entries featuring the interjection tag, which appears to be especially
important for predicting the positive class.  On the other hand, the
system of~\citet{Mohammad:13} seems to rely more on token and
character $n$-grams, as nine out of ten attributes belong to either of
these two categories. The only outlier in this respect is the
\texttt{Last\%QMarkCnt} attribute (line 2), which denotes the presence
of a question mark and is apparently a good clue of neutral
microblogs.  Finally, the classifier of~\citet{Guenther:14} almost
exclusively prefers lexical $n$-grams, as it has nine unigrams and one
bigram among its top-ten entries.

\begin{table}[hbt]
  \begin{center}
    \bgroup\setlength\tabcolsep{0.47\tabcolsep}\scriptsize
    \begin{tabular}{>{\centering\arraybackslash}p{0.05\columnwidth} 
        *{9}{>{\centering\arraybackslash}p{0.092\columnwidth}}} 
      \toprule
      \multirow{2}{0.05\columnwidth}{Rank} & \multicolumn{3}{c}{\bfseries GMN} & %
                      \multicolumn{3}{c}{\bfseries MHM} & %
                      \multicolumn{3}{c}{\bfseries GNT}\\\cmidrule(lr){2-4}\cmidrule(lr){5-7}\cmidrule(lr){8-10}
      & Feature & Label & Weight & Feature & Label & Weight %
      & Feature & Label & Weight\\\midrule
          1 & NK-ITJ| & POS & 0.457 & * & NEUT & 0.131 & hate & NEG & 1.86 \\
          2 & DM-ITJ| & POS & 0.334 & Last\-\%QMark\-Cnt & NEUT & 0.088 & sick & NEG & 1.7\\
          3 & V-DM-I & POS & 0.244 & s-c & NEG & 0.079 & kahretsinn & NEG & 1.69\\
          4 & N-NK-I & POS & 0.24 & *-\%possmiley & POS & 0.067 & dasisaberschade & NEG & 1.69\\
          5 & MO-ITJ| & POS & 0.211 & c-h-e-i-s & NEG & 0.064 & Anziehen & POS & 1.67\\
          6 & A-DM-I & POS & 0.196 & h-a-h & POS & 0.064 & \textbackslash{}x016434 & POS & 1.65\\
          7 & A-MO-I & POS & 0.191 & t-\textvisiblespace{}-. & NEG & 0.064 & p\"archenabend & POS & 1.65\\
          8 & NK-ITJ & POS & 0.165 & geil & POS & 0.062 & derien\heart\heart{} & POS & 1.65\\
          9 & NK-\$. & NEUT & 0.16 & *-? & NEUT & 0.062 & sch\"on-nicht & POS & 1.56\\
          10 & DM-ITJ & POS & 0.157 & ? & NEUT & 0.061 & applause & POS & 1.5\\\bottomrule
    \end{tabular}
    \egroup{}
    \caption[Top-10 features learned by MLSA classifiers]{Top-10
      features learned by ML-based MLSA methods\\{\small (sorted by
        the absolute values of their weights)}}\label{fgsa:tbl:ml:to10-features}
  \end{center}
\end{table}

\subsection{Classifiers}\label{subsec:cgsa:ml-methods:classifiers-analysis}

Another important factor that could significantly affect the quality
of ML-based approaches was the underlying classification method, which
was used to optimize the feature weights and make the final
predictions.  Although most of the previous studies agree on the
superior performance of support vector machines for this
task~\cite[see ][]{Pang:02,Gamon:04,Mohammad:13}, we decided to
question these conclusions as well and reran our experiments,
replacing the linear SVC predictor with the Na\"{\i}ve Bayes and
Logistic Regression algorithms.

Somewhat surprisingly, these changes indeed resulted in an
improvement, especially in the case of the logistic classifier, which
yielded the best macro- and micro-averaged scores for the systems of
\citet{Mohammad:13} and \citet{Guenther:14} on the PotTS corpus (see
Table~\ref{snt-cgsa:tbl:ml-res-classifiers}) and also produced the
highest micro-\F{} results for these two approaches on the SB10k
dataset.  Nevertheless, the SVM algorithm still remains a competitive
option, in particular for the feature-sparse method
of~\citet{Gamon:04}, but also with respect to the macro-\F{} of
Mohammmad's and G\"unther's analyzers.

\begin{table}[h]
  \begin{center}
    \bgroup\setlength\tabcolsep{0.1\tabcolsep}\scriptsize
    \begin{tabular}{p{0.2\columnwidth} 
        *{6}{>{\centering\arraybackslash}p{0.13097\columnwidth}}}
      \toprule
      \multirow{2}{0.15\columnwidth}{%
      \bfseries Classifier} & %
      \multicolumn{6}{c}{\bfseries System Scores}\\
      & \multicolumn{2}{c}{\bfseries GMN} & \multicolumn{2}{c}{\bfseries MHM} %
      & \multicolumn{2}{c}{\bfseries GNT}\\%
      \cmidrule(lr){2-3}\cmidrule(lr){4-5}\cmidrule(lr){6-7}

      & Macro\newline \F{}$^{+/-}$ & Micro\newline \F{} %
      & Macro\newline \F{}$^{+/-}$ & Micro\newline \F{} %
      & Macro\newline \F{}$^{+/-}$ & Micro\newline \F{}\\\midrule

      \multicolumn{7}{c}{\cellcolor{cellcolor}PotTS}\\
      SVM & \textbf{0.453} & \textbf{0.617} & 0.674 & 0.727 & \textbf{0.624} & 0.673\\
      Na\"{\i}ve Bayes & 0.432 & 0.577 & 0.635 & 0.675 & 0.567 & 0.59\\
      Logistic Regression & 0.431 & 0.612 & \textbf{0.677} & \textbf{0.741} & \textbf{0.624} & \textbf{0.688}\\

      \multicolumn{7}{c}{\cellcolor{cellcolor}SB10k}\\
      SVM & 0.329 & \textbf{0.699} & \textbf{0.564} & 0.752 & 0.491 & 0.724\\
      Na\"{\i}ve Bayes & \textbf{0.351} & 0.637 & 0.516 & 0.755 & 0.453 & 0.675\\
      Logistic Regression & 0.309 & 0.693 & 0.553 & \textbf{0.772} & \textbf{0.512} & \textbf{0.75}\\\bottomrule
    \end{tabular}
    \egroup{}
    \caption{
      Results of ML-based MLSA methods with different classifiers}\label{snt-cgsa:tbl:ml-res-classifiers}
  \end{center}
\end{table}

Even though our results contradict previous claims in the literature,
we would advise against premature conclusions at this point and stress
the fact that different classifiers might have fairly varying results
on different datasets.  Therefore, higher scores of the logistic
regression on our corpora do not preclude better SVM results on the
official SemEval data.

\subsection{Error Analysis}\label{subsec:cgsa:ml-methods:err-analysis}

As in our previous experiments, we also decided to have a closer look
at errors produced by each tested system.  For this purpose, we again
collected misclassifications that were unique to only one of the
classifiers, and provide some examples of these errors below.

The first wrong result shown in
Example~\ref{snt:cgsa:exmp:gamon-error} was produced by the system
of~\citet{Gamon:04}.
\begin{example}[An Error Made by the System of~\citeauthor{Gamon:04}]\label{snt:cgsa:exmp:mohammad-error}
  \noindent\textup{\bfseries\textcolor{darkred}{Tweet:}} {\upshape Das ist das zynische. \"Uber Themen labern, Leute schlecht machen. Wenn nicht der Papst damit Thema w\"are, kein Wort. Ich hasse das.}\\
  \noindent It's cynical.  To babble about topics, to talk people
  down.  If the topic wouldn't be the Pope, no word. I hate
  this.\\[\exampleSep]
  \noindent\textup{\bfseries\textcolor{darkred}{Gold Label:}}\hspace*{4.3em}\textbf{%
    \upshape\textcolor{midnightblue}{negative}}\\
 \noindent\textup{\bfseries\textcolor{darkred}{Predicted Label:}}\hspace*{2em}\textbf{%
    \upshape\textcolor{green3}{positive*}}\label{snt:cgsa:exmp:gamon-error}
\end{example}
\noindent{} In this case, the classifier incorrectly assigned the
positive label to a clearly negative microblog despite the presence of
multiple negatively connoted terms (``zynische'' [\emph{cynical}],
``labern'' [\emph{to babble}], ``schlecht machen'' [\emph{to talk
    down}], and ``hasse'' [\emph{to hate}]).  The reason for this
decision is quite simple: As we already noted in the foregoing
description, this method is completely unlexicalized and relies only
on grammatical information while making its predictions.  In
particular, for this microblog, the top-5 most important features
(ranked by the absolute values of their coefficients) are:
\begin{enumerate}
\item \texttt{PD-ADJA} (\textcolor{black}{neutral}): -0.62896911412,
\item \texttt{---VVINF} (\textcolor{midnightblue}{negative}): 0.517300341184,
\item \texttt{PD-ADJA} (\textcolor{green3}{positive}): 0.505413668274,
\item \texttt{---VVINF} (\textcolor{green3}{positive}): -0.346990702756,
\item \texttt{CJ-VVINF} (\textcolor{green3}{positive}):
  0.303311030403.
\end{enumerate}
As we can see, none of these attributes reflects any information about
the lexical terms appearing in the message, and the system simply
prefers the positive class based on the presence of a predicate
adjective (\texttt{PD-ADJA}) and coordinately conjoined infinitive
(\texttt{CJ-VVINF}).

Another error shown in Example~\ref{snt:cgsa:exmp:mohammad-error} was
made by the system of~\citet{Mohammad:13}.  This time, a positive
tweet was misclassified as neutral.  But the reason for this erroneous
decision is completely different.  As we can see from the list of the
highest ranked features given below:
\begin{enumerate}
\item \texttt{*} (\textcolor{black}{neutral}): 0.131225868029,
\item \texttt{*} (\textcolor{midnightblue}{negative}): -0.0840804221845,
\item \texttt{\%PoS-CARD} (\textcolor{black}{neutral}): 0.0833658576233,
\item \texttt{\%PoS-ADJD} (\textcolor{black}{neutral}): -0.069745190018,
\item \texttt{t-\textvisiblespace{}-n} (\textcolor{green3}{positive}): 0.0556721202587;
\end{enumerate}
this analyzer makes its decision based on rather general, but
extremely heavy-weighted features, such as placeholder token
\texttt{*} or the PoS-tag features (\texttt{\%PoS-CARD} and
\texttt{\%PoS-ADJD}).  As a result, its prediction succumbs to the
neutral bias of these general attributes.

\begin{example}[An Error Made by the System of~\citeauthor{Mohammad:13}]\label{snt:cgsa:exmp:mohammad-error}
  \noindent\textup{\bfseries\textcolor{darkred}{Tweet:}} {\upshape das
    klingt richtig gut! Was f\"ur eine hast du denn? (uvu) \%PosSmiley3}\\
  \noindent It sounds really great.  Which one do you have? (uvu) \%PosSmiley3\\[\exampleSep]
  \noindent\textup{\bfseries\textcolor{darkred}{Gold Label:}}\hspace*{4.3em}\textbf{%
    \upshape\textcolor{green3}{positive}}\\
 \noindent\textup{\bfseries\textcolor{darkred}{Predicted Label:}}\hspace*{2em}\textbf{%
    \upshape\textcolor{black}{neutral*}}
\end{example}

Finally, the last example (\ref{snt:cgsa:exmp:guenther-error}) shows
another wrong decision, where a negative microblog was incorrectly
analyzed as positive by the method of~\citet{Guenther:14}, even though
the polar term ``borniert'' (\emph{narrow-minded}) was present in both
utilized sentiment lexicons (ZPL and Linear Projection) as a negative
item.  This again can be explained by the prevalence of general
features (\eg{} \texttt{8}, \texttt{nicht-nur\_NEG},
\texttt{nur\_NEG}, etc.) and their strong bias towards the majority
class in the PotTS dataset.

\begin{example}[An Error Made by the System of~\citeauthor{Guenther:14}]\label{snt:cgsa:exmp:guenther-error}
  \noindent\textup{\bfseries\textcolor{darkred}{Tweet:}} {\upshape Den CDU-W\"ahlern traue ich durchaus zu der FDP 8 bis 9\% zu bescheren! Die sind so borniert, nicht nur in Niedersachsen!}\\
  \noindent I don't put giving 8 to 9\% to the FDP past the CDU-voters!  They are so narrow-minded, not only in Lower Saxony!\\[\exampleSep]
  \noindent\textup{\bfseries\textcolor{darkred}{Gold Label:}}\hspace*{4.3em}\textbf{%
    \upshape\textcolor{midnightblue}{negative}}\\
 \noindent\textup{\bfseries\textcolor{darkred}{Predicted Label:}}\hspace*{2em}\textbf{%
    \upshape\textcolor{green3}{positive}}
\end{example}


\section{Deep-Learning Methods}\label{sec:cgsa:dl-based}

Even though traditional ML-based approaches still show competitive
results and play an important role in the sentiment analysis of social
media, they are gradually giving place to allegedly more powerful and
in a certain sense more intuitive deep learning (DL) methods.  As we
already mentioned in the previous chapter, in contrast to the standard
supervised techniques with human-engineered features, DL systems
induce the best feature representation completely automatically, and
in some cases might produce even better features than the ones devised
by human experts.  Another important advantage of this paradigm is its
more straightforward way to implement the ``compositionality'' of
language \cite{Frege:1892}: Whereas conventional classifiers usually
consider each instance as a \emph{bag of features} and predict its
label based on the sum of these features' values multiplied with their
respective weights, DL approaches try to \emph{combine} the
representation of each part of that instance (be it tokens or
sentences) into a single whole and then deduce the final class from
this joint embedding.

Among the first who explicitly incorporated the compositionality
principle into a DL-based sentiment application were
\citet{Yessenalina:11}.  In their proposed matrix-space approach, the
authors represented each word $w$ of an input phrase $\mathbf{x}^i =
w^i_1, w^i_2, \ldots, w^i_{|\mathbf{x}^i|}$ as a matrix
$W_{w}\in\mathbb{R}^{m\times m}$ and computed the sentiment score
$\xi^i$ of this phrase as the product of its token matrices,
multiplying the final result with two auxiliary model parameters
$\vec{u}$ and $\vec{v}\in\mathbb{R}^m$ to get a scalar value:
\begin{align*}
  \xi^i =& \vec{u}^\top\left(\prod_{j=1}^{|x^i|}W_{w^i_j}\right)\vec{v}.
\end{align*}
After computing this term, they predicted the intensity and polarity
of the phrase on a five-level sentiment scale (ranging from very
negative to very positive) by comparing $\xi^i$ with automatically
derived thresholds.  With this system, \citeauthor{Yessenalina:11}
attained a ranking loss of 0.6375 on the MPQA corpus~\cite{Wiebe:05},
outperforming the traditional PRank algorithm \cite{Crammer:01} and
bag-of-words ordered logistic regression.

Almost simultaneously with this work, \citet{Socher:11} introduced a
deep recursive autoencoder (RAE), in which they obtained a fixed-width
vector representation for a complex phrases $\vec{w}_p$ by recursively
merging the vectors of its tokens over a binarized dependency tree,
first multiplying these vectors with a compositional matrix $W$ and
then applying a non-linear function ($softmax$) to the resulting
product:
\begin{align}
  \vec{w}_p &= softmax\left(W\begin{bmatrix}
      \vec{w}_l\\
      \vec{w}_r
  \end{bmatrix}\right),\label{cgsa:eq:socher-11}
\end{align}
where $\vec{w}_l$ and $\vec{w}_r$ represent the embeddings of the left
and right dependents respectively.  By applying a max-margin
classifier to the final phrase vector, the authors could improve the
state of the art on predicting sentence-level polarity of user's blog
posts~\cite{Potts:10} and also outperformed the system
of~\citet{Nasukawa:03} on the MPQA dataset~\cite{Wiebe:05}, achieving
86.4\% accuracy on predicting contextual polarity of opinionated
expressions.

Later on, \citet{Socher:12} further improved this approach by
associating an additional matrix $W_w$ with each vocabulary word $w$
and performing the inference simultaneously over both vector and
matrix representations:
\begin{align*}
  \vec{w}_p = \tanh\left(W_v \begin{bmatrix}W_r\vec{w}_l\\
      W_l \vec{w}_r\end{bmatrix} \right),\\
  W_p = W_m \begin{bmatrix}W_l;\\
    W_r\end{bmatrix};
\end{align*}
where $\vec{w}_p\in\mathbb{R}^n$ stands for the embedding of the
parent node, $\vec{w}_l$ and $\vec{w}_r$ represent the embeddings of
its left and right dependents, and $W_p, W_l, W_r \in
\mathbb{R}^{n\times n}$ denote the respective matrices associated with
these vertices.  The compositionality matrices
$W_v\in\mathbb{R}^{n\times 2n}$ and $W_m\in\mathbb{R}^{n\times 2n}$
were shared across all instances and learned along with the vector
embeddings.  This model, called Matrix-Vector Recursive Neural Network
(MVRNN), surpassed the RAE system on the IMDB movie review
dataset~\cite{Pang:05}, attaining 0.91 Kullback-Leibler divergence
between the assigned scores and probabilities of correct labels.

Yet another improvement, a Recursive Neural Tensor Network (RNTN), was
presented by~\citet{Socher:13}.  In this system, the authors again
opted for a vector representation of words, but enhanced the original
matrix-vector product from Equation~\ref{cgsa:eq:socher-11} with an
additional tensor multiplication:
\begin{align*}
  \vec{w}_p &= softmax\left(\begin{bmatrix}
  \vec{w}_l\\
  \vec{w}_r
  \end{bmatrix}^{\top}V^{[1:d]}\begin{bmatrix}
  \vec{w}_l\\
  \vec{w}_r
  \end{bmatrix}
            + W\begin{bmatrix}
  \vec{w}_l\\
  \vec{w}_r
\end{bmatrix}\right),\label{cgsa:eq:socher-13}
\end{align*}
where $\vec{w}_p, \vec{w}_l, \vec{w}_r\in\mathbb{R}^n$, and
$W\in\mathbb{R}^{n\times 2n}$ are defined as before; and $V$
represents a $2n\times 2n\times n$-dimensional tensor.  By increasing
this way the number of parameters in comparison with the RAE approach,
but significantly reducing it with respect to the MVRNN method, the
authors gained a significant improvement of the results, boosting the
classification accuracy on their own Stanford Sentiment Treebank from
82.9 to 85.4\%.

A real breakthrough in the use of deep learning methods for sentiment
analysis of Twitter happened with the work~\citet{Severyn:15}, whose
proposed feed-forward DL system ranked first in SemEval-2015
Subtask~10~A (phrase-level polarity prediction) \cite{Rosenthal:15}
and achieved second place (0.6459~\F$^{+/-}$) in Subtask~10~B
(message-level classification) of this competition.  Drawing on the
ideas of~\citet{Kalchbrenner:14}, the authors devised a simple
convolutional network in which they multiplied pretrained word
embeddings with 300 distinct convolutional kernels each of width 5,
pooled the maximum value of this multiplication for each kernel, and
then passed the results of this pooling to a piecewise linear ReLU
filter with a densely connected softmax layer.  An important aspect of
this approach, which accounted for a huge part of its success, was a
special multi-stage training scheme that was used to optimize the
parameters: In the initial stage of this scheme,
\citeauthor{Severyn:15} first computed Twitter-specific word
embeddings by applying the word2vec algorithm to a large Twitter
corpus.  Afterwards, they pretrained the complete system including the
word vectors, convolutional filters, and inter-layer matrices on a big
set of noisily labeled microblogs from this collection, and, finally,
fine-tuned the parameters of the model on the official SemEval
dataset.

Later on, this system was further improved by \citet{Deriu:16}, who
increased the number of convolutional layers (applying two layers
instead of one) and simultaneously trained two such models (using
word2vec vectors as input for the first one and passing GloVe
embeddings to the second), joining their output at the end and
achieving this way 0.671~\F$^{+/-}$ on the SemEval-2015 test set.  A
similar enhancement was also proposed by~\citet{Rouvier:16}, who used
three different types of embeddings (word2vec, word2vec specific to
particular parts of speech, and sentiment-tailored vectors), training
separate sets of convolutions for each of these types.

Although convolutional approaches still show competitive scores and
are hard to outperform in practice, in recent time, they are gradually
being superseded by recurrent neural networks~\cite{Xu:16,Wang:15}.
One of the most prominent such systems has been recently proposed
by~\citet{Baziotis:17}.  In their submission to
SemEval~2017~\cite{Rosenthal:17}, the authors used two successive
bidirectional LSTM units (BiLSTMs).  In each of these units, they
concatenated the results of the left-to-right recurrence
($\vec{h}^{(l)}_{i_{\rightarrow}}\in\mathbb{R}^{150}$) with the
respective outputs of the right-to-left loop
($\vec{h}^{(l)}_{i_{\leftarrow}}\in\mathbb{R}^{150}$) and then passed
the result of this concatenation ($\vec{h}_i^{(l)} =
[\vec{h}^{(l)}_{i_{\rightarrow}},
  \vec{h}^{(l)}_{i_{\leftarrow}}]\in\mathbb{R}^{300}$) to the next
layer of the network.  After getting the output of the second BiLSTM,
they united the states of this unit from all time steps $i$ into a
single vector $\vec{a}$ with the help of a special attention
mechanism, in which they first multiplied each BiLSTM state
$\vec{h}_i$ with the respective globally normalized attention score
$a_i$ and then took the sum of these weighted vectors over all $i$
positions:
\begin{align}
  \vec{a} =&
  \sum_{i=1}^{|\mathbf{x}|}a_i\vec{h}^{(2)}_i,\nonumber\\
  \mbox{where }a_i =&
  \frac{\exp(e_i)}{\sum_{j=1}^{|\mathbf{x}|}\exp(e_j)},\nonumber\\
  \textrm{s.t. }e_i =&
  \tanh\left(\vec{\alpha}\vec{h}^{(2)}_i + \beta_i\right).
\end{align}\label{eq:cgsa:baziotis-attention}%
The $\vec{\alpha}$ and $\beta$ terms in the above equations denote the
attention parameters (score and bias), which are optimized during the
training process.  To make the final prediction,
\citeauthor{Baziotis:17} multiplied the attention vector $\vec{a}$
with matrix $W$ and computed element-wise softmax of this product,
getting probability scores for each of the three polarity classes and
choosing the label with the maximum score:
\begin{align}
  \hat{y} =& \argmax\left(softmax(W^\top\vec{a})\right).
\end{align}
\noindent With this approach, the authors attained the first first
place in Task~4 of SemEval-2017 (0.675~\F{}$^{+/-}$), being on a par
with the system of~\citet{Cliche:17} and even outperforming the method
of~\citet{Rouvier:17} despite the fact that both of these competitors
used ensembles of LSTMs and convolutional networks.

\begin{figure*}[htbp!]
{
  \centering
  \includegraphics[width=\linewidth]{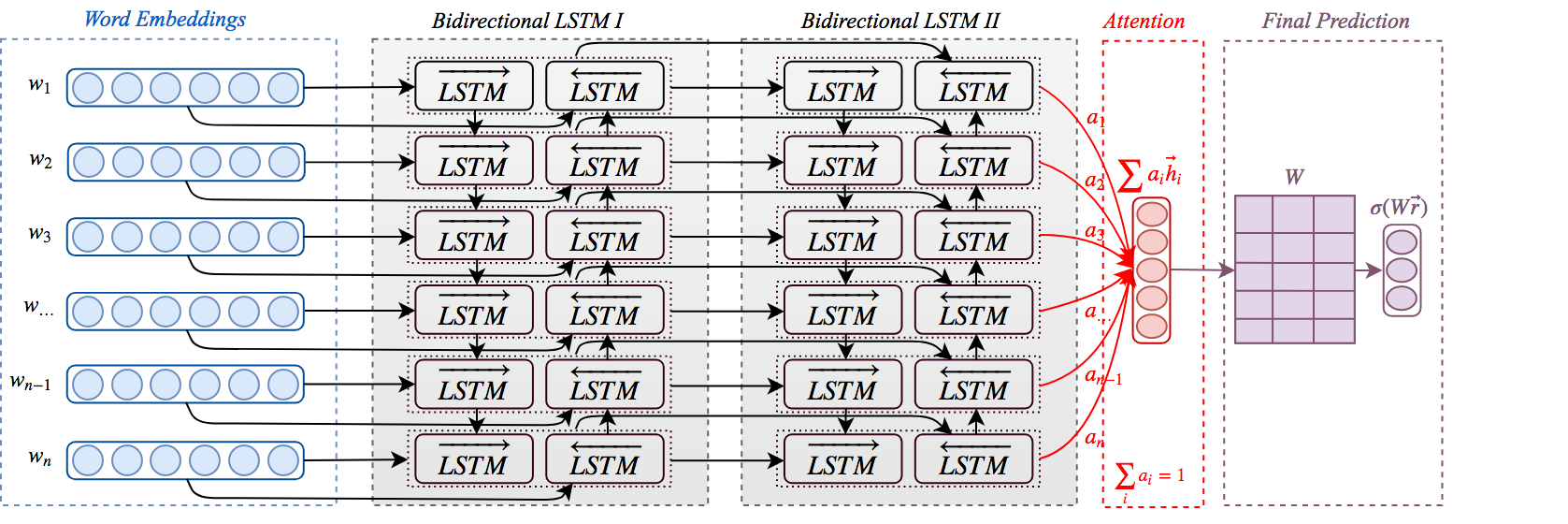}
}
\caption[Neural network of \citet{Baziotis:17}]{Architecture of the
  neural network proposed
  by~\citet{Baziotis:17}}\label{cgsa:fig:baziotis}
\end{figure*}

\subsection{Lexicon-Based Attention}

Even though the approach of~\citet{Baziotis:17} represents the current
state of the art in sentiment analysis of Twitter and yields
extraordinarily good results, in our opinion, this method has yet some
potential for improvements.  This, first of all, concerns the way how
attention coefficients are computed.  As we can see from
Equation~\ref{eq:cgsa:baziotis-attention}, the magnitude of the
attention score $a_i$ primarily depends on the absolute value of the
BiLSTM outputs and the bias term at the $i$-th position.  Albeit this
strategy is definitely plausible, assuming the fact that LSTMs shall
produce higher scores for polar tokens and presupposing that polar
terms near the end of the message will usually have a greater
influence on the net polarity of the tweet than subjective words at
its beginning, a crucial prerequisite for this strategy to work is
\begin{inparaenum}[(i)]
\item that the LSTM layer can already provide sufficiently reliable
  results and
\item that the bias terms do not overly boost the importance of
  irrelevant tokens that just accidentally appeared at favored
  positions.
\end{inparaenum}
Unfortunately, both of these prerequisites are rarely fulfilled in
practice.

In order to overcome these deficiencies, we augmented the original
architecture of~\citet{Baziotis:17} shown in
Figure~\ref{cgsa:fig:baziotis} with two additional types of attention:
\emph{lexicon-} and \emph{context-based} one.  In the former type, we
estimated the importance weight $b_i$ for position $i$ as the polarity
score of the word $w_i$, obtaining this value from our Linear
Projection lexicon and normalizing it by the sum of polarity scores
for all tweet tokens:
\begin{align*}
  \vec{b} =& \sum_{i=1}^{|\mathbf{x}|}b_i\vec{h}_i,\\
  b_i =& \frac{\exp(f_i)}{\sum_{j=1}^{|\mathbf{x}|}\exp(f_j)},\\
  \mbox{s.t. }f_i
  =& \left\{
  \begin{array}{ll}
    \tanh(abs(V[{w_i}]) + \epsilon) & \textrm{ if } w_i\in V\\
    \tanh(\epsilon) & \, \textrm{otherwise.} \\
  \end{array}
  \right.
\end{align*}\label{cgsa:eq:lba}%
This way, we hoped to force the network to pay more attention to the
BiLSTM outputs that were produced at the positions of polar terms
rather than favoring arbitrary words in the message.

Another important factor that could notably affect the polarity of a
microblog were the so-called \emph{valence
  shifters}~\cite[][]{Polanyi:06}---words and phrases such as ``kaum''
(\emph{hardly}) or ``nicht'' (\emph{not}) that could significantly
change (or even reverse) the semantic orientation of polar terms.  To
account for these phenomena, we added another type of attention---a
\emph{context-based} one, whose goal was to identify such shifters in
the message and give them bigger weights in the recursion.  To discern
these elements, we introduced a linear classifier that had to predict
the modifying power of a token $w_i$, given its original word
embedding $\vec{w}_i$ and the LSTM output of its parent in the
dependency tree times the lexicon-based attention score of that parent
($\vec{b}_p := b_p\vec{h}_p$).  To keep the resulting attention scores
within an appropriate range, we again used the same $\tanh$
transformation and global normalization over all positions as we did
in the previous two types:
\begin{align*}
  \vec{c} =& \sum_{i=1}^{|\mathbf{x}|}c_i\vec{h}_i,\\ c_i =&
  \frac{\exp(g_i)}{\sum_{j=1}^{|\mathbf{x}|}\exp(g_j)},\\ g_i =&
  \tanh\left(C [\vec{w}_i, \vec{b}_p]^{\top}\right).
\end{align*}
The $C$ term in the above equation represents a context-based
attention matrix $\mathbb{R}^{200 \times 100}$; the $\vec{w}_i$
variable denotes the word embedding of the $i$-th token; and the
$\vec{b}_p$ term stands for the value of vector $\vec{b}$ (the result
of lexicon-based attention from Equation~\ref{cgsa:eq:lba}) at
position $p$ (the index of syntactic parent of $w_i$).  With this
classifier, we hoped to amplify the importance of shifting words in
the cases when the immediate syntactic ancestors of these tokens were
highly subjective expressions (\eg{} ``Er hat die Pr\"ufung kaum
bestanden'' [\emph{He hardly passed the exam}] or ``Ich mag den neuen
Bundesminister nicht'' [\emph{I do not like the new federal
    minister}]), but ignore them when they did not relate to any
subjective term.

At last, to make the final prediction, we concatenated the outputs of
the three attention layers into a single matrix $A\in\mathbb{R}^{3
  \times 100}$ and multiplied it with a vector
$\vec{w}\in\mathbb{R}^{1\times{}100}$, applying softmax normalization
at the end:
\begin{align*}
  \vec{o} =& softmax\left(A\vec{w}^\top\right),\textrm{ where}\\
  A =& \begin{bmatrix}
    \vec{a}\\
    \vec{b}\\
    \vec{c}\end{bmatrix}.
\end{align*}

\begin{figure*}[htbp!]
{ \centering \includegraphics[width=1.3\linewidth]{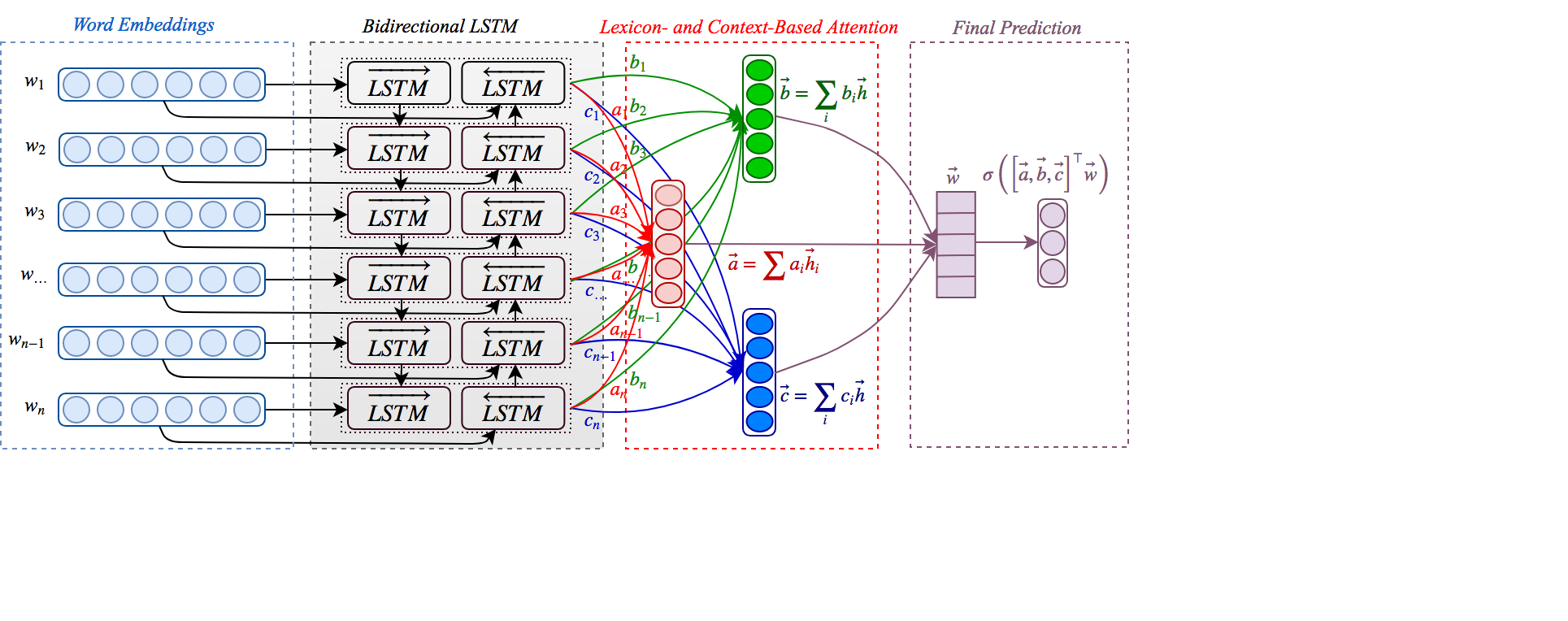} }
\caption[Neural network with lexicon-based attention]{Architecture of
  the neural network with lexicon- and context-based
  attention}\label{cgsa:fig:lba}
\end{figure*}

Since introducing additional attention types increased the number of
model parameters, we removed one of the intermediate Bi-LSTM layers in
the network to counterbalance this effect and report our results for
both settings: using one and two Bi-LSTM units (denoted as LBA$^{(1)}$
and LBA$^{(2)}$, respectively).  The final architecture of our
approach is shown in Figure~\ref{cgsa:fig:lba}.



To evaluate the performance of the previously presented methods and to
compare our lexicon-based attention system with these solutions, we
reimplemented the approaches of~\citet{Yessenalina:11},
\citet{Socher:11,Socher:12,Socher:13}, \citet{Severyn:15},
and~\citet{Baziotis:17}.  For the sake of uniformity and simplicity,
we used task-specific word embeddings of size~$\mathbb{R}^{100}$ in
all systems, optimizing these vectors along with other network
parameters during the training.  Moreover, we also unified the final
activation parts and cost functions of all networks, using a densely
connected softmax layer as the last component of each classifier and
optimizing their weights w.r.t. the categorical hinge loss on the
training data, picking the values that yielded the highest accuracy on
the development set.

\begin{table}[h]
  \begin{center}
    \bgroup\setlength\tabcolsep{0.1\tabcolsep}\scriptsize
    \begin{tabular}{p{0.162\columnwidth} 
        *{9}{>{\centering\arraybackslash}p{0.074\columnwidth}} 
        *{2}{>{\centering\arraybackslash}p{0.068\columnwidth}}} 
      \toprule
      \multirow{2}*{\bfseries Method} & %
      \multicolumn{3}{c}{\bfseries Positive} & %
      \multicolumn{3}{c}{\bfseries Negative} & %
      \multicolumn{3}{c}{\bfseries Neutral} & %
      \multirow{2}{0.068\columnwidth}{\bfseries\centering Macro\newline \F{}$^{+/-}$} & %
      \multirow{2}{0.068\columnwidth}{\bfseries\centering Micro\newline \F{}}\\
      \cmidrule(lr){2-4}\cmidrule(lr){5-7}\cmidrule(lr){8-10}

      & Precision & Recall & \F{} & %
      Precision & Recall & \F{} & %
      Precision & Recall & \F{} & & \\\midrule

      \multicolumn{12}{c}{\cellcolor{cellcolor}PotTS}\\




      Y\&C & 0.45 & \textbf{1.0} & 0.62 & %
      0.0 & 0.0 & 0.0 & %
      0.0 & 0.0 & 0.0 & %
      0.308 & 0.446\\




      RAE & 0.64 & 0.78 & 0.7 & %
      0.38 & 0.04 & 0.08 & %
      0.57 & 0.68 & 0.62 & %
      0.389 & 0.605\\




      MVRNN & 0.45 & \textbf{1.0} & 0.62 & %
      0.0 & 0.0 & 0.0 & %
      0.0 & 0.0 & 0.0 & %
      0.308 & 0.446\\




      RNTN & 0.45 & 0.87 & 0.59 & %
      0.19 & 0.02 & 0.03 & %
      0.32 & 0.1 & 0.15 & %
      0.312 & 0.428\\




      SEV & 0.73 & 0.79 & 0.76 & %
      \textbf{0.41} & \textbf{0.52} & \textbf{0.46} & %
      \textbf{0.72} & 0.55 & 0.62 & %
      \textbf{0.608} & 0.651\\




      BAZ & 0.45 & \textbf{1.0} & 0.62 & %
      0.0 & 0.0 & 0.0 & %
      0.0 & 0.0 & 0.0 & %
      0.308 & 0.446\\

      LBA$^{(1)}$ & \textbf{0.82} & 0.73 & \textbf{0.77} & %
      0.0 & 0.0 & 0.0 & %
      0.56 & \textbf{0.92} & \textbf{0.69} & %
      0.387 & \textbf{0.662}\\

      LBA$^{(2)}$ & 0.45 & \textbf{1.0} & 0.62 & %
      0.0 & 0.0 & 0.0 & %
      0.0 & 0.0 & 0.0 & %
      0.308 & 0.446\\

      \multicolumn{12}{c}{\cellcolor{cellcolor}SB10k}\\




      Y\&C & 0.0 & 0.0 & 0.0 & %
      0.0 & 0.0 & 0.0 & %
      0.62 & \textbf{1.0} & 0.77 & %
      0.0 & 0.622\\




      RAE & 0.63 & 0.57 & 0.6 & %
      0.0 & 0.0 & 0.0 & %
      \textbf{0.75} & 0.94 & 0.83 & %
      0.299 & 0.721\\

      MVRNN & 0.0 & 0.0 & 0.0 & %
      0.0 & 0.0 & 0.0 & %
      0.62 & \textbf{1.0} & 0.77 & %
      0.0 & 0.622\\

      RNTN & 0.2 & 0.03 & 0.05 & %
      0.07 & 0.01 & 0.02 & %
      0.62 & 0.94 & 0.75 & %
      0.033 & 0.594\\

      %
      %
      %
      SEV & 0.0 & 0.0 & 0.0 & %
      0.0 & 0.0 & 0.0 & %
      0.62 & \textbf{1.0} & 0.77 & %
      0.0 & 0.622\\




      BAZ & 0.75 & 0.47 & 0.58 & %
      0.0 & 0.0 & 0.0 & %
      0.71 & 0.98 & 0.83 & %
      0.291 & 0.72\\

      LBA$^{(1)}$ & 0.72 & \textbf{0.58} & \textbf{0.64} & %
      0.0 & 0.0 & 0.0 & %
      0.74 & 0.97 & \textbf{0.84} & %
      \textbf{0.321} & \textbf{0.737}\\

      LBA$^{(2)}$ & \textbf{0.76} & 0.49 & 0.6 & %
      0.0 & 0.0 & 0.0 & %
      0.72 & 0.98 & 0.83 & %
      0.298 & 0.723\\\bottomrule
    \end{tabular}
    \egroup{}
    \caption[Results of DL-based MLSA methods]{Results of
      deep-learning--based MLSA methods\\ {\small Y\&C~--~\citet{Yessenalina:11},
        RAE~--~Recursive Auto-Encoder \cite{Socher:11},
        MVRNN~--~Matrix-Vector RNN \cite{Socher:12}, RNTN~--~Recursive
        Neural-Tensor Network \cite{Socher:13},
        SEV~--~\citet{Severyn:15}, BAZ~--~\citet{Baziotis:17},
        LBA$^{(1)}$~--~lexicon-based attention with one Bi-LSTM
        layer, LBA$^{(2)}$~--~lexicon-based attention with two Bi-LSTM
        layers}}\label{snt-cgsa:tbl:dl-res}
  \end{center}
\end{table}

The results of this evaluation are shown in
Table~\ref{snt-cgsa:tbl:dl-res}.  As we can see from the figures, the
LBA method performs fairly well, especially on the positive and
neutral classes where it achieves the best \F-benchmarks on both
datasets and also attains the highest overall micro-averaged \F-scores
on all test samples (0.662 on PotTS and 0.737 on SB10k).  Even though
our approach also yields the best macro-averaged result on the SB10k
set~(0.321~\F), it seems to face a major difficulty with the extreme
label skewness of this corpus, failing to predict any negative tweet
in the test set.  This problem, in general, appears to be an
insurmountable hurdle for almost all other compared systems,
especially the matrix-space, MVRNN, and convolutional approaches,
which eventually end up predicting only the most common neutral label
for all messages in this dataset.  A single notable exception to this
tendency is the recursive neural tensor approach of~\citet{Socher:13},
which succeeds in classifying some of the negative instances and also
predicts positive and neutral labels, but whose precision and recall
are still far below an acceptable level.

A similar, though less severe situation is also observed on the PotTS
corpus.  This time, the Y\&C, MVRNN, BAZ, and LBA$^{(2)}$ methods
lapse into always predicting only the most frequent positive class.
Other systems, however, perform much better, especially the approach
of~\citet{Severyn:15}, which does an extraordinarily good job at
classifying negative messages, reaching remarkable 0.46~\F{} on this
subset and also attaining the best macro-average score (0.608) on all
tweets due to its competitive performance on positive and neutral
microblogs.  Nevertheless, even the best-performing DL systems (SEV
and LBA) lag far behind the traditional supervised machine-learning
method of~\citet{Mohammad:13}, and barely outperform the lexicon-based
approach of~\citet{Hu:04} in terms of the micro-averaged \F{} on
SB10k.  Two possible explanations for these mediocre scores could be a
bad starting point of the parameters, which prevented the optimizers
from finding the optimal solution to the optimization objective, or an
insufficient amount of training data, which caused an extreme
overfitting of the training set, but poor generalization to unseen
examples.  We will now investigate both of these factors in detail.

\subsection{Word Embeddings}

As in the previous chapters, we decided to replace randomly
initialized word vectors in the very first layer of vector-based
neural networks with pretrained word2vec embeddings, keeping this
parameter fixed during the optimization.  As we can see from the
figures in Table~\ref{snt-cgsa:tbl:dl-res-word2vec}, this operation
leads to a significant improvement of the results for almost all
classifiers except for the recursive auto-encoder and convolutional
approach of~\citet{Severyn:15}, where it slightly lowers the
micro-averaged \F-score in the former case (from 0.605 to 0.55) and
considerably worsens the macro-averaged \F{} (from 0.608 to 0.36~\F)
of the latter system.  Nonetheless, even despite these exceptional
setbacks, the best observed macro-score increases from 0.608 to 0.64
on the PotTS dataset and almost doubles from 0.321 to 0.53 on the
SB10k data.  A similar situation is observed with the micro-averaged
\F{}, which rises from 0.662 to 0.69 on PotTS and also improves from
0.737 to 0.75 on the SB10k corpus.  Unfortunately, these improvements
usually come at the expense of a lower recall of the majority classes
(positive and neutral respectively), but the gains in the overall
metrics are generally much higher and, first of all, more important
than the losses in these single aspects.

\begin{table}[h]
  \begin{center}
    \bgroup \setlength\tabcolsep{0.1\tabcolsep}\scriptsize
    \begin{tabular}{p{0.162\columnwidth} 
        *{9}{>{\centering\arraybackslash}p{0.074\columnwidth}} 
        *{2}{>{\centering\arraybackslash}p{0.068\columnwidth}}} 
      \toprule
      \multirow{2}*{\bfseries Method} & %
      \multicolumn{3}{c}{\bfseries Positive} & %
      \multicolumn{3}{c}{\bfseries Negative} & %
      \multicolumn{3}{c}{\bfseries Neutral} & %
      \multirow{2}{0.068\columnwidth}{\bfseries\centering Macro\newline \F{}$^{+/-}$} & %
      \multirow{2}{0.068\columnwidth}{\bfseries\centering Micro\newline \F{}}\\
      \cmidrule(lr){2-4}\cmidrule(lr){5-7}\cmidrule(lr){8-10}

      & Precision & Recall & \F{} & %
      Precision & Recall & \F{} & %
      Precision & Recall & \F{} & & \\\midrule

      \multicolumn{12}{c}{\cellcolor{cellcolor}PotTS}\\
      RAE & 0.58\negdelta{0.06} & 0.74\negdelta{0.04} & 0.65\negdelta{0.05} & %
      0.34\negdelta{0.04} & 0.26\posdelta{0.22} & 0.29\posdelta{0.21} & %
      0.59\posdelta{0.02} & 0.46\negdelta{0.22} & 0.52\negdelta{0.1} & %
      0.47\posdelta{0.08} & 0.55\negdelta{0.06}\\

      RNTN & 0.48\posdelta{0.03} & 0.77\negdelta{0.1} & 0.59 & %
      0.33\posdelta{0.14} & 0.03\posdelta{0.01} & 0.06\posdelta{0.03} & %
      0.46\posdelta{0.14} & 0.33\posdelta{0.23} & 0.38\posdelta{0.01} & %
      0.33\posdelta{0.02} & 0.47\posdelta{0.04}\\

      SEV & 0.69\negdelta{0.04} & 0.74\negdelta{0.05} & 0.72\negdelta{0.04} & %
      0.0\negdelta{0.41} & 0.0\negdelta{0.52} & 0.0\negdelta{0.46} & %
      0.58\negdelta{0.14} & 0.84\posdelta{0.29} & 0.69\posdelta{0.07} & %
      0.36\negdelta{0.25} & 0.64\negdelta{0.01}\\

      BAZ & 0.85\posdelta{0.4} & 0.61\negdelta{0.39} & 0.71\posdelta{0.09} & %
      0.57\posdelta{0.57} & 0.32\posdelta{0.32} & 0.41\posdelta{0.41} & %
      0.55\posdelta{0.55} & 0.87\posdelta{0.87} & 0.68\posdelta{0.68} & %
      0.56\posdelta{0.25} & 0.65\posdelta{0.2}\\

      LBA$^{(1)}$ & 0.86\posdelta{0.04} & 0.6\negdelta{0.13} & 0.71\negdelta{0.06} & %
      0.61\posdelta{0.61} & 0.46\posdelta{0.46} & 0.53\posdelta{0.53} & %
      0.6\posdelta{0.04} & 0.89\negdelta{0.03} & 0.72\posdelta{0.03} & %
      0.62\posdelta{0.23} & 0.68\posdelta{0.02}\\

      LBA$^{(2)}$ & 0.84\posdelta{0.39} & 0.65\negdelta{0.35} & 0.73\posdelta{0.11} & %
      0.57\posdelta{0.57} & 0.54\posdelta{0.54} & 0.55\posdelta{0.55} & %
      0.63\posdelta{0.63} & 0.82\posdelta{0.82} & 0.72\posdelta{0.72} & %
      0.64\posdelta{0.33} & 0.69\posdelta{0.24}\\

      \multicolumn{12}{c}{\cellcolor{cellcolor}SB10k}\\
      RAE & 0.61\negdelta{0.02} & 0.56\negdelta{0.01} & 0.58\negdelta{0.02} & %
      0.29\posdelta{0.29} & 0.01\posdelta{0.01} & 0.02\posdelta{0.02} & %
      0.74\negdelta{0.01} & 0.92\negdelta{0.02} & 0.82\negdelta{0.01} & %
      0.3 & 0.71\negdelta{0.01}\\

      RNTN & 0.54\posdelta{0.34} & 0.02\negdelta{0.01} & 0.04\negdelta{0.01} & %
      0.0\negdelta{0.07} & 0.0\negdelta{0.01} & 0.0\negdelta{0.02} & %
      0.63\posdelta{0.01} & 1.0\posdelta{0.06} & 0.77\posdelta{0.02} & %
      0.02\negdelta{0.01} & 0.62\posdelta{0.03}\\

      SEV & 0.72\posdelta{0.72} & 0.5\posdelta{0.5} & 0.59\posdelta{0.59} & %
      0.49\posdelta{0.49} & 0.27\posdelta{0.27} & 0.35\posdelta{0.35} & %
      0.75\negdelta{0.13} & 0.92\negdelta{0.08} & 0.82\posdelta{0.05} & %
      0.47\posdelta{0.47} & 0.73\posdelta{0.11}\\

      BAZ & 0.78\posdelta{0.03} & 0.51\posdelta{0.04} & 0.61\posdelta{0.03} & %
      0.49\posdelta{0.49} & 0.42\posdelta{0.42} & 0.45\posdelta{0.45} & %
      0.78\posdelta{0.07} & 0.91\negdelta{0.07} & 0.84\posdelta{0.01} & %
      0.53\posdelta{0.24} & 0.75\posdelta{0.03}\\

      LBA$^{(1)}$ & 0.84\posdelta{0.12} & 0.42\negdelta{0.16} & 0.56\negdelta{0.08} & %
      0.5\posdelta{0.5} & 0.28\posdelta{0.28} & 0.36\posdelta{0.36} & %
      0.74 & 0.96\negdelta{0.01} & 0.84 & %
      0.46\posdelta{0.14} & 0.73\posdelta{0.01}\\

      LBA$^{(2)}$ & 0.79\posdelta{0.03} & 0.45\negdelta{0.04} & 0.57\negdelta{0.03} & %
      0.57\posdelta{0.57} & 0.23\posdelta{0.23} & 0.33\posdelta{0.33} & %
      0.74\posdelta{0.02} & 0.96\negdelta{0.02} & 0.84\posdelta{0.01} & %
      0.45\posdelta{0.15} & 0.74\posdelta{0.02}\\\bottomrule

    \end{tabular}
    \egroup
    \caption[Results of DL-based MLSA methods with pretrained word2vec
      vectors]{Results of deep-learning--based MLSA methods with
      pretrained word2vec vectors}
    \label{snt-cgsa:tbl:dl-res-word2vec}
  \end{center}
\end{table}

In order to see whether these changes would be different if we
optimized word representations as well, we reran our experiments once
again, initializing word vectors with word2vec embeddings as before,
but allowing them to be updated during the training.  Moreover, to
approximate task-specific representations of words that were missing
from the training set, we also computed the optimal transformation
matrix for converting the original word2vec vectors into optimized
sentiment embeddings using the method of the ordinary least squares,
as we did in the previous chapters, and used this matrix to derive
task-specific vectors during the testing.

As suggested by the results in Table~\ref{snt-cgsa:tbl:dl-res-lstsq},
these modifications improve the results even further, setting a new
record of the macro-averaged \F-scores on the PotTS corpus (0.69~\F),
and pushing our LBA$^{(1)}$ system even above its most challenging
competitors.  A similar effect is also observed with other systems,
first of all BAZ and LBA$^{(2)}$, which yield similarly good results
for all polarities.  Nevertheless, like in the previous case, these
improvements usually cause a drop in the recall for the most frequent
class of the respective dataset, which is especially severe for the
system of \citeauthor{Baziotis:17} on the PotTS data (-0.28 on
positive messages) and the LBA$^{(1)}$ approach on the SB10k test set
(-0.17 on neutral tweets).  Furthermore, the convolutional system of
\citeauthor{Severyn:15} and recursive neural tensor approach
of~\citet{Socher:13} fail to predict any negative tweet on PotTS and
SB10k, respectively, which also leads to a notable drop of their
overall macro-\F--values.  These drops, however, are rather
exceptional, as the same system of \citeauthor{Severyn:15} shows an
extraordinary big boost of the results on the SB10k corpus (+0.45
macro-\F{} and +0.1 micro-\F--score), and the macro-averaged \F-values
of all recurrent methods also become twice as high as in the case of
randomly initialized word vectors.

\begin{table}[h]
  \begin{center}
    \bgroup\setlength\tabcolsep{0.1\tabcolsep}\scriptsize
    \begin{tabular}{p{0.162\columnwidth} 
        *{9}{>{\centering\arraybackslash}p{0.074\columnwidth}} 
        *{2}{>{\centering\arraybackslash}p{0.068\columnwidth}}} 
      \toprule
      \multirow{2}*{\bfseries Method} & %
      \multicolumn{3}{c}{\bfseries Positive} & %
      \multicolumn{3}{c}{\bfseries Negative} & %
      \multicolumn{3}{c}{\bfseries Neutral} & %
      \multirow{2}{0.068\columnwidth}{\bfseries\centering Macro\newline \F{}$^{+/-}$} & %
      \multirow{2}{0.068\columnwidth}{\bfseries\centering Micro\newline \F{}}\\
      \cmidrule(lr){2-4}\cmidrule(lr){5-7}\cmidrule(lr){8-10}

      & Precision & Recall & \F{} & %
      Precision & Recall & \F{} & %
      Precision & Recall & \F{} & & \\\midrule

      \multicolumn{12}{c}{\cellcolor{cellcolor}PotTS}\\
      RAE & 0.61\negdelta{0.03} & 0.61\negdelta{0.17} & 0.61\negdelta{0.09} & %
      0.22\negdelta{0.16} & 0.01\negdelta{0.03} & 0.03\negdelta{0.05} & %
      0.48\negdelta{0.09} & 0.72\negdelta{0.04} & 0.57\negdelta{0.05} & %
      0.32\negdelta{0.07} & 0.54\negdelta{0.07}\\

      RNTN & 0.45 & 0.82\negdelta{0.05} & 0.59 & %
      0.24\posdelta{0.05} & 0.06\negdelta{0.04} & 0.1\negdelta{0.07} & %
      0.43\posdelta{0.09} & 0.17\posdelta{0.07} & 0.24\posdelta{0.09} & %
      0.34\posdelta{0.03} & 0.44\negdelta{0.01}\\

      SEV & 0.73 & 0.74\negdelta{0.05} & 0.74\negdelta{0.02} & %
      0.0\negdelta{0.41} & 0.0\negdelta{0.52} & 0.0\negdelta{0.46} & %
      0.56\negdelta{0.16} & 0.84\posdelta{0.29} & 0.68\posdelta{0.06} & %
      0.37\negdelta{0.24} & 0.64\negdelta{0.01}\\

      BAZ & 0.82\posdelta{0.37} & 0.72\negdelta{0.28} & 0.77\posdelta{0.15} & %
      0.62\posdelta{0.62} & 0.49\posdelta{0.49} & 0.55\posdelta{0.55} & %
      0.68\posdelta{0.68} & 0.85\posdelta{0.85} & 0.76\posdelta{0.76} & %
      0.66\posdelta{0.35} & 0.73\posdelta{0.28}\\

      LBA$^{(1)}$ & 0.76\negdelta{0.06} & 0.84\posdelta{0.11} & 0.79\posdelta{0.02} & %
      0.6\posdelta{0.6} & 0.56\posdelta{0.56} & 0.58\posdelta{0.58} & %
      0.75\posdelta{0.19} & 0.68\negdelta{0.24} & 0.72\posdelta{0.03} & %
      0.69\posdelta{0.3} & 0.73\posdelta{0.07}\\

      LBA$^{(2)}$ & 0.84\posdelta{0.39} & 0.73\negdelta{0.27} & 0.78\posdelta{0.16} & %
      0.57\posdelta{0.57} & 0.48\posdelta{0.48} & 0.53\posdelta{0.53} & %
      0.66\posdelta{0.66} & 0.82\posdelta{0.82} & 0.73\posdelta{0.73} & %
      0.65\posdelta{0.34} & 0.72\posdelta{0.27}\\

      \multicolumn{12}{c}{\cellcolor{cellcolor}SB10k}\\
      RAE & 0.5\negdelta{0.13} & 0.73\posdelta{0.16} & 0.59\negdelta{0.01} & %
      0.35\posdelta{0.35} & 0.06\posdelta{0.06} & 0.1\posdelta{0.1} & %
      0.8\posdelta{0.05} & 0.8\negdelta{0.14} & 0.8\negdelta{0.03} & %
      0.35\posdelta{0.15} & 0.68\negdelta{0.04}\\

      RNTN & 0.0\negdelta{0.02} & 0.0\negdelta{0.03} & 0.0\negdelta{0.05} & %
      0.0\negdelta{0.07} & 0.0\negdelta{0.01} & 0.0\negdelta{0.02} & %
      0.62 & 1.0\negdelta{0.06} & 0.77\negdelta{0.02} & %
      0.0\negdelta{0.03} & 0.62\posdelta{0.03}\\

      SEV & 0.64\posdelta{0.64} & 0.58\posdelta{0.58} & 0.61\posdelta{0.61} & %
      0.51\posdelta{0.51} & 0.21\posdelta{0.21} & 0.3\posdelta{0.3} & %
      0.76\posdelta{0.14} & 0.89\negdelta{0.11} & 0.82\posdelta{0.05} & %
      0.45\posdelta{0.45} & 0.72\posdelta{0.1}\\

      BAZ & 0.72\posdelta{0.03} & 0.59\posdelta{0.12} & 0.65\posdelta{0.07} & %
      0.53\posdelta{0.53} & 0.33\posdelta{0.33} & 0.41\posdelta{0.41} & %
      0.79\posdelta{0.08} & 0.91\negdelta{0.07} & 0.84\posdelta{0.01} & %
      0.53\posdelta{0.24} & 0.75\posdelta{0.03}\\

      LBA$^{(1)}$ & 0.6\negdelta{0.12} & 0.72\posdelta{0.14} & 0.66\posdelta{0.02} & %
      0.47\posdelta{0.47} & 0.42\posdelta{0.42} & 0.44\posdelta{0.44} & %
      0.84\posdelta{0.1} & 0.8\negdelta{0.17} & 0.82\posdelta{0.02} & %
      0.55\posdelta{0.23} & 0.73\negdelta{0.01}\\

      LBA$^{(2)}$ & 0.72\negdelta{0.04} & 0.57\posdelta{0.08} & 0.64\posdelta{0.04} & %
      0.55\posdelta{0.55} & 0.39\posdelta{0.39} & 0.46\posdelta{0.46} & %
      0.79\posdelta{0.07} & 0.9\negdelta{0.08} & 0.84\posdelta{0.01} & %
      0.55\posdelta{0.25} & 0.75\posdelta{0.03}\\\bottomrule
    \end{tabular}
    \egroup{}
    \caption[Results of DL-based MLSA methods with least-squares
      embeddings]{Results of deep-learning--based MLSA methods with
      least-squares embeddings}\label{snt-cgsa:tbl:dl-res-lstsq}
  \end{center}
\end{table}

\subsection{Error Analysis}

Before we proceed with the evaluation of the second factor (larger
training set), let us first analyze some errors that were specific to
each of the classifiers trained with the least-squares embeddings.

Since interpreting and understanding the results of deep learning
systems is a complex task due to a big number of model parameters and
unobvious correlations between them, we decided to use the
\textsc{Lime} package~\cite{Ribeiro:16}, a recently proposed
model-agnostic interpretation tool, to get a better intuition about
the reasons of the classifiers' decisions.  To derive an explanation
for a particular prediction, \textsc{Lime} randomly removes or
perturbs parts of the input (in our case, tokens), estimating which of
these modifications lead to the biggest changes in the output, and
assigns corresponding class-specific association scores to each of the
changed parts.  The higher this score, the more predictive is the
given feature for that particular label.  For the sake of vividness,
we have highlighted all tokens that, according to \textsc{Lime}, were
associated with the neutral class as white, marked negative attributes
with the \colorbox{blue!30}{blue} background, and highlighted
positively connoted words in \colorbox{green!30}{green}, reflecting
the respective association strength with a higher color brightness.

The first incorrect prediction shown in
Example~\ref{snt:cgsa:exmp:rae-error} was made by the RAE system
of~\citet{Socher:11}.  As we can see from the visualization, the model
correctly recognized the positive term ``gef\"allt'' (\emph{to like}),
but, unfortunately, this word is the only one which contributes to the
right decision, and its learned weight is obviously not enough to
outdo the effect of multiple neutral and negative items, such as
``Gr\"un'' (\emph{green}), ``Schwarz'' (\emph{black}), and most
surprisingly ``PosSmiley\%'' (\emph{PosSmiley\%}), which unexpectedly
is stronger associated with the negative semantic orientation than
with the positive class.  As a consequence of this, the classifier
erroneously predicts the \textsc{neutral} label for the whole message,
falling against the prevalence of allegedly objective terms.

\begin{example}[An Error Made by the RAE System]\label{snt:cgsa:exmp:rae-error}
  \noindent\textup{\bfseries\textcolor{darkred}{Tweet:}} {\upshape
    \colorbox{white!5}{Gr\"un}\colorbox{blue!11}{-}\colorbox{white!25.7}{Schwarz}
    \colorbox{white!5}{in} \colorbox{white!2.5!blue!1}{meinem}
    Bundesland. \colorbox{green!10}{Gef\"allt}
    \colorbox{white!5!blue!4.5}{mir} \colorbox{white!5!blue!5}{doch}
    \colorbox{white!4.6!blue!5}{sehr}
    \colorbox{white!5!blue!5}{\%PosSmiley}}\\
  \noindent \colorbox{white!5}{Green}\colorbox{blue!11}{-}\colorbox{white!25.7}{Black} \colorbox{white!5}{in} \colorbox{white!2.5!blue!1}{my} state.  \colorbox{white!5!blue!5}{Yet}, \colorbox{white!5!blue!4.5}{I} \colorbox{green!10}{like} it \colorbox{white!4.6!blue!5}{so much} \colorbox{white!5!blue!5}{\%PosSmiley}\\[\exampleSep]
  \noindent\textup{\bfseries\textcolor{darkred}{Gold Label:}}\hspace*{4.3em}\textbf{%
    \upshape\textcolor{green3}{positive}}\\
 \noindent\textup{\bfseries\textcolor{darkred}{Predicted Label:}}\hspace*{2em}\textbf{%
    \upshape\textcolor{black}{neutral*}}
\end{example}

A similar situation is also observed with the recurrent neural tensor,
whose sample error is shown in
Example~\ref{snt:cgsa:exmp:socher13-error}.  As we can see from the
analysis, the bias towards the neutral class is even more pronounced
this time, as virtually all of the terms in the tweet are highlighted
in white.  The only word which shows a minimal negative connotation is
``tumblr,'' which indeed appeared twice in a negative tweet, two times
in neutral messages, and once in a positive microblog in the training
corpus.  Nonetheless, even for this term the skewness towards the
neutral orientation is still ten times bigger than its association
with the negative polarity ($\expnumber{1.4}{-4}$ versus
$\expnumber{1.5}{-5}$), which can be explained by the general
prevalence of neutral messages in SB10k.

\begin{example}[An Error Made by the RNTN System]\label{snt:cgsa:exmp:socher13-error}
  \noindent\textup{\bfseries\textcolor{darkred}{Tweet: }} {\upshape
    \colorbox{blue!1.4!white!5}{tumblr} \colorbox{white!19}{people} \colorbox{white!32}{sind} \colorbox{white!5}{meine} \colorbox{white!30}{lieblings} \colorbox{white!19}{people} \colorbox{white!8}{\%PosSmiley}}\\
  \noindent \colorbox{blue!1.4!white!5}{tumblr} \colorbox{white!19}{people} \colorbox{white!32}{are} \colorbox{white!5}{my} \colorbox{white!30}{favorite} \colorbox{white!19}{people} \colorbox{white!8}{\%PosSmiley}\\[\exampleSep]
  \noindent\textup{\bfseries\textcolor{darkred}{Gold Label:}}\hspace*{4.3em}\textbf{%
    \upshape\textcolor{green3}{positive}}\\
 \noindent\textup{\bfseries\textcolor{darkred}{Predicted Label:}}\hspace*{2em}\textbf{%
    \upshape\textcolor{black}{neutral*}}
\end{example}

A slightly different behavior is shown by the method
of~\citet{Severyn:15} on the PotTS corpus.  This time, we can see at
least two clearly positive words (``ist'' [\emph{is}] and ``Freund''
[\emph{friend}]).  However, the former of these terms is an auxiliary
copular verb, which can hardly express any polarity, since it usually
plays an auxiliary role and lacks any distinct lexical meaning.
Nevertheless, the latter word (``Freund'' [\emph{friend}]) indeed
conveys a positive feeling of its prepositional argument (``Iran'')
towards the subject of the sentence (``Syrien'' [\emph{Syria}]), but
this positive effect is nullified by the author's statement that this
friendship poses a problem. Unfortunately, the word ``Problem''
(\emph{problem}) is recognized only as a neutral marker, just like
many other terms in this microblog.

\begin{example}[An Error Made by the SEV System]\label{snt:cgsa:exmp:severyn-error}
  \noindent\textup{\bfseries\textcolor{darkred}{Tweet:}} {\upshape
    \colorbox{white!7.4}{Syrien} \colorbox{green!8}{ist}
    \colorbox{green!11}{Freund} \colorbox{white!6.6}{von}
    \colorbox{white!40.5}{Iran}, das \colorbox{green!8}{ist}
    \colorbox{white!12.7}{das}
    \colorbox{green!1.5}{Problem}\colorbox{green!3}{!}
    \colorbox{blue!0.000005!white!8}{annewill}}\\
  \noindent \colorbox{white!7.4}{Syria} \colorbox{green!8}{is} a \colorbox{green!11}{friend} \colorbox{white!6.6}{of} \colorbox{white!40.5}{Iran}. That\colorbox{green!8}{'s} \colorbox{white!12.7}{the} \colorbox{green!1.5}{problem}\colorbox{green!3}{!} \colorbox{blue!0.000005!white!8}{annewill}\\[\exampleSep]
  \noindent\textup{\bfseries\textcolor{darkred}{Gold Label:}}\hspace*{4.3em}\textbf{%
    \upshape\textcolor{midnightblue}{negative}}\\
 \noindent\textup{\bfseries\textcolor{darkred}{Predicted Label:}}\hspace*{2em}\textbf{%
    \upshape\textcolor{black}{neutral*}}
\end{example}

In Example~\ref{snt:cgsa:exmp:baziotis-error}, we can see another
error made by the system of~\citet{Baziotis:17}.  This time, again, we
observe the prevalence of positive and neutral items, with the only
exception being the possessive pronoun ``meinen'' (\emph{my}), which,
according to the classifier, indicates negative polarity.  Apart from
this term, we also can notice several inaccuracies at recognizing
positive and neutral features: For example, the pronominal adverb
``darin'' (\emph{in it}) is the strongest positive trait, whose
predictiveness is even higher than the scores of the words ``singen''
(\emph{to sing}) and ``Liebeslied'' (\emph{love song}).  This
contradicts the fact that pronominal adverbs by themselves do not
express any semantic orientation, all the more as in this case the
antecedent of the adverb (the noun ``Kleiderschrank''
[\emph{wardrobe}]) is recognized as a neutral item.  On the other
hand, the modal verb ``wollte'' (\emph{wanted}) is considered as an
objective term, although it has a slight positive connotation as it
expresses a wish of the author.

\begin{example}[An Error Made by the BAZ System]\label{snt:cgsa:exmp:baziotis-error}
  \noindent\textup{\bfseries\textcolor{darkred}{Tweet:}} {\upshape \colorbox{white!1.4}{Wollte} \colorbox{blue!7.7}{meinen} \colorbox{white!3.6}{Kleiderschrank} \colorbox{blue!1.7}{aufr\"aumen} \ldots \colorbox{white!1.2}{sitze} \colorbox{green!4.6}{nun} \colorbox{green!31.5}{darin} \colorbox{green!2.8}{und} \colorbox{green!29.7}{singe} \colorbox{green!15.2}{Liebeslieder}\ldots}\\
  \noindent \colorbox{white!1.4}{Wanted} to \colorbox{blue!1.7}{clean up} \colorbox{blue!7.7}{my} \colorbox{white!3.6}{wardrobe}\ldots \colorbox{green!4.6}{Now} \colorbox{white!1.2}{sitting} \colorbox{green!31.5}{in it} \colorbox{green!2.8}{and} \colorbox{green!29.7}{singing} \colorbox{green!15.2}{love songs}\ldots\\[\exampleSep]
  \noindent\textup{\bfseries\textcolor{darkred}{Gold Label:}}\hspace*{4.3em}\textbf{%
    \upshape\textcolor{black}{neutral}}\\
 \noindent\textup{\bfseries\textcolor{darkred}{Predicted Label:}}\hspace*{2em}\textbf{%
    \upshape\textcolor{green3}{positive*}}
\end{example}

Finally, Example~\ref{snt:cgsa:exmp:lba-error} shows an incorrect
prediction of our lexicon-based attention system.  In contrast to the
previous two methods, the positive information is much more condensed
in this case and represented by a single term ``super.''
Surprisingly, this term outweighs a whole bunch of neutral features
such as ``gerade'' (\emph{right now}), ``Lust haben'' (\emph{to be up
  to}), ``was'' (\emph{something}) etc.  Admittedly, the first part of
this message indeed expresses a positive attitude of the author, but
this effect is invalidated by the second clause, which shows the
impossibility of that wish.

\begin{example}[An Error Made by the LBA System]\label{snt:cgsa:exmp:lba-error}
  \noindent\textup{\bfseries\textcolor{darkred}{Tweet:}} {\upshape
    \colorbox{green!0.5!blue!0.4}{Gerade} \colorbox{green!89}{super} \colorbox{blue!0.3}{Lust}, mit \colorbox{white!2}{Carls} Haaren \colorbox{white!0.6}{was} zu \colorbox{green!1}{machen} \colorbox{green!0.3}{aber} \colorbox{white!2}{ca} 300 \colorbox{white!1}{km}
    \colorbox{white!1}{Distanz} halten \colorbox{blue!0.3}{mich} davon \colorbox{white!1}{ab}.}\\
  \noindent\colorbox{green!89}{Super} \colorbox{blue!0.3}{up to}
  \colorbox{green!1}{do} \colorbox{white!0.6}{something} with
  \colorbox{white!2}{Carl}'s hair \colorbox{green!0.5!blue!0.4}{right
    now}, \colorbox{green!0.3}{but} \colorbox{white!2}{ca.} 300
  \colorbox{white!1}{km} \colorbox{white!1}{distance} keep
  \colorbox{blue!0.3}{me} \colorbox{white!1}{off} from
  this.\\[\exampleSep]
  \noindent\textup{\bfseries\textcolor{darkred}{Gold Label:}}\hspace*{4.3em}\textbf{%
    \upshape\textcolor{black}{neutral}}\\
 \noindent\textup{\bfseries\textcolor{darkred}{Predicted Label:}}\hspace*{2em}\textbf{%
    \upshape\textcolor{green3}{positive*}}
\end{example}

\section{Evaluation}

Now that we have familiarized ourselves with the peculiarities and
results of the most prominent sentiment analysis approaches from all
method groups (lexicon-, machine-learning-- and deep-learning--based
ones), let us have a closer look at how changing different common
parameters of these methods might affect their performance.  In
particular, we would like to see whether increasing the amount of the
training data, switching to a different type of sentiment lexicon, or
using unnormalized text as input would improve or, vice versa, lower
the classification scores.

\subsection{Weak Supervision}

The first avenue that we are going to explore in this evaluation is
the effect of weakly supervised data---an additional collection of
training tweets that have been automatically labeled with sentiment
tags based on the occurrence of some sufficiently reliable formal
criteria, such as emoticons or hashtags.

Among the first who proposed the idea of training a sentiment
classifier on a larger corpus of automatically annotated messages was
\citet{Read:05}, who gathered a set of 766,000 Usenet posts containing
frownies or smileys, assigned a polarity label to each of these posts,
judging by the type of the emoticons, and subsequently used a subset
of these documents (22,000 posts) to optimize a Na\"{\i}ve Bayes and
SVM system.  Even though these classifiers could achieve a
considerable accuracy (up to 70\%) on predicting noisy labels of the
remaining posts, they could not generalize to texts from other genres
(movie reviews and newswire articles) where they hardly outperformed
the random-chance baseline. With the onset of the Twitter era, this
idea of weak supervision has experienced its renaissance with the
works of \citet{Go:09}, \citet{Pak:10}, and \citet{Barbosa:10}.

In order to check the effect of such noisily annotated data on our
tested methods, we also automatically labeled all messages from the
German Twitter Snapshot~\cite{Scheffler:14} based on the occurrences
of smileys: In particular, we considered a microblog as positive if
its normalized version contained the token \texttt{\%PositiveSmiley}
with no other facial expressions.  Likewise, we regarded a message as
negative if the only emoticon in this tweet was
\texttt{\%NegativeSmiley}.  We skipped all posts that contained both
types of smileys, and assigned the rest of the messages to the neutral
class. (A detailed breakdown of the final distribution is given in
Table~\ref{snt-cgsa:tbl:corp-dist} at the beginning of this chapter.)

Since it was impossible to utilize the whole snapshot for the training
due to limited computational resources (only reading the dataset into
memory would require 9.3Gb RAM, not to mention the space required for
storing the embeddings and features), we confined ourselves to one
sixth of these data, which still resulted in 4~M messages.
Furthermore, to mitigate the extreme skewness of this corpus, we
downsampled positive and neutral tweets to get an equal number of
instances for all classes (59,000 microblogs for each polarity) and
used these examples in addition to the manually analyzed PotTS and
SB10k tweets.

Since lexicon-based approaches were mostly independent of the training
set, we decided to rerun our experiments only with ML- and fastest
DL-based methods (RNN, SEV, BAZ, and LBA),\footnote{In all subsequent
  evaluation experiments with DL-based systems, we will use
  pre-trained word2vec vectors (if applicable) with the least-squares
  fallback, and compare the results of these approaches to the
  respective scores in Table~\ref{snt-cgsa:tbl:dl-res-lstsq}.} which
still incurred running times up to five days for some systems.  The
results of this evaluation are shown in
Table~\ref{snt-cgsa:tbl:weak-supervision}.

\begin{table}[h]
  \begin{center}
    \bgroup\setlength\tabcolsep{0.1\tabcolsep}\scriptsize
    \begin{tabular}{p{0.162\columnwidth} 
        *{9}{>{\centering\arraybackslash}p{0.074\columnwidth}} 
        *{2}{>{\centering\arraybackslash}p{0.068\columnwidth}}} 
      \toprule
      \multirow{2}*{\bfseries Method} & %
      \multicolumn{3}{c}{\bfseries Positive} & %
      \multicolumn{3}{c}{\bfseries Negative} & %
      \multicolumn{3}{c}{\bfseries Neutral} & %
      \multirow{2}{0.068\columnwidth}{\bfseries\centering Macro\newline \F{}$^{+/-}$} & %
      \multirow{2}{0.068\columnwidth}{\bfseries\centering Micro\newline \F{}}\\
      \cmidrule(lr){2-4}\cmidrule(lr){5-7}\cmidrule(lr){8-10}

      & Precision & Recall & \F{} & %
      Precision & Recall & \F{} & %
      Precision & Recall & \F{} & & \\\midrule

      \multicolumn{12}{c}{\cellcolor{cellcolor}PotTS}\\

      GMN & 0.8\posdelta{0.13} & 0.34\negdelta{0.39} & 0.48\negdelta{0.22} & %
       0.2\negdelta{0.15} & 0.29\posdelta{0.14} & 0.24\negdelta{0.03} & %
       0.53\negdelta{0.07} & 0.79\posdelta{0.07} & 0.63\negdelta{0.03} & %
       0.36\negdelta{0.01} & 0.49\negdelta{0.12}\\


      MHM & 0.86\posdelta{0.07} & 0.59\negdelta{0.18} & 0.7\negdelta{0.08} & %
      0.31\negdelta{0.27} & 0.39\negdelta{0.17} & 0.35\negdelta{0.22} & %
      0.55\negdelta{0.18} & 0.68\negdelta{0.08} & 0.61\negdelta{0.13} & %
      0.52\negdelta{0.15} & 0.59\negdelta{0.14}\\


      GNT & 0.86\posdelta{0.15} & 0.6\negdelta{0.2} & 0.71\negdelta{0.04} & %
      0.26\negdelta{0.29} & 0.31\negdelta{0.14} & 0.28\negdelta{0.22} & %
      0.53\negdelta{0.15} & 0.68\negdelta{0.05} & 0.59\negdelta{0.06} & %
      0.5\negdelta{0.12} & 0.57\negdelta{0.1}\\

      RAE & 0.68\posdelta{0.07} & 0.31\negdelta{0.3} & 0.43\negdelta{0.18} & %
      0.25\posdelta{0.03} & 0.46\posdelta{0.45} & 0.32\posdelta{0.29} & %
      0.49\posdelta{0.01} & 0.61\negdelta{0.11} & 0.54\negdelta{0.03} & %
      0.38\posdelta{0.06} & 0.45\negdelta{0.09}\\

      SEV & 0.87\posdelta{0.14} & 0.51\negdelta{0.23} & 0.64\negdelta{0.1} & %
      0.27\posdelta{0.27} & 0.49\posdelta{0.49} & 0.35\posdelta{0.35} & %
      0.55\negdelta{0.01} & 0.58\negdelta{0.26} & 0.56\negdelta{0.12} & %
      0.49\posdelta{0.12} & 0.53\negdelta{0.11}\\

      BAZ & 0.0\negdelta{0.82} & 0.0\negdelta{0.72} & 0.0\negdelta{0.77} & %
      0.19\negdelta{0.43} & 1.0\posdelta{0.51} & 0.32\negdelta{0.23} & %
      0.0\negdelta{0.68} & 0.0\negdelta{0.85} & 0.0\negdelta{0.76} & %
      0.16\negdelta{0.5} & 0.19\negdelta{0.43}\\

      LBA$^{(1)}$ & 0.48\negdelta{0.28} & 0.88\posdelta{0.04} & 0.62\negdelta{0.17} & %
      0.25\negdelta{0.35} & 0.23\negdelta{0.33} & 0.24\negdelta{0.34} & %
      0.0\negdelta{0.75} & 0.0\negdelta{0.68} & 0.0\negdelta{0.72} & %
      0.43\negdelta{0.26} & 0.44\negdelta{0.29}\\

      LBA$^{(2)}$ & 0.91\posdelta{0.07} & 0.08\negdelta{0.65} & 0.14\negdelta{0.64} & %
      0.19\negdelta{0.38} & 0.99\posdelta{0.51} & 0.32\negdelta{0.21} & %
      0.0\negdelta{0.66} & 0.0\negdelta{0.82} & 0.0\negdelta{0.73} & %
      0.23\negdelta{0.42} & 0.22\negdelta{0.5}\\

      \multicolumn{12}{c}{\cellcolor{cellcolor}SB10k}\\


      GMN & 0.71\posdelta{0.06} & 0.27\negdelta{0.18} & 0.4\negdelta{0.13} & %
      0.24\negdelta{0.14} & 0.11\posdelta{0.03} & 0.15\posdelta{0.02} & %
      0.71\negdelta{0.01} & 0.96\posdelta{0.03} & 0.82\negdelta{0.01} & %
      0.27\negdelta{0.06} & 0.68\negdelta{0.02}\\

      MHM & 0.77\posdelta{0.06} & 0.4\negdelta{0.25} & 0.53\negdelta{0.15} & %
      0.61\negdelta{0.1} & 0.1\negdelta{0.3} & 0.18\negdelta{0.27} & %
      0.71\negdelta{0.09} & 0.97\negdelta{0.1} & 0.82\negdelta{0.02} & %
      0.35\negdelta{0.21} & 0.71\negdelta{0.04}\\


      GNT & 0.77\posdelta{0.1} & 0.39\negdelta{0.23} & 0.52\negdelta{0.12} & %
      0.25\negdelta{0.19} & 0.13\negdelta{0.15} & 0.17\negdelta{0.17} & %
      0.71\negdelta{0.07} & 0.92\posdelta{0.05} & 0.8\negdelta{0.02} & %
      0.34\negdelta{0.15} & 0.68\negdelta{0.04}\\

      RAE & 0.44\negdelta{0.06} & 0.27\negdelta{0.51} & 0.34\negdelta{0.25} & %
      0.24\negdelta{0.11} & 0.59\posdelta{0.53} & 0.34\posdelta{0.24} & %
      0.78\negdelta{0.02} & 0.62\negdelta{0.18} & 0.69\negdelta{0.11} & %
      0.34\negdelta{0.01} & 0.54\negdelta{0.14}\\

      SEV & 0.64 & 0.39\negdelta{0.19} & 0.49\negdelta{0.12} & %
      0.34\negdelta{0.17} & 0.12\negdelta{0.09} & 0.18\negdelta{0.12} & %
      0.7\negdelta{0.06} & 0.9\posdelta{0.01} & 0.78\negdelta{0.04} & %
      0.33\negdelta{0.12} & 0.69\negdelta{0.03}\\

      BAZ & 0.24\negdelta{0.48} & 1.0\posdelta{0.41} & 0.38\negdelta{0.27} & %
      0.0\negdelta{0.53} & 0.0\negdelta{0.33} & 0.0\negdelta{0.41} & %
      0.0\negdelta{0.79} & 0.0\negdelta{0.91} & 0.0\negdelta{0.84} & %
      0.19\negdelta{0.34} & 0.24\negdelta{0.51}\\

     LBA$^{(1)}$ & 0.64\posdelta{0.04} & 0.43\negdelta{0.29} & 0.52\negdelta{0.14} & %
      0.59\posdelta{0.12} & 0.09\negdelta{0.33} & 0.16\negdelta{0.28} & %
      0.71\negdelta{0.13} & 0.93\posdelta{0.13} & 0.8\negdelta{0.02} & %
      0.34\negdelta{0.21} & 0.69\negdelta{0.04}\\

      LBA$^{(2)}$ & 0.0\negdelta{0.72} & 0.0\negdelta{0.57} & 0.0\negdelta{0.64} & %
      0.14\negdelta{0.41} & 1.0\posdelta{0.61} & 0.25\negdelta{0.21} & %
      0.0\negdelta{0.79} & 0.0\negdelta{0.9} & 0.0\negdelta{0.84} & %
      0.12\negdelta{0.43} & 0.14\negdelta{0.61}\\\bottomrule
    \end{tabular}
    \egroup{}
    \caption[Results of MLSA methods with weak supervision]{
      Results of MLSA methods with weakly supervised data}\label{snt-cgsa:tbl:weak-supervision}
  \end{center}
\end{table}

As we can see from the scores, apart from improved precision of
positive tweets and higher recall of negative microblogs, adding
noisily labeled messages to the training set has a strong negative
effect on the results of all methods, with the biggest drops
demonstrated by the approach of~\citeauthor{Baziotis:17} ($-0.5$
macro-\F{} and $-0.43$ micro-\F{} on the PotTS corpus; $-0.34$
macro-\F{} and $-0.51$ micro-\F{} on the SB10k dataset) and our own
LBA$^{(2)}$ solution ($-0.42$ macro-\F{}-score and $-0.5$ micro-\F{}
on the PotTS test set; $-0.43$ macro-\F{} and $-0.61$ micro-\F{} on
the SB10k data), which both fail to predict any neutral message on
PotTS and always assign the same polarity to all SB10k tweets.  Less
severe, but still substantial degradation is also observed with the
machine-learning systems of \citeauthor{Mohammad:09} and
\citeauthor{Guenther:13} as well as our DL-based LBA$^{(1)}$ method,
whose macro-averaged \F{}-scores go down by 0.15, 0.12, and 0.26
points on the former corpus and sink by 0.21, 0.15, 0.21,
respectively, on the latter dataset.  The micro-averaged \F{}-results
of these methods, however, decrease to a much smaller degree, since
the main drops happen on the negative class, which is by far the least
represented polarity in both corpora.  The micro-averages of the
remaining systems seem to be affected even less, but are still worse
than the results obtained without the snapshot data.  We hypothesize
that the main reason for this decrease is a substantial difference
between the class distributions in noisily annotated training tweets
and manually labeled test sets, which overly bias classifiers'
predictions.

\subsection{Lexicons}\label{cgsa:subsec:eval:lexicons}

Another factor that could significantly affect the results of some
systems was the sentiment lexicon that these systems used either
directly, for computing the polarity of a message (\eg{} lexicon-based
approaches), or indirectly, as features or attention scores (\eg{} ML-
and DL-based techniques).  To estimate the effect of this resource, we
successively replaced the lexicons that we used in our previous
experiments with other polarity lists presented in
Chapter~\ref{chap:snt:lex}, and recomputed the scores of the tested
systems.

\begin{figure*}
{
\centering
\begin{subfigure}{.5\textwidth}
  \centering
  \includegraphics[width=\linewidth]{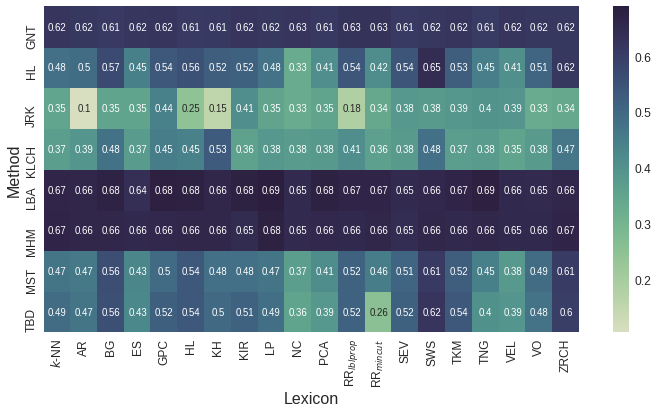}
  \caption{\texttt{Macro-\F{}}}\label{cgsa:fig:potts-lexicon-macro}
\end{subfigure}%
\begin{subfigure}{.5\textwidth}
  \centering
  \includegraphics[width=\linewidth]{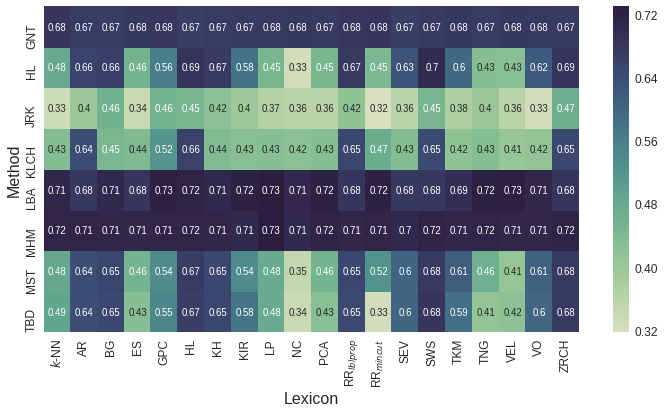}
  \caption{\texttt{Micro-\F{}}}\label{cgsa:fig:potts-lexicon-micro}
\end{subfigure}
}
\caption[MLSA results on PotTS with different lexicons]{Results of
  MLSA methods with different lexicons on the PotTS
  corpus}\label{cgsa:fig:potts-lexicon-effect}
\end{figure*}

As we can see in Figure~\ref{cgsa:fig:potts-lexicon-effect}, the
system of~\citet{Mohammad:13} and our own lexicon-based attention
approach clearly outperform all other competitors on the PotTS corpus
independent of the lexicon they use.  The only method that comes at
least close to their results is the ML-based classifier
of~\citet{Guenther:14}, which is still almost 5\% below the average
macro-\F{} of these two classifiers.  The same also applies to the
micro-\F-scores, where the solution of~\citet{Guenther:14} loses
almost 3\% on average to the two top performers.  Regarding the
differences between the MHM and LBA themselves, we can observe a
rather mixed relation: The approach of~\citet{Mohammad:13} yields
better macro-averaged \F{}-results with the lexicons
of~\citet{Esuli:05}, \citet{Vo:16}, and \citet{Clematide:10}, but
falls against LBA when used with the polarity lists
of~\citet{Blair-Goldensohn:08}, \citet{Waltinger:10}, \citet{Hu:04},
\citet{Kiritchenko:14}, \citet{Rao:09}, \citet{Takamura:05},
\citet{Tang:14}, and \citet{Velikovich:10} as well as the NWE-based
\textsc{LinProj} and \textsc{PCA} lexicons.  Moreover, when trained
with the polarity list of \citeauthor{Tang:14} and our
\textsc{LinProj} lexicon, the LBA system achieves the best overall
macro-\F{} on this corpus.

These results, however, look slightly differently when we consider the
micro-averaged scores. This time, the system
of~\citeauthor{Mohammad:13} outperforms our solution in eight out of
twenty cases, but performs worse than LBA with four other polarity
lists (GPC, KIR, RR$_{\textrm{mincut}}$, and VEL).  Nevertheless, our
approach still reaches the best overall observed score (0.73) with
three tested resources (GPC, \textsc{LinProj}, and VEL).

Regarding the performance of single lexicons, we can see that the best
results are achieved with the manually curated
\textsc{SentiWS}~\cite{Remus:10} and Zurich Polarity
List~\cite{Clematide:10}, followed by the dictionary-based approaches
of~\citet{Blair-Goldensohn:08} and \citet{Rao:09}.  The method of the
nearest centroids vice versa appears to be of the lowest utility for
almost all systems, even though it demonstrated quite acceptable
scores in our initial intrinsic evaluation.

A similar situation also holds for the SB10k corpus, where the
ML-based approaches of~\citet{Mohammad:13} and \citet{Guenther:14} and
our proposed LBA system outperform all other methods in terms of both
macro- and micro-averaged \F{}-scores.  This time, however, the
average difference between the macro-results of LBA and GNT is much
smaller and amounts to only 0.02\% in favor of LBA, which again
achieves the best overall macro-\F{} (0.58) in combination with the
min-cut lexicon of~\citet{Rao:09}.  Unfortunately, our system clearly
falls against the latter classifier with respect to the micro-averaged
scores, performing worse than it in 16 out of 20 experiments.

The effect of single lexicons is also less pronounced than in the
PotTS case, as all of the tested polarity lists show a more or less
similar behavior, especially regarding the macro-averaged \F{}-score.
In terms of the micro-\F{}, however, we can observe that
dictionary-based lists, especially those of~\citet{Awadallah:10},
\citet{Blair-Goldensohn:08}, \citet{Hu:04}, and \citet{Kim:04}, lead
to generally better scores than corpus- and NWE-based resources.

\begin{figure*}
{
\centering
\begin{subfigure}{.5\textwidth}
  \centering
  \includegraphics[width=\linewidth]{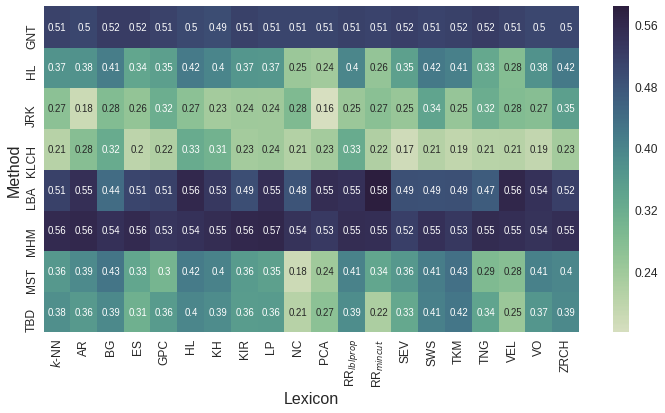}
  \caption{\texttt{Macro-\F}}\label{cgsa:fig:sb10k-lexicon-macro}
\end{subfigure}%
\begin{subfigure}{.5\textwidth}
  \centering
  \includegraphics[width=\linewidth]{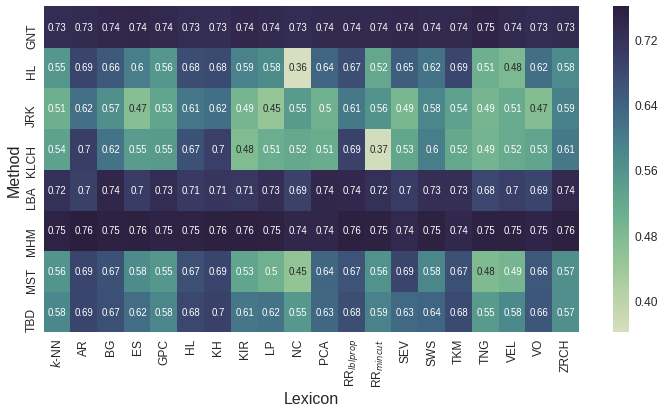}
  \caption{\texttt{Micro-\F}}\label{cgsa:fig:sb10k-lexicon-micro}
\end{subfigure}
}
\caption[MLSA results on SB10k with different lexicons]{Results of MLSA
  methods with different lexicons on the SB10k
  corpus}\label{cgsa:fig:sb10k-lexicon-effect}
\end{figure*}

\subsection{Text Normalization}

Finally, the last aspect that we are going to analyze in this
evaluation is the effect of the text normalization, which we applied
to the input messages before passing them to the classifiers.  To
verify the utility of this step, we rerun all experiments from the
initial sections, using the original Twitter messages instead of their
preprocessed forms, and recalculated the results of the tested
systems.

As we can see from the figures in
Table~\ref{snt-cgsa:tbl:res-no-normalization}, switching off the
normalization has a strong negative effect on the scores of almost all
approaches except for the methods of~\citet{Yessenalina:10} and
\citet{Socher:12,Socher:13}, which notoriously keep predicting the
majority class in most of the cases in the same way as they did
before.  Apart from this, we can notice that the lexicon-based systems
(HL, JRK, KLCH, MST, and TBD) suffer the greatest loss in terms of
both macro- and micro-averaged \F{}-scores on the PotTS corpus (up to
$-0.25$ macro- and and $-0.22$ micro-\F{}).  A closer look at their
errors revealed that this deterioration is mostly due to the increased
variety of different emoticons in the dataset (which were typically
unified during the preprocessing) and the absence of these forms in
the utilized polarity list.  The second biggest quality drop is
demonstrated by the DL-based approaches BAZ, LBA$^{(1)}$, and
LBA$^{(2)}$, which apparently also got confused by the higher lexical
variety of the input and failed to optimize all their internal
parameters to properly fit this diversity.  The remaining DL- and
ML-based classifiers (especially those of MHM, GNT, and RNTN) seem to
be more resistant to the introduced changes, but still show a decrease
by up to 0.04 macro- and 0.08 micro-\F{}.  The only exception in this
case is the MVRNN system of~\citet{Socher:12}, which slightly improves
on the negative and neutral classes, leaving the majority class
pitfall.  Unfortunately, this increase appears to be too small to
positively influence the overall statistics of this method.

Regarding the breakdown of single polarity classes, we can see that
most of the rare improvements affect the recall of positive and
neutral messages, with the biggest gains demonstrated by the RAE and
RNTN approaches ($+0.37$ and $+0.11$, respectively).  Other positive
changes are fairly sporadic and produced by only few classifiers
(first of all, MVRNN).  Nevertheless, even in these exceptional cases,
the improvements are typically so small that they hardly outweigh the
decreased scores on other aspects and have virtually no effect on the
net results for all classes.

\begin{table}[htb!]
  \begin{center}
    \bgroup\setlength\tabcolsep{0.1\tabcolsep}\scriptsize
    \begin{tabular}{p{0.162\columnwidth} 
        *{9}{>{\centering\arraybackslash}p{0.074\columnwidth}} 
        *{2}{>{\centering\arraybackslash}p{0.068\columnwidth}}} 
      \toprule
      \multirow{2}*{\bfseries Method} & %
      \multicolumn{3}{c}{\bfseries Positive} & %
      \multicolumn{3}{c}{\bfseries Negative} & %
      \multicolumn{3}{c}{\bfseries Neutral} & %
      \multirow{2}{0.068\columnwidth}{\bfseries\centering Macro\newline \F{}$^{+/-}$} & %
      \multirow{2}{0.068\columnwidth}{\bfseries\centering Micro\newline \F{}}\\
      \cmidrule(lr){2-4}\cmidrule(lr){5-7}\cmidrule(lr){8-10}

      & Precision & Recall & \F{} & %
      Precision & Recall & \F{} & %
      Precision & Recall & \F{} & & \\\midrule

      \multicolumn{12}{c}{\cellcolor{cellcolor}PotTS}\\

      HL & 0.63\negdelta{0.12} & 0.3\negdelta{0.46} & 0.4\negdelta{0.36} & %
      0.46\negdelta{0.07} & 0.29\negdelta{0.14} & 0.36\negdelta{0.11} & %
      0.41\negdelta{0.26} & 0.77\posdelta{0.04} & 0.54\negdelta{0.15} & %
      0.38\negdelta{0.24} & 0.464\negdelta{0.22}\\


      TBD & 0.65\negdelta{0.12} & 0.24\negdelta{0.47} & 0.36\negdelta{0.38} & %
      0.46\negdelta{0.08} & 0.27\negdelta{0.12} & 0.34\negdelta{0.11} & %
      0.41\negdelta{0.22} & 0.83\posdelta{0.06} & 0.55\negdelta{0.14} & %
      0.348\negdelta{0.25} & 0.457\negdelta{0.22}\\

      MST & 0.63\negdelta{0.12} & 0.29\negdelta{0.43} & 0.4\negdelta{0.34} & %
      0.47\negdelta{0.01} & 0.34\negdelta{0.13} & 0.39\negdelta{0.09} & %
      0.42\negdelta{0.26} & 0.77\posdelta{0.05} & 0.54\negdelta{0.16} & %
      0.4\negdelta{0.21} & 0.47\negdelta{0.21}\\

      JRK & 0.44\negdelta{0.16} & 0.22\negdelta{0.09} & 0.29\negdelta{0.12} & %
      0.14\negdelta{0.28} & 0.06\negdelta{0.14} & 0.08\negdelta{0.19} & %
      0.36\negdelta{0.07} & 0.7\negdelta{0.1} & 0.47\negdelta{0.09} & %
      0.19\negdelta{0.15} & 0.36\negdelta{0.11}\\

      KLCH & 0.61\negdelta{0.1} & 0.23\negdelta{0.49} & 0.33\negdelta{0.38} & %
      0.33\negdelta{0.01} & 0.21\posdelta{0.04} & 0.26\posdelta{0.04} & %
      0.41\negdelta{0.25} & 0.82 & 0.55\negdelta{0.18} & %
      0.3\negdelta{0.17} & 0.44\negdelta{0.21}\\


      GMN & 0.59\negdelta{0.08} & 0.77\posdelta{0.04} & 0.66\negdelta{0.04} & %
      0.37\negdelta{0.02} & 0.14\negdelta{0.01} & 0.2\negdelta{0.01} & %
      0.57\negdelta{0.03} & 0.55\negdelta{0.17} & 0.56\negdelta{0.1} & %
      0.43\negdelta{0.02} & 0.57\negdelta{0.05}\\

      MHM & 0.78\negdelta{0.01} & 0.76\negdelta{0.01} & 0.77\negdelta{0.01} & %
      0.59\posdelta{0.01} & 0.54\negdelta{0.02} & 0.56\negdelta{0.01} & %
      0.7\negdelta{0.03} & 0.74\negdelta{0.02} & 0.72\negdelta{0.02} & %
      0.67\negdelta{0.006} & 0.71\negdelta{0.007}\\

      GNT & 0.68\negdelta{0.03} & 0.8 & 0.73\negdelta{0.02} & %
      0.55 & 0.43\negdelta{0.02} & 0.48\negdelta{0.02} & %
      0.67\negdelta{0.01} & 0.59\negdelta{0.04} & 0.62\negdelta{0.03} & %
      0.61\negdelta{0.017} & 0.65\negdelta{0.02}\\

      Y\&C & 0.45 & 1.0 & 0.62 & %
      0.0 & 0.0 & 0.0 & %
      0.0 & 0.0 & 0.0 & %
      0.31 & 0.45\\

      RAE & 0.46\negdelta{0.15} & 0.98\posdelta{0.37} & 0.62\posdelta{0.01} & %
      0.0\negdelta{0.22} & 0.0\negdelta{0.01} & 0.0\negdelta{0.03} & %
      0.63\posdelta{0.15} & 0.05\negdelta{0.67} & 0.09\negdelta{0.48} & %
      0.31\negdelta{0.01} & 0.46\negdelta{0.08}\\

      MVRNN & 0.45 & 0.92\negdelta{0.08} & 0.6\negdelta{0.02} & %
      0.08\posdelta{0.08} & 0.01\posdelta{0.01} & 0.01\posdelta{0.01} & %
      0.26\posdelta{0.26} & 0.03\posdelta{0.03} & 0.06\posdelta{0.06} & %
      0.31 & 0.43\negdelta{0.02}\\

      RNTN & 0.45 & 0.93\posdelta{0.11} & 0.61\posdelta{0.02} & %
      0.29\posdelta{0.05} & 0.01\negdelta{0.05} & 0.01\negdelta{0.09} & %
      0.4\negdelta{0.03} & 0.07\negdelta{0.1} & 0.12\negdelta{0.12} & %
      0.31\negdelta{0.03} & 0.45\negdelta{0.01}\\

      SEV & 0.56\negdelta{0.17} & 0.79\posdelta{0.05} & 0.66\negdelta{0.08} & %
      0.0 & 0.0 & 0.0 & %
      0.57\posdelta{0.01} & 0.57\negdelta{0.27} & 0.57\negdelta{0.11} & %
      0.33\negdelta{0.04} & 0.56\negdelta{0.08}\\

      BAZ & 0.65\negdelta{0.17} & 0.59\negdelta{0.13} & 0.62\negdelta{0.15} & %
      0.62 & 0.22\negdelta{0.27} & 0.32\negdelta{0.23} & %
      0.5\negdelta{0.18} & 0.74\negdelta{0.11} & 0.6\negdelta{0.16} & %
      0.47\negdelta{0.19} & 0.57\negdelta{0.16}\\

      LBA$^{(1)}$ & 0.58\negdelta{0.18} & 0.77\negdelta{0.07} & 0.66\negdelta{0.13} & %
      0.54\negdelta{0.06} & 0.53\negdelta{0.03} & 0.54\negdelta{0.04} & %
      0.63\negdelta{0.12} & 0.37\negdelta{0.31} & 0.46\negdelta{0.26} & %
      0.6\negdelta{0.09} & 0.58\negdelta{0.15}\\

      LBA$^{(2)}$ & 0.67\negdelta{0.17} & 0.52\negdelta{0.21} & 0.59\negdelta{0.19} & %
      0.51\negdelta{0.06} & 0.44\negdelta{0.04} & 0.47\negdelta{0.06} & %
      0.52\negdelta{0.14} & 0.7\negdelta{0.12} & 0.6\negdelta{0.13} & %
      0.53\negdelta{0.12} & 0.57\negdelta{0.15}\\

      \multicolumn{12}{c}{\cellcolor{cellcolor}SB10k}\\

      HL & 0.41\negdelta{0.08} & 0.42\negdelta{0.2} & 0.42\negdelta{0.13} & %
      0.24\negdelta{0.03} & 0.28\negdelta{0.06} & 0.26\negdelta{0.04} & %
      0.66\negdelta{0.07} & 0.63\negdelta{0.01} & 0.65\negdelta{0.02} & %
      0.34\negdelta{0.08} & 0.53\negdelta{0.05}\\

      TBD & 0.41\negdelta{0.07} & 0.37\negdelta{0.23} & 0.39\negdelta{0.14} & %
      0.21\negdelta{0.03} & 0.24\negdelta{0.03} & 0.22\negdelta{0.03} & %
      0.65\negdelta{0.07} & 0.66\posdelta{0.03} & 0.66\negdelta{0.01} & %
      0.31\negdelta{0.08} & 0.53\negdelta{0.04}\\

      MST & 0.4\negdelta{0.05} & 0.32\negdelta{0.17} & 0.35\negdelta{0.12} & %
      0.26\negdelta{0.03} & 0.3\negdelta{0.05} & 0.28\negdelta{0.04} & %
      0.65\negdelta{0.05} & 0.68\negdelta{0.04} & 0.67 & %
      0.32\negdelta{0.08} & 0.54\negdelta{0.03}\\

      JRK & 0.4\negdelta{0.01} & 0.42\negdelta{0.03} & 0.41\negdelta{0.01} & %
      0.36 & 0.26 & 0.3 & %
      0.69 & 0.72\negdelta{0.03} & 0.71\negdelta{0.01} & %
      0.36\posdelta{0.01} & 0.59\negdelta{0.006}\\

      KLCH & 0.42\posdelta{0.03} & 0.21\negdelta{0.01} & 0.28 & %
      0.25\negdelta{0.09} & 0.13 & 0.17\negdelta{0.02} & %
      0.66 & 0.86 & 0.75 & %
      0.23\negdelta{0.005} & 0.6\negdelta{0.002}\\


      GMN & 0.48\negdelta{0.17} & 0.31\negdelta{0.14} & 0.37\negdelta{0.16} & %
      0.27\negdelta{0.11} & 0.07\negdelta{0.01} & 0.11\negdelta{0.02} & %
      0.69\negdelta{0.03} & 0.9\negdelta{0.03} & 0.78\negdelta{0.03} & %
      0.24\negdelta{0.09} & 0.64\negdelta{0.06}\\

      MHM & 0.67\negdelta{0.04} & 0.62\negdelta{0.03} & 0.65\negdelta{0.03} & %
      0.59\negdelta{0.08} & 0.42\negdelta{0.02} & 0.49\negdelta{0.04} & %
      0.8 & 0.88\negdelta{0.01} & 0.84 & %
      0.56\negdelta{0.002} & 0.75\negdelta{0.001}\\

      GNT & 0.42\negdelta{0.25} & 0.21\negdelta{0.41} & 0.28\negdelta{0.36} & %
      0.25\negdelta{0.19} & 0.13\negdelta{0.15} & 0.17\negdelta{0.17} & %
      0.66\negdelta{0.12} & 0.86\negdelta{0.01} & 0.75\negdelta{0.07} & %
      0.22\negdelta{0.2} & 0.604\negdelta{0.12}\\

      Y\&C & 0.0 & 0.0 & 0.0 & %
      0.0 & 0.0 & 0.0 & %
      0.62 & 1.0 & 0.77 & %
      0.0 & 0.62\\

      RAE & 0.46\negdelta{0.04} & 0.62\negdelta{0.11} & 0.53\negdelta{0.06} & %
      0.18\negdelta{0.17} & 0.02\negdelta{0.04} & 0.03\negdelta{0.07} & %
      0.77\negdelta{0.03} & 0.82\posdelta{0.02} & 0.79\negdelta{0.01} & %
      0.28\negdelta{0.07} & 0.66\negdelta{0.02}\\

      MVRNN & 0.19 & 0.01 & 0.03 & %
      0.0 & 0.0 & 0.0 & %
      0.62 & 0.97 & 0.76 & %
      0.01 & 0.61\\

      RNTN & 0.0 & 0.0 & 0.0 & %
      0.0 & 0.0 & 0.0 & %
      0.62 & 1.0 & 0.77 & %
      0.0 & 0.62\\

      SEV & 0.58\negdelta{0.06} & 0.39\negdelta{0.19} & 0.47\negdelta{0.14} & %
      0.23\negdelta{0.28} & 0.05\negdelta{0.16} & 0.08\negdelta{0.22} & %
      0.7\negdelta{0.06} & 0.92\posdelta{0.03} & 0.8\negdelta{0.02} & %
      0.27\negdelta{0.18} & 0.67\negdelta{0.05}\\

      BAZ & 0.69\negdelta{0.03} & 0.54\negdelta{0.16} & 0.6\negdelta{0.05} & %
      0.36\negdelta{0.17} & 0.49\posdelta{0.16} & 0.41 & %
      0.79 & 0.79\negdelta{0.12} & 0.79\negdelta{0.05} & %
      0.51\negdelta{0.02} & 0.69\negdelta{0.06}\\

      LBA$^{(1)}$ & 0.24\negdelta{0.36} & 0.86\posdelta{0.14} & 0.38\negdelta{0.28} & %
      0.45\negdelta{0.02} & 0.45\posdelta{0.03} & 0.45\posdelta{0.01} & %
      0.69\negdelta{0.15} & 0.01\negdelta{0.79} & 0.02\negdelta{0.8} & %
      0.41\negdelta{0.14} & 0.27\negdelta{0.46}\\

      LBA$^{(2)}$ & 0.74\negdelta{0.02} & 0.42\negdelta{0.15} & 0.54\negdelta{0.1} & %
      0.62\posdelta{0.07} & 0.25\negdelta{0.14} & 0.35\negdelta{0.11} & %
      0.73\negdelta{0.06} & 0.95\posdelta{0.05} & 0.82\negdelta{0.02} & %
      0.45\negdelta{0.1} & 0.72\negdelta{0.03}\\\bottomrule
    \end{tabular}
    \egroup{}
    \caption[Results of MLSA methods without text normalization]{
      Results of MLSA methods without text normalization}\label{snt-cgsa:tbl:res-no-normalization}
  \end{center}
\end{table}

A similar situation also happens on the SB10k corpus, where we can see
even fewer improvements (in 10 out of 176 cases).  The biggest
increase this time ($+0.16$ recall) is demonstrated by the approach
of~\citet{Baziotis:17} on the negative class.  The remaining growths,
however, are much smaller and typically range between one and seven
percent.  On the other hand, three of the tested methods (Y\&C, MVRNN,
and RNTN) have exactly the same results as they did previously with
normalized messages, although, most of the time, these classifiers
only predict the majority label anyway.  As to the rest of the
systems, we can see that their scores are notably lower than in our
initial experiments, but the decrease is much smaller in comparison
with the PotTS corpus.  A sad exception in this case is a major drop
of the recall of neutral messages ($-0.79$) demonstrated by our
LBA$^{(1)}$ system, which, in turn, results in a significant decrease
of its macro- and micro-averaged \F{}-scores ($-0.14$ and $-0.46$,
respectively).  Other approaches (including the sibling method
LBA$^{(2)}$) behave much more stable in this regard and their average
decrease amounts to $-0.06$ macro- and $-0.03$ micro-\F{}.

Similar to the results on the PotTS data, most of the gains are
concentrated at the recall of the neutral class (four out of ten
improvements), with the other positive changes being rather sporadic
and affecting only a few classifiers.  Nevertheless, unlike in the
previous case, this time, we can even observe a slight improvement of
the macro-averaged \F-measure for one of the systems (the
lexicon-based approach of~\citeauthor{Jurek:15}), but its
micro-averaged metric remains mainly unaffected by this increase.  In
general, however, the vast majority of macro- and micro-\F-scores show
an obvious decline on both datasets, which once again proves the
advantage of preprocessing.

\section{Summary and Conclusions}\label{slsa:subsec:conclusions}

Now the we have reached the end of the chapter, we would like to
remind the reader that in this part of the thesis we have made the
following findings and contributions:
\begin{itemize}
  \item we have compared three major families of message-level
    sentiment analysis methods: lexicon-, machine-learning-- and
    deep-learning--based ones, finding that the last two groups
    significantly outperform lexicon-driven systems;
  \item surprisingly, among all compared lexicon methods, the most
    simple one (the classifier of~\citeauthor{Hu:04}
    [\citeyear{Hu:04}]) produced the best macro- and micro-averaged
    \F{}-results on the PotTS corpus (0.615 and 0.685, respectively)
    and also yielded the highest macro \F{}-measure on the SB10k
    dataset (0.421).  Other systems, however, could have improved
    their scores if they better handled the negation of polar terms
    (after switching off the negation component in the method
    of~\citeauthor{Musto:14}, its macro-\F{} on the PotTS corpus
    increased to 0.641, surpassing the benchmark
    of~\citeauthor{Hu:04});
  \item as expected, the ML-based system of~\citet{Mohammad:13}---the
    winner of the inaugural run of SemEval task in sentiment analysis
    of Twitter~\cite{Nakov:13}---also surpassed other ML competitors,
    achieving highly competitive results: 0.674 macro- and 0.727
    micro-\F{} on the PotTS data, and 0.564 macro- and 0.752
    micro-averaged \F{}-measure on the SB10k test set;
  \item as in the previous case, however, these results could have
    been improved if the classifier dispensed with character-level and
    part-of-speech features and used logistic regression instead of
    SVM;
  \item a much more varied situation was observed with
    deep-learning--based systems, which frequently simply fell into
    always predicting the majority class for all tweets, but sometimes
    yielded extraordinarily good results as it was the case with our
    proposed lexicon-based attention system, which attained 0.69
    macro-\F{} on the PotTS corpus and 0.55 macro \F{}-score on the
    SB10k dataset (0.73 and 0.75 micro-\F{}, respectively), setting a
    new state of the art for the former data;
  \item speaking of word embeddings, we should note that almost all
    DL-based approaches showed fairly low scores when they used
    randomly initialized task-specific embeddings, but notably
    improved their results after switching to pre-trained word2vec
    vectors, and benefited even more from the least-squares fallback;
  \item against our expectations, we could not overcome the majority
    class pitfall of DL-based systems after adding more weakly
    supervised training data, which, in general, only lowered the
    scores of both ML- and DL-based methods.  Since this result
    contradicts the findings of other authors, we hypothesize that
    this degradation is primarily due to the differences in the class
    distributions between automatically and manually labeled tweets;
  \item on the other hand, we could see that using more qualitative
    sentiment lexicons (especially manually curated and
    dictionary-based ones) resulted in further improvements for the
    systems that relied on this lexical resource;
  \item last but not least, we proved the utility of the text
    normalization step, which brought about significant improvements
    for all tested methods, as confirmed by our last ablation test.
\end{itemize}


\chapter{Discourse-Aware Sentiment Analysis}\label{chap:discourse}

Although message-level sentiment analysis methods do a fairly good job
at classifying the overall polarity of a message,
a crucial limitation of all these systems is that they completely
overlook the structural nature of their input by either considering it
as a single whole (\eg{} bag-of-features approaches) or analyzing it
as a monotone sequence of equally important elements (\eg{} recurrent
neural methods).  Unfortunately, both of these solutions violate the
hierarchical principle of language~\cite{Saussure:90,Hjelmslev:70},
which states that complex linguistic units are formed from smaller
language elements in the bottom-up way, \eg{} words are created by
putting together morphemes, sentences are made of several words, and
discourses are composed of multiple coherent sentences.  Moreover,
apart from this inherent structural heterogeneity, even units of the
same linguistic level might play a different role and be of unequal
importance when joined syntagmatically into the higher-level whole.
For example, in words, the root morpheme typically conveys more
lexical meaning than the affixes; in sentences, the syntactic head
usually dominates its grammatical dependents; and, in discourse, one
of the sentences frequently expresses more relevant ideas than the
rest of the text.

Exactly the lack of discourse information was one of the main reasons
for the misclassifications made by the systems of \citet{Severyn:15},
\citet{Baziotis:17}, and our own LBA method in
Examples~\ref{snt:cgsa:exmp:severyn-error},\ \ref{snt:cgsa:exmp:baziotis-error},
and~\ref{snt:cgsa:exmp:lba-error}.  Since none of these approaches
explicitly took discourse structure into account, we decided to check
whether making the last of these solutions (the LBA classifier) aware
of discourse phenomena would improve its results.  But before we
present these experiments, we first would like to make a short
digression into the theory of discourse and give an overview of the
most popular approaches to text-level analysis that exist in the
literature nowadays.  Afterwards, in Section~\ref{sec:dasa:data}, we
will describe the way how we inferred discourse information for PotTS
and SB10k tweets.  Then, in Section~\ref{sec:dasa:methods}, we will
summarize the current state of the art in discourse-aware sentiment
analysis (DASA) and also present our own methods, evaluating them on
the aforementioned datasets.  After analyzing the effects of various
common factors (such as the impact of the underlying sentiment
classifier and the amenability of various discourse relation schemes
to different DASA approaches), we will recap the results and summarize
our findings in the last part of this chapter.

\section{Discourse Analysis}\label{sec:dasa:theory}

Since the main focus of our experiments will be on \emph{discourse
  analysis}, we first need to clarify what discourse analysis actually
means and which common ways there are to represent and analyze
discourse automatically.

In a nutshell, discourse analysis is an area of research which
explores and analyzes language phenomena beyond the sentence
level~\cite{Stede:11}.  Although the scope of this research can be
quite large, ranging from the use of pronouns in a sentence to the
logical composition of the whole document, in our work we will
primarily concentrate on the coherence structure of a text, \ie{} its
segmentation into \emph{elementary discourse units} (typically single
propositions) and induction of hierarchical \emph{coherence relations}
(semantic or pragmatic links) between these EDUs.

Although the idea of splitting the text into smaller meaningful pieces
and inferring semantic relationships between these parts is anything
but new, dating back to the very origins of general
linguistics~\cite{Aristotle:10} and in particular its structuralism
branch~\cite{Saussure:90}, an especially big surge of interest in this
field happened in the 1970-s with the fundamental works of
\citet{vanDijk:72} and \citet{vanDijk:83}, who introduced the notion
of local and global coherence, defining the former as a set of ``rules
and conditions for the well-formed concatenation of pairs of sentences
in a linearly ordered sequence'' and specifying the latter as
constraints on the macro-structure of the
narrative~\cite[see][]{Hoey:83}.  Similar ideas were also proposed
by~\citet{Longacre:79,Longacre:96}, who considered the paragraph as a
unit of tagmemic grammar that was composed of multiple sentences
according to a predefined set of compositional principles.  Almost
contemporary with these works, \citet{Winter:77} presented an
extensive study of various lexical means that could connect two
sentences and grouped these means into two major categories:
\textsc{Matching} and \textsc{Logical Sequence}, depending on whether
they introduced sentences that were giving more details on the
preceding content (\textsc{Matching}) or adding new information to the
narrative (\textsc{Logical Sequence}).

The increased interest of traditional linguistics in text-level
analysis has rapidly spurred the attention of the broader NLP
community.  Among the first who stressed the importance of discourse
phenomena for automatic generation and understanding of texts was
\citet{Hobbs:79}, who argued that semantic ties between sentences were
one the most important component for building a coherent discourse.
Similarly to \citeauthor{Winter:77}, \citeauthor{Hobbs:79} also
proposed a classification of inter-sentence relations, dividing them
into \textsc{Elaboration}, \textsc{Parallel}, and \textsc{Contrast}.
Albeit this taxonomy was obviously too small to accommodate all
possible semantic and pragmatic relationships that could exist between
two clauses, this division had laid the foundations for many
successful approaches to automatic discourse analysis that appeared in
the following decades.

\paragraph{RST.}

One of the best-known such approaches, \emph{Rhetorical Structure
  Theory} or \emph{RST}, was presented by~\citet{Mann:88}.  Besides
revising \citeauthor{Hobbs:79}' inventory of discourse relations and
expanding it to 23 elements (including new items such as
\textsc{Antithesis}, \textsc{Circumstance}, \textsc{Evidence}, and
\textsc{Elaboration}), the authors also grouped all coherence links
into nucleus-satellite (hypotactic) and multinuclear (paratactic)
ones, depending on whether the arguments of these edges were of
different or equal importance to the content of the whole text.  Based
on this grouping, they formally described each relation as a set of
constraints on the \emph{Nucleus} (N), \emph{Satellite} (S), \emph{the
  N+S combination}, and \emph{the effect} of the whole combination on
the reader (R).  An excerpt from the original description of the
\textsc{Antithesis} relation is given in
Example~\ref{dasa:exmp:rst-evidence}

\begin{example}[Definition of the \textsc{Antithesis} Relation]\label{dasa:exmp:rst-evidence}
  \textbf{Relation Name:} \textsc{Antithesis}

  \textbf{Constraints on N:} W has positive regard for the situation
  presented in N

  \textbf{Constraints on S:} None

  \textbf{Constraints on the N+S Combination:} the situations presented
  in N and S are in contrast (\ie{} are
  \begin{inparaenum}[(a)]
  \item comprehended as the same in many respects,
  \item comprehended as differing in a few respects and
  \item compared with respect to one or more of these differences
  \end{inparaenum}); because of an incompatibility that arises from the contrast, one
  cannot have positive regard for both the situations presented in N
  and S\@; comprehending S and the incompatibility between the
  situations presented in N and S increases R's positive regard for
  the situation presented in N

  \textbf{Effect:} R's positive regard for N is increased

  \textbf{Locus of the Effect:} N
\end{example}
The authors then defined the general structure of discourse as a
projective (constituency) tree whose nodes were either elementary
discourse units or subtrees, which were connected to each other via
discourse relations.

You can see an example of such a discourse tree from the original
Rhetorical Structure Treebank~\cite{Carlson:01a} in
Figure~\ref{dasa:fig:rst-tree}.

\begin{figure*}[htb!]\label{dasa:fig:rst-tree}
  \input{rst.tex}
\end{figure*}

Despite its immense popularity and practical utility~\cite[see
][]{Marcu:98,Yoshida:14,Bhatia:15,Goyal:16}, RST has often been
criticized for the rigidness of the imposed tree
structure~\cite{Wolf:05} and unclear distinction between discourse
relations~\cite{Nicholas:94,Miltsakaki:04}.  As a result of this
criticism, two alternative approaches to automatic discourse analysis
were proposed in later works.

\paragraph{PDTB.}

One of these approaches, \emph{PDTB} (named so after the Penn
Discourse Treebank [\citeauthor{Prasad:04}, \citeyear{Prasad:04}]),
was developed by a research group at University of
Pennsylvania~\cite{Miltsakaki:04,Miltsakaki:04a,Prasad:08}.  Instead
of fully specifying the hierarchical structure of the whole text and
providing an all-embracing set of discourse relations, the authors of
this theory mainly focused on the grammatical and lexical means that
could connect two sentences (\emph{connectives}) and express a
semantic relationship (\emph{sense}) between these predicates.
Typical such means are coordinating or subordinating conjunctions
(\eg{} \emph{and}, \emph{because}, \emph{since}) and discourse
adverbials (\eg{} \emph{however}, \emph{otherwise}, \emph{as a
  result}), which can denote a \textsc{Comparison}, a
\textsc{Contingency}, or some other sense\footnote{In particular, the
  authors of PDTB distinguished four major senses
  (\textsc{Comparison}, \textsc{Contingency}, \textsc{Expansion}, and
  \textsc{Temporal}), and subdivided each of these categories into
  further subtypes, \eg{} \textsc{Comparison} included
  \textsc{Concession} and \textsc{Contrast}, whereas
  \textsc{Contingency} sense was further divided into \textsc{Cause}
  and \textsc{Condition}.} between two sentential arguments
(\textsc{Arg1} and \textsc{Arg2}).


Apart from \emph{explicitly} mentioned connectives, \citet{Prasad:04}
also allowed for situations where a connective was missing but could
be easily inferred from the text.  They called such cases
\emph{implicit} discourse relations and demanded the arguments of such
structures be determined as well.  Furthermore, if there was no
connective at all, the authors of PDTB distinguished three different
possibilities:
\begin{itemize}
\item the coherence relation was either expressed by an alternative
  lexical means, which made the connective redundant
  (\textsc{AltLex}),
\item or it was achieved by referring to the same entities in both
  arguments (\textsc{EntRel}),
\item or there was no coherence relation at all (\textsc{NoRel});
\end{itemize}
and also provided a special \textsc{Attribution} label for marking the
authors of reported speech.

Example~\ref{dasa:exmp:pdtb-analysis} shows the previous fragment of
the Rhetorical Treebank now annotated according to the PDTB scheme.

As we can see from the analysis, PDTB is indeed more flexible than
RST, as it allows its discourse units (arguments) to overlap, be
disjoint or even embedded into other segments.  The assignment of
sense relations is also more straightforward and mainly determined by
the connectives that link the arguments.  But, at the same time, the
structure of this annotation is completely flat so that we can neither
infer which of the sentences plays a more prominent role nor see the
modification scope of other supplementary statements.

\begin{example}[Example of PDTB Analysis]\label{dasa:exmp:pdtb-analysis}
  \fbox{Analysts said,} \argone[1]{profit for the dozen or so big drug
    makers, as a group, is estimated to have climbed between 11\% and
    14\%.}  \connective[1]{\textsc{implicit}$:=$in fact}
  \argtwo[1]{\connective[2]{\textsc{explicit}$:=$While}
    \argtwo[2]{that's not spectacular}}, \fbox{Neil Sweig, an analyst
    with Prudential Bache, said} \argtwo[1]{\argone[2]{\argone[3]{that
        the rate of growth will ``look especially good as compared to
        other companies} \connective[3]{\textsc{explicit}:
        if}\argtwo[3]{the economy turns downward}}}.''
\end{example}

\paragraph{SDRT.}

Another alternative to RST, \emph{Segmented Discourse Representation
  Theory} or \emph{SDRT}, was proposed by \citet{Lascarides:01}.
Although developed from a completely different angle of view (the
authors of SDRT mainly drew their inspiration from predicate logic,
dynamic semantics, and anaphora theory), this theory shares many of
its features with Rhetorical Structure Theory, as it also assumes a
graph-like structure of text and distinguishes between coordinating
and subordinating relations.  However, unlike RST, Segmented Discourse
Representation explicitly allows the text structure to be a multigraph
and not only tree (\ie{} a discourse node can have multiple parents
and can also be connected via multiple links to the same vertex),
provided that it does not have crossing dependencies (\ie{} does not
violate the right-frontier constraint).

\begin{figure}[htbp]
  \begin{center}
    \begin{tikzpicture}[>=triangle 45,semithick]
      \tikzstyle{edu}=[]; \tikzstyle{cdu}=[draw,shape=rectangle];
      \node[edu] (1a) at (1,0) {$\pi_{1a}$}; \node[edu] (1b) at (1,-2)
           {$\pi_{1b}$};

           \node[edu] (p'') at (7,0)  {$\pi''$};
           \node[edu] (p') at (5.5,-2)  {$\pi'$};
           \node[edu] (1g) at (8.5,-2)  {$\pi_{1g}$};

           \node[edu] (1e) at (4,-4)  {$\pi_{1e}$};
           \node[edu] (1f) at (7,-4)  {$\pi_{1f}$};

           \node[edu] (1c) at (2,-2)  {$\pi_{1c}$};
           \node[edu] (1d) at (4,-2)  {$\pi_{1d}$};

           \draw[->] (1a)  to node [auto] {Source} (1b);
           \draw[->] (1a)  to node [auto] {Narration} (p'');
           \draw[-]  (p'') to node [auto] {} (p');
           \draw[-]  (p'') to node [auto] {} (1g);
           \draw[->] (p')  to node [auto] {Precondition} (1g);
           \draw[-]  (p')  to node [auto] {} (1e);
           \draw[-]  (p')  to node [auto] {} (1f);
           \draw[->] (1e)  to node [auto] {Contrast} (1f);
           \draw[->] (p'') to node [xshift=-8mm,yshift=-0.35mm] {Commentary} (1c);
           \draw[->] (p'') to node [xshift=-0mm,yshift=0mm] {Source} (1d);
    \end{tikzpicture}
    \caption{Example of an SDRT graph}\label{dasa:fig:sdrt-graph}
  \end{center}
\end{figure}
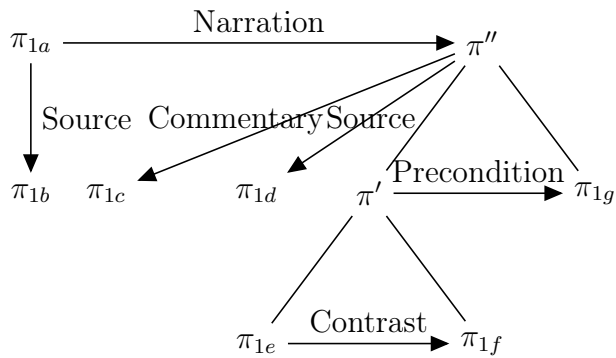

We can also notice the relatedness of the two theories by looking at
the SDRT analysis of the previous RST fragment in
Example~\ref{dasa:fig:sdrt-graph}.  Although the names of the
relations in the presented graph differ from those used in Rhetorical
Structure Theory, many of these links have the same (or at least
similar) meaning as the respective edges in the first analysis: for
example, the \textsc{Source} relation in SDRT almost completely
corresponds to the \textsc{Attribution} edge in
Example~\ref{dasa:fig:rst-tree}, and the \textsc{Contrast} link is
similar to the \textsc{Comparison} relation defined by
\citet{Carlson:01b}.

\paragraph{Final choice.}

Because it was unclear which of these approaches (RST, PDTB, or SDRT)
would be more amenable to our sentiment experiments, we have made our
decision by considering the following theoretical and practical
aspects: From theoretical perspective, we wanted to have a strictly
hierarchical discourse structure for each analyzed tweet so that we
could infer the semantic orientation of that message by recursively
accumulating polarity scores of its elementary discourse segments.
From practical point of view, since there was no discourse parser
readily available for German, we wanted to have a maximal assortment
of such systems available for English so that we could pick one that
would be easiest to retrain on German data.  Fortunately, both of
these concerns have lead us to the same solution---Rhetorical
Structure Theory, which was the only formalism that explicitly
guaranteed a single root for each analyzed text and also offered a
wide variety of open-source parsing
systems~\cite[\eg][]{Hernault:10,Feng:14,Ji:14,Yoshida:14,Joty:15}.

\section{Data Preparation}\label{sec:dasa:data}

To prepare the data for our experiments, we split all microblogs from
\begin{figure*}[htb]
  \centering { \centering
        \begin{subfigure}{0.7\textwidth}
          \centering
          \includegraphics[width=\linewidth]{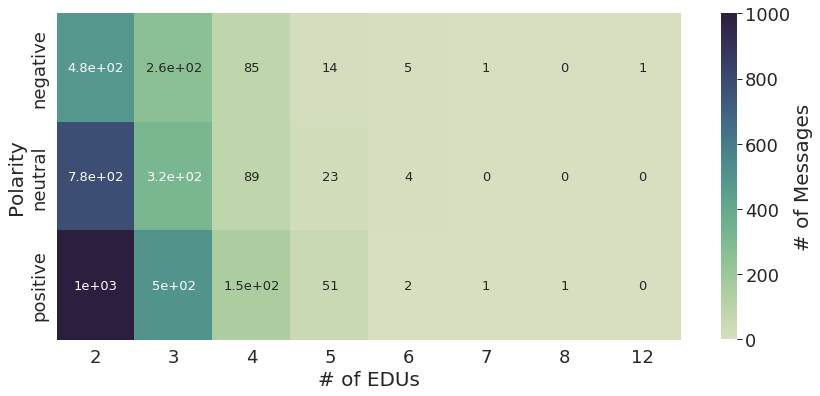}
          \caption{PotTS}\label{dasa:fig:data-distribution-potts}
        \end{subfigure}
      }
      \centering
          {
            \centering
            \begin{subfigure}{0.7\textwidth}
              \centering
              \includegraphics[width=\linewidth]{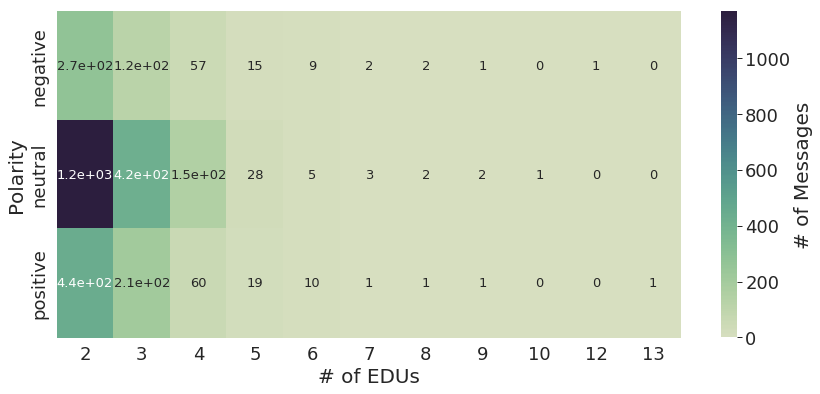}
              \caption{SB10k}\label{dasa:fig:data-distribution-sb10k}
            \end{subfigure}
          }
          \caption[EDU distribution in PotTS and SB10k]{Distribution
            of elementary discourse units and polarity classes in the
            training and development sets of PotTS and
            SB10k}\label{dasa:fig:data-distribution}
\end{figure*}

the PotTS and SB10k corpora into elementary discourse units using the
ML-based discourse segmenter of \citet{Sidarenka:15}, which had been
previously trained on the Potsdam Commentary Corpus~\cite[PCC~2.0;
][]{Stede:14}.  After filtering out all tweets that had only one
EDU,\footnote{Since the focus of this chapter is mainly on discourse
  phenomena, we skip all messages that consist of a single discourse
  segment, because their overall polarity is unaffected by the
  discourse structure and can be normally determined with the standard
  discourse-unaware sentiment techniques.} we obtained 4,771 messages
(12,137 segments) for PotTS and 3,763 posts (9,625 segments) for the
SB10k corpus.  In the next step, we assigned polarity scores to the
segments of these microblogs with the help of our lexicon-based
attention classifier, analyzing each elementary unit in isolation,
independently of the rest of the tweet.  We again used the same
70--10--20 split into training, development, and test sets as we did
in the previous chapters, considering message-level labels inferred
from the annotation of the second expert as gold standard for the
PotTS corpus and using provided manual sentiment labels for tweets as
reference for the SB10k data.

As we can see from the statistics in
Figure~\ref{dasa:fig:data-distribution}, most tweets that consist of
multiple EDUs typically have two or three segments, whereas messages
with more than three discourse units are extremely rare.  This is also
not surprising regarding that the maximum length of a microblog is
constrained to 140 characters.  Nonetheless, even with this severe
length restriction, there still are a few messages that have up to 13
EDUs.  Since it was somewhat surprising for us to see that many
segments in a single tweet, we decided to have a closer look at these
cases.  As it turned out, such high number of discourse units
typically resulted from spurious punctuation marks, which were
carelessly used by Twitter users and evidently confused the segmenter
(see Example~\ref{dasa:exmp:many-segments}).

\begin{example}[SB10k Tweet with 13 EDUs]\label{dasa:exmp:many-segments}
  \noindent\textup{\bfseries\textcolor{darkred}{Tweet:}} {\upshape
    [Guinness on Wheelchairs :]$_1$ [Das .]$_2$ [Ist .]$_3$ [Verdammt
      .]$_4$ [Noch .]$_5$ [Mal .]$_6$ [Einer .]$_7$ [Der .]$_8$
    [Besten .]$_9$ [Werbespots .]$_{10}$ [Des .]$_{11}$ [Jahrzehnts
      .]$_{12}$ [( Auch ...]$_{13}$ }\\
                  {\textup{[}Guinness on
                      Wheelchairs :\textup{]$_1$} \textup{[}This .\textup{]$_2$}
                    \textup{[}Is .\textup{]$_3$} \textup{[}Gosh .\textup{]$_4$}
                    \textup{[}Darn .\textup{]$_5$} \textup{[}It .\textup{]$_6$}
                    \textup{[}One .\textup{]$_7$} \textup{[}Of .\textup{]$_8$}
                    \textup{[}The best .\textup{]$_9$} \textup{[}Commercials
                      .\textup{]$_{10}$} \textup{[}Of .\textup{]$_{11}$} \textup{[}The
                      Decade .\textup{]$_{12}$} \textup{[}( Also ...\textup{]$_{13}$}}
\end{example}

Another noticeable trend that we can see in the data is that the
distribution of polar classes in messages with multiple segments
largely corresponds to the frequencies of these polarities in the
complete datasets: For example, the positive semantic orientation
still dominates the PotTS corpus, whereas the neutral polarity
constitutes the vast majority of the SB10k set.  At the same time,
negative microblogs again are the least represented class in both
cases and account for only 22\% of the former corpus and for 16\% of
the latter data.

To obtain RST trees for these messages, we retrained the DPLP
discourse parser of~\citet{Ji:14} on PCC, after converting all
discourse relations to the binary scheme $\{$\textsc{Contrastive},
\textsc{Non-Contrastive}$\}$ as suggested
by~\citet{Bhatia:15}.\footnote{See Table~\ref{dasa:tbl:rst-rel-sets}
  for more details regarding this mapping.}  In contrast to the
original DPLP implementation though, we did not use Brown
clusters~\cite{Brown:92}, because this resource was not available for
German, nor did we apply the linear projection of the features,
because the released parser code was missing this component either.
In part due to these modifications, but mostly because of the
specifics of the German language (richer morphology, higher lexical
variety, and syntactic ambiguity) and a skewed distribution of
discourse relation, the results of the retrained model were
considerably lower than the figures reported for the English treebank,
amounting to 0.777, 0.512, and 0.396~\F{} for span, nuclearity, and
relation classification on PCC~2.0 versus corresponding 82.08, 71.13,
and 61.63~\F{} on the RST Treebank.\footnote{Following \citet{Ji:14},
  we use the span-based evaluation metric of~\citet{Marcu:00}.}
\begin{figure*}[htb]
  \input{twitter-rst.tex}
\end{figure*}

\begin{figure*}[bht]
  \centering
      {
        \centering
        \begin{subfigure}{0.7\textwidth}
          \centering
          \includegraphics[width=\linewidth]{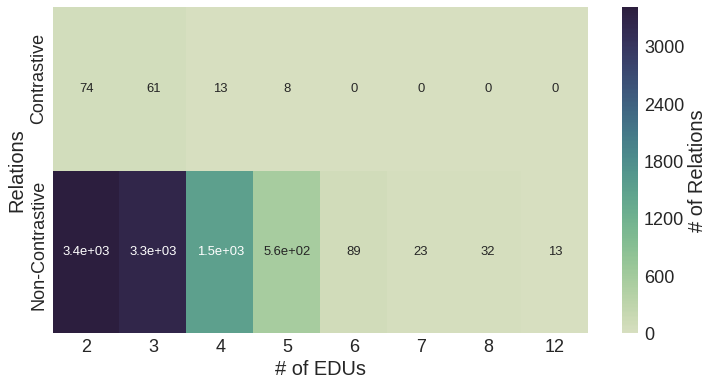}
          \caption{PotTS}\label{dasa:fig:relation-distribution-potts}
        \end{subfigure}
      }
      \centering
          {
            \centering
            \begin{subfigure}{0.7\textwidth}
              \centering
              \includegraphics[width=\linewidth]{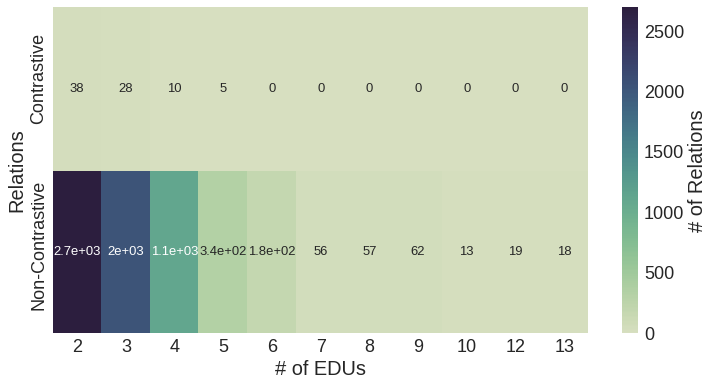}
              \caption{SB10k}\label{dasa:fig:relation-distribution-sb10k}
            \end{subfigure}
          }
          \caption[Relation distribution in PotTS and
            SB10k]{Distribution of discourse relations in the training
            and development sets of PotTS and
            SB10k}\label{dasa:fig:relation-distribution}
\end{figure*}

An example of an automatically induced RST tree is shown in
Figure~\ref{dasa:fig:twitter-rst-tree}.  As we can see from this
picture, the adapted parser can correctly distinguish between
contrastive and non-contrastive relations in the analyzed tweet (even
though it only predicts the former class for two percent of all edges
on the PotTS and SB10k data [see
  Figure~\ref{dasa:fig:relation-distribution}]), but apparently
struggles with the disambiguation of the nuclearity status, assigning
the highest importance in this example to the initial discourse
segment (``Mooooiiinn.''  [\emph{Hellloooo!}]), which is merely a
greeting, and weighing the second EDU (``Gegen solche N\"achte hilft
die beste Kur nicht.''  [\emph{Even the best cure won't help against
    such nights.}]) less than the third one (``Aber Kaffee!''
[\emph{But coffee!}]), although traditional RST would rather consider
both units as equally relevant and join them via the multi-nuclear
\textsc{Contrast} link.

\section{Discourse-Aware Sentiment Analysis}\label{sec:dasa:methods}



Now before we use these data in our sentiment experiments, let us
first revise the most prominent approaches to discourse-aware
sentiment analysis that exist in the literature nowadays.

As it turns out, even the very first works on opinion mining already
pointed out the importance of discourse phenomena for classification
of the overall polarity of a text.  For example, in the seminal paper
of~\citet{Pang:02}, where the authors tried to predict the semantic
orientation of movie reviews, they quickly realized the fact that it
was insufficient to rely on the mere presence or even the majority of
polarity clues in the text, because these clues could any time be
reversed by a single counter-argument of the critic (see
Example~\ref{disc-snt:exmp-pang02}).  This observation was also
confirmed by \citet{Polanyi:06}, who ranked discourse relations among
the most important factors that could significantly affect the
intensity and polarity of a sentiment.  To prove this claim, they gave
several convincing examples, where a concessive statement considerably
weakened the strength of a polar opinion, and vice versa, an
elaboration notably increased its persuasiveness.

\citet{Pang:04} were also among the first who incorporated a
discourse-aware component into a document-level sentiment classifier.
For this purpose, they developed a two-stage system in which the first
predictor distinguished between subjective and objective statements by
constructing a graph of all sentences (linking each sentence to its
neighbors and also connecting it to two abstract polarity nodes) and
then partitioning this graph into two clusters (subjective and
objective) based on its minimum cut; the second classifier then
inferred the overall polarity of the text by only looking at the
sentences from the first (subjective) group.  With this method,
\citeauthor{Pang:04} achieved a statistically significant improvement
(86.2\% versus 85.2\% for the Na\"{\i}ve Bayes system and 86.15\%
versus 85.45\% for SVM) over classifiers that analyzed all text
sentences at once, without any filtering.

\begin{example}[Polarity reversal via discourse antithesis]\label{disc-snt:exmp-pang02}
  \noindent\upshape This film should be brilliant.  It sounds like a
  great plot, the actors are first grade, and the supporting cast is
  good as well, and Stallone is attempting to deliver a good
  performance.  However, it can't hold up.~\cite{Pang:02}
\end{example}

Although an oversimplification, the core idea that locally adjacent
sentences are likely to share the same subjective orientation
(\emph{local coherence}) was dominating the following DASA research
for almost a decade.  For example, \citet{Riloff:03} also improved the
accuracy of their Na\"{\i}ve Bayes predictor of subjective expressions
by almost two percent after adding a set of local coherence features.
Similarly, \citet{Hu:04} could better disambiguate users' attitudes to
particular product attributes by taking the semantic orientation of
previous sentences into account.

At the same time, another line of discourse-aware sentiment research
concentrated on the joint classification of all opinions in the text,
where in addition to predicting each sentiment in isolation, the
authors also sought to maximize the ``total happiness'' (\emph{global
  coherence}) of these assignments, ensuring that related subjective
statements received agreeing polarity scores.  Notable works in this
direction were done by \citet{Snyder:07}, who proposed the Good Grief
algorithm for predicting users' satisfaction with different restaurant
aspects, and \citet{Somasundaran:08a,Somasundaran:08}, who introduced
the concept of \emph{opinion frames} (OF), a special data structure
for capturing the relations between opinions in discourse.  Depending
on the type of these opinions (arguing~[\emph{A}] or
sentiment~[\emph{S}]), their polarity towards the target
(positive~[\emph{P}] or negative~[\emph{N}]), and semantic
relationship between these targets (alternative~[\emph{Alt}] or the
same~[\emph{same}]), the authors distinguished 32 types of possible
frames (\emph{SPSPsame}, \emph{SPSNsame}, \emph{APAPalt}, etc.),
dividing them into reinforcing and non-reinforcing ones.  In later
works, \citet{Somasundaran:09a,Somasundaran:09b} also presented two
joint inference frameworks (one based on the iterative classification
and another one relying on integer linear programming) for determining
the best configuration of all frames in text, achieving 77.72\%
accuracy on frame prediction in the AMI meeting
corpus~\cite{Carletta:05}.

An attempt to unite local and global coherence was made by
\citet{McDonald:07}, who tried to simultaneously predict the polarity
of a document and classify semantic orientations of its sentences.
For this purpose, the authors devised an undirected probabilistic
graphical model based on the structured linear
classifier~\cite{Collins:02}.  Similarly to \citet{Pang:04}, they
connected the label nodes of each sentence to the labels of its
neighboring clauses and also linked these nodes to the overarching
vertex representing the polarity of the text.  After optimizing this
model with the MIRA learning algorithm~\cite{Crammer:03},
\citeauthor{McDonald:07} achieved an accuracy of 82.2\% for
document-level classification and 62.6\% for sentence-level prediction
on a corpus of online product reviews, outperforming pure document and
sentence classifiers by up to four percent.  A crucial limitation of
this system though was that its optimization required the gold labels
of sentences and documents to be known at the training time, which
considerably limited its applicability to other domains with no such
data.


Another significant drawback of all previous approaches is that they
completely ignored traditional discourse theory and, as a result,
severely oversimplified discourse structure.  Among the first who
tried to overcome this omission were \citet{Voll:07}, who proposed two
discourse-aware enhancements of their lexicon-based sentiment
calculator (SO-CAL).  In the first method, the authors let the SO-CAL
analyze only the topmost nucleus EDU of each sentence, whereas in the
second approach, they expanded its input to all clauses that another
classifier had considered as relevant to the main topic of the
document.  Unfortunately, the former solution did not work out as well
as expected, yielding 69\% accuracy on the corpus of Epinion
reviews~\cite{Taboada:06}, but the latter system could perform much
better, achieving 73\% on this two-class prediction task.

Other ways of adding discourse information to a sentiment system were
explored by \citet{Heerschop:11}, who experimented with three
different approaches:
\begin{inparaenum}[(i)]
\item increasing the polarity scores of words that appeared near the
  end of the document,
\item assigning higher weights to nucleus tokens, and finally
\item learning separate scores for nuclei and satellites using a
  genetic algorithm.
\end{inparaenum}
An evaluation of these methods on the movie review corpus
of~\citet{Pang:04} showed better performance of the first option
(60.8\% accuracy and 0.597 macro-\F), but the authors could
significantly improve the results of the last classifier at the end by
adding an offset to the decision boundary of this method, which
increased both its accuracy and macro-averaged \F{} to 0.72.

Further notable contributions to RST-based sentiment analysis were
made by \citet{Zhou:11}, who used a set of heuristic rules to infer
polarity shifts of discourse units based on their nuclearity status
and outgoing relation links; \citet{Zirn:11}, who used a lexicon-based
sentiment system to predict the polarity scores of elementary
discourse units and then enforced consistency of these assignments
over the RST tree with the help of Markov logic constraints; and,
finally, \citet{Wang:13}, who determined the semantic orientation of a
document by taking a linear combination of the polarity scores of its
EDUs and multiplying these scores with automatically learned
coefficients.


Among the most recent advances in RST-aware sentiment research, we
should especially emphasize the work of \citet{Bhatia:15}, who
proposed two different DASA systems:
\begin{itemize}
\item discourse-depth reweighting (DDR)
\item and rhetorical recursive neural network (R2N2).
\end{itemize}
In the former approach, the authors estimated the relevance
$\lambda_i$ of each elementary discourse unit $i$ as:
\begin{equation*}
  \lambda_i = \max\left(0.5, 1 - d_i/6\right),
\end{equation*}
where $d_i$ stands for the depth of the $i$-th EDU in the document's
discourse tree.  Afterwards, they computed the sentiment score
$\sigma_i$ of that unit by taking the dot product of its binary
feature vector $\mathbf{w}_i$ (token unigrams) with polarity scores
$\boldsymbol{\theta}$ of these unigrams:
\begin{equation*}
  \sigma_i = \boldsymbol{\theta}^{\top}\mathbf{w}_i;
\end{equation*}
and then calculated the overall semantic orientation of the
document~$\Psi$ as the sum of sentiment scores for all units,
multiplying these scores by their respective discourse-depth factors:
\begin{equation*}
  \Psi = \sum_i\lambda_i\boldsymbol{\theta}^T\mathbf{w}_i = \boldsymbol{\theta}^T\sum_i\lambda_i\mathbf{w}_i,
\end{equation*}
In the R2N2 system, the authors largely adopted the RNN method
of~\citet{Socher:13} by recursively computing the polarity scores of
discourse units as:
\begin{equation*}
  \psi_i = \tanh\left(K_n^{(r_i)} \psi_{n(i)} + K_s^{(r_i)}\psi_{s(i)} \right),
\end{equation*}
where $K_n^{(r_i)}$ and $K_s^{(r_i)}$ stand for the nucleus and
satellite coefficients associated with the rhetorical relation $r_i$,
and $\psi_{n(i)}$ and $\psi_{s(i)}$ represent sentiment scores of the
nucleus and satellite of the $i$-th vertex.  This approach achieved
84.1\% two-class accuracy on the movie review corpus
of~\citet{Pang:04} and reached 85.6\% on the dataset
of~\citet{Socher:13}.

For the sake of completeness, we should also note that there exist
discourse-aware sentiment approaches that build upon PDTB and SDRT\@.
For example, \citet{Trivedi:13} proposed a method based on latent
structural SVM~\cite{Yu:09}, where they represented each sentence as a
vector of features produced by a feature function $\mathbf{f}(y,
\mathbf{x}_i, h_i)$, in which $y\in\{-1, +1\}$ denotes the potential
polarity of the whole document, $h_i \in \{0, 1\}$ stands for the
assumed subjectivity class of sentence $i$, and $\mathbf{x}_i$
represents the surface form of that sentence; and then tried to infer
the most likely semantic orientation of the document $\hat{y}$ over
all possible assignments $\mathbf{h}$, \ie{}:
\begin{equation*}
  \hat{y} =
  \argmax_y\left(\max_{\mathbf{h}}\mathbf{w}^{\top}\mathbf{f}(y,
  \mathbf{x}, \mathbf{h})\right).
\end{equation*}
To ensure that these assignments were still coherent, the authors
additionally extended their feature space with special
\emph{transitional} attributes, which indicated whether two adjacent
sentences were likely to share the same subjectivity given the
discourse connective between them.  With the help of these features,
\citeauthor{Trivedi:13} could improve the accuracy of the
connector-unaware model on the movie review corpus of~\citet{Maas:11}
from 88.21 to 91.36\%.

The first step towards an SDRT-based sentiment approach was made by
\citet{Asher:08}, who presented an annotation scheme and a pilot
corpus of English and French texts that were analyzed according to the
SDRT theory and enriched with additional sentiment information.
Specifically, the authors asked the annotators to ascribe one of four
opinion categories (reporting, judgment, advice, or sentiment) along
with their subclasses (\eg{} inform, assert, blame, recommend) to each
discourse unit that had at least one opinionated word from a sentiment
lexicon.  Afterwards, they showed that with a simple set of rules, one
could easily propagate opinions through SDRT graphs, increasing the
strengths or reversing the polarity of the sentiments, depending on
the type of the discourse relation that was linking two segments.

In general, however, PDTB- and SDRT-based sentiment systems are much
less common than RST-inspired solutions.  Because of this fact and due
to the reasons described in Section~\ref{sec:dasa:theory}, we will
primarily concentrate on the RST-based of methods.  In particular, for
the sake of comparison, we replicated the linear combination approach
of \citet{Wang:13} and also reimplemented the DDR and R2N2 systems
of~\citet{Bhatia:15}.  Furthermore, to see how these techniques would
perform in comparison with much simpler baselines, we additionally
created two methods that predicted the polarity of a message by only
considering its last or topmost nucleus EDU (henceforth \textsc{Last}
and \textsc{Root}), and also estimated the results of our original LBA
classifier without any discourse-related modifications (henceforth
\textsc{No-Discourse}).

Apart from the above baselines and existing methods, we propose
several novel DASA solutions, which will be briefly described below.

\subsection{Latent CRF}

In the first of these solutions, called \emph{Latent Conditional
  Random Fields} or \emph{LCRFs}, we consider the problem of
message-level sentiment analysis as an inference task over an
undirected graphical model, where the nodes of the model represent
polarity probabilities of elementary discourse units and the structure
of the graph reflects the RST dependency tree of the
message.\footnote{Drawing on the work of~\citet{Bhatia:15}, we obtain
  this representation using the DEP-DT algorithm of~\citet{Hirao:13}
  with a minor modification that we do not follow any satellite
  branches while computing the heads of abstract RST nodes in Step 1
  of this procedure~\cite[see][pp.~1516--1517]{Hirao:13}.}  In
particular, we define CRF graph $\mathcal{G}=(\mathcal{V},
\mathcal{E})$ as a set of vertices $\mathcal{V}=
\mathcal{Y}\cup\mathcal{X}$, in which $\mathcal{Y}=\{y_{(i, j)}\mid
i\in\{\text{\textsc{Root}}, 1, 2, \ldots, T\}, j
\in\{\text{\textsc{Negative}, \textsc{Neutral},
  \textsc{Positive}}\}\}$ represents (partially observed) random
variables (with $T$ standing for the number of EDUs in the tweet), and
$\mathcal{X}=\{x_{(i, j)}\mid i\in\{\text{\textsc{Root}}, 1, 2,
\ldots, T\}, j \in[0,\ldots, 3]\}$ denotes the respective features of
these nodes (three polarity scores returned by the LBA classifier plus
an additional offset feature whose value is always \texttt{1}
irrespectively of the input).  Since the \textsc{Root} vertex,
however, does not have a corresponding discourse segment in the RST
tree, we use the polarity scores predicted by the LBA classifier for
the whole message as features for this node.

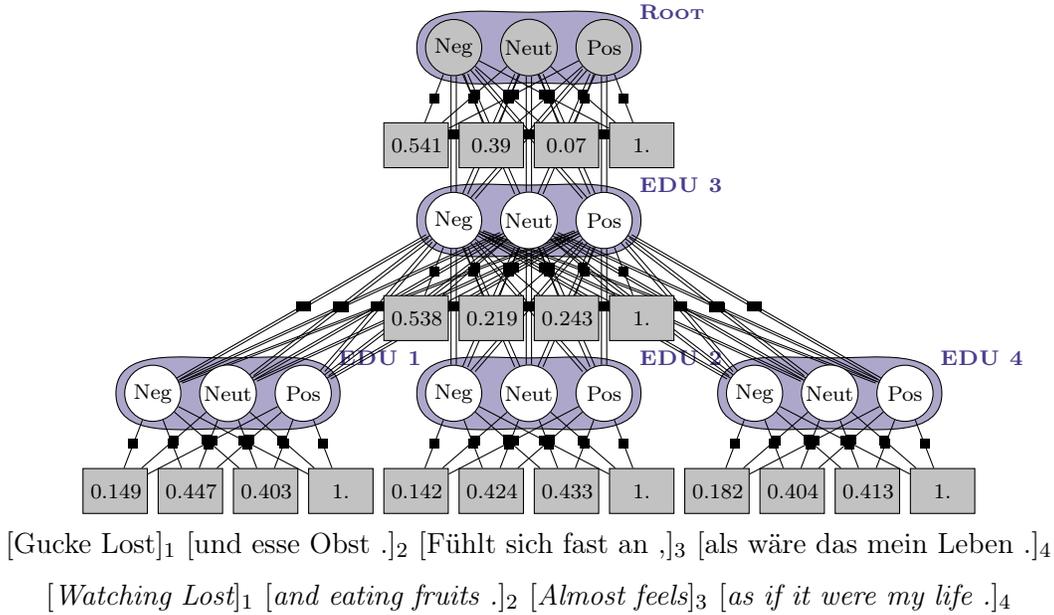
\begin{figure*}[thb]
  \centering {\scriptsize
  \begin{tikzpicture}
    \tikzstyle{xnode}=[rectangle,draw,fill=gray76,align=center,minimum size=2em,text width=2.2em] %
    \tikzstyle{lnode}=[circle,draw,fill=white,inner sep=1pt,minimum size=2.5em] %
    \tikzstyle{ynode}=[circle,draw,fill=gray76,inner sep=1pt,minimum size=2.5em] %
    \tikzstyle{label}=[] %
    \tikzstyle{extext}=[] %
    \tikzstyle{factor}=[rectangle,fill=black,midway,inner sep=0pt,minimum size=0.4em] %

    \node[ynode] (0Neg) at (5, 6) {Neg};
    \node[ynode] (0Neut) at (6, 6) {Neut};
    \node[ynode] (0Pos) at (7, 6) {Pos};
    \hyperNode{0Neg}{0Neut}{0Pos}{Root};
    \crfFeatures{0/4.5/0.541, 1/5.5/0.39, 2/6.5/0.07, 3/7.5/1.}{0}{4.7};

    \node[lnode] (1Neg) at (5, 3.7) {Neg};
    \node[lnode] (1Neut) at (6, 3.7) {Neut};
    \node[lnode] (1Pos) at (7, 3.7) {Pos};
    \hyperNode{1Neg}{1Neut}{1Pos}{EDU 3};
    \crfEdgesLatent{0}{1};             
    \crfFeatures{0/4.5/0.538, 1/5.5/0.219, 2/6.5/0.243, 3/7.5/1.}{1}{2.4};

    \node[lnode] (2Neg) at (1, 1.4) {Neg};
    \node[lnode] (2Neut) at (2, 1.4) {Neut};
    \node[lnode] (2Pos) at (3, 1.4) {Pos};
    \hyperNode{2Neg}{2Neut}{2Pos}{EDU 1};
    \crfEdgesLatent{1}{2};
    \crfFeatures{0/0.5/0.149, 1/1.5/0.447, 2/2.5/0.403, 3/3.5/1.}{2}{0.1};

    \node[lnode] (3Neg) at (5, 1.4) {Neg};
    \node[lnode] (3Neut) at (6, 1.4) {Neut};
    \node[lnode] (3Pos) at (7, 1.4) {Pos};
    \hyperNode{3Neg}{3Neut}{3Pos}{EDU 2};
    \crfEdgesLatent{1}{3};
    \crfFeatures{0/4.5/0.142, 1/5.5/0.424, 2/6.5/0.433, 3/7.5/1.}{3}{0.1};

    \node[lnode] (4Neg) at (9, 1.4) {Neg};
    \node[lnode] (4Neut) at (10, 1.4) {Neut};
    \node[lnode] (4Pos) at (11, 1.4) {Pos};
    \hyperNode{4Neg}{4Neut}{4Pos}{EDU 4};
    \crfEdgesLatent{1}{4};
    \crfFeatures{0/8.5/0.182, 1/9.5/0.404, 2/10.5/0.413, 3/11.5/1.}{4}{0.1};

    \node[extext] (GText) [below=4.5em of 3Neut] {\small $[$Gucke Lost$]_1$ $[$und esse Obst .$]_2$ $[$F\"uhlt sich fast an ,$]_3$ $[$als w\"are das mein Leben .$]_4$};
    \node[extext] (EText) [below=0.3em of GText] {\small\itshape $[$Watching Lost$]_1$ $[$and eating fruits .$]_2$ $[$Almost feels$]_3$ $[$as if it were my life .$]_4$};
\end{tikzpicture}
}
  \caption[Example of an RST-based Latent-CRF]{Example of an
    automatically constructed RST-based latent-CRF tree\\ {\small
      (random variables are shown as circles, fixed input parameters
      appear as rectangles, and observed values are displayed in
      gray)}}\label{dasa:fig:latent-crf}
\end{figure*}

Graph edges $\mathcal{E}$ connect random variables to their
corresponding features and also link every pair of vertices
$(v_{(k,\cdot)},v_{(i,\cdot)})$ if node $k$ appears as the parent of
node $i$ in the RST dependencies.\footnote{In fact, we use two edges
  to connect each child to its parent: one for the
  \textsc{Contrastive} relation and another one for the
  \textsc{Non-Contrastive} link.}  You can see an example of such
automatically induced CRF tree in Figure~\ref{dasa:fig:latent-crf}.

Now before we describe the training of our model, let us briefly
recall that in the standard CRF optimization we typically try to find
optimal parameters $\boldsymbol{\theta}^*$ that maximize the
log-likelihood of all label sequences $\mathbf{y}^{(i)}$ on the
training set $\mathcal{D}=\left\{\left(\mathbf{x}^{(i)},
\mathbf{y}^{(i)}\right)\right\}_{i=1}^{N}$, \ienocomma:
\begin{equation*}
  \boldsymbol{\theta}^* = \argmax_{\boldsymbol{\theta}}\ell(\boldsymbol{\theta}) = \sum_{i=1}^{N}\log\left(p\left(\mathbf{y}^{(i)}\vert\mathbf{x}^{(i)}; \boldsymbol{\theta}\right)\right),\label{dasa:eq:crf-objective}
\end{equation*}
where the conditional likelihood is normally estimated as:
\begin{equation*}
  p\left(\mathbf{y}^{(i)}\vert\mathbf{x}^{(i)}; \boldsymbol{\theta}\right) =
  \frac{\exp\left(\sum_{t=1}^{T_i}\sum_k\boldsymbol{\theta}_k\mathbf{f}_k\left(\mathbf{x}^{(i)}_t,\mathbf{y}^{(i)}_{t-t},\mathbf{y}^{(i)}_{t}\right)\right)}{Z}.
\end{equation*}
Adapting this equation to our RST-based CRF structures, we obtain:
\begin{equation}
  p\left(\mathbf{y}^{(i)}\vert\mathbf{x}^{(i)}; \boldsymbol{\theta}\right) =
  \frac{\exp\left(\sum_{t=0}^{T_i}\left[%
      \sum_v\boldsymbol{\theta}_v\mathbf{f}_v\left(\mathbf{x}^{(i)}_t,\mathbf{y}^{(i)}_{t}\right)
      + \sum_{c\in
        ch(t)}\sum_e\boldsymbol{\theta}_e\mathbf{f}_e\left(\mathbf{y}^{(i)}_{t},
      \mathbf{y}^{(i)}_{c}\right)\right]\right)}{Z},\label{dasa:eq:tree-crf}
\end{equation}
where $ch(t)$ denotes the children of node $t$, $v$ stands for the
indices of node features, and $e$ represents the indices of edge
attributes.

A crucial problem with this formulation though is that in our task,
only a small subset of labels from $\mathbf{y}^{(i)}$ (namely those of
the root node) are actually observed at the training time, whereas the
rest of the tags (those which pertain to EDUs) are unknown.  We will
denote these observed and hidden subsets as $\mathbf{y}_o^{(i)}$ and
$\mathbf{y}_h^{(i)}$ respectively.  Using this notation, we can
redefine the training objective of our model as finding such
parameters $\boldsymbol{\theta}^*$ that maximize the log-likelihood of
\emph{observed} labels, \ienocomma:
\begin{equation*}
  \boldsymbol{\theta}^* =
  \argmax_{\boldsymbol{\theta}}\sum_{i=1}^{N}\log\left(p\left(\mathbf{y}_o^{(i)}\vert\mathbf{x}^{(i)};
  \boldsymbol{\theta}\right)\right).
\end{equation*}
With this formulation, however, it is still unclear what we should do
with hidden tags $\mathbf{y}_h^{(i)}$, because the values of their
features remain undefined.

One possible way to approach the problem of unobserved states in the
input is to assume that any label sequence $\mathbf{y}_h^{(i)}$ might
be true, and then try to maximize the parameters along the path that
leads to the maximum probability of the correct observed tag,
\ienocomma:
\begin{align}
  \begin{split}
    \mathbf{y}^{(i)}&=[\mathbf{y}_o^{(i)}, \mathbf{y}_h^{*(i)}]\text{, where}\\\label{dasa:eq:y_i}
    \mathbf{y}_h^{*(i)}&=\argmax_{\mathbf{y}_h^{(i)}}p\left(\mathbf{y}_o^{(i)}\vert\mathbf{x}^{(i)}\right),
  \end{split}
\end{align}
and which we can easily find using standard Viterbi decoding.

Unfortunately, if we simply consider label sequence $\mathbf{y}^{(i)}$
from Equation~\ref{dasa:eq:y_i} as the ground truth and penalize all
labels that disagree with this sequence, we might overly commit
ourselves to the model's guess of unknown tags and unduly discriminate
against other possible hidden label assignments.  To mitigate this
effect, we can instead penalize only one other sequence, namely the
one that maximizes the probability of an incorrect label at the
observed state:
\begin{align*}
  \mathbf{y}^{'(i)}&=\argmax_{\mathbf{y}_o^{'(i)}\neq\mathbf{y}_o^{(i)}}p\left([\mathbf{y}_o^{'(i)},
    \mathbf{y}_h^{*(i)}]\vert\mathbf{x}^{(i)}\right)\text{,
    where}\\
  \mathbf{y}_h^{*(i)}&=\argmax_{\mathbf{y}_h^{(i)}}p\left(\mathbf{y}_o^{'(i)}\vert\mathbf{x}^{(i)}\right).
\end{align*}
Correspondingly, we reformulate our objective and instead of
maximizing the log-likelihood of the training set will now maximize
the difference between the log-probabilities of the correct and most
likely wrong assignments:
\begin{align}
  \begin{split}
    \boldsymbol{\theta}^* &= \argmax_{\boldsymbol{\theta}}\sum_{i=1}^{N}\log\left(p\left(\mathbf{y}^{(i)}\right)\right) - \log\left(p\left(\mathbf{y}^{'(i)}\right)\right)\\
    &= \argmax_{\boldsymbol{\theta}}\sum_{i=1}^{N}\log\left(\exp\left(\boldsymbol{\theta}^{\top}\mathbf{f}(\mathbf{x}^{(i)},\mathbf{y}^{(i)})\right)\right) - \log\left(\exp\left(\boldsymbol{\theta}^{\top}\mathbf{f}(\mathbf{x}^{(i)},\mathbf{y}^{'(i)})\right)\right)\\
    &= \argmax_{\boldsymbol{\theta}}\sum_{i=1}^{N}\boldsymbol{\theta}^{\top}\left(\mathbf{f}(\mathbf{x}^{(i)},\mathbf{y}^{(i)}) - \mathbf{f}(\mathbf{x}^{(i)},\mathbf{y}^{'(i)})\right),\label{dasa:eq:hcrf-objective}
  \end{split}
\end{align}
where $\mathbf{f}(\mathbf{x}^{(i)},\mathbf{y}^{(i)})$ and
$\mathbf{f}(\mathbf{x}^{(i)},\mathbf{y}^{'(i)})$ mean all features
associated with label sequences $\mathbf{y}^{(i)}$ and
$\mathbf{y}^{'(i)}$ respectively.

The only thing that we now need to do to the above objective is to
introduce a regularization term
$\frac{1}{2}\norm{\boldsymbol{\theta}}^2$ in order to prevent its
divergence to infinity in the cases when
$\mathbf{f}(\mathbf{x}^{(i)},\mathbf{y}^{(i)})$ and
$\mathbf{f}(\mathbf{x}^{(i)},\mathbf{y}^{'(i)})$ are perfectly
separable.  This brings us to the final formulation:
\begin{align}
  \boldsymbol{\theta}^* &=
  \argmin_{\boldsymbol{\theta}}\frac{1}{2}\norm{\boldsymbol{\theta}}^2 -
  \sum_{i=1}^{N}\boldsymbol{\theta}^{\top}\left(\mathbf{f}(\mathbf{x}^{(i)},\mathbf{y}^{(i)})
  - \mathbf{f}(\mathbf{x}^{(i)},\mathbf{y}^{'(i)})\right)
\end{align}
At this point, we can notice that the resulting function is identical
to the unconstrained minimization problem of structural
SVM~\cite{Taskar:03}, and we indeed can piggyback on one of the many
efficient SVM-optimization techniques to learn the parameters of our
model.  In particular, we use the block-coordinate Frank-Wolfe
algorithm~\cite{Lacoste-Julien:13}, running it for 1,000 epochs or
until convergence, whichever of these events occurs first.

\subsection{Latent-Marginalized CRF}

Another way to tackle unobserved labels is to estimate the probability
of observed tags by marginalizing (summing) out hidden variables from
the joint distribution, \ienocomma:
\begin{align*}
  p\left(\mathbf{Y}_o{=}\mathbf{y}_o\right) &=
  \sum_{\mathbf{y}_h} p\left(\mathbf{Y}_o{=}\mathbf{y}_o,
  \mathbf{Y}_h{=}\mathbf{y}_h\right).
\end{align*}
Applying this formula to Equation~\ref{dasa:eq:tree-crf}, we get:
\begin{align*}
  \begin{split}
    p\left(\mathbf{y}_o^{(i)}\vert\mathbf{x}^{(i)}; \boldsymbol{\theta}\right) &=%
    \sum_{\mathbf{y}_h^{(i)}} p\left([\mathbf{y}_o^{(i)}, \mathbf{y}_h^{(i)}]%
    \vert\mathbf{x}^{(i)}; \boldsymbol{\theta}\right)\\
    &= \frac{\sum_{\mathbf{y}_h^{(i)}}\exp\left(\sum_{t=0}^{T_i}\left[%
        \sum_v\boldsymbol{\theta}_v\mathbf{f}_v\left(\mathbf{x}^{(i)}_t,\mathbf{y}^{(i)}_{t}\right)
        + \sum_{c\in
          ch(t)}\sum_e\boldsymbol{\theta}_e\mathbf{f}_e\left(\mathbf{y}^{(i)}_{t},
        \mathbf{y}^{(i)}_{c}\right)\right]\right)}{Z},
  \end{split}
\end{align*}
where $\mathbf{y}^{(i)}$ in the numerator is defined as before:
$\mathbf{y}^{(i)}=[\mathbf{y}_o^{(i)}, \mathbf{y}_h^{(i)}]$.

This time again, we would like to maximize the probability of the
correct assignment, setting it apart from its closest competitor by
some margin.  Unfortunately, due to the summation over all
$\mathbf{y}_h^{(i)}$, we cannot avail ourselves of the log-exp
cancellation trick, which we used previously in
Equation~\ref{dasa:eq:hcrf-objective}.  Instead of this, we replace
the difference of the log-likelihoods by the ratio of marginal
probabilities:
\begin{align}
  \begin{split}
    \boldsymbol{\theta}^* &= \argmax_{\boldsymbol{\theta}}\sum_{i=1}^{N}\frac{p(\mathbf{y}^{(i)})}{p(\mathbf{y}^{'(i)})}\\
    &= \argmax_{\boldsymbol{\theta}}\sum_{i=1}^{N}
    \frac{\sum_{\mathbf{y}_h^{(i)}}\exp\left(\sum_{t=0}^{T_i}\left[%
        \sum_v\boldsymbol{\theta}_v\mathbf{f}_v\left(\mathbf{x}^{(i)}_t,\mathbf{y}^{(i)}_{t}\right)
        + \sum_{c\in
          ch(t)}\sum_e\boldsymbol{\theta}_e\mathbf{f}_e\left(\mathbf{y}^{(i)}_{t},
        \mathbf{y}^{(i)}_{c}\right)\right]\right)}{\sum_{\mathbf{y}_h^{(i)}}\exp\left(\sum_{t=0}^{T_i}\left[%
        \sum_v\boldsymbol{\theta}_v\mathbf{f}_v\left(\mathbf{x}^{(i)}_t,\mathbf{y}^{'(i)}_{t}\right)
        + \sum_{c\in
          ch(t)}\sum_e\boldsymbol{\theta}_e\mathbf{f}_e\left(\mathbf{y}^{'(i)}_{t},
        \mathbf{y}^{'(i)}_{c}\right)\right]\right)}\label{dasa:eq:hmcrf-objective}
  \end{split}
\end{align}
To simplify this expression, we can introduce the following
abbreviations:
\begin{align*}
  a &\defeq \exp\left(\sum_{t=0}^{T_i}\left[
    \sum_v\boldsymbol{\theta}_v\mathbf{f}_v\left(\mathbf{x}^{(i)}_t,\mathbf{y}^{(i)}_{t}\right)
    + \sum_{c\in
      ch(t)}\sum_e\boldsymbol{\theta}_e\mathbf{f}_e\left(\mathbf{y}^{(i)}_{t},
    \mathbf{y}^{(i)}_{c}\right)\right]\right),\\
  b &\defeq \exp\left(\sum_{t=0}^{T_i}\left[
    \sum_v\boldsymbol{\theta}_v\mathbf{f}_v\left(\mathbf{x}^{(i)}_t,\mathbf{y}^{'(i)}_{t}\right)
    + \sum_{c\in
      ch(t)}\sum_e\boldsymbol{\theta}_e\mathbf{f}_e\left(\mathbf{y}^{'(i)}_{t},
    \mathbf{y}^{'(i)}_{c}\right)\right]\right).
\end{align*}
Now we estimate the derivatives of functions $a$ and $b$ w.r.t.~a
single parameter $\boldsymbol{\theta}_v$ as:
\begin{align*}
  \frac{\partial{}a}{\partial\boldsymbol{\theta}_v} &= a\sum_{t=0}^{T_i}\mathbf{f}_v\left(\mathbf{x}^{(i)}_t,\mathbf{y}^{(i)}_{t}\right)\propto\mathbb{E}_{\mathbf{y}^{(i)}}\left[\mathbf{f}_v\right],\\
  \frac{\partial{}b}{\partial\boldsymbol{\theta}_v} &= b\sum_{t=0}^{T_i}\mathbf{f}_v\left(\mathbf{x}^{(i)}_t,\mathbf{y}^{(i)}_{t}\right)\propto\mathbb{E}_{\mathbf{y}^{'(i)}}\left[\mathbf{f}_v\right];
\end{align*}
and analogously obtain:
\begin{align*}
  \frac{\partial{}a}{\partial\boldsymbol{\theta}_e} &= a\sum_{t=0}^{T_i}\sum_{c\in ch(t)}\mathbf{f}_e\left(\mathbf{y}^{(i)}_{t}, \mathbf{y}^{(i)}_{c}\right)\propto\mathbb{E}_{\mathbf{y}^{(i)}}\left[\mathbf{f}_e\right],\\
  \frac{\partial{}b}{\partial\boldsymbol{\theta}_e} &= b\sum_{t=0}^{T_i}\sum_{c\in ch(t)}\mathbf{f}_e\left(\mathbf{y}^{'(i)}_{t}, \mathbf{y}^{'(i)}_{c}\right)\propto\mathbb{E}_{\mathbf{y}^{'(i)}}\left[\mathbf{f}_e\right].
\end{align*}
With the help of these expressions, we can easily compute the gradient
of the objective function w.r.t. $\boldsymbol{\theta}$ by observing
that:
\begin{align}
  \nabla_{\boldsymbol{\theta}} &=
  \sum_{i=1}^{N}\frac{\sum_{\mathbf{y}_h^{(i)}}\nabla_{\boldsymbol{\theta}}a\sum_{\mathbf{y}_h^{(i)}}b
    -
    \sum_{\mathbf{y}_h^{(i)}}a\sum_{\mathbf{y}_h^{(i)}}\nabla_{\boldsymbol{\theta}}b}{\left(\sum_{\mathbf{y}_h^{(i)}}b\right)^{2}}.\label{dasa:eq:lmcrf-gradient}
\end{align}
We again use the block-coordinate Frank-Wolfe algorithm to optimize
the parameters of our model, but instead of pushing these parameters
in the direction
$\boldsymbol{\psi}=\mathbf{f}(\mathbf{x}^{(i)},\mathbf{y}^{(i)})-\mathbf{f}(\mathbf{x}^{(i)},\mathbf{y}^{'(i)})$
(which is the derivative of latent CRFs, see Algorithm~2 in
[\citeauthor{Lacoste-Julien:13}, \citeyear{{Lacoste-Julien:13}}]), we
now maximize them along the gradient from
Equation~\ref{dasa:eq:lmcrf-gradient}.

\begin{figure*}[thb]
  \centering
  \begin{subfigure}[t]{0.4\textwidth}
    \centering
    \resizebox{\hcrfwidth}{\hcrfheight}{
      {\scriptsize
  \begin{tikzpicture}
    \tikzstyle{xnode}=[rectangle,draw,fill=gray76,align=center,minimum size=1.5em,text width=1.5em] %
    \tikzstyle{lnode}=[circle,draw,fill=white,inner sep=1pt,minimum size=2.5em] %
    \tikzstyle{ynode}=[circle,draw,fill=gray76,inner sep=1pt,minimum size=2.5em] %
    \tikzstyle{highlight}=[draw=red,line width=0.5mm] %
    \tikzstyle{label}=[] %
    \tikzstyle{extext}=[] %
    \tikzstyle{factor}=[rectangle,fill=black,midway,inner sep=0pt,minimum size=0.4em] %

    \node[ynode] (0Neg) at (5, 6) {Neg};
    \node[ynode,highlight] (0Neut) at (6, 6) {Neut};
    \node[ynode] (0Pos) at (7, 6) {Pos};
    \hyperNode{0Neg}{0Neut}{0Pos}{Root};
    \crfColoredFeatures{0/4.5/0.54, 1/5.5/0.39, 2/6.5/0.07, 3/7.5/1.}{0}{4.7}{Neut};

    \node[lnode] (1Neg) at (5, 3.7) {Neg};
    \node[lnode,highlight] (1Neut) at (6, 3.7) {Neut};
    \node[lnode] (1Pos) at (7, 3.7) {Pos};
    \hyperNode{1Neg}{1Neut}{1Pos}{EDU 3};
    \crfColoredEdgesLatent{0}{1}{0Neut}{1Neut}; 
    \crfColoredFeatures{0/4.5/0.54, 1/5.5/0.22, 2/6.5/0.24, 3/7.5/1.}{1}{2.4}{Neut};

    \node[lnode] (2Neg) at (1, 1.4) {Neg};
    \node[lnode,highlight] (2Neut) at (2, 1.4) {Neut};
    \node[lnode] (2Pos) at (3, 1.4) {Pos};
    \hyperNode{2Neg}{2Neut}{2Pos}{EDU 1};
    \crfColoredEdgesLatent{1}{2}{1Neut}{2Neut};
    \crfColoredFeatures{0/0.5/0.15, 1/1.5/0.45, 2/2.5/0.4, 3/3.5/1.}{2}{0.1}{Neut};

    \node[lnode] (3Neg) at (5, 1.4) {Neg};
    \node[lnode,highlight] (3Neut) at (6, 1.4) {Neut};
    \node[lnode] (3Pos) at (7, 1.4) {Pos};
    \hyperNode{3Neg}{3Neut}{3Pos}{EDU 2};
    \crfColoredEdgesLatent{1}{3}{1Neut}{3Neut};
    \crfColoredFeatures{0/4.5/0.14, 1/5.5/0.42, 2/6.5/0.43, 3/7.5/1.}{3}{0.1}{Neut};

    \node[lnode] (4Neg) at (9, 1.4) {Neg};
    \node[lnode] (4Neut) at (10, 1.4) {Neut};
    \node[lnode,highlight] (4Pos) at (11, 1.4) {Pos};
    \hyperNode{4Neg}{4Neut}{4Pos}{EDU 4};
    \crfColoredEdgesLatent{1}{4}{1Neut}{4Pos};
    \crfColoredFeatures{0/8.5/0.18, 1/9.5/0.4, 2/10.5/0.41, 3/11.5/1.}{4}{0.1}{Pos};

\end{tikzpicture}
}
    }
    \caption{Computational path of the probability of the correct
      label in latent CRF}
  \end{subfigure}
  ~
  \begin{subfigure}[t]{0.4\textwidth}
    \centering
    \resizebox{\hcrfwidth}{\hcrfheight}{
      {\scriptsize
  \begin{tikzpicture}
    \tikzstyle{xnode}=[rectangle,draw,fill=gray76,align=center,minimum size=1.5em,text width=1.5em] %
    \tikzstyle{lnode}=[circle,draw,fill=white,inner sep=1pt,minimum size=2.5em] %
    \tikzstyle{ynode}=[circle,draw,fill=gray76,inner sep=1pt,minimum size=2.5em] %
    \tikzstyle{highlight}=[draw=blue,line width=0.5mm] %
    \tikzstyle{label}=[] %
    \tikzstyle{extext}=[] %
    \tikzstyle{factor}=[rectangle,fill=black,midway,inner sep=0pt,minimum size=0.4em] %

    \node[ynode,highlight] (0Neg) at (5, 6) {Neg};
    \node[ynode] (0Neut) at (6, 6) {Neut};
    \node[ynode] (0Pos) at (7, 6) {Pos};
    \hyperNode{0Neg}{0Neut}{0Pos}{Root};
    \crfColoredFeatures{0/4.5/0.54, 1/5.5/0.39, 2/6.5/0.07, 3/7.5/1.}{0}{4.7}{Neg};

    \node[lnode,highlight] (1Neg) at (5, 3.7) {Neg};
    \node[lnode] (1Neut) at (6, 3.7) {Neut};
    \node[lnode] (1Pos) at (7, 3.7) {Pos};
    \hyperNode{1Neg}{1Neut}{1Pos}{EDU 3};
    \crfColoredEdgesLatent{0}{1}{0Neg}{1Neg}; 
    \crfColoredFeatures{0/4.5/0.54, 1/5.5/0.22, 2/6.5/0.24, 3/7.5/1.}{1}{2.4}{Neg};

    \node[lnode] (2Neg) at (1, 1.4) {Neg};
    \node[lnode,highlight] (2Neut) at (2, 1.4) {Neut};
    \node[lnode] (2Pos) at (3, 1.4) {Pos};
    \hyperNode{2Neg}{2Neut}{2Pos}{EDU 1};
    \crfColoredEdgesLatent{1}{2}{1Neg}{2Neut};
    \crfColoredFeatures{0/0.5/0.15, 1/1.5/0.45, 2/2.5/0.4, 3/3.5/1.}{2}{0.1}{Neut};

    \node[lnode] (3Neg) at (5, 1.4) {Neg};
    \node[lnode,highlight] (3Neut) at (6, 1.4) {Neut};
    \node[lnode] (3Pos) at (7, 1.4) {Pos};
    \hyperNode{3Neg}{3Neut}{3Pos}{EDU 2};
    \crfColoredEdgesLatent{1}{3}{1Neg}{3Neut};
    \crfColoredFeatures{0/4.5/0.14, 1/5.5/0.42, 2/6.5/0.43, 3/7.5/1.}{3}{0.1}{Neut};

    \node[lnode] (4Neg) at (9, 1.4) {Neg};
    \node[lnode,highlight] (4Neut) at (10, 1.4) {Neut};
    \node[lnode] (4Pos) at (11, 1.4) {Pos};
    \hyperNode{4Neg}{4Neut}{4Pos}{EDU 4};
    \crfColoredEdgesLatent{1}{4}{1Neg}{4Neut};
    \crfColoredFeatures{0/8.5/0.18, 1/9.5/0.4, 2/10.5/0.41, 3/11.5/1.}{4}{0.1}{Neut};

\end{tikzpicture}
}
    }
    \caption{Computational path of the probability of a wrong label in
      latent CRF}
  \end{subfigure}\\[1em]
  \begin{subfigure}[t]{0.4\textwidth}
    \centering
    \resizebox{\hcrfwidth}{\hcrfheight}{
      {\scriptsize
  \begin{tikzpicture}
    \tikzstyle{xnode}=[rectangle,draw,fill=gray76,align=center,minimum size=1.5em,text width=1.5em] %
    \tikzstyle{lnode}=[circle,draw,fill=white,inner sep=1pt,minimum size=2.5em] %
    \tikzstyle{ynode}=[circle,draw,fill=gray76,inner sep=1pt,minimum size=2.5em] %
    \tikzstyle{highlight}=[draw=red,line width=0.5mm] %
    \tikzstyle{label}=[] %
    \tikzstyle{extext}=[] %
    \tikzstyle{factor}=[rectangle,fill=black,midway,inner sep=0pt,minimum size=0.4em] %

    \node[ynode] (0Neg) at (5, 6) {Neg};
    \node[ynode,highlight] (0Neut) at (6, 6) {Neut};
    \node[ynode] (0Pos) at (7, 6) {Pos};
    \hyperNode{0Neg}{0Neut}{0Pos}{Root};
    \crfColoredFeatures{0/4.5/0.54, 1/5.5/0.39, 2/6.5/0.07, 3/7.5/1.}{0}{4.7}{Neut};

    \node[lnode,highlight] (1Neg) at (5, 3.7) {Neg};
    \node[lnode,highlight] (1Neut) at (6, 3.7) {Neut};
    \node[lnode,highlight] (1Pos) at (7, 3.7) {Pos};
    \hyperNode{1Neg}{1Neut}{1Pos}{EDU 3};
    \crfColoredEdgesLatent{0}{1}{0Neut}{ANY}; 
    \crfColoredFeatures{0/4.5/0.54, 1/5.5/0.22, 2/6.5/0.24, 3/7.5/1.}{1}{2.4}{ANY};

    \node[lnode,highlight] (2Neg) at (1, 1.4) {Neg};
    \node[lnode,highlight] (2Neut) at (2, 1.4) {Neut};
    \node[lnode,highlight] (2Pos) at (3, 1.4) {Pos};
    \hyperNode{2Neg}{2Neut}{2Pos}{EDU 1};
    \crfColoredEdgesLatent{1}{2}{ANY}{ANY};
    \crfColoredFeatures{0/0.5/0.15, 1/1.5/0.45, 2/2.5/0.4, 3/3.5/1.}{2}{0.1}{ANY};

    \node[lnode,highlight] (3Neg) at (5, 1.4) {Neg};
    \node[lnode,highlight] (3Neut) at (6, 1.4) {Neut};
    \node[lnode,highlight] (3Pos) at (7, 1.4) {Pos};
    \hyperNode{3Neg}{3Neut}{3Pos}{EDU 2};
    \crfColoredEdgesLatent{1}{3}{ANY}{ANY};
    \crfColoredFeatures{0/4.5/0.14, 1/5.5/0.42, 2/6.5/0.43, 3/7.5/1.}{3}{0.1}{ANY};

    \node[lnode,highlight] (4Neg) at (9, 1.4) {Neg};
    \node[lnode,highlight] (4Neut) at (10, 1.4) {Neut};
    \node[lnode,highlight] (4Pos) at (11, 1.4) {Pos};
    \hyperNode{4Neg}{4Neut}{4Pos}{EDU 4};
    \crfColoredEdgesLatent{1}{4}{ANY}{ANY};
    \crfColoredFeatures{0/8.5/0.18, 1/9.5/0.4, 2/10.5/0.41, 3/11.5/1.}{4}{0.1}{ANY};

\end{tikzpicture}
}
    }
    \caption{Computational path of the probability of the correct
      label in latent-marginalized CRF}
  \end{subfigure}
  ~
  \begin{subfigure}[t]{0.4\textwidth}
    \centering
    \resizebox{\hcrfwidth}{\hcrfheight}{
      {\scriptsize
  \begin{tikzpicture}
    \tikzstyle{xnode}=[rectangle,draw,fill=gray76,align=center,minimum size=1.5em,text width=1.5em] %
    \tikzstyle{lnode}=[circle,draw,fill=white,inner sep=1pt,minimum size=2.5em] %
    \tikzstyle{ynode}=[circle,draw,fill=gray76,inner sep=1pt,minimum size=2.5em] %
    \tikzstyle{highlight}=[draw=blue,line width=0.5mm] %
    \tikzstyle{label}=[] %
    \tikzstyle{extext}=[] %
    \tikzstyle{factor}=[rectangle,fill=black,midway,inner sep=0pt,minimum size=0.4em] %

    \node[ynode,highlight] (0Neg) at (5, 6) {Neg};
    \node[ynode] (0Neut) at (6, 6) {Neut};
    \node[ynode] (0Pos) at (7, 6) {Pos};
    \hyperNode{0Neg}{0Neut}{0Pos}{Root};
    \crfColoredFeatures{0/4.5/0.54, 1/5.5/0.39, 2/6.5/0.07, 3/7.5/1.}{0}{4.7}{Neg};

    \node[lnode,highlight] (1Neg) at (5, 3.7) {Neg};
    \node[lnode,highlight] (1Neut) at (6, 3.7) {Neut};
    \node[lnode,highlight] (1Pos) at (7, 3.7) {Pos};
    \hyperNode{1Neg}{1Neut}{1Pos}{EDU 3};
    \crfColoredEdgesLatent{0}{1}{0Neg}{ANY}; 
    \crfColoredFeatures{0/4.5/0.54, 1/5.5/0.22, 2/6.5/0.24, 3/7.5/1.}{1}{2.4}{ANY};

    \node[lnode,highlight] (2Neg) at (1, 1.4) {Neg};
    \node[lnode,highlight] (2Neut) at (2, 1.4) {Neut};
    \node[lnode,highlight] (2Pos) at (3, 1.4) {Pos};
    \hyperNode{2Neg}{2Neut}{2Pos}{EDU 1};
    \crfColoredEdgesLatent{1}{2}{ANY}{ANY};
    \crfColoredFeatures{0/0.5/0.15, 1/1.5/0.45, 2/2.5/0.4, 3/3.5/1.}{2}{0.1}{ANY};

    \node[lnode,highlight] (3Neg) at (5, 1.4) {Neg};
    \node[lnode,highlight] (3Neut) at (6, 1.4) {Neut};
    \node[lnode,highlight] (3Pos) at (7, 1.4) {Pos};
    \hyperNode{3Neg}{3Neut}{3Pos}{EDU 2};
    \crfColoredEdgesLatent{1}{3}{ANY}{ANY};
    \crfColoredFeatures{0/4.5/0.14, 1/5.5/0.42, 2/6.5/0.43, 3/7.5/1.}{3}{0.1}{ANY};

    \node[lnode,highlight] (4Neg) at (9, 1.4) {Neg};
    \node[lnode,highlight] (4Neut) at (10, 1.4) {Neut};
    \node[lnode,highlight] (4Pos) at (11, 1.4) {Pos};
    \hyperNode{4Neg}{4Neut}{4Pos}{EDU 4};
    \crfColoredEdgesLatent{1}{4}{ANY}{ANY};
    \crfColoredFeatures{0/8.5/0.18, 1/9.5/0.4, 2/10.5/0.41, 3/11.5/1.}{4}{0.1}{ANY};

\end{tikzpicture}
}
    }
    \caption{Computational path of the probability of a wrong label in
      latent-marginalized CRF}
  \end{subfigure}
  \caption[Computational paths in LCRF and
    LMCRF]{Confronted computational paths in latent and
    latent-marginalized conditional random
    fields}\label{dasa:fig:lcrf-vs-lmcrf}
\end{figure*}
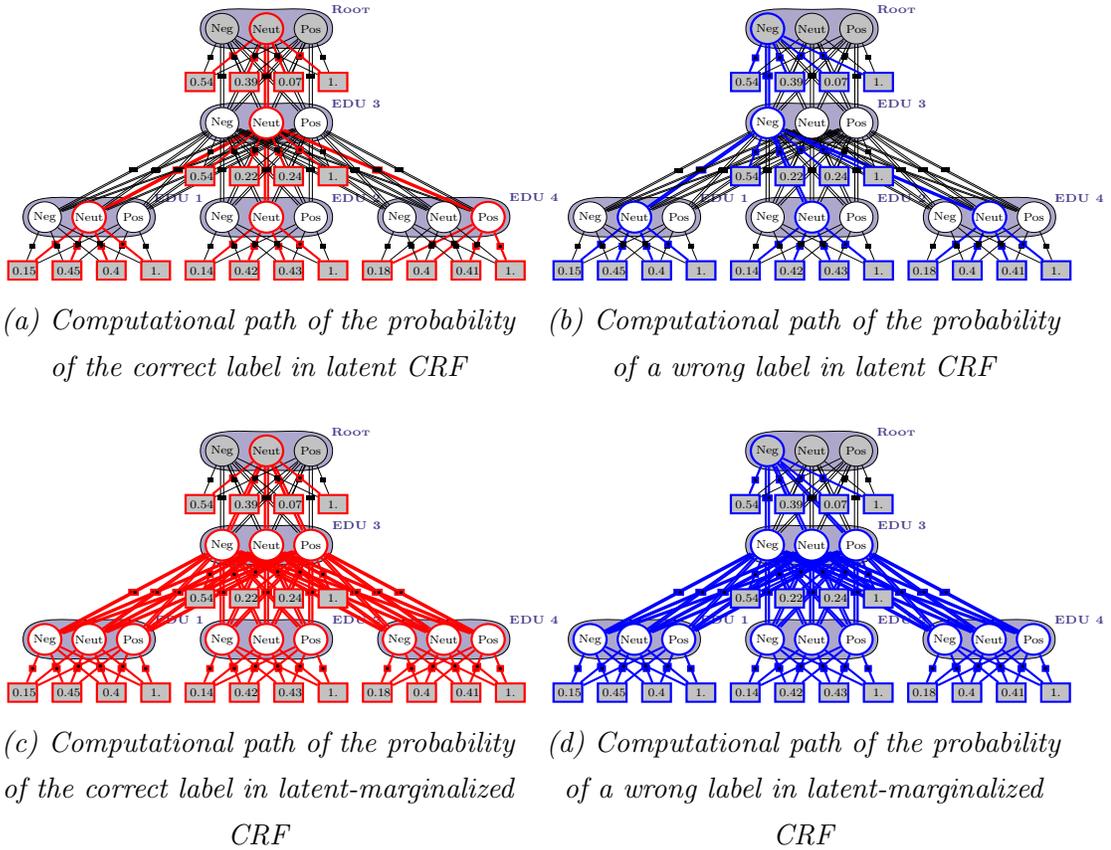

It is probably easier to realize the difference between the two CRF
methods (latent and latent-marginalized CRFs) more vividly by looking
at Figure~\ref{dasa:fig:lcrf-vs-lmcrf}, in which we highlighted the
paths that are used to compute the probabilities of correct and wrong
labels in both systems.  As we can see from this picture, LCRF only
considers one label sequence that leads to the maximum probability of
the correct tag (\textsc{Neut}) at the single observed \textsc{Root}
node and then compares this sequence with the path that maximizes the
probability of an incorrect tag (in this case \textsc{NEG}) at the
same node.  In contrast to this, LMCRF considers all possible label
configurations of elementary discourse units and uses this total
cumulative mass to estimate the probability of both (correct and
wrong) observed tags.

\subsection{Recursive Dirichlet Process}

Finally, the last method that we present in this chapter,
\emph{Recursive Dirichlet Process} or \emph{RDP}, goes a further step
in the probabilistic direction by assuming that not only the
probabilities of discourse units but also the parameters via which
these probabilities are computed represent random variables.

In particular, we associate a variable
$\mathbf{z}_j\in\mathbb{R}_+^3$, s.t. $\norm{\mathbf{z}}_1 = 1$, with
every RST node $j$ (which in this case can be either an elementary
discourse segment or an abstract span).\footnote{In contrast to the
  previous CRF approaches, this time, we depart from the dependency
  tree representation and adopt the discourse tree structure proposed
  by~\citet{Bhatia:15} for their R2N2 method.  In this structure, we
  keep all abstract nodes from the original RST tree, but relink all
  satellites to the abstract parents of their nuclei.}  This variable
specifies the multivariate probability of the three polarities
(\textsc{Negative}, \textsc{Neutral}, and \textsc{Positive}) for the
$j$-th node.  Since every element of $\mathbf{z}_j$ has to be
non-negative and their total sum must add up to one, it is natural to
assume that the value of this variable is drawn from a Dirichlet
distribution:
\begin{align*}
  \mathbf{z}_j \sim Dir(\boldsymbol{\alpha}).
\end{align*}
The only parameter accepted by this distribution, which simultaneously
controls both the mean and the variance of its outcomes, is vector
$\boldsymbol{\alpha}$.  Consequently, our primary goal in this method
is to find a way how to compute this parameter automatically for each
node.

An obvious starting point for this computation is the polarity scores
predicted by the base classifier for every elementary discourse unit,
which we will henceforth denote as
$\boldsymbol{z}_{j_0}\in\mathbb{R}^3_+$. Since these scores, however,
are only available for elementary segments, we initialize the
corresponding variables of the abstract spans to zeroes with the only
exception being the root node, to which we again assign the scores
returned by the LBA classifier for the whole message.

To compute the posterior distribution of the root
($\mathbf{z}_{\text{\textsc{Root}}}$), we sort all nodes of the RST
tree in reverse topological order and estimate the polarities of the
spans from the bottom up by joining the $\mathbf{z}$-scores of their
children.  But before we do this joining, we multiply the
$\mathbf{z}$-vector of each child $k$ with a special matrix $M_r$,
where $r \in \{\{\text{\textsc{Nucleus},
  \textsc{Satellite}}\}\times\{\text{\textsc{Con\-tra\-sti\-ve},
  \textsc{Non-\-Con\-tra\-sti\-ve}}\}\}$ is the discourse relation
holding between that child and its parent, and project the result of
this multiplication back to the probability simplex using the
sparsemax operation~\cite{Martins:16}:
\begin{align}
  \mathbf{z}^*_k&=
  sparsemax\left(M_r\mathbf{z}_k^{\top}\right).\label{dasa:eq:z-asterisk}
\end{align}

The main goal of matrix $M_r$ is to reflect contextual polarity
changes that might be conveyed by discourse relations: for example, a
contrastive link might stronger affect the polarity of the parent than
a non-contrastive one (compare, for instance, the contrastive
\emph{Many people support Trump, but he behaves like an alpha male}
with the non-contrastive \emph{Many people support Trump, because he
  behaves like an alpha male}).  Because this parameter also
represents a random variable, we sample it from a multivariate normal
distribution:
\begin{align*}
  M_r \sim \mathcal{N}_{3\times3}(\boldsymbol{\mu}_r, \mathbf{\Sigma}_r),
\end{align*}
setting the mean of this distribution to:\footnote{Before we do the
  actual sampling, we unroll this parameter to a vector and then
  reshape the sampled value back to a $3\times3$ matrix.}
\begin{equation*}
  \boldsymbol{\mu}_r=\begin{bmatrix}
  1 & 0 & 0\\
  0 & 0.3 & 0\\
  0 & 0 & 1
  \end{bmatrix},
\end{equation*}
and initializing its covariance matrix to all ones:
\begin{equation*}
  \boldsymbol{\Sigma}_r=\begin{bmatrix}
  1 & 1 & 1\\
  1 & 1 & 1\\
  1 & 1 & 1
  \end{bmatrix}.
\end{equation*}
With this choice of parameters, we hope to dampen the effect of
neutral EDUs\footnote{As you can see from
  Equation~\ref{dasa:eq:z-asterisk}, the middle row of the $M_r$
  matrix is responsible for propagating the neutral score of the
  $j$-the node, and by setting this row to $[0, 0.3, 0]$ we
  effectively reduce the neutral polarity by two thirds.} in order to
prevent situations where multiple objective segments vanquish the
meaning of a single polar discourse unit.

Afterwards, when seeing the $k$-th child of the $j$-th node in the RST
tree, we compute the $\boldsymbol{\alpha}$ parameter of this node as
follows:
\begin{align}
  \boldsymbol{\alpha}_{j_k}&= \boldsymbol{\beta}\odot\mathbf{z}^*_k +
  (\mathbf{1} -
  \boldsymbol{\beta})\odot\mathbf{z}_{j_{k-1}},\label{dasa:eq:alpha}
\end{align}
where $\boldsymbol{\beta}\in\mathbb{R}^3$ is another multivariate
random variable sampled from the Beta distribution $B(5., 5.)$, which
controls the amount of information we want to pass from child to its
parent; $\mathbf{z}_{j_{k-1}}$ is the value of the $\mathbf{z}$-vector
for the $j$-th node after seeing its previous ($k-1$-th) child; and
$\odot$ means elementwise multiplication.

The only thing that we now need to do to the above
$\boldsymbol{\alpha}_{j_k}$ term before drawing the actual probability
$\boldsymbol{z}_{j_k}$ is to scale this vector by a certain amount in
order to reduce the variance of the resulting Dirichlet
distribution.\footnote{Because if we keep the
  $\boldsymbol{\alpha}_{j_k}$ vector from Equation~\ref{dasa:eq:alpha}
  unchanged, most of its values will be in the range $[0,\ldots,1]$
  which will lead to an extremely high variance of the Dirichlet
  distribution.} In particular, we compute this scaling factor as
follows:
\begin{align*}
  scale&= \frac{\xi \times \left(0.1 + \cos\left(\mathbf{z}^*_k,
    \mathbf{z}_{j_{k-1}}\right)\right)}{H\left(\boldsymbol{\alpha}_{j_k}\right)};
\end{align*}
where $\xi$ is a model parameter sampled from a $\chi^2$-distribution:
$\xi\sim\chi^2(34)$; 0.1 is a constant used to prevent zero scales in
the cases when $\cos\left(\mathbf{z}^*_k, \mathbf{z}_{j_{k-1}}\right)$
is zero; and $H\left(\boldsymbol{\alpha}_{j_k}\right)$ stands for the
entropy of the $\boldsymbol{\alpha}_{j_k}$ vector.  Although this
expression looks somewhat complicated, the intuition behind it is very
simple: The $\xi$ term encodes our prior belief in the correctness of
model's prediction (the higher its value, the more we trust the
model); the cosine measures the similarity between the probabilities
of parent and child (the more similar these probabilities, the greater
will be the scale); and, finally, the entropy in the denominator tells
us how uniform the vector $\boldsymbol{\alpha}_{j_k}$ is (the more
equal its scores, the less confident we will be in the final outcome).

\begin{figure}[htb!]
  {\centering
    \includegraphics[width=\linewidth]{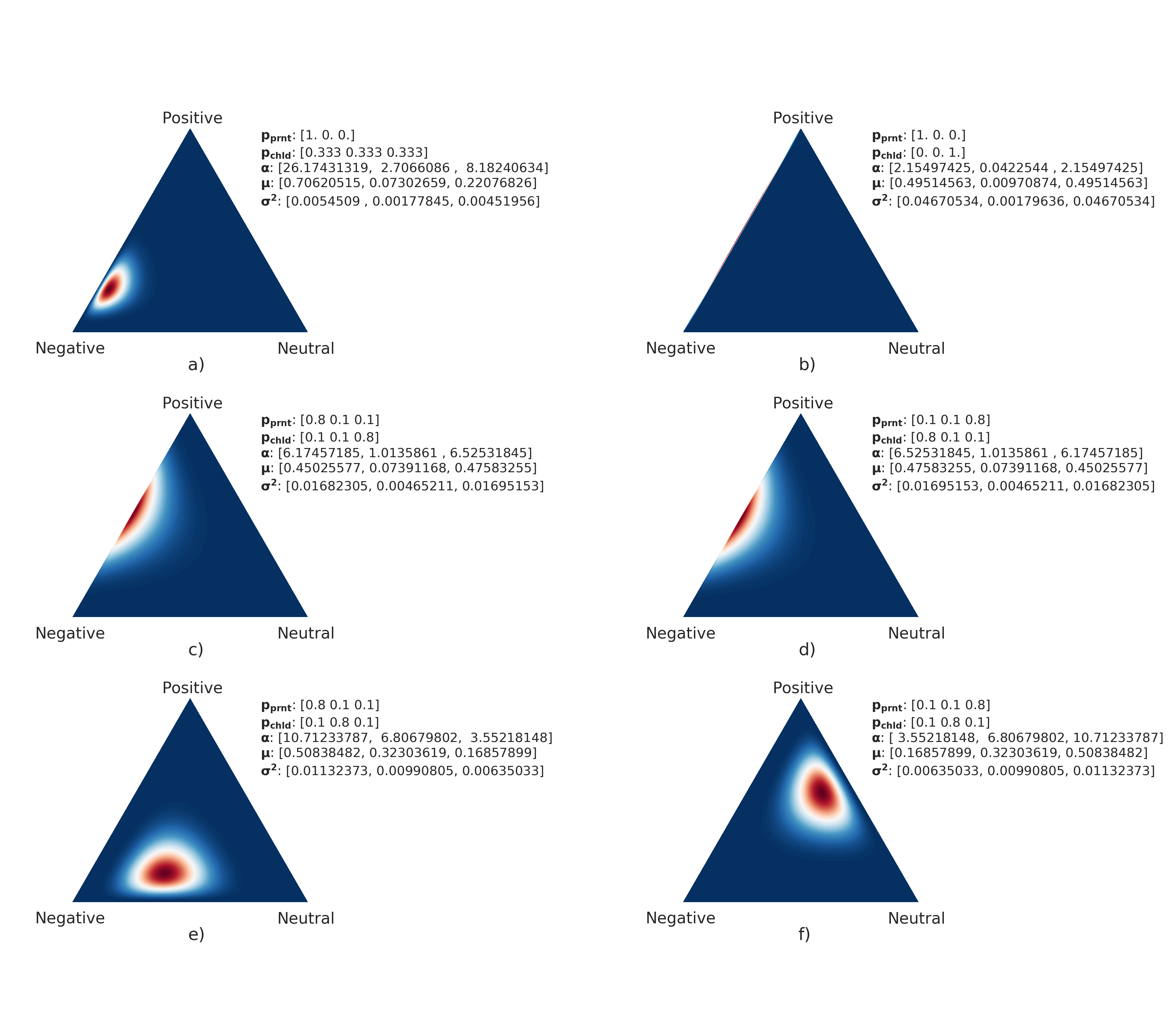} }
  \caption[Probability distributions computed by RDP]{Probability
    distributions of polar classes computed by the Recursive Dirichlet
    Process\\ {(higher probability regions are highlighted in red;
      \small $\mathbf{p}_{prnt}$ means the probability of the parent
      node [the values in the vector represent the scores for the
        negative, neutral, and positive polarities respectively];
      $\mathbf{p}_{chld}$ denotes the probability of the child; and
      $\boldsymbol{\alpha}$, $\boldsymbol{\mu}$, and
      $\boldsymbol{\sigma}^2$ represent the parameters of the
      resulting joint distribution shown in the
      simplices)}}\label{dasa:fig:rdp-alpha}
\end{figure}

With the $scale$ and $\boldsymbol{\alpha}_{j_k}$ terms at hand, we are
all set to compute the updated probability of polar classes for the
$j$-th node after considering its $k$-th child:
\begin{align*}
  \mathbf{z}_{j_{k}}\sim Dir(scale \times \boldsymbol{\alpha}_{j_k}).
\end{align*}

You can see some examples of this computation in
Figure~\ref{dasa:fig:rdp-alpha}, where we plotted different
configurations of parent and child probabilities
($\mathbf{z}_{j_{k-1}}$ and $\mathbf{z}_{k}$, shown to the right of
each picture) and the resulting Dirichlet distributions (represented
as simplices).  For instance, in the top-left figure, we show a
situation where the parent has a very strong probability of the
negative class ($[1, 0, 0]$), but the probability of the child is
absolutely uniform ($[0.33, 0.33, 0.33]$); in this case, the model
keeps to the negative polarity, heaping almost all probability mass in
this corner.  At the same, to account for the uncertainty about the
child, RDP slightly moves the crest of the probability hill (\ie{} its
mean) towards the positive class and makes the slopes of this hill
lower along all three axes (\ie{} increases its variance). On the
other hand, if parent and child have completely opposite semantic
orientations (say \textsc{Positive} and \textsc{Negative}), which the
base classifier is perfectly sure about, as shown in Subfigure~b, RDP
uniformly distributes the whole probability just along the
\textsc{Positive}--\textsc{Negative} edge.  Another situation is
depicted in the middle row, where parent and child again have opposite
polarities, but the base predictor is less sure about its decisions
and also admits a small chance that either of these nodes is neutral.
In this case, RDP still spreads most probability along the main polar
edge, but places the mean of this distribution right in-between the
two polar corners and also screeds some part of that mass towards the
center of the simplex.
Finally, in the last row, we can see our intended discrimination of
the neutral orientation: This time, the parent node is strictly polar
(negative on the left and positive on the right), whereas its child is
neutral.  In contrast to the previous two examples, the mean of the
resulting distribution is located closer to the polar corner and not
in-between the two juxtaposed classes as before.
\begin{figure}[htb]
  \begin{center}
    \includegraphics[height=15em,width=0.4\linewidth]{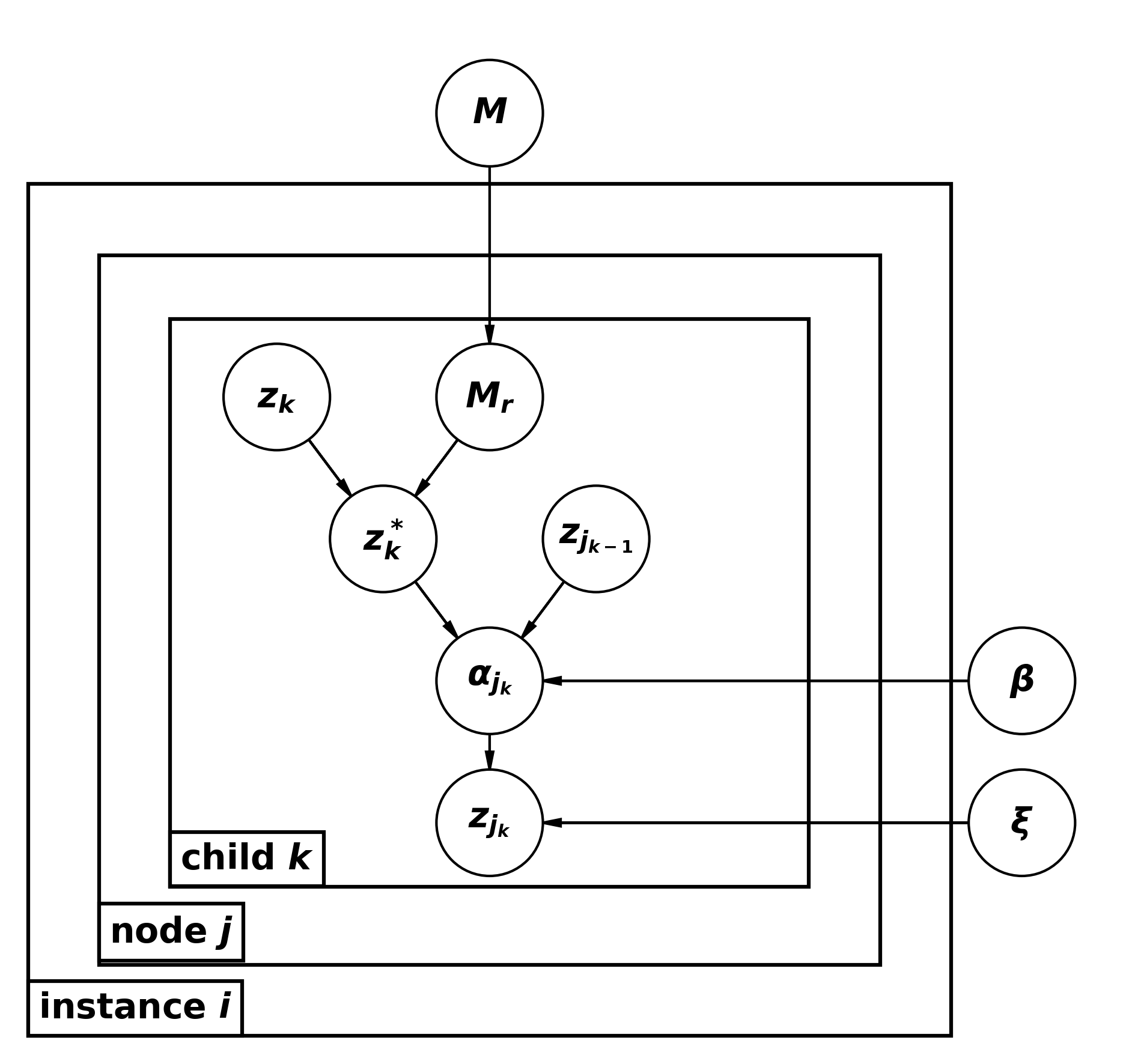}
    \caption[A plate diagram of the Recursive Dirichlet Process]{A plate
      diagram of the Recursive Dirichlet Process\\{\small (without the
        final categorical draw)}}\label{dasa:fig:rdp-plate}
  \end{center}
\end{figure}

Returning back to our model, after processing all $K$ children of the
$j$-th node, we regard the last outcome $\mathbf{z}_{j_{K}}$ as the
final polarity distribution of that node and use this value to
estimate the probabilities of the remaining ancestors in the RST tree.
Finally, after finishing processing all descendants of the root, we
use the resulting $\mathbf{z}_{\text{\textsc{Root}}_K}$ vector as a
parameter of a categorical distribution from which we draw the final
prediction label $y$:
\begin{align*}
  y \sim Cat(\mathbf{z}_{\text{\textsc{Root}}}).
\end{align*}

Using this manually defined model as a starting point, we can estimate
our prior belief in the joint probability of hidden and observed
variables $p(y, \mathbf{z})$.  As it turns out, knowing this belief is
enough to derive another probability $q(\mathbf{z})$, which best
approximates the distribution of only the latent nodes.  In
particular, we define the structure of $q(\mathbf{z})$ to be the same
as in $p(y, \mathbf{z})$, but deprive it of the last step (drawing of
the observed label) and optimize the parameters $\boldsymbol{\theta}$
of this model ($\boldsymbol{\mu}_r$, $\boldsymbol{\Sigma}_r$, and the
parameters of the Beta and $\chi^2$ distributions) by maximizing the
evidence lower bound between $p$ and $q$, using stochastic gradient
descent\cite[see][]{Ranganath:14}:
\begin{align*}
  \mathcal{L}\left(\boldsymbol{\theta}\right)
  &=\mathbb{E}_{q_{\boldsymbol{\theta}}(\mathbf{z})}%
  \left[\log\left(p(y, \mathbf{z})\right) -
    \log\left(q(\mathbf{z})\right)\right].
\end{align*}
We perform this optimization for 100 epochs, picking the parameters
that yield the best macro-averaged \F{}-score on the set-aside
development data.

The results of our proposed and baseline methods are shown in
Table~\ref{dasa:tbl:res}.
\begin{table}[hbt]
  \begin{center}
    \bgroup\setlength\tabcolsep{0.1\tabcolsep}\scriptsize
    \begin{tabular}{p{0.162\columnwidth} 
        *{9}{>{\centering\arraybackslash}p{0.074\columnwidth}} 
        *{2}{>{\centering\arraybackslash}p{0.068\columnwidth}}} 
      \toprule
      \multirow{2}*{\bfseries Method} & %
      \multicolumn{3}{c}{\bfseries Positive} & %
      \multicolumn{3}{c}{\bfseries Negative} & %
      \multicolumn{3}{c}{\bfseries Neutral} & %
      \multirow{2}{0.068\columnwidth}{\bfseries\centering Macro\newline \F{}} & %
      \multirow{2}{0.068\columnwidth}{\bfseries\centering Micro\newline \F{}}\\
      \cmidrule(lr){2-4}\cmidrule(lr){5-7}\cmidrule(lr){8-10}

      & Precision & Recall & \F{} & %
      Precision & Recall & \F{} & %
      Precision & Recall & \F{} & & \\\midrule

      \multicolumn{12}{c}{\cellcolor{cellcolor}PotTS}\\

      LCRF & 0.76 & 0.79 & \textbf{0.77} & %
      \textbf{0.61} & 0.53 & 0.56 & %
      0.7 & 0.71 & 0.71 & %
      0.67 & 0.709\\

      LMCRF & \textbf{0.77} & 0.77 & \textbf{0.77} & %
      \textbf{0.61} & 0.54 & 0.57 & %
      0.69 & \textbf{0.74} & \textbf{0.72} & %
      0.671 & \textbf{0.712}\\

      RDP & 0.73 & 0.82 & \textbf{0.77} & %
      \textbf{0.61} & 0.56 & \textbf{0.58} & %
      \textbf{0.73} & 0.65 & 0.69 & %
      \textbf{0.678} & 0.706\\

      DDR & 0.73 & 0.77 & 0.75 & %
      0.54 & \textbf{0.59} & 0.56 & %
      0.69 & 0.61 & 0.65 & %
      0.655 & 0.674\\

      R2N2 & 0.74 & 0.78 & 0.76 & %
      0.59 & 0.53 & 0.56 & %
      0.68 & 0.68 & 0.68 & %
      0.657 & 0.692\\

      WNG & 0.58 & 0.79 & 0.67 & %
      \textbf{0.61} & 0.21 & 0.31 & %
      0.61 & 0.57 & 0.59 & %
      0.487 & 0.59\\

      \textsc{Last} & 0.52 & \textbf{0.83} & 0.64 & %
      0.57 & 0.17 & 0.26 & %
      0.61 & 0.43 & 0.5 & %
      0.453 & 0.549\\

      \textsc{Root} & 0.56 & 0.73 & 0.64 & %
      0.58 & 0.22 & 0.32 & %
      0.55 & 0.54 & 0.54 & %
      0.481 & 0.56\\

      \textsc{No-Discourse} & 0.73 & 0.82 & \textbf{0.77} & %
      \textbf{0.61} & 0.56 & \textbf{0.58} & %
      0.72 & 0.66 & 0.69 & %
      0.677 & 0.706\\

      \multicolumn{12}{c}{\cellcolor{cellcolor}SB10k}\\

      LCRF & \textbf{0.64} & \textbf{0.69} & 0.66 & %
      0.45 & \textbf{0.45} & 0.45 & %
      \textbf{0.82} & 0.79 & \textbf{0.8} & %
      0.557 & 0.713\\

      LMCRF & \textbf{0.64} & \textbf{0.69} & \textbf{0.67} & %
      0.45 & \textbf{0.45} & 0.45 & %
      \textbf{0.82} & 0.79 & \textbf{0.8} & %
      \textbf{0.56} & \textbf{0.715}\\

      RDP & \textbf{0.64} & \textbf{0.69} & 0.66 & %
      0.45 & \textbf{0.45} & 0.45 & %
      0.82 & 0.79 & \textbf{0.8} & %
      0.557 & 0.713\\

      DDR & 0.59 & 0.63 & 0.61 & %
      \textbf{0.48} & 0.44 & \textbf{0.46} & %
      0.77 & 0.76 & 0.77 & %
      0.534 & 0.681\\

      R2N2 & \textbf{0.64} & \textbf{0.69} & 0.66 & %
      0.46 & \textbf{0.45} & 0.45 & %
      0.81 & 0.79 & \textbf{0.8} & %
      0.559 & 0.713\\

      WNG & 0.61 & 0.63 & 0.62 & %
      0.46 & 0.29 & 0.36 & %
      0.76 & \textbf{0.82} & 0.79 & %
      0.488 & 0.693\\

      \textsc{Last} & 0.56 & 0.55 & 0.56 & %
      0.46 & 0.29 & 0.36 & %
      0.73 & 0.8 & 0.76 & %
      0.459 & 0.661\\

      \textsc{Root} & 0.51 & 0.55 & 0.53 & %
      0.4 & 0.3 & 0.35 & %
      0.74 & 0.76 & 0.75 & %
      0.438 & 0.64\\

      \textsc{No-Discourse} & \textbf{0.64} & \textbf{0.69} & 0.66 & %
      0.45 & \textbf{0.45} & 0.45 & %
      \textbf{0.82} & 0.79 & \textbf{0.8} & %
      0.557 & 0.713\\\bottomrule
    \end{tabular}
    \egroup{}
    \caption[Evaluation of DASA methods]{Results of discourse-aware
      sentiment analysis methods\\ {\small LCRF~--~latent conditional
        random fields, LMCRF~--~latent-marginalized conditional random
        fields, RDP~--~recursive Dirichlet process,
        DDR~--~discourse-depth reweighting~\cite{Bhatia:15},
        R2N2~--~rhetorical recursive neural network~\cite{Bhatia:15},
        WNG~--~\citet{Wang:13}, \textsc{Last}~--~polarity determined
        by the last EDU, \textsc{Root}~--~polarity determined by the
        root EDU(s), \textsc{No-Discourse}~--~discourse-unaware
        classifier}}\label{dasa:tbl:res}
  \end{center}
\end{table}

As we can see from the table, our approaches perform fairly well in
comparison with other systems, outperforming them in terms of macro-
and macro-averaged $F_1$ on both datasets.  Especially the
latent-marginalized CRF shows fairly strong scores, yielding the best
$F_1$-results for the positive and neutral classes on the PotTS and
SB10k data, which in turn leads to the highest overall micro-averaged
$F_1$-measure on these corpora.  This solution is closely followed by
the Recursive Dirichlet Process, whose $F_1$ for the positive class on
the PotTS test set is identical to that attained by LMCRF and the
$F$-score for the negative class is even one percent higher, which
allows it to reach the best macro-average on this test set.

As it turns out, the strongest competitors to our systems are the
\textsc{No-Discourse} approach and the R2N2 method by
\citet{Bhatia:15}. The former solution outperforms the latter on the
PotTS corpus on both metrics (macro- and micro-$F_1$), but falls
against it with respect to the macro-\F{} on the SB10k set.  The DDR
and WNG methods get sixth and seventh places respectively, followed by
the simplest solutions, \textsc{Last} and \textsc{Root}.
Interestingly enough, the \textsc{Last} approach beats the
\textsc{Root} method on the SB10k data, but shows worse scores on the
PotTS corpus, which is mostly due to the lower recall of the negative
class.

\subsection{Error Analysis}

Although our methods performed quite competitive, we decided to still
look at their remaining errors in order to understand the reasons for
their potential weaknesses.

The first such error shown in Example~\ref{snt:dasa:exmp:lcrf-error}
was made by the latent CRF system, which erroneously considered a
negative tweet as neutral.  But as we can see from the picture of the
automatic RST tree in this example, we can hardly expect the right
decision in this case anyway, because neither EDUs nor the root node
of this message were correctly classified as negative by the LBA
classifier.  Nevertheless, even in this apparently hopeless situation,
messages propagated from leaves to the root during the max-product
inference still tell the latter node that the predicted class better
be negative. (We inspected the belief propagation messages passed in
the forward direction and found that the total score for the negative
class amounts to 0.597, whereas the belief in the positive class [its
  closest rival] only runs up to 0.462.)  Unfortunately, these
messages cannot outweigh the high score of the neutral class that
results from the node features (the state score for this polarity is
equal to 0.524, whereas the negative class only obtains a score of
-0.118).\footnote{All scores for this example are given in the
  logarithm domain.}
\begin{example}[An Error Made by the \textsc{LCRF} System\newline]\label{snt:dasa:exmp:lcrf-error}
  \noindent\textup{\bfseries\textcolor{darkred}{Tweet:}} {\upshape
    $[$Boah , also doch wieder ein Mann , oder ?$]_1$ $[$ODER ?$]_2$
    $[$papst$]_3$}\\
  \noindent $[$Boah, a man again, isn't it ?$]_1$ $[$ISN'T ?$]_2$
  $[$pope$]_3$\\[\exampleSep]
  \noindent\textup{\bfseries\textcolor{darkred}{Gold Label:}}\hspace*{4.3em}\textbf{%
    \upshape\textcolor{midnightblue}{negative}}\\
  \noindent\textup{\bfseries\textcolor{darkred}{Predicted Label:}}\hspace*{2em}\textbf{%
    \upshape\textcolor{black}{neutral*}}%
    {
      \begin{center}
        {\scriptsize
  \begin{tikzpicture}
    \tikzstyle{xnode}=[rectangle,draw,fill=gray76,align=center,minimum size=2em,text width=2.2em] %
    \tikzstyle{lnode}=[circle,draw,fill=white,inner sep=1pt,minimum size=2.5em] %
    \tikzstyle{ynode}=[circle,draw,fill=gray76,inner sep=1pt,minimum size=2.5em] %
    \tikzstyle{label}=[] %
    \tikzstyle{extext}=[] %
    \tikzstyle{factor}=[rectangle,fill=black,midway,inner sep=0pt,minimum size=0.4em] %

    \node[ynode] (0Neg) at (5, 6) {Neg};
    \node[ynode] (0Neut) at (6, 6) {Neut};
    \node[ynode] (0Pos) at (7, 6) {Pos};
    \hyperNode{0Neg}{0Neut}{0Pos}{Root};
    \crfFeatures{0/4.5/0.352, 1/5.5/0.474, 2/6.5/0.174, 3/7.5/1.}{0}{4.7};

    \node[lnode] (3Neg) at (5, 3.7) {Neg};
    \node[lnode] (3Neut) at (6, 3.7) {Neut};
    \node[lnode] (3Pos) at (7, 3.7) {Pos};
    \hyperNode{3Neg}{3Neut}{3Pos}{EDU 3};
    \crfEdgesLatent{0}{3};
    \crfFeatures{0/4.5/0.158, 1/5.5/0.39, 2/6.5/0.452, 3/7.5/1.}{3}{2.4};

    \node[lnode] (1Neg) at (1., 3.7) {Neg};
    \node[lnode] (1Neut) at (2., 3.7) {Neut};
    \node[lnode] (1Pos) at (3., 3.7) {Pos};
    \hyperNode{1Neg}{1Neut}{1Pos}{EDU 1};
    \crfEdgesLatent{0}{1};             
    \crfFeatures{0/0.5/0.398, 1/1.5/0.421, 2/2.5/0.181, 3/3.5/1.}{1}{2.4};

    \node[lnode] (2Neg) at (9, 3.7) {Neg};
    \node[lnode] (2Neut) at (10, 3.7) {Neut};
    \node[lnode] (2Pos) at (11, 3.7) {Pos};
    \hyperNode{2Neg}{2Neut}{2Pos}{EDU 2};
    \crfEdgesLatent{0}{2};
    \crfFeatures{0/8.5/0.155, 1/9.5/0.338, 2/10.5/0.507, 3/11.5/1.}{2}{2.4};
\end{tikzpicture}
}

      \end{center}
    }
\end{example}

As it turns out, high neutral node scores of the root are also the
main reason for the misclassification in
Example~\ref{snt:dasa:exmp:lmcrf-error}, where the LMCRF system also
confuses the negative polarity with the neutral class.  This time,
however, messages coming from the leaves suggest almost equal
probabilities for both semantic orientations, so that feature scores
of the root completely call the shots in the final decision.
\begin{example}[An Error Made by the \textsc{LMCRF} System\newline]\label{snt:dasa:exmp:lmcrf-error}
  \noindent\textup{\bfseries\textcolor{darkred}{Tweet:}} {\upshape '
    $[$Wissen ?$]_1$ $[$Igitt geh weg damit !$]_2$}\\
  \noindent $[$Knowledge ?$]_1$ $[$Yuck , go away with it$]_2$\\[\exampleSep]
  \noindent\textup{\bfseries\textcolor{darkred}{Gold Label:}}\hspace*{4.3em}\textbf{%
    \upshape\textcolor{midnightblue}{negative}}\\
  \noindent\textup{\bfseries\textcolor{darkred}{Predicted Label:}}\hspace*{2em}\textbf{%
    \upshape\textcolor{black}{neutral*}}%
    {
      \begin{center}
        {\scriptsize
  \begin{tikzpicture}
    \tikzstyle{xnode}=[rectangle,draw,fill=gray76,align=center,minimum size=2em,text width=2.2em] %
    \tikzstyle{lnode}=[circle,draw,fill=white,inner sep=1pt,minimum size=2.5em] %
    \tikzstyle{ynode}=[circle,draw,fill=gray76,inner sep=1pt,minimum size=2.5em] %
    \tikzstyle{label}=[] %
    \tikzstyle{extext}=[] %
    \tikzstyle{factor}=[rectangle,fill=black,midway,inner sep=0pt,minimum size=0.4em] %

    \node[ynode] (0Neg) at (5, 3.7) {Neg};
    \node[ynode] (0Neut) at (6, 3.7) {Neut};
    \node[ynode] (0Pos) at (7, 3.7) {Pos};
    \hyperNode{0Neg}{0Neut}{0Pos}{Root};
    \crfFeatures{0/4.5/0.001, 1/5.5/0.997, 2/6.5/0.001, 3/7.5/1.}{0}{2.4};

    \node[lnode] (1Neg) at (2.5, 1.4) {Neg};
    \node[lnode] (1Neut) at (3.5, 1.4) {Neut};
    \node[lnode] (1Pos) at (4.5, 1.4) {Pos};
    \hyperNode{1Neg}{1Neut}{1Pos}{EDU 1};
    \crfEdgesLatent{0}{1};             
    \crfFeatures{0/2/0.001, 1/3/0.997, 2/4/0.001, 3/5/1.}{1}{0.1};

    \node[lnode] (2Neg) at (7.5, 1.4) {Neg};
    \node[lnode] (2Neut) at (8.5, 1.4) {Neut};
    \node[lnode] (2Pos) at (9.5, 1.4) {Pos};
    \hyperNode{2Neg}{2Neut}{2Pos}{EDU 2};
    \crfEdgesLatent{0}{2};
    \crfFeatures{0/7/0.999, 1/8/0.001, 2/9/0.0, 3/10/1.}{2}{0.1};
\end{tikzpicture}
}

      \end{center}
  }
\end{example}

Unfortunately, the recursive Dirichlet process cannot withstand the
erroneous predictions of the base classifier either.  For instance, in
Example~\ref{snt:dasa:exmp:rdp-error}, LBA assigns the highest scores
to the positive class in three out of four EDUs, even though each of
these units by itself expresses a negative attitude of the author.
Alas, the only case where the base classifier correctly predicts the
negative label (``Das is noch lange nicht ausdiskutiert !''
[\emph{It's no way been talked out !}]) drowns at the very beginning
of the score propagation.  (As it turned out, the learned
$\boldsymbol{\beta}$ parameter, which controls the amount of
information passed from child to its parent in
Equation~\ref{dasa:eq:alpha}, is extremely low for the negative class,
amounting to only 0.097, whereas for the positive and negative
polarities it runs up to 0.212 and 0.279.  Due to this low value, only
one tenth of the negative score from the third EDU arrives at the
parent when the model computes the polarity scores of the abstract
span 2.)
\begin{example}[An Error Made by the \textsc{RDP} System\newline]\label{snt:dasa:exmp:rdp-error}
  \noindent\textup{\bfseries\textcolor{darkred}{Tweet:}} {\upshape
    $[$Prima , was sind das f\"ur Idioten im DFB ?$]_1$ $[$Das is eine
    Muppetsshow auf LSD !$]_2$ $[$Das is noch lange nicht ausdiskutiert !$]_3$
    $[$Kiessling ist ein Depp !$]_4$}\\
  \noindent $[$Great, who are these idiots in the DFB ? $]_1$ $[$It is
    a muppet show on LSD$]_2$ $[$It's no way been talked out !$]_3$
  $[$Kiessling is a goof !$]_4$\\[\exampleSep]
  \noindent\textup{\bfseries\textcolor{darkred}{Gold Label:}}\hspace*{4.3em}\textbf{%
    \upshape\textcolor{midnightblue}{negative}}\\
  \noindent\textup{\bfseries\textcolor{darkred}{Predicted Label:}}\hspace*{2em}\textbf{%
    \upshape\textcolor{green3}{positive*}}%
    {
      \begin{center}
        {\scriptsize
  \begin{tikzpicture}
    \tikzstyle{xnode}=[rectangle,draw,fill=gray76,align=center,minimum size=2em,text width=2.2em] %
    \tikzstyle{lnode}=[circle,draw,fill=white,inner sep=1pt,minimum size=2.5em] %
    \tikzstyle{ynode}=[circle,draw,fill=gray76,inner sep=1pt,minimum size=2.5em] %
    \tikzstyle{label}=[] %
    \tikzstyle{extext}=[] %
    \tikzstyle{factor}=[rectangle,fill=black,midway,inner sep=0pt,minimum size=0.4em] %

    \node[ynode] (0Neg) at (3.25, 6) {Neg};
    \node[ynode] (0Neut) at (4.25, 6) {Neut};
    \node[ynode] (0Pos) at (5.25, 6) {Pos};
    \hyperNode{0Neg}{0Neut}{0Pos}{Root};
    \crfFeatures{0/3.25/0.003, 1/4.25/0.0, 2/5.25/0.997}{0}{5};

    \node[lnode] (1Neg) at (1, 4) {Neg};
    \node[lnode] (1Neut) at (2, 4) {Neut};
    \node[lnode] (1Pos) at (3, 4) {Pos};
    \hyperNode{1Neg}{1Neut}{1Pos}{EDU 1};
    \crfEdgesLatent{0}{1};             
    \crfFeatures{0/1/0.0, 1/2/0.001, 2/3/0.999}{1}{3};

    \node[lnode] (5Neg) at (6.25, 4) {Neg};
    \node[lnode] (5Neut) at (7.25, 4) {Neut};
    \node[lnode] (5Pos) at (8.25, 4) {Pos};
    \hyperNode{5Neg}{5Neut}{5Pos}{Span 1};
    \crfEdgesLatent{0}{5};

    \node[lnode] (2Neg) at (4, 2.7) {Neg};
    \node[lnode] (2Neut) at (5, 2.7) {Neut};
    \node[lnode] (2Pos) at (6, 2.7) {Pos};
    \hyperNode{2Neg}{2Neut}{2Pos}{EDU 2};
    \crfEdgesLatent{5}{2};
    \crfFeatures{0/4/0.001, 1/5/0.0, 2/6/0.999}{2}{1.7};

    \node[lnode] (6Neg) at (8.5, 2.7) {Neg};
    \node[lnode] (6Neut) at (9.5, 2.7) {Neut};
    \node[lnode] (6Pos) at (10.5, 2.7) {Pos};
    \hyperNode{6Neg}{6Neut}{6Pos}{Span 2};
    \crfEdgesLatent{5}{6};

    \node[lnode] (3Neg) at (7, 1.4) {Neg};
    \node[lnode] (3Neut) at (8, 1.4) {Neut};
    \node[lnode] (3Pos) at (9, 1.4) {Pos};
    \hyperNode{3Neg}{3Neut}{3Pos}{EDU 3};
    \crfEdgesLatent{6}{3};
    \crfFeatures{0/7/1., 1/8/0., 2/9/0.}{3}{0.4};

    \node[lnode] (4Neg) at (10, 1.4) {Neg};
    \node[lnode] (4Neut) at (11, 1.4) {Neut};
    \node[lnode] (4Pos) at (12, 1.4) {Pos};
    \hyperNode{4Neg}{4Neut}{4Pos}{EDU 4};
    \crfEdgesLatent{6}{4};
    \crfFeatures{0/10/0., 1/11/0., 2/12/1.}{4}{0.4};
\end{tikzpicture}
}

      \end{center}
    }
\end{example}

Another interesting error shown in
Example~\ref{snt:dasa:exmp:root-error} was made by the baseline
\textsc{Root} system, which similarly to CRF-based approaches confused
the negative class with the neutral polarity.  This time, however, the
misclassification is due to the discourse structure itself rather than
wrong predictions of the underlying sentiment method.  Because LBA
correctly recognizes that the negative smiley at the end of tweet has
a strictly negative semantic orientation, but the discourse-aware
baseline does not see this EDU at all, as it only considers the
segment at the top of the tree, which merely expresses a factual
hypothesis, free of any polar connotation.
\begin{example}[An Error Made by the \textsc{Root} System\newline]\label{snt:dasa:exmp:root-error}
  \noindent\textup{\bfseries\textcolor{darkred}{Tweet:}} {\upshape
    $[$Die NSA weiss auch von dir \ldots$]_1$ $[$N\"utzt uns auch nichts .$]_2$ $[$ \%NegSmiley$]_3$}\\
  \noindent $[$The NSA also knows about you \ldots$]_1$ $[$It doesn't help us
    either$]_2$ $[$ \%NegSmiley$]_3$\\[\exampleSep]
  \noindent\textup{\bfseries\textcolor{darkred}{Gold Label:}}\hspace*{4.3em}\textbf{%
    \upshape\textcolor{midnightblue}{negative}}\\
  \noindent\textup{\bfseries\textcolor{darkred}{Predicted Label:}}\hspace*{2em}\textbf{%
    \upshape\textcolor{black}{neutral*}}%
    {
      \begin{center}
        {
\scriptsize
\begin{tikzpicture}
\tikzstyle{xnode}=[rectangle,draw,fill=gray76,align=center,minimum size=2em,text width=2.2em] %
\tikzstyle{lnode}=[circle,draw,fill=white,inner sep=1pt,minimum size=2.5em] %
\tikzstyle{ynode}=[circle,draw,fill=gray76,inner sep=1pt,minimum size=2.5em] %
\tikzstyle{label}=[] %
\tikzstyle{extext}=[] %
\tikzstyle{factor}=[rectangle,fill=black,midway,inner sep=0pt,minimum size=0.4em] %

\node[lnode] (1Neg) at (5 , 3.7) {Neg};
\node[lnode] (1Neut) at (6, 3.7) {Neut};
\node[lnode] (1Pos) at (7, 3.7) {Pos};
\hyperNode{1Neg}{1Neut}{1Pos}{EDU 1};
\crfFeatures{0/5/0.007, 1/6/0.985, 2/7/0.007}{1}{2.4};

\node[lnode] (2Neg) at (2.5, 1.4) {Neg};
\node[lnode] (2Neut) at (3.5, 1.4) {Neut};
\node[lnode] (2Pos) at (4.5, 1.4) {Pos};
\hyperNode{2Neg}{2Neut}{2Pos}{EDU 2};
\crfEdgesLatent{1}{2};
\crfFeatures{0/2.5/0.001, 1/3.5/0.998, 2/4.5/0.001}{2}{0.1};

\node[lnode] (3Neg) at (7.5, 1.4) {Neg};
\node[lnode] (3Neut) at (8.5, 1.4) {Neut};
\node[lnode] (3Pos) at (9.5, 1.4) {Pos};
\hyperNode{3Neg}{3Neut}{3Pos}{EDU 3};
\crfEdgesLatent{1}{3};
\crfFeatures{0/7.5/1., 1/8.5/0., 2/9.5/0.}{3}{0.1};
\end{tikzpicture}
}

      \end{center}
    }
\end{example}

Finally, the last example (\ref{snt:dasa:exmp:last-error}) shows an
error made by the \textsc{Last} baseline system, which predicts the
neutral label for a negative tweet based on the polarity of its
right-most EDU\@.  This unit indeed admits some positive moments with
regard to the sad news expressed in the first segment, but in contrast
to the movie description from Example~\ref{disc-snt:exmp-pang02},
where the last sentence completely overturned the polarity of the
whole text, this time, the final opinion does not alter the general
negative mood of the message, but only dampens its effect.
\begin{example}[An Error Made by the \textsc{Last} System]\label{snt:dasa:exmp:last-error}
  \noindent\textup{\bfseries\textcolor{darkred}{Tweet:}} {\upshape
    $[$' ( :'( :'( Die letzte Aussprache war wohl das schwerste
      Telefonat meines gesamten Lebens :'( :'( :'($]_1$ $[$Aber wir
      gehen friedlich und als F $\ldots]_2$}\\
  \noindent $[$' ( :'( :'( The last talk was probably the most
    difficult call in my entire life :'( :'( :'($]_1$ $[$But we go
    apart peacefully and as f $\ldots]_1$\\[\exampleSep]
  \noindent\textup{\bfseries\textcolor{darkred}{Gold Label:}}\hspace*{4.3em}\textbf{%
    \upshape\textcolor{midnightblue}{negative}}\\
  \noindent\textup{\bfseries\textcolor{darkred}{Predicted Label:}}\hspace*{2em}\textbf{%
    \upshape\textcolor{black}{neutral*}}
\end{example}

\section{Evaluation}

As we could see from the examples in Section~\ref{sec:dasa:methods},
the results of our proposed methods were significantly limited by two
key factors:
\begin{inparaenum}[(i)]
\item scores predicted by the base sentiment system for tweets and
  EDUs and
\item the structure of RST trees constructed for these messages.
\end{inparaenum}
In order to estimate the effect of these factors more precisely, we
decided to rerun our experiments, trying alternative solutions for
each of these aspects.

\begin{figure*}[hbt!]
{ \centering
\begin{subfigure}{\textwidth}
  \centering
  \includegraphics[width=\linewidth]{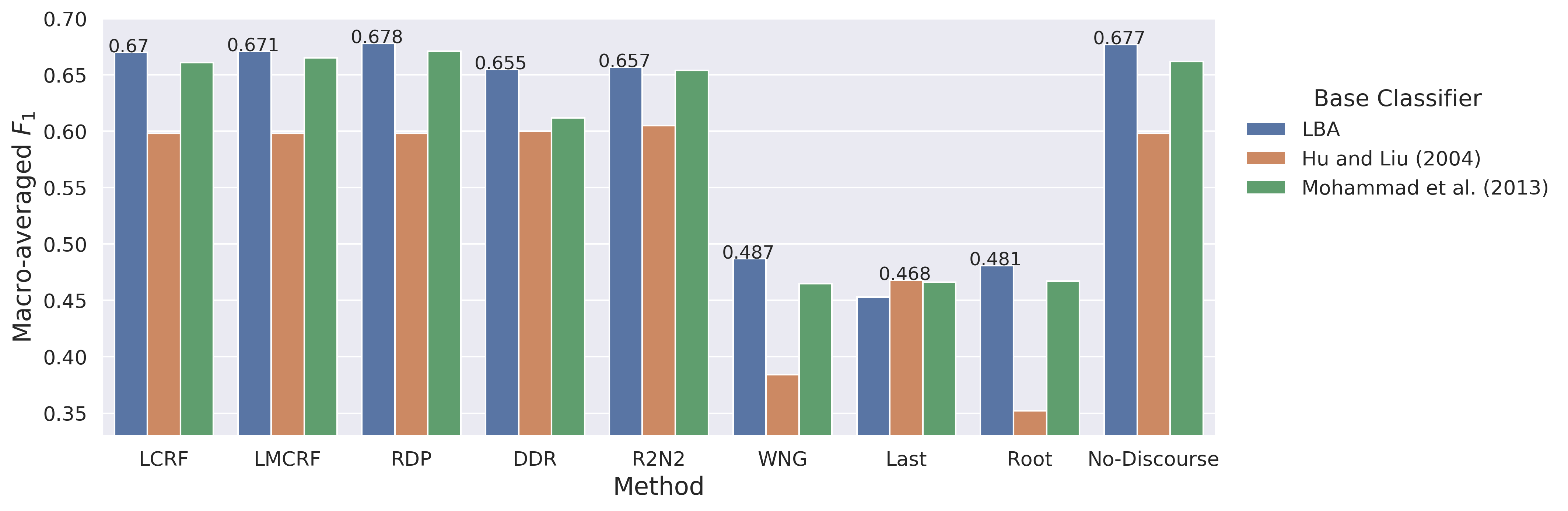}
  \caption{\texttt{Macro-\F{}}}\label{dasa:fig:potts-base-classifier-macro-F1}
\end{subfigure}\\
\begin{subfigure}{\textwidth}
  \centering
  \includegraphics[width=\linewidth]{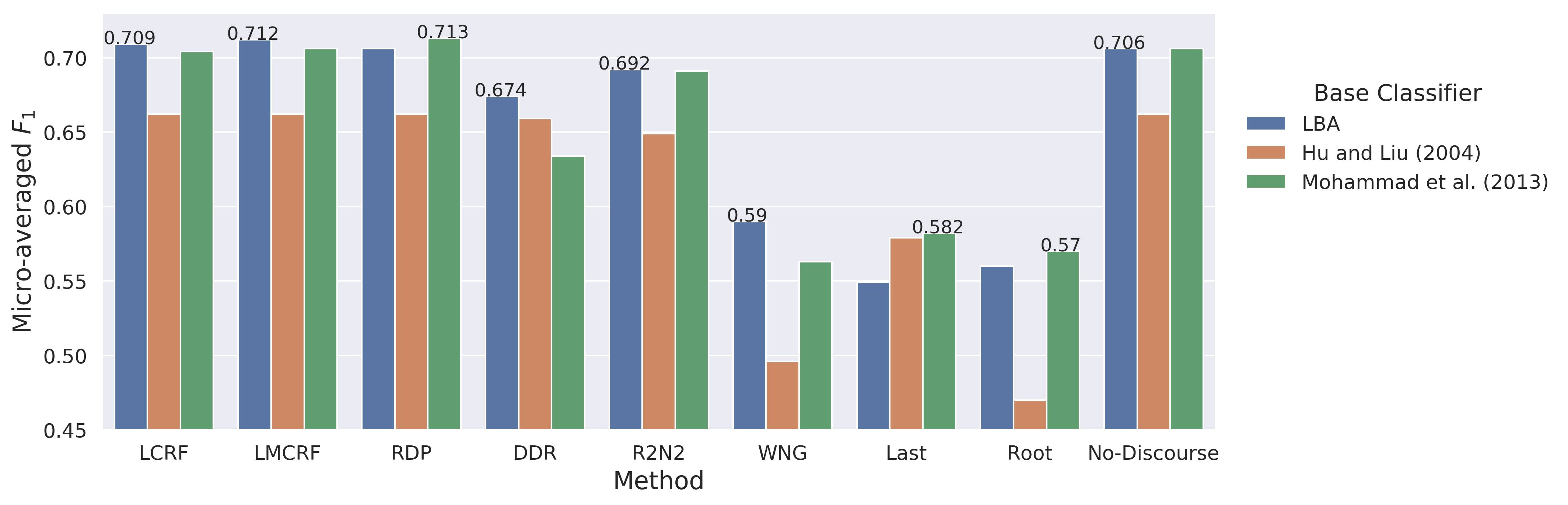}
  \caption{\texttt{Micro-\F{}}}\label{dasa:fig:potts-base-classifier-micro-F1}
\end{subfigure}
}
\caption[PotTS results of discourse-aware classifiers with different
  base classifiers]{Results of discourse-aware sentiment analysis
  methods with different base classifiers on the PotTS
  corpus}\label{dasa:fig:potts-base-classifier}
\end{figure*}

\subsection{Base Classifier}

\begin{figure*}[htb!]
{ \centering
\begin{subfigure}{\textwidth}
  \centering
  \includegraphics[width=\linewidth]{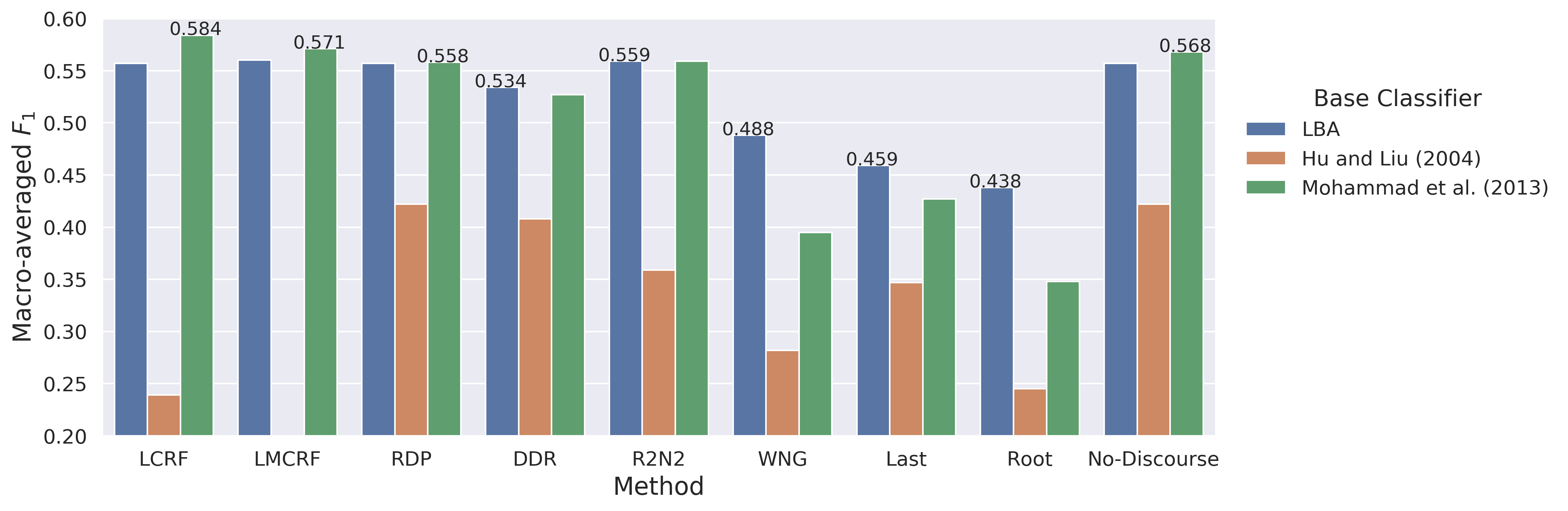}
  \caption{\texttt{Macro-\F{}}}\label{dasa:fig:sb10k-base-classifier-macro-F1}
\end{subfigure}\\
\begin{subfigure}{\textwidth}
  \centering
  \includegraphics[width=\linewidth]{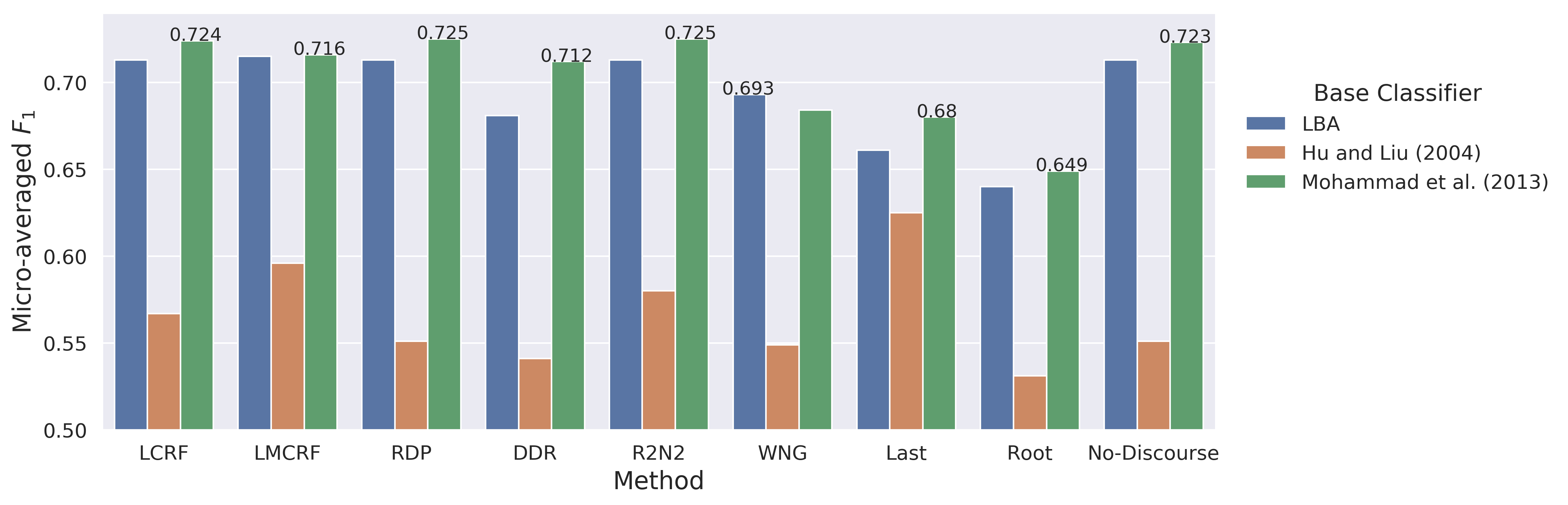}
  \caption{\texttt{Micro-\F{}}}\label{dasa:fig:sb10k-base-classifier-micro-F1}
\end{subfigure}
}
\caption[SB10k results of discourse-aware classifiers with different
  base classifiers]{Results of discourse-aware sentiment analysis
  methods with different base classifiers on the SB10k
  corpus}\label{dasa:fig:sb10k-base-classifier}
\end{figure*}

To assess the impact of the former factor (the quality of the base
sentiment classifier), we replaced all polarity scores produced by the
LBA system with the respective values predicted by the best lexicon-
and machine-learning--based MLSA methods (the systems of
\citeauthor{Hu:04} [\citeyear{Hu:04}] and \citeauthor{Mohammad:13}
           [\citeyear{Mohammad:13}] respectively) and retrained all
           DASA approaches on the updated data, subsequently
           evaluating them on the PotTS and SB10k test sets.  The
           results of this evaluation are shown in
           Figures~\ref{dasa:fig:potts-base-classifier}
           and~\ref{dasa:fig:sb10k-base-classifier}.

As we can see from the first figure, our initially chosen LBA approach
is indeed a more amenable basis to almost all discourse-aware
sentiment methods on the PotTS corpus.  A few exceptions to this
general rule are the macro-averaged \F{}-score of the \textsc{Last}
baseline, which surprisingly improves in combination with the
lexicon-based system, and the micro-average of the \textsc{RDP} and
\textsc{Last} methods, which attain their best results (0.713 and
0.582) in conjunction with the SVM classifier of \citet{Mohammad:13}.

A slightly different situation is observed on the SB10k corpus though.
On this dataset, LBA still leads to higher macro-\F{}--scores for
\textsc{DDR}, \textsc{R2N2}, \textsc{WNG}, \textsc{Last}, and
\textsc{Root}; but the approach of \citet{Mohammad:13} improves the
results of \textsc{LCRF}, \textsc{LMCRF}, \textsc{RDP}, and
\textsc{No-Discourse}.  The SVM classifier is also the unequivocal
leader in terms of the micro-averaged \F, yielding the highest scores
for all systems except WNG\@.  Unfortunately, the lexicon-based
predictor of \citet{Hu:04} performs much weaker than SVM and LBA:\ the
highest macro- and micro-averaged \F{}-scores achieved with this
approach run up to 0.422 (\textsc{RDP}) and 0.625 (\textsc{Last})
respectively.  The most disappointing result for us, however, is that
the LMCRF system completely fails to predict any polar class except
\textsc{Neutral} on the SB10k test set when trained with the scores of
this method (see
Figure~\ref{dasa:fig:sb10k-base-classifier-macro-F1}).  Similarly,
LCRF yields considerably lower scores in combination with this
solution, reaching only 0.239 macro-\F{}.

\subsection{Parsing Quality and Relation Scheme}

Another factor that could significantly influence the results of
discourse-aware methods was the quality of automatic RST parsing and
the set of discourse relations distinguished by the parser system.
Although improving the results of DPLP let alone manually annotating
the complete PotTS and SB10k datasets was beyond the scope of our
dissertation (even though we have made such attempt, see
[\citeauthor{Sidarenka:15a}, \citeyear{Sidarenka:15a}]), we decided to
check whether at least evaluating the DASA methods on manually
annotated data would improve their results.  For this purpose, we
asked a student assistant to segment and parse 88\% of the tweets from
the PotTS test set\footnote{Unfortunately, due to the limited
  availability of the student, we could not annotate the whole test
  set.} and tested all DASA approaches on these hand-crafted RST data.
\begin{table}[bht!]
  \begin{center}
    \bgroup\setlength\tabcolsep{0.1\tabcolsep}\scriptsize
    \begin{tabular}{p{0.162\columnwidth} 
        *{9}{>{\centering\arraybackslash}p{0.074\columnwidth}} 
        *{2}{>{\centering\arraybackslash}p{0.068\columnwidth}}} 
      \toprule
      \multirow{2}*{\bfseries Method} & %
      \multicolumn{3}{c}{\bfseries Positive} & %
      \multicolumn{3}{c}{\bfseries Negative} & %
      \multicolumn{3}{c}{\bfseries Neutral} & %
      \multirow{2}{0.068\columnwidth}{\bfseries\centering Macro\newline \F{}} & %
      \multirow{2}{0.068\columnwidth}{\bfseries\centering Micro\newline \F{}}\\
      \cmidrule(lr){2-4}\cmidrule(lr){5-7}\cmidrule(lr){8-10}

      & Precision & Recall & \F{} & %
      Precision & Recall & \F{} & %
      Precision & Recall & \F{} & & \\\midrule

      \multicolumn{12}{c}{\cellcolor{cellcolor}PotTS}\\

      LCRF & 0.82 & 0.82 & \textbf{0.82} & %
       \textbf{0.66} & 0.55 & 0.6 & %
       0.69 & 0.75 & 0.72 & %
       0.71 & 0.747\\

     LMCRF & \textbf{0.83} & 0.81 & \textbf{0.82} & %
       0.65 & 0.55 & 0.6 & %
       0.69 & \textbf{0.78} & \textbf{0.73} & %
       0.709 & 0.749\\

      RDP & 0.8 & 0.84 & \textbf{0.82} & %
       0.64 & 0.58 & 0.61 & %
       \textbf{0.72} & 0.71 & 0.72 & %
       \textbf{0.718} & 0.751\\

      DDR & 0.78 & 0.75 & 0.77 & %
       0.58 & \textbf{0.66} & \textbf{0.62} & %
       0.66 & 0.63 & 0.64 & %
       0.693 & 0.698\\

      R2N2 & 0.81 & 0.82 & 0.81 & %
       0.64 & 0.53 & 0.58 & %
       0.68 & 0.74 & 0.71 & %
       0.697 & 0.737\\

      WNG & 0.58 & 0.74 & 0.65 & %
       0.63 & 0.19 & 0.29 & %
       0.51 & 0.51 & 0.51 & %
       0.47 & 0.558\\

      \textsc{Last} & 0.55 & \textbf{0.86} & 0.67 & %
       0.51 & 0.11 & 0.18 & %
       0.56 & 0.35 & 0.43 & %
       0.426 & 0.55\\

      \textsc{Root} & 0.58 & 0.56 & 0.57 & %
       0.58 & 0.25 & 0.35 & %
       0.43 & 0.6 & 0.5 & %
       0.46 & 0.513\\

      \textsc{No-Discourse} & 0.81 & 0.84 & \textbf{0.82} & %
       0.65 & 0.57 & 0.61 & %
       \textbf{0.72} & 0.73 & \textbf{0.73} & %
       0.716 & \textbf{0.753}\\\bottomrule
    \end{tabular}
    \egroup{}
    \caption[Results of DASA methods on manually annotated RST
      trees]{Results of discourse-aware sentiment analysis methods on
      the PotTS corpus with manually annotated RST trees}\label{dasa:tbl:res-gold}
  \end{center}
\end{table}

As we can see from the results in Table~\ref{dasa:tbl:res-gold}, the
scores of all systems except \textsc{WNG}, \textsc{Last}, and
\textsc{Root} increase by three to four percent.  Even the
macro-averaged \F{}-measure of the discourse-unaware classifier
improves from 0.677 to 0.716, as does its micro-\F{}--score, which
rises from 0.706 to 0.753 \F{}.  These last changes, however, are
exclusively due to the reduced size of the test data (on which the
base classifier performs better than on the full test set), since the
discourse-unaware method does not take RST trees into account.
Unfortunately, this time, \textsc{No-Discourse} also outperforms all
discourse-aware approaches in terms of the micro-averaged \F{},
achieving an accuracy of 75,3\%, although it still loses to the
Recursive Dirichlet Process on the macro-averaged metric, yielding a
0.2\% worse result than RDP (0.716 versus 0.718 macro-\F{}).
Another surprising finding for us is that in the gold discourse
annotation, EDUs that determine the actual polarity of the tweet are
unlikely to appear either at the end of a message or at the top of its
RST tree, which leads to the degradation of the scores for the
\textsc{Last} and \textsc{Root} baselines.

Although manually annotated RST trees do improve the results of most
discourse-aware sentiment methods, this fact is of little help to us
if we are bound to the output of an automatic parser. A common way to
improve the quality of automatic RST analysis and ease the task of
DASA methods is to reduce the number of discourse relations
distinguished by the parsing system.  Drawing on the work of~\citet{Bhatia:15}, we also used this approach, projecting all
discourse relations from the Potsdam Commentary Corpus~\cite{Stede:14}
to the binary set of \textsc{Contrastive} and \textsc{Non-Contrastive}
ones.  Although similar approximations were made in almost all other
discourse-aware solutions~\cite[cf.
][]{Chenlo:13,Heerschop:11,Zhou:11}, we were not sure whether the
subset that we used was indeed optimal and sufficient to reflect all
possible discourse interactions that could play an important role in
sentiment composition.

To answer this question, we retrained the DPLP parser on the PCC,
using the subsets of relations proposed by \citet{Chenlo:13},
\citet{Heerschop:11}, and \citet{Zhou:11}, and also tried the original
set of all RST links from the Potsdam Commentary Corpus.  A detailed
overview of these sets is given in Table~\ref{dasa:tbl:rst-rel-sets}.
\begin{table}[hbt]
  \begin{center}
    \bgroup{}
    \setlength\tabcolsep{0.8\tabcolsep}\scriptsize
    \begin{tabular}{p{0.135\columnwidth} 
        *{1}{>{\centering\arraybackslash}p{0.4\columnwidth}}
        *{1}{>{}p{0.4\columnwidth}}} 
      \toprule
      \textbf{Scheme} & \textbf{Relation Set} & {\centering\textbf{Equivalence Classes}}\\\midrule

      \citeauthor{Bhatia:15} & \{\textsc{Contrastive},
      \textsc{\bfseries Non-Contrastive}\} & \textsc{Contrastive} $\defeq$
      \{\textsc{Antithesis}, \textsc{Antithesis-E},
      \textsc{Comparison}, \textsc{Concession},
      \textsc{Consequence-S}, \textsc{Contrast},
      \textsc{Problem-Solution}\}.\\

      \citeauthor{Chenlo:13} & \{\textsc{Attribution},
      \textsc{Background}, \textsc{Cause}, \textsc{Comparison},
      \textsc{Condition}, \textsc{Consequence}, \textsc{Contrast},
      \textsc{Elaboration}, \textsc{Enablement}, \textsc{Evaluation},
      \textsc{Explanation}, \textsc{Joint}, \textsc{Otherwise},
      \textsc{Temporal}, \textsc{\bfseries Other}\} & \\

      \citeauthor{Heerschop:11} & \{\textsc{Attribution},
      \textsc{Background}, \textsc{Cause}, \textsc{Condition},
      \textsc{Contrast}, \textsc{Elaboration}, \textsc{Enablement},
      \textsc{Explanation}, \textsc{\bfseries Other}\} & \\

      \textsc{PCC} & \{\textsc{Antithesis}, \textsc{Background},
      \textsc{Cause}, \textsc{Circumstance}, \textsc{Concession},
      \textsc{Condition}, \textsc{Conjunction}, \textsc{Contrast},
      \textsc{Disjunction}, \textsc{E-Elaboration},
      \textsc{Elaboration}, \textsc{Enablement},
      \textsc{Evaluation-N}, \textsc{Evaluation-S}, \textsc{Evidence},
      \textsc{Interpretation}, \textsc{Joint}, \textsc{Justify},
      \textsc{List}, \textsc{Means}, \textsc{Motivation},
      \textsc{Otherwise}, \textsc{Preparation}, \textsc{Purpose},
      \textsc{Reason}, \textsc{Restatement}, \textsc{Restatement-MN},
      \textsc{Result}, \textsc{Sequence}, \textsc{Solutionhood},
      \textsc{Summary}, \textsc{Unconditional}, \textsc{Unless},
      \textsc{Unstated-Relation}\} & \\

      \citeauthor{Zhou:11} & \{\textsc{Contrast}, \textsc{Condition},
      \textsc{Continuation}, \textsc{Cause}, \textsc{Purpose},
      \textsc{\bfseries Other}\} & \textsc{Contrast} $\defeq$
      \{\textsc{Antithesis}, \textsc{Concession}, \textsc{Contrast},
      \textsc{Otherwise}\};\newline \textsc{Continuation} $\defeq$
      \{\textsc{Continuation}, \textsc{Parallel}\};\newline
      \textsc{Cause} $\defeq$ \{\textsc{Evidence}, \textsc{Nonvolitional-Cause},
      \textsc{Nonvolitional-Result}, \textsc{Volitional Cause},
      \textsc{Volitional-Result}\};\\\bottomrule
    \end{tabular}
    \egroup{}
    \caption[RST relations used in different discourse-aware sentiment
      methods]{RST relations used in the original Potsdam Commentary
      Corpus and different discourse-aware sentiment methods\\ {\small
        (default relation, which subsumes the rest of the links, is
        shown in \textbf{boldface})}}\label{dasa:tbl:rst-rel-sets}
  \end{center}
\end{table}

To check whether cardinalities of these sets indeed correlated with
the quality of automatic RST parsing, we evaluated each retrained
system on the held-out PCC test data and present the results of this
evaluation in Table~\ref{dasa:tbl:dplp-results}.  As is evident from
the scores, coarser relation schemes in fact improve parsing quality,
especially in terms of relation \F{}.  In the most extreme case (\eg{}
\citeauthor{Bhatia:15}, which has only two links, versus PCC, which
comprises 34 relations), these gains can reach up to seven
percent. However, with respect to other metrics (span and nuclearity
\F{}), the gaps are notably smaller and might even be in favor of the
richer relation set (cf.\ nuclearity \F{} for PCC).
\begin{table}[htb]
  \begin{center}
    \bgroup\setlength\tabcolsep{0.1\tabcolsep}\scriptsize
    \begin{tabular}{p{0.22\columnwidth} 
        *{3}{>{\centering\arraybackslash}p{0.25\columnwidth}}} \toprule

      \textbf{Relation Scheme} & \textbf{Span \F{}} &
      \textbf{Nuclearity \F{}} & \textbf{Relation \F{}}\\\midrule

      \citeauthor{Bhatia:15} & \textbf{0.777} & 0.512 & \textbf{0.396}\\

      \citeauthor{Chenlo:13} & 0.769 & 0.505 & 0.362\\

      \citeauthor{Heerschop:11} & 0.774 & 0.51 & 0.361\\

      \textsc{PCC} & 0.776 & \textbf{0.534} & 0.326\\

      \citeauthor{Zhou:11} & 0.776 & 0.501 & 0.388\\\bottomrule
    \end{tabular}
    \egroup{}
    \caption[Results of the DPLP parser on PCC~2.0]{Results of the
      DPLP parser on PCC~2.0 with different relation schemes\\}\label{dasa:tbl:dplp-results}
  \end{center}
\end{table}

\begin{figure*}[bh!]
{
\centering
\begin{subfigure}{\textwidth}
  \centering
  \includegraphics[width=\linewidth]{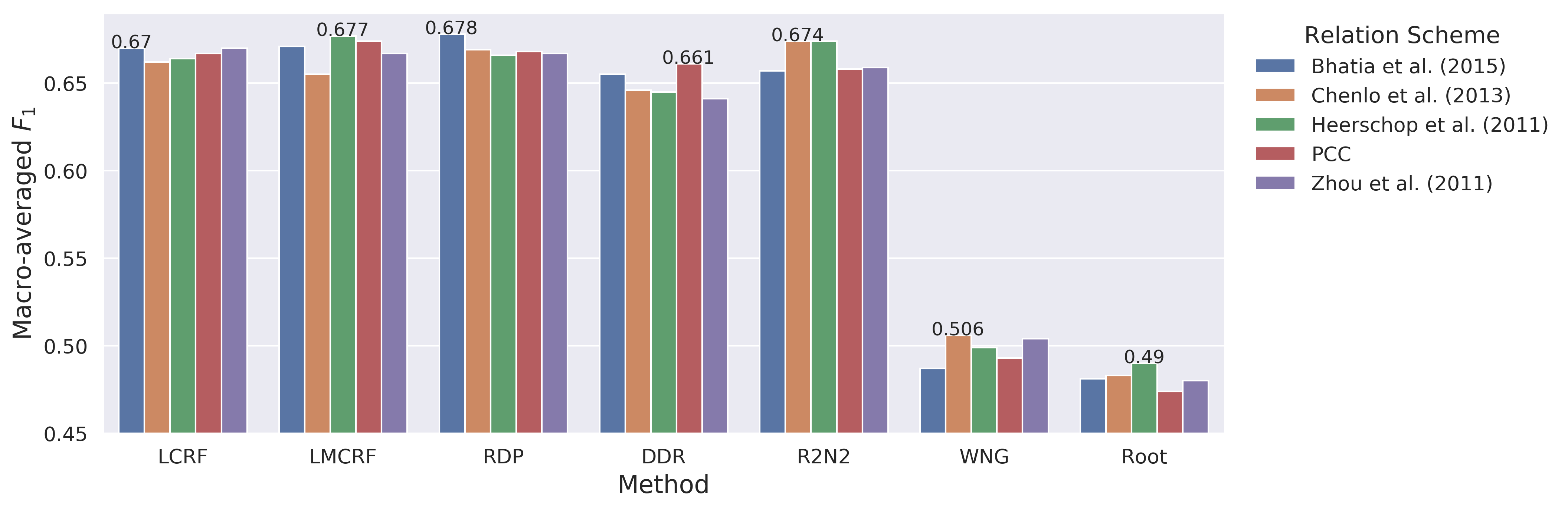}
  \caption{\texttt{Macro-\F{}}}\label{dasa:fig:potts-rel-schemes-macro-F1}
\end{subfigure}\\
\begin{subfigure}{\textwidth}
  \centering
  \includegraphics[width=\linewidth]{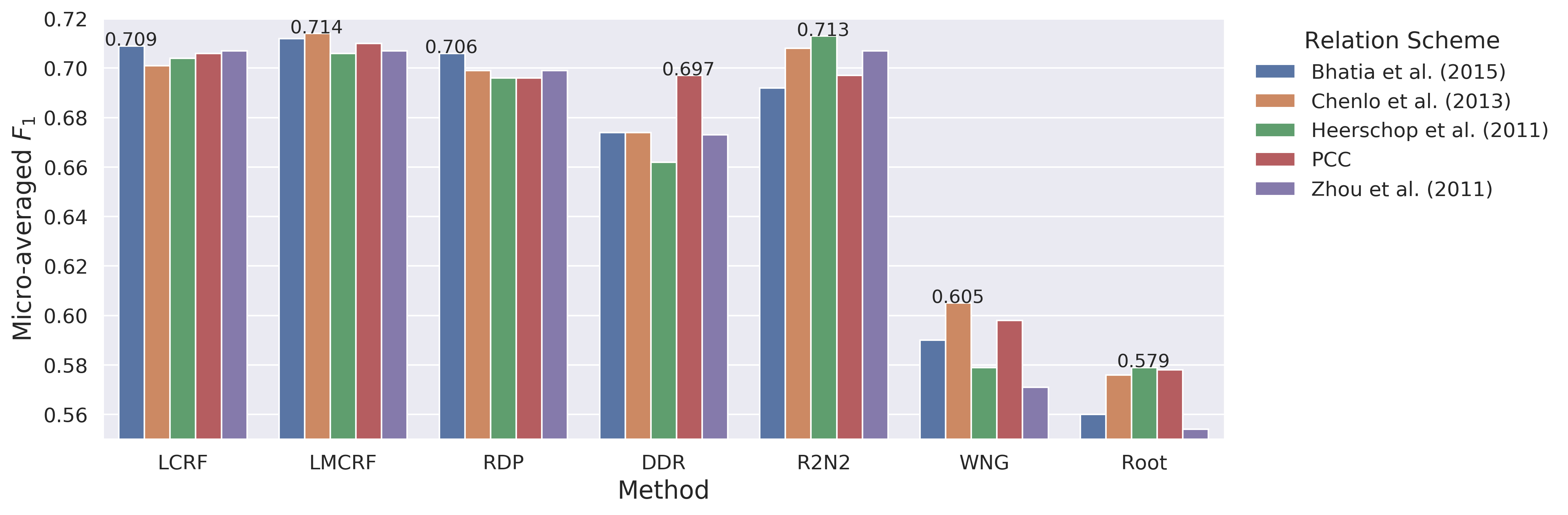}
  \caption{\texttt{Micro-\F{}}}\label{dasa:fig:potts-rel-schemes-micro-F1}
\end{subfigure}
}
\caption[PotTS results of discourse-aware classifiers with different
  relation schemes]{Results of discourse-aware sentiment classifiers
  for different relation schemes on the PotTS
  corpus}\label{dasa:fig:potts-rel-schemes}
\end{figure*}

\begin{figure*}[htb!]
{ \centering
\begin{subfigure}{\textwidth}
  \centering
  \includegraphics[width=\linewidth]{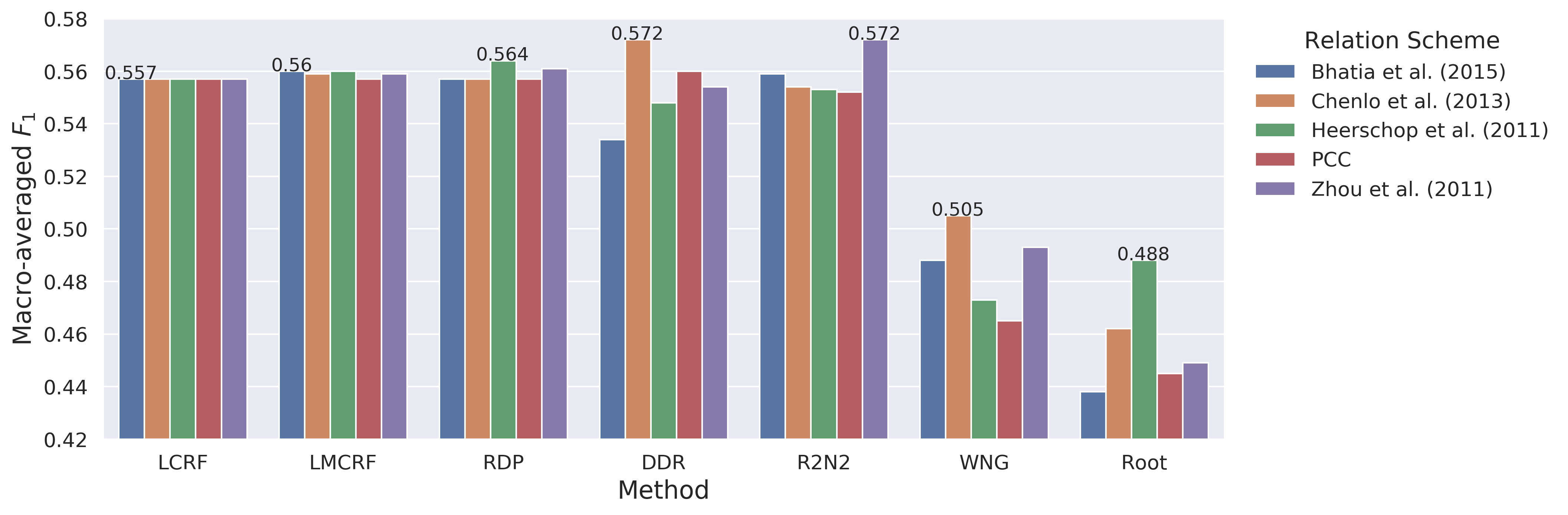}
  \caption{\texttt{Macro-\F{}}}\label{dasa:fig:sb10k-rel-schemes-macro-F1}
\end{subfigure}\\
\begin{subfigure}{\textwidth}
  \centering
  \includegraphics[width=\linewidth]{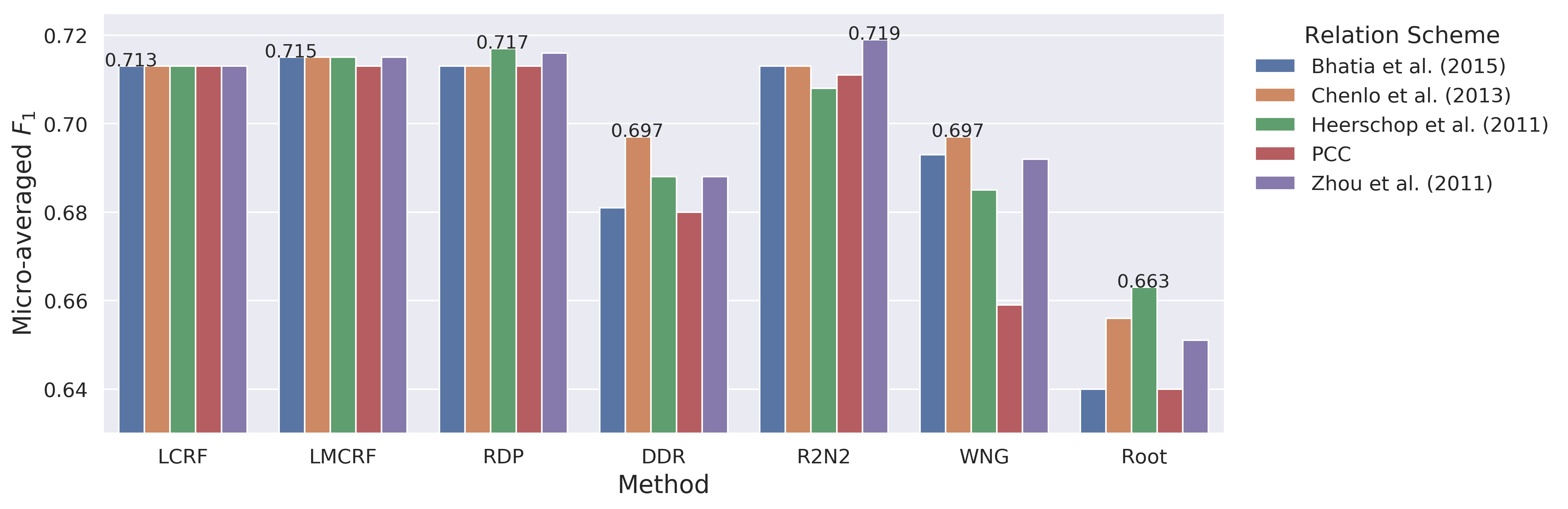}
  \caption{\texttt{Micro-\F{}}}\label{dasa:fig:sb10k-rel-schemes-micro-F1}
\end{subfigure}
}
\caption[SB10k results of discourse-aware classifiers for different
  relation schemes]{Results of discourse-aware sentiment classifiers
  for different relation schemes on the SB10k
  corpus}\label{dasa:fig:sb10k-rel-schemes}
\end{figure*}

To see how this varying quality affected the net results of
discourse-aware sentiment methods, we re-evaluated all DASA approaches
on the updated automatic RST trees and show the results of this
evaluation in Figures~\ref{dasa:fig:potts-rel-schemes}
and~\ref{dasa:fig:sb10k-rel-schemes}.

As it turns out, latent-marginalized CRF can still hold the overall
record in both macro- and micro-averaged \F{} on the PotTS corpus,
although its margin to the closest competitor (R2N2) is relatively
small, amounting to only 0.1 percent.  Interestingly enough, both
top-performing methods (LMCRF and R2N2) achieve their best results
with richer relation sets than the one we used in our initial
experiment: For example, LMCRF attains its highest macro-score in
combination with the relation scheme of~\citet{Heerschop:11} and
yields the best micro-\F{} when used with the scheme
of~\citet{Chenlo:14}.  The rhetorical recursive neural network, vice
versa, attains its highest macro-average with the latter relation set
and reaches its best micro-\F{} in conjunction with the former subset.

A different situation is observed with other DASA approaches though.
For example, LCRF and RDP perform best when used with the initially
chosen set of \citet{Bhatia:15}.  On the other hand, discourse-depth
reweighting strongly benefits from the full unconstrained set of PCC
relations, which is probably due to the better nuclearity
classification achieved with this scheme.  Finally, \textsc{WNG} and
\textsc{Root} reach their best results with the relation subsets
proposed by~\citeauthor{Chenlo:14} and \citeauthor{Heerschop:11},
respectively.

A much more uniform situation is observed on the SB10k corpus (see
Figure~\ref{dasa:fig:sb10k-rel-schemes}), where the \F-scores of our
methods vary only slightly across different relation schemes.  The
only significant improvements that we can notice this time are higher
macro- and micro-averaged \F{}s achieved by the RDP approach in
combination with the \citeauthor{Heerschop:11}'s subset.  This subset
is also most amenable to the \textsc{Root} baseline, which reaches
0.488 macro-\F{} and 0.663 micro-\F{}, significantly improving on its
initial results.  At the same time, discourse-depth reweighting and
the approach of~\citeauthor{Wang:13} capitalize on the relations
defined by~\citeauthor{Chenlo:13} so much that the former system even
achieves the highest overall macro-\F{}--score (0.572), being on a par
with the R2N2 system.

\section{Summary and Conclusions}

At this point, our chapter has come to an end and, concluding it, we
would like to recap that in this part of the thesis:
\begin{itemize}
  \item we have presented an overview of the most popular approaches
    to automatic discourse analysis (RST, PDTB, and SDRT) and
    explained why we think that one of these frameworks (Rhetorical
    Structure Theory) would be more amenable to the purposes of
    discourse-aware sentiment analysis than the others;
  \item to substantiate our claims and to see whether the
    lexicon-based attention system introduced in the previous chapter
    would indeed benefit from information on discourse structure, we
    segmented all microblogs from the PotTS and SB10k corpora into
    elementary discourse units using the SVM-based segmenter
    of~\citet{Sidarenka:15} and parsed these messages with the RST
    parser of~\citet{Ji:14}, which had been previously retrained on
    the Potsdam Commentary Corpus~\cite{Stede:14};
  \item afterwards, we estimated the results of existing
    discourse-aware sentiment methods (the systems
    of~\citeauthor{Wang:15}~[\citeyear{Wang:15}] and
    \citeauthor{Bhatia:15}~[\citeyear{Bhatia:15}]) and also evaluated
    two simpler baselines (in which we the predicted semantic
    orientation of a tweet by taking the polarity of its last and root
    EDUs), getting the best results with the R2N2 solution
    of~\citet{Bhatia:15} (0.657 and 0.559 macro-\F{} on PotTS and
    SB10k respectively);
  \item we could, however, improve on these scores and also outperform
    the plain LBA system (although by a not very large margin) with
    our three proposed discourse-aware sentiment solutions: latent and
    latent-marginalized conditional random fields and Recursive
    Dirichlet Process; pushing the macro-averaged \F{}-score on PotTS
    up to 0.678 and increasing the result on SB10k to 0.56 macro-\F{};
  \item a subsequent evaluation of these approaches with different
    settings showed that the results of all discourse-aware methods
    largely correlated with the scores of the base sentiment
    classifier and also revealed an important drawback of the
    latent-marginalized CRFs, which failed to predict any positive or
    negative instance on the test set of the SB10k corpus when trained
    in combination with the lexicon-based approach of~\citet{Hu:04};
  \item nevertheless, almost all DASA solutions could improve their
    scores when tested on manually annotated RST trees or used with a
    richer set of discourse relations.
\end{itemize}

\chapter*{Afterword}
\markboth{\textsc{AFTERWORD}}{}
\addcontentsline{toc}{chapter}{Afterword}

It is hard to believe, but at this point we have finally reached the
home stretch of our \thepage-page long marathon and, preparing the
final spurt, we should first recall the main milestones that we have
seen along this way:
\begin{itemize}
\item As you might remember, we started off by summarizing the history
  of sentiment analysis, going back to its very origins in the ancient
  Greek philosophy and tracing its development to the present day;

\item Afterwards, to see what the current state of the art in opinion
  mining would yield on German Twitter, we created a corpus of
  $\approx8,000$ German tweets, collecting these messages for four
  different topics (federal elections, papal conclave, general
  political discussions, and casual everyday conversations).  To
  ensure a good recall of opinionated statements in the resulting
  dataset, we grouped all microblogs into three formal categories
  (tweets with a polar term from the SentiWS lexicon, messages
  containing a smiley, and all remaining microblogs) and sampled an
  equal number of tweets (666) for each of the four topics from each
  of these three categories.  After annotating the corpus in three
  steps (initial, adjudication, and final), we attained a reliable
  level of inter-annotator agreement for all elements (sentiments,
  sources, targets, polar terms, downtoners, negations, and
  intensifiers), finding that both selection criteria (topics and
  formal traits) significantly affected the distribution of sentiments
  and polar terms and the reliability of their annotation;

\item Then, at the first checkpoint, we compared existing German
  sentiment lexicons, which were translated from English resources and
  revised by human experts, with lexicons that were generated
  automatically from scratch with the help of state-of-the-art
  dictionary\mbox{-,} corpus\mbox{-,} and word-embedding--based
  methods.  An evaluation of these approaches on our corpus showed
  that semi-automatically translated polarity lists were generally
  better than the automatically induced ones, reaching 0.587
  macro-\F{} and attaining 0.955 micro-\F{}--score on the prediction
  of polar terms.  Furthermore, among fully automatic methods,
  dictionary-based systems showed stronger results than their corpus-
  and word-embedding--based competitors, yielding 0.479 macro-\F{} and
  0.962 micro-\F{}.  We could, however, improve on the latter metric
  (pushing it to 0.963) with our proposed linear projection solution,
  in which we first found a line that maximized the mutual distance
  between the projections of seed vectors with opposite semantic
  orientations and then projected the embeddings of all remaining
  words on that line, considering the distance of these projections to
  the median as polarity scores of respective terms;

\item In Chapter~\ref{chap:fgsa}, we turned our attention to the
  fine-grained sentiment analysis, in which we tried to predict the
  spans of sentiments, targets, and holders of opinions using two most
  popular approaches to this task: conditional random fields and
  recurrent neural networks.  We obtained our best results (0.287
  macro-\F{}) with the first-order linear-chain CRFs.  We could,
  however, increase these scores by using alternative topologies of
  CRFs (second-order linear-chain and semi-Markov CRFs) and also boost
  the macro-averaged \F{} to 0.38 by taking a narrower interpretation
  of sentiment spans (in which we only assigned the \textsc{Sentiment}
  tag to polar terms).  Further evaluation of these methods proved the
  utility of the text normalization step (which raised the macro-\F{}
  of the CRF-method by almost 3\%) and task-specific word embeddings
  with the least-squares fallback (which improved the
  macro-\F{}--score of the GRU system by 1.4\%);

\item Afterwards, in Chapter~\ref{chap:cgsa}, we addressed one of the
  most popular objective in contemporary sentiment
  analysis---message-level sentiment analysis (MLSA).  To get a better
  overview of the numerous existing systems, we compared three larger
  families of MLSA methods: dictionary-, machine-learning--, and
  deep-learning--based ones; finding that the last two groups
  performed significantly better than the lexicon-based approaches
  (the best macro-\F{}--scores of machine- and deep-learning methods
  run up to 0.677 and 0.69 respectively, whereas the best
  lexicon-based solution [\citeauthor{Hu:04}, \citeyear{Hu:04}] only
  reached 0.641 macro-\F{}).  Apart from this, we improved the results
  of many reimplemented approaches by changing their default
  configuration (\eg{} abandoning polarity changing rules of
  lexicon-based systems, using alternative classifiers in ML-based
  systems, or taking the least-squares embeddings for DL-based
  methods).  In addition to the numerous reimplementations of popular
  existing algorithms, we also proposed our own
  solution---lexicon-based attention (LBA), in which we tried to unite
  the lexicon and deep-learning paradigms by taking a bidirectional
  LSTM network and explicitly pointing its attention to polar terms
  that appeared in the analyzed messages.  With this solution, we not
  only outperformed all alternative DL systems but also improved on
  the scores of ML-based classifiers, attaining 0.69 macro-\F{} and
  0.73 micro-\F{} on the PotTS corpus.  Similarly to our findings of
  the previous chapter, we observed a strong positive effect of text
  normalization and task-specific embeddings with the least-squares
  approximation;

\item Finally, in the last part, we tried to improve the results of
  the proposed LBA method by making it aware of the discourse
  structure.  For this purpose, we segmented all microblogs from the
  PotTS and SB10k corpora into elementary discourse units,
  individually analyzing each of these segments with our MLSA
  classifier, and then estimated the overall polarity of a tweet by
  joining the polarity scores of its EDUs over the RST tree.  We
  proposed three different ways of doing this joining: latent CRFs,
  latent-marginalized CRFs, and Recursive Dirichlet Process; obtaining
  better results than existing discourse-aware sentiment methods and
  also outperforming the original discourse-unaware baseline.  In the
  concluding experiments, we further improved these scores by using
  manually annotated RST trees and richer subsets of discourse
  relations.
\end{itemize}

\section*{Conclusions}

Now that we have gone past all these landmarks, it is time to unbag
the questions which we had asked ourselves at the beginning of this
endeavor, and try to answer them again, equipped with all knowledge
that we have acquired during our run.  Here we go:

\begin{itemize}
  \item\textbf{Can we apply opinion mining methods devised for
    standard English to German Twitter?}

    Yes, we can, but the success of these approaches might
    significantly vary depending on the task, the size, and the
    reliability of the training data, as well as the evaluation metric
    that we use. For example, dictionary-based lexicon methods
    achieved fairly good results on their objective, but this success
    was mostly due to the high quality of the \textsc{GermaNet}
    annotation.  On the other hand, our manually labeled PotTS corpus
    was evidently too small for fine-grained sentiment systems, which
    failed to generalize to unseen tweets despite their very high
    scores on the training set.  Message-level sentiment approaches,
    vice versa, seemed to be quite happy with the size of the training
    dataset, attaining good results on both corpora (PotTS and SB10k).
    Nevertheless, we again experienced a lack of data while working on
    discourse-aware enhancements, many of which hit the same ceiling
    of the macro-averaged \F{}-scores.

    Apart from these difficulties arising from insufficient data, we
    also noticed a significant degradation of the scores for systems
    whose original tasks and evaluation metrics were different from
    ours.  For example, the lexicon generation method of
    \citet{Esuli:05} was originally designed to assign polarity scores
    to all \emph{synsets} found in the \textsc{WordNet} and not to
    produce a list of polar \emph{words}.  Similarly, the RNTN
    approach of \citet{Socher:13} was trained and evaluated on all
    syntactic subtrees of a document and not only at the top text
    level.  Likewise, the system of~\citet{Yessenalina:11} was devised
    for doing ordinal logistic regression and not polarity
    classification, as in our case.  As a result, all these approaches
    showed lower scores than their competitors in our evaluation, even
    though they are undoubtedly well suited for their original data
    and tasks.

    Due to the high diversity of methods, metrics, and tasks, it is
    difficult to provide a general recipe for transferring existing
    English sentiment systems to German Twitter, but we still would
    like to formulate at least a few rules of thumb, which came up
    during our experiments:
    \begin{itemize}
      \item\textbf{Prefer methods that are closest to your training
        objective} and that were trained under similar conditions
        w.r.t.\ the amount of data, their class distribution and
        domain;
      \item\textbf{Put every single setting of these methods into
        question}---bear in mind that things that work well in the
        original cases are not guaranteed to work in your
        situation.\footnote{In this respect, it is important to
          realize that every classification task is merely an attempt
          to solve a system of equations, so that methods that are
          good at solving one system might completely fail to solve
          another set of equations.}  The more options you try, the
        better will be your results;
      \item\textbf{Try using manually labeled resources for your
        target domain}, if they are available, but pay attention to
        the quality of their annotation---it often matters more than
        the corpus size;
      \item If there are manually annotated data, \textbf{prefer
        machine-learning methods to hard-coded rules}---they will
        penalize their bad components automatically by themselves;
      \item\textbf{Do not use randomly initialized word embeddings for
        deep-learning systems}---initialize them with language-model
        vectors (which are cheap to obtain).  Otherwise, your model
        might get stuck in a very bad local optimum.
    \end{itemize}

  \item\textbf{Which groups of approaches are best suited for which
    sentiment tasks?}

    Based on our evaluation, we answer this question as follows:
    \begin{itemize}
      \item\emph{Sentiment lexicon generation} is more amenable to
        dictionary-based solutions, provided that there exists a
        sufficiently big, reliably annotated lexical taxonomy for
        these systems.  If there is no such resource, one should
        better resort to word-embedding--based algorithms;

      \item With a limited amount of training data, \emph{fine-grained
        sentiment analysis} can be better addressed with probabilistic
        graphical models, such as conditional random fields with
        hand-crafted features;

      \item On the other hand, plain \emph{message-level sentiment
        analysis} can be efficiently tackled with both machine- and
        deep-learning algorithms, such as SVM, logistic regression, or
        RNN\@;

      \item But probabilistic graphical models strike back at
        \emph{discourse-aware sentiment methods}, where they might
        even outperform pure neural-network solutions, although the
        margin of these improvements is not that large.
    \end{itemize}

    Thus, probabilistic model can still hold their ground when it
    comes to structured prediction, but the difference of these
    algorithms from and their improvements upon neural networks are
    gradually vanishing.

  \item\textbf{How much do word- and discourse-level analyses affect
    message-level sentiment classification?}

    Our evaluation in Section~\ref{cgsa:subsec:eval:lexicons} showed
    that the macro-averaged \F{}-scores of our proposed lexicon-based
    attention system varied by up to 14\% (from 0.64 to 0.69
    macro-\F{} on the PotTS corpus, and from 0.44 to 0.58 on SB10k)
    depending on the lexicon used by this approach.  At the same,
    discourse enhancements could only improve the results of LBA by at
    most 1.5\% percent (from 0.677 to 0.678 on PotTS, and from 0.557
    to 0.572 on SB10k).  Although it appears as if the lexicon
    component were more important to a sentiment system, we would like
    to preclude such incorrect conclusion, because
    \begin{inparaenum}[(a)]
      \item a full-fledged sentiment solution should take into account
        both linguistic levels (words and discourse) and
      \item these relative results might look different if we expand
        the analyzed domain to longer documents or apply
        discourse-aware methods to complete discussion threads.
    \end{inparaenum}

  \item\textbf{Does text normalization help analyze sentiments?}

    Yes, it definitely does.  As we could see in
    Chapters~\ref{chap:fgsa} and~\ref{chap:cgsa}, normalization
    significantly improves the quality of fine-grained and
    message-level sentiment analyses, boosting the results on the
    former task by up to 4\% (see
    Table~\ref{snt-fgsa:tbl:normalization}) and improving the
    macro-averaged \F{}-measure of message-level sentiment methods by
    up to 25\% (see Table~\ref{snt-cgsa:tbl:res-no-normalization}).

    The only question that remained unanswered in this context is
    which normalization steps exactly improve the scores of sentiment
    systems.  To make up for this omission, we separately deactivated
    each individual step of our text normalization pipeline
    (unification of Twitter phenomena, spelling correction, and
    normalization of slang terms) and rerun our message-level
    classification experiments using the lexicon-based attention
    system.  As we can see from the results in
    Table~\ref{afterword:tbl:lba-normalization-steps}, the
    micro-averaged \F{}-scores on both datasets benefit most from the
    unification of Twitter-specific phenomena, sinking by almost 19\%
    when this component is deactivated.  This step is also most useful
    for the macro-\F{} on the SB10k corpus, whereas the macro-average
    on PotTS mostly capitalizes on the normalization of slang terms.
    \begin{table}[htb!]
      \begin{center}
        \bgroup\setlength\tabcolsep{0.1\tabcolsep}\scriptsize
        \begin{tabular}{p{0.162\columnwidth} 
            *{9}{>{\centering\arraybackslash}p{0.074\columnwidth}} 
            *{2}{>{\centering\arraybackslash}p{0.068\columnwidth}}} 
          \toprule
          \multirow{2}*{\bfseries Method} & %
          \multicolumn{3}{c}{\bfseries Positive} & %
          \multicolumn{3}{c}{\bfseries Negative} & %
          \multicolumn{3}{c}{\bfseries Neutral} & %
          \multirow{2}{0.068\columnwidth}{\bfseries\centering Macro\newline \F{}$^{+/-}$} & %
          \multirow{2}{0.068\columnwidth}{\bfseries\centering Micro\newline \F{}}\\
          \cmidrule(lr){2-4}\cmidrule(lr){5-7}\cmidrule(lr){8-10}

          & Precision & Recall & \F{} & %
          Precision & Recall & \F{} & %
          Precision & Recall & \F{} & & \\\midrule

          \multicolumn{12}{c}{\cellcolor{cellcolor}PotTS}\\
          with normalization & 0.76 & 0.84 & 0.79 & %
          0.6 & 0.56 & 0.58 & %
          0.75 & 0.68 & 0.72 & %
          0.69 & 0.73\\
          w/o unification of Twitter phenomena & 0.51\negdelta{0.25} & 0.87\posdelta{0.03} & 0.64\negdelta{0.05} & %
          0.57\negdelta{0.03} & 0.4\negdelta{0.16} & 0.47\negdelta{0.11} & %
          0.68\negdelta{0.07} & 0.22\negdelta{0.46} & 0.34\negdelta{0.38} & %
          0.56\negdelta{0.13} & 0.54\negdelta{0.19}\\
          w/o spelling correction & 0.67\negdelta{0.09} & 0.84 & 0.74\negdelta{0.05} & %
          0.61\posdelta{0.01} & 0.34\negdelta{0.22} & 0.44\negdelta{0.14} & %
          0.74\negdelta{0.01} & 0.68 & 0.71\negdelta{0.01} & %
          0.59\negdelta{0.1} & 0.69\negdelta{0.04}\\
          w/o slang normalization & 0.59\negdelta{0.17} & 0.87\posdelta{0.03} & 0.7\negdelta{0.09} & %
          0.6 & 0.17\negdelta{0.39} & 0.26\negdelta{0.32} & %
          0.72\negdelta{0.03} & 0.6\negdelta{0.08} & 0.65\negdelta{0.07} & %
          0.48\negdelta{0.21} & 0.64\negdelta{0.09}\\

          \multicolumn{12}{c}{\cellcolor{cellcolor}SB10k}\\
          with normalization & 0.6 & 0.72 & 0.66 & %
          0.47 & 0.42 & 0.44 & %
          0.84 & 0.8 & 0.82 & %
          0.55 & 0.73\\
          w/o unification of Twitter phenomena & 0.36\negdelta{0.24} & 0.85\posdelta{0.13} & 0.5\negdelta{0.16} &%
          0.6\posdelta{0.13} & 0.25\negdelta{0.17} & 0.35\negdelta{0.09} & %
          0.84 & 0.51\negdelta{0.29} & 0.63\negdelta{0.19} & %
          0.43\negdelta{0.12} & 0.55\negdelta{0.18}\\

          w/o spelling correction & 0.54\negdelta{0.06} & 0.71\negdelta{0.01} & 0.61\negdelta{0.05} & %
          0.54\posdelta{0.07} & 0.26\negdelta{0.16} & 0.35\negdelta{0.09} & %
          0.79\negdelta{0.05} & 0.79\negdelta{0.01} & 0.79\negdelta{0.03} & %
          0.48\negdelta{0.07} & 0.7\negdelta{0.03}\\

          w/o slang normalization & 0.55\negdelta{0.05} & 0.71\negdelta{0.01} & 0.62\negdelta{0.04} & %
          0.64\posdelta{0.17} & 0.2\negdelta{0.22} & 0.3\negdelta{0.14} & %
          0.78\negdelta{0.06} & 0.82\posdelta{0.02} & 0.8\negdelta{0.02} & %
          0.46\negdelta{0.09} & 0.7\negdelta{0.03}\\\bottomrule
        \end{tabular}
        \egroup{}
        \caption{LBA$^{(1)}$ results without single text normalization
          steps}\label{afterword:tbl:lba-normalization-steps}
      \end{center}
    \end{table}

  \item\textbf{Can we do better than existing approaches?}

    Yes, we can:
    \begin{itemize}
    \item we improved the macro-averaged results of existing
      lexicon-generation methods with our proposed linear-projection
      algorithm;
    \item we increased the scores of fine-grained analysis by
      redefining the topologies of CRFs;
    \item our lexicon-based attention network outperformed many of
      its competitors on message-level classification;
    \item and, finally, we surpassed the discourse-unware baseline and
      other existing discourse-aware sentiment solutions with the
      proposed latent-marginalized CRFs and Recursive Dirichlet
      Process.
    \end{itemize}
\end{itemize}

\section*{Contributions}

Apart from answering the above questions and pushing the state of the
art for several major sentiment tasks on the PotTS and SB10k corpora,
we have also paved the way for other researchers who want to work on
the same topics by releasing the data and the code that we used in our
experiments:
\begin{itemize}
\item the Potsdam Twitter Sentiment (PotTS) corpus is available at:\\
  \url{https://github.com/WladimirSidorenko/PotTS};
\item scripts and executables used in our lexicon generation chapter
  can be downloaded
  from:\\ \url{https://github.com/WladimirSidorenko/SentiLex};
\item for our text normalization pipeline and fine-grained sentiment
  methods, please refer to:\\
  \url{https://github.com/WladimirSidorenko/TextNormalization};
\item furthermore, you can find our MLSA approaches
  at:\\ \url{https://github.com/WladimirSidorenko/CGSA};
\item and get all discourse-aware solutions
  from:\\ \url{https://github.com/WladimirSidorenko/DASA};
\item last but not least, we have released the discourse segmenter,
  the adapted DPLP parser, and our modified version of
  \textsc{RST-Tool}, which was adjusted to the annotation of
  multilogues, at:
  \begin{itemize}
  \item\url{https://github.com/WladimirSidorenko/DiscourseSegmenter},
  \item\url{https://github.com/WladimirSidorenko/RSTParser}, and
  \item\url{https://github.com/WladimirSidorenko/RSTTool}, respectively.
  \end{itemize}
\end{itemize}

In addition to open-sourcing all projects, we have also made a few
attempts to increase the visibility of our research with the following
publications:
\begin{itemize}
  \item the rule-based text normalization was described
    in~\cite{Sidarenka:13};
  \item the PotTS corpus was presented in~\cite{Sidarenka:16};
  \item in~\cite{Sidarenka:16b}, we summarized the evaluation of
    existing sentiment lexicons (a separate paper on the linear
    projection algorithm was withdrawn due to a mistake in the initial
    implementation);
  \item in~\cite{Sidarenka:16a} and~\cite{Sidarenka:17}, we described
    our initial experiments on message-level classification;
  \item furthermore, we introduced the SVM-based discourse segmenter
    in~\cite{Sidarenka:15};
  \item and sketched our pilot study of discourse annotation in
    Twitter in~\cite{Sidarenka:15a}.
\end{itemize}
Unfortunately, due to the lack of experience at the initial stage of
working on this dissertation and limited time at the concluding stage,
I\footnote{Throughout this work we have been using the scientific
  ``we,'' considering the reader as a companion in our marathon.  But
  because at this point I start describing the limitations of this
  work, which I am the only person responsible for, I would like to
  exclude the reader from this criticism by switching to the solitary
  ``I''.} was not able to publish more or at higher-level venues.  I
apologize for that.

\section*{Limitations}

Much to my regret, the initial lack of academic expertise has also
prevented me from running this scientific marathon faster, better,
and, most sadly, along more exciting places.  Alas, in the ``Sentiment
Analysis of German Twitter'', I have concentrated more on the
``Sentiment Analysis'' part, much at the cost of ``German Twitter''.
I wish I had tried out more sophisticated cross-lingual methods for
adapting English methods to German, elaborated more on linguistic
traits of microblogs, and addressed the social aspect of the Twitter
network.  The main reason why I have not done all these things is
that, six years ago, when I started working on this dissertation
completely from scratch, with neither code, nor data, nor any proper
plan in my head, I was so overwhelmed by the abundance of works on
opinion mining and text normalization that I decided to answer the
questions whether existing sentiment methods work, whether text
normalization helps, and whether I could improve on these methods,
first.  Regrettably, these questions have pretty much preoccupied me
since then.  Another reason why I refrained from addressing certain
topics (why, for example, I did not extend the Rhetorical Structure
Theory to multilogues) is because properly handling these problems
would require another dissertation and would certainly first need to
be done for English in order to get any international attention.

\section*{Final Remarks}

Nevertheless, I believe that with this thesis I have basically built a
theme park on the moon.  Although this park is still missing a slot
machine and a few carousels, it does have a nice card table and many
other kinds of funny amusements.  Another good thing about it is that
new carousels are now easier to build; existing amusement rides work
for free and better than in many other places; and, finally, the park
itself might still entertain its occasional guests.  You might have
enjoyed this theme park as well, or you might have not---I appreciate
either of your decisions and would like to thank you in any case for
your visit.
\begin{figure*}[htb]
  \centering \includegraphics[height=5em]{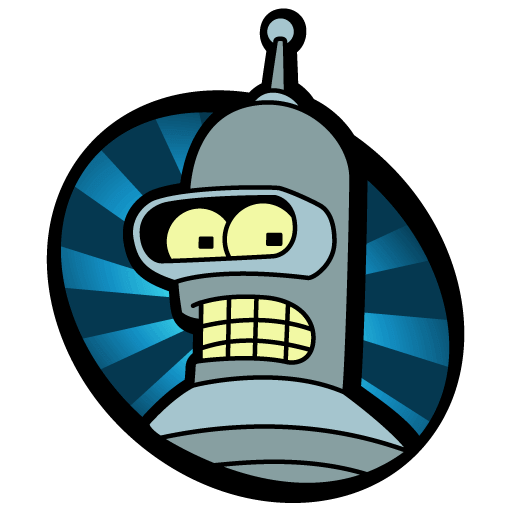}
\end{figure*}

\appendix

\chapter{Annotation Guidelines of the Sentiment Corpus}\label{chap:apdx:corp-guidelines}

{
  \setlength{\parindent}{0ex}
\tocless\section{Introduction}

In this assignment, your task is to annotate sentiments in a corpus of
Twitter messages.  We define \emph{sentiments} as polar (either
positive or negative) evaluative opinions about some persons,
entities, or events.  Your goal is to annotate both: text spans
denoting the opinions (\emph{sentiments}) and text spans signifying
the evaluated entities and events (\emph{sentiment targets}).  In
addition to that, you also have to label opinions' holders
(\emph{sentiment sources}) and lexical elements that might
significantly affect the polarity or the intensity of a sentiment.
These elements are:
\begin{itemize}
  \item\emph{polar terms}, which are words or phrases that
    unequivocally possess an evaluative lexical meaning in and of
    themselves (these are typically words like \emph{hassen}
    [\emph{hate}], \emph{bewundern} [\emph{admire}], \emph{sch\"on}
    [\emph{nice}] etc.);
  \item\emph{intensifiers} and \emph{diminishers} (or
    \emph{downtoners}), which are words and expressions that increase
    or decrease the evaluative sense of a polar term.  Examples of
    intensifiers are words like \emph{sehr} (\emph{very}),
    \emph{besonders} (\emph{especially}), or \emph{insbesondere}
    (\emph{particularly}).  Typical examples of diminishers are
    \emph{ein wenig} (\emph{a little}), \emph{ein bisschen} (\emph{a
      bit}), \emph{gewisserma\ss{}en} (\emph{to a certain degree}),
    etc.;
    \item and, finally, \emph{negations}, which are words or
      expressions that completely flip the polarity of a polar term or
      sentiment to the opposite (\eg{} \emph{nicht} gut [\emph{not}
        good] or \emph{kein} Talent [\emph{not} a talent]).
\end{itemize}

\tocless\section{Annotation Tool}

For annotating this corpus, you need to install \texttt{MMAX2}, a
freely available annotation tool, which you can download at:

\small\url{http://sourceforge.net/projects/mmax2/files/mmax2/mmax2_1.13.003/MMAX2_1.13.003b.zip/download}

After you have downloaded this file, unzip the received archive,
change to the newly created directory \texttt{1.13.003/MMAX2} in your
shell and execute the following commands: \code{chmod u+x~./mmax2.sh\\
  nohup~./mmax2.sh \&}
An \texttt{MMAX2} window will then appear on your screen.
If you have never used \texttt{MMAX2} before, please read its user
manual \texttt{mmax2quickstart.pdf}, which you can find in the
subdirectory \texttt{MMAX2/Docs} of the downloaded archive.

\tocless\section{Corpus Files}

You should also have received a copy of corpus files either as a
tar-gzipped archive or via a version control system.  In the former
case, you need to unpack the downloaded \texttt{.tgz} file using the
following command: \code{tar -xzf archive-name.tgz} After that, a
directory called \corpusDir{} will appear in your current folder.

You can find your annotation files in the subdirectory
\texttt{\corpusDir{}/corpus/annotator-ANNOTATOR\_ID}, where
ANNOTATOR\_ID is the ID number that has been previously assigned to
you by the supervisor.  In order to load an annotation file into your
\texttt{MMAX2} program, click on the menu \texttt{File -> Load}.  In
the displayed pop-up window, select the path to the
\corpusDir{}\texttt{/anno\-ta\-tor-ANNOTATOR\_ID} directory, and click
on one of the \texttt{*.mmax} files in this folder.


\tocless\section{Tags and Attributes}\label{sec:markables}

Below, you can find a short list of all labels and their possible
attributes that will be used in this assignment:
\begin{multicols}{2}
  \begin{enumerate}
  \item \texttt{sentiment}s with the attributes:
    \begin{enumerate}
    \item \texttt{polarity},
    \item \texttt{intensity},
    \item \texttt{sarcasm};
    \end{enumerate}
  \item \texttt{target}s with the attributes:
    \begin{enumerate}
    \item \texttt{preferred},
    \item \texttt{anaph-ref},
    \item \texttt{sentiment-ref};
    \end{enumerate}
  \item \texttt{source}s with the attributes:
    \begin{enumerate}
    \item \texttt{anaph-ref},
    \item \texttt{sentiment-ref};
    \end{enumerate}
  \item \texttt{polar-term}s with the attributes:
    \begin{enumerate}
    \item \texttt{polarity},
    \item \texttt{intensity},
    \item \texttt{sarcasm},
    \item \texttt{sentiment-ref};
    \end{enumerate}
  \item \texttt{intensifier}s with the attributes:
    \begin{enumerate}
    \item \texttt{degree},
    \item \texttt{polar-term-ref};
    \end{enumerate}
  \item \texttt{diminisher}s with the attributes:
    \begin{enumerate}
    \item \texttt{degree},
    \item \texttt{polar-term-ref};
    \end{enumerate}
  \item and, finally, \texttt{negation}s with the single attribute:
    \begin{enumerate}
    \item \texttt{polar-term-ref}.
    \end{enumerate}
  \end{enumerate}
\end{multicols}
A more detailed description of these attributes is given in the
following sections.

\tocless\subsection{sentiment}\label{sec:sentiment}
\paragraph{Definition.} \emph{Sentiments} are polar subjective
evaluative opinions about people, entities, or events.

According to this definition, a sentiment must always fulfill the
following three criteria:
\begin{itemize}
\item it has to be \textbf{polar}, \ie{} it must always reflect either
  positive or negative attitude to its respective target.  Neutral,
  non-evaluative statements such as \textit{Ich glaube, er wird heute
    fr\"uher kommen} (\textit{I think he will be earlier today}) must
  not be marked as \texttt{sentiment}s;

\item it has to be \textbf{subjective}, \ie{} you must not assign this
  tag to statements of objective facts, such as \textit{Beim Angriff
    wurden 14 Glasscheiben besch\"adigt} (\textit{14 glass plates were
    broken during the attack}), even if you have a personal polar
  attitude to such events.  Sentiments should always reflect \emph{the
    personal opinion of their holder, not yours};

\item a sentiment has to be \textbf{evaluative}, which means that it
  must always refer to an explicit target and judge about its
  properties.  You should not regard cases like \textit{Ich bin heute
    so gl\"ucklich} (\textit{I am so happy today}) as
  \texttt{sentiment}s, because such statements do not evaluate
  anything in particular, but only express general mood of the author.
\end{itemize}

\paragraph{Example.} Typical examples of sentiments are evaluative
sentences similar to the one shown below.
\begin{example}
  \sentiment{Ich mag den neuen James Bond Film nicht.}

  (\sentiment{I don't like the new James Bond movie.})\label{ex:sentiment}
\end{example}
This example expresses a personal subjective evaluation; the opinion
is strictly negative; and it also has an explicit evaluation
target---the \textit{movie}.  Therefore, we enclose this sentence in
the \texttt{sentiment} tags.

We also consider contrastive comparisons as a special type of
evaluations.  But unlike other sentiments, comparisons typically
express a relative subjective judgment, \ie{} an object is regarded as
better or worse than another, but we usually do not know whether the
author actually likes or dislikes any of them.  To distinguish such
cases, we have introduced a special \texttt{comparison} value for the
\texttt{polarity} attribute of this element, which you should use to
distinguish such cases.

You should not label as \texttt{sentiments} polar opinions whose truth
status is unknown.  These are sentences like \textit{Ich wei\ss{}
  nicht, ob ich meinen Bruder mag} (\textit{I don't know whether I
  like my brother}), where neither we nor the author actually know
whether the author likes or dislikes her brother.  Exceptions from
this rule are cases like \textit{Ich zweifle, dass er ein guter Mensch
  ist} (\textit{I doubt that he is a good man}) or \textit{Ich glaube
  nicht, dass er diesen Preis verdient hat} (\textit{I don't think
  that he has deserved this award}), which express author's
disagreement with positive evaluations and, consequently, acts as a
negative judgment.  Special care should be taken when dealing with
questions and subjunctive sentences though (see FAQ Section in the
extended
version\footnote{\url{https://github.com/WladimirSidorenko/PotTS/blob/master/docs/annotation_guidelines.pdf}}
of these guidelines).

\paragraph{Boundaries.} \texttt{sentiment} tags should
enclose both the evaluated object (target) and the evaluative
expression (typically a polar-term), \ie{} you should put these tags
around the \emph{minimal complete syntactic or discourse-level unit in
  which both (target and evaluation expression) appear together}.

In Example~\ref{exmp:book}, for instance, the evaluated object is
\textit{Buch} (\textit{book}), the evaluative expression is
\textit{langweiliges} (\textit{boring}), and the minimal syntactic
unit that simultaneously comprises both of these elements is the noun
phrase \textit{ein langweiliges Buch} (\textit{a boring book}).
Therefore, we annotate the noun phrase with the \texttt{sentiment}
tags, but do not enclose anything else inside these labels.
\begin{example}
  Auf dem Tisch lag \sentiment{ein langweiliges Buch}.

  (There was \sentiment{a boring book} on the table.)\label{exmp:book}
\end{example}
Sentiments are not restricted to just noun phrases, they can also be
expressed by complete clauses or even multiple sentences (discourse
units).  In these cases, a \texttt{sentiment} span still has to be
\emph{complete}, \ie{} it should capture the common syntactic or
discourse-level ancestor of the evaluative expression and its target,
as well as all other descendents of that common ancestor element; and
it has to be \emph{minimal}, \ie{} it should only enclose the closest
possible ancestor, without including its parent or sibling elements.

Example~\ref{exmp:petterson} demonstrates a sentiment expressed by a
clause:
\begin{example}
  Wir akzeptieren das, weil \sentiment{wir alle ein bisschen in
    Petterson verliebt sind}.

  (We accept this because \sentiment{we all are a little bit in love
    with Petterson}.)\label{exmp:petterson}
\end{example}
In this sentence, the evaluative statement is made about
\textit{Petterson}, who acts as sentiment's target; the author says
that they all \textit{in ihn verliebt sind} (\textit{are in love with
  him}), which is a subjective evaluation.  Both (target and
evaluative expression) appear in one verb phrase, whose head is the
link verb \textit{sein} (\textit{to be}).  Consequently, we enclose
the complete verb phrase including its grammatical subject
\textit{wir} (\textit{we}) in the \texttt{sentiment} tags.


\paragraph{Attributes.} After you have annotated the \texttt{sentiment} span,
you should next set the values of its attributes, which are summarized
in Table~\ref{tbl:sentiment}.
\begin{center}
  \begin{table}[htb!]
    \begin{tabular}{m{0.25\clmnwidth}>{\centering\arraybackslash}m{0.25\clmnwidth}m{0.92\clmnwidth}}
      \toprule
      \textbf{Attribute} & \textbf{Value} & \multicolumn{1}{c}{\textbf{Meaning}}\\\midrule

      \multirow[c]{3}{*}[-2.5cm]{polarity} & \multirow[c]{1}{*}{\textit{positive}} & sentiment expresses a positive attitude to
      its respective target, \eg{} \textit{Es war ein fantastischer
        Abend (It was a fantastic evening)};\\\cmidrule{2-3}

      & \textit{\shortstack{negative\\(default)}} & sentiment expresses a
      negative attitude to its respective target, \eg{} \textit{Seine
        Schwester ist einfach unausstehlich (His sister is simply
        obnoxious)}\\\cmidrule{2-3}

      & \textit{comparison} & sentiment
      expresses a comparison of two objects with preference given to
      one of them, \eg{} \textit{Mir gef\"allt das rote Kleid mehr als
        das blaue (I like the red dress more than the blue
        one)}\\\midrule


      \multirow[c]{3}{*}[-1.5cm]{intensity} & \textit{weak} & sentiment expresses a weak evaluative opinion,
      \eg{} \textit{Der Auftritt war mehr oder weniger gut (The
        appearance was more or less good)}\\\cmidrule{2-3}

      & \textit{\shortstack{medium\\(default)}} & sentiment has a middle
      emotional expressivity, \eg{} \textit{Mir hat das neue Album gut
        gefallen (I enjoyed the new album)}\\\cmidrule{2-3}

      & \textit{strong} & sentiment
      expresses a very emotional polar statement, \eg{} \textit{Dieses
        Festival war einfach umwerfend!!! (This festival was simply
        terrific!!!)}\\\midrule

       & \textit{true} & the opinion is
      derisive, \ie{} its actual polarity is the opposite of its
      literal meaning, although there are no immediate modifiers in
      the nearby context.  An example of a sarcastic sentiment is the
      following passage: \textit{Mein J\"ungerer ist in der Pr\"ufung
        durchgefallen.  Klasse! (My youngest has failed his exam.
        Well done!)}  In this case, you should set the polarity
      attribute of the sentiment to \texttt{negative} and the value of
      the \texttt{sarcasm} attribute to \texttt{true}.\\\cmidrule{2-3}

      \multirow[c]{-2}{*}[2.7cm]{sarcasm} &
      \textit{\shortstack{false\\(default)}} & no sarcasm is
      present---polar attitude has its literal meaning.\\\bottomrule
    \end{tabular}
    \caption{Attributes of \texttt{sentiment}s}\label{tbl:sentiment}
  \end{table}
\end{center}

\tocless\subsection{target}
\paragraph{Definition.} \emph{Targets} are objects or events
that are evaluated by a sentiment.

Because sentiments are required to be evaluative, there always must be
at least one target for each \texttt{sentiment} element.

\paragraph{Example.} An example of a sentiment target is given in
sentence~\ref{exmp:target}:
\begin{example}
Mein Bruder ist nicht begeistert von \target{dem neuen Call of Duty}.

(My brother is not impressed by \target{the new Call of
  Duty}.)\label{exmp:target}
\end{example}
In this message, the author tells us about the opinion of her brother
regarding the new version of a computer game.  The computer game is
the object of this evaluation, so you shall label it as a
\texttt{target}.

\paragraph{Boundaries.} As for \texttt{sentiment}s, you have to put
the \texttt{target} tags around the minimal complete syntactic or
discourse-level unit that denotes the evaluated entity or event.
These are usually noun phrases (\eg{} \textit{Mir wird's schlecht,
  wenn ich \target{diese Werbung} im Fernsehen sehe} [\textit{I feel
    sick when I see this \target{ad} on TV}]) or clauses (\eg{}
\textit{Ich hasse wenn \target{Voldemort mein Shampoo benutzt}.}
       [\textit{I hate when \target{Voldemort is using my shampoo}}]).

If a sentiment has multiple targets, you shall label each one of them
separately (see Example~\ref{exmp:trg-conj}).
\begin{example}
  Meiner Mutter haben \target{Nelken} und \target{Dahlien} immer gefallen.

  (My mother has always liked \target{carnations} and
  \target{dahlias}.)\label{exmp:trg-conj}
\end{example}
Similarly, in comparisons, you have to annotate each compared object
with a separate tag.  In addition to that, you should also set the
value of the \texttt{preferred} attribute to \texttt{false} for the
object that is dispreferred in the comparison (see Example~\ref{exmp:trg-comp}).
\begin{example}
  Ich mag \target[preferred=true]{Domino-Eis} mehr als
  \target[preferred=false]{Magnum}.

  (I like \target[preferred=true]{Domino ice cream} more than
  \target[preferred=false]{Magnum}.)\label{exmp:trg-comp}
\end{example}

\paragraph{Attributes.} Further possible attributes of \texttt{target}s
are given in Table~\ref{tbl:target}.
\begin{center}
  \begin{table}[htb!]
    \begin{tabular}{m{0.25\clmnwidth}>{\centering\arraybackslash}m{0.25\clmnwidth}m{0.92\clmnwidth}}\toprule
      \textbf{Attribute} & \textbf{Value} & \multicolumn{1}{c}{\textbf{Meaning}}\\\midrule

      \multirow{-2}{*}[-2cm]{preferred} &
      \textit{\shortstack{true\\(default)}} & in comparisons,
      this value means that the respective target is considered better
      than another compared object, \eg{} \textit{\emph{Die neue
          Frisur} passt ihr garantiert besser als die alte (\emph{The
          new hairstyle} suits her definitely better than the old
        one)};\\\cmidrule{2-3}

      & \textit{false} & in comparisons,
      this value signifies the target element that is considered worse
      than its counterpart, \eg{} \textit{Die zweite Saison von
        Breaking Bad war viel spannender als \emph{die dritte} (The
        second season of Breaking Bad was much more exciting than
        \emph{the third one})};\\\midrule

      sentiment-ref & \textit{\shortstack{$\longrightarrow$\\(directed
        edge)}} & a directed edge pointing from \texttt{target} to its
      respective \texttt{sentiment}.  You need to draw this edge in
      two cases:
      \begin{itemize}
      \item when the \texttt{target} is located at intersection of two
        different \texttt{sentiment}s (in this case, you should draw
        an edge from \texttt{target} to \texttt{sentiment}, which this
        \texttt{target} actually belongs to),

      \item when the target of an opinion is expressed outside the
        \texttt{sentiment} span;
      \end{itemize}\\\midrule

      anaph-ref & \textit{\shortstack{$\longrightarrow$\\(directed edge)}} &
      a directed edge pointing from \texttt{target} expressed by a
      pronoun or pronominal adverb to its respective non-pronominal
      antecedent (in order to draw this edge, you also need to
      annotate the antecedent as \texttt{target})\\\bottomrule
    \end{tabular}
    \caption{Attributes of \texttt{target}s}\label{tbl:target}
  \end{table}
\end{center}

\tocless\subsection{source}
\paragraph{Definition.} Sentiment \emph{sources} are immediate author(s) 
or holder(s) of evaluative opinions.  These are typically the author 
of a message, or officials whose opinion is cited.

If sentiment's holder is not explicitly mentioned in the tweet, it is
implicitly assumed that it is the user who wrote that microblog, and
you need not annotate anything as a source in this case.

\paragraph{Example.} An example of an explicitly mentioned source
is the pronoun \textit{Sie} (\textit{she}) in the following sentence.
\begin{example}
  \source{Sie} mag die neue Farbe nicht

  (\source{She} doesn't like the new color)\label{exmp:source}
\end{example}

Note that in citations you should only label the immediate person or
the institution whose original opinion is cited, but should not
annotate the citing person as a \texttt{source} (see
Example~\ref{exmp:source-citation}).
\begin{example}
  Laut Staatsanwalt soll die \source{Angeklagte} sich missbilligend \"uber
  ihren Vorgesetzten ge\"au\ss{}ert haben.

  (According to the attorney, the \source{defendant} had made
  disapproving remarks about her boss.)\label{exmp:source-citation}
\end{example}

\paragraph{Boundaries.} For determining the boundaries of
\texttt{source}s, you should proceed in a similar way as you did for
\texttt{target}s and \texttt{sentiment}s, \ie{} only annotate complete
minimal syntactic units.  Sources are most commonly expressed by noun
phrases.  As with \texttt{target}s, if the source of a sentiment is
expressed by multiple separate noun phrases, you should label each of
them separately (see Example~\ref{exmp:source2}).

\begin{example}
  \source{Ihr} und \source{ihrer Mutter} gef\"allt die neue Farbe
  nicht.\\ (Neither \source{she} and \source{her mother} likes the new
  color)\label{exmp:source2}
\end{example}

\paragraph{Attributes.} The attributes of the \texttt{source}
tag are fully identical to the attributes of the \texttt{target}
elements and are recapped in Table~\ref{tbl:source}.
\begin{center}
  \begin{table}[htb!]
    \begin{tabular}{m{0.25\clmnwidth}>{\centering\arraybackslash}m{0.25\clmnwidth}m{0.92\clmnwidth}}\toprule
      \textbf{Attribute} & \textbf{Value} & \multicolumn{1}{c}{\textbf{Meaning}}\\\midrule

      sentiment-ref & \textit{\shortstack{$\longrightarrow$\\(directed
        edge)}} & see Table~\ref{tbl:target}\\\midrule

      anaph-ref & \textit{\shortstack{$\longrightarrow$\\(directed edge)}} &
      see Table~\ref{tbl:target}\\\bottomrule
    \end{tabular}
    \caption{Attributes of \texttt{source}s}\label{tbl:source}
  \end{table}
\end{center}

\tocless\subsection{polar-term}
\paragraph{Definition.} \emph{polar-terms} are words or
phrases that have an inherent evaluative meaning.

\paragraph{Example.} An example of a polar-term is the word
\textit{ekelhaft} (\textit{disgusting}) in sentence~\ref{exmp:emo-expr1}.
\begin{example}
  Beim Aufr\"aumen des Zimmers haben wir einen
  \emoexpression{ekelhaften} Teller mit verschimmeltem Essen unter dem
  Bett gefunden.

  (When we cleaned the room, we found a \emoexpression{disgusting}
  plate with moldy food under the bed.)\label{exmp:emo-expr1}
\end{example}

In contrast to \texttt{source}s and \texttt{target}s, which should
only be annotated in the presence of a \texttt{sentiment}, you always
have to label polar terms in the text irrespective of any other
tags.

Note, however, that because many words and idioms are ambiguous and
can have several different meanings, it can often be the case that
only some of these meanings are evaluative and subjective.  In such
cases, you should only label such words if their actual sense in the
given context is polar.  If these words denote an objective entity or
fact, you must not use this tag.
\begin{example}
  Dieser Wein ist ein echtes \emoexpression{Juwel} in meiner
  Kollektion.

  (This wine is a real \emoexpression{jewel} in my collection.)

  Koh-i-Noor ist das teuerste Juwel heutzutage.

  (Koh-i-Noor is the most expensive jewel nowadays.)\label{exmp:polar-term-jewel}
\end{example}
In Example~\ref{exmp:polar-term-jewel}, for instance, the meaning of
the word \textit{Juwel} (\textit{jewel}) is metaphoric and subjective
in the first sentence, but literal and objective in the second
statement.  So you should only annotate this word as
\texttt{polar-term} in the former case, but disregard it in the
latter.

\paragraph{Boundaries.} \texttt{polar-term}s are typically
expressed by:
\begin{itemize}
  \item nouns, \eg{} \textit{Held (hero)}, \textit{Ideal (ideal)},
    \textit{Betr\"uger (fraudster)};

  \item adjectives or adverbs, \eg{} \textit{sch\"on (nice)},
    \textit{zuverl\"assig (reliably)}, \textit{hinterh\"altig
      (devious)}, \textit{heimt\"uckisch (insidiously)};

  \item verbs, \eg{} \textit{lieben (to love)}, \textit{bewundern (to
    admire)}, \textit{hassen (to hate)};

  \item idioms, \eg{} \textit{auf die Nerven gehen (to get on one's
    nerves)};

  \item smileys, \eg{} :), :-(, \smiley{}, \frownie{}. 
\end{itemize}
If a \texttt{polar-term} represents an idiomatic phrase, you shall
always annotate the complete idiom.  If a verb has an evaluative sense
only in conjunction with certain prepositions (\eg{} \textit{to go for
  sth.} in the sense of \textit{to like}), you shall annotate
both the verb and the preposition with a single pair of tags (check
the \texttt{MMAX} manual to see how to annotate discontinuous spans).

\paragraph{Attributes.}

When determining the \texttt{polarity} of a \texttt{polar-term}, you
should disregard any possible contextual modifiers such as
intensifiers or negations and set the value of this attribute to the
lexical (\ie{} \emph{prior}) polarity of that term (see
Example~\ref{exmp:polar-term-polarity}).
\begin{example}
Es war keine \emoexpression[polarity=positive]{gute} Idee.

(It was not a \emoexpression[polarity=positive]{good} idea.)\label{exmp:polar-term-polarity}
\end{example}
Apart from that, when determining the value of the \texttt{polarity}
attribute of a \texttt{polar-term}, you should analyze its polarity
from the perspective of the holder of the opinion towards the
evaluated object.  This means that in cases like \textit{Ich vermisse
  meine Freundin} (\textit{I miss my girlfriend}), the polarity of the
polar-term \textit{vermissen} (\textit{to miss}) is still positive
because the author has a positive attitude to his girlfriend, and
consequently feels sad about of her absence.

Further attributes of \texttt{polar-term}s include \texttt{intensity},
\texttt{sarcasm}, and \texttt{sentiment-ref}; their possible values
are summarized in Table~\ref{tbl:polar-term}.
\begin{center}
  \begin{table}[htb!]
    \begin{tabular}{m{0.25\clmnwidth}>{\centering\arraybackslash}m{0.25\clmnwidth}m{0.92\clmnwidth}}\toprule
      \textbf{Attribute} & \textbf{Value} & \multicolumn{1}{c}{\textbf{Meaning}}\\\midrule

      \multirow{2}{*}[-0.7cm]{polarity} & \textit{positive} & polar term has a positive evaluative
      meaning, \eg{} \textit{gut (good), verhimmeln (to ensky),
        Prachtkerl (corker)} etc.\\\cmidrule{2-3}

      & \textit{\shortstack{negative\\(default)}}
      & polar term expresses a negative evaluation of its target,
      \eg{} \textit{versauen (to botch up), rotzig (snotty),
        Dreckskerl (scum)} etc.\\\midrule


      \multirow{3}{*}[-1.1cm]{intensity} & \textit{weak} & polar-term has a weak evaluative sense, \eg{}
      \textit{solala (so-so), nullachtf\"unfzehn (vanilla),
        durchschnittlich (mediocre)} etc.\\\cmidrule{2-3}

      & \textit{\shortstack{medium\\(default)}} & polar-term has middle
      stylistic expressivity, \eg{} \textit{gut (good), schlecht
        (bad), robust (tough)} etc.\\\cmidrule{2-3}

      & \textit{strong} & polar-term
      expresses a very strong positive or negative evaluation, \eg{}
      \textit{allerbeste (bettermost), zum Kotzen (to make one puke),
        Kacke (shit)} etc.\\\midrule


      \multirow{2}{*}[-0.4cm]{sarcasm} & \textit{true} & polar-term is
      derisive, \ie{} its actual polarity is the opposite of its
      primary lexical sense even though there are no negations in the
      surrounding context\\\cmidrule{2-3}

      & \textit{\shortstack{false\\(default)}} & no sarcasm is present---the
      term has its literal polar meaning; this is the
      default\\\midrule


      sentiment-ref & \textit{\shortstack{$\longrightarrow$\\(directed
        edge)}} & an arrow pointing to the \texttt{sentiment} that this
      \texttt{polar-term} belongs to.  You should only draw this edge
      if a \texttt{polar-term} is located at an intersection of two
      \texttt{sentiment}s or outside of the \texttt{sentiment} span
      that it belongs to\\\bottomrule
    \end{tabular}
    \caption{Attributes of \texttt{polar-term}s}\label{tbl:polar-term}
  \end{table}
\end{center}

\tocless\subsection{intensifier}
\paragraph{Definition.} \emph{Intensifiers} are elements that increase
the expressivity or the evaluative sense of a polar term.

\paragraph{Example.}An example of intensifier is the word
\textit{sehr} (\textit{very}) in sentence~\ref{exmp:intensifier}.
\begin{example}
  Wir suchen eine \intensifier{sehr} zuverl\"assige Polin als
  Haushaltshilfe.

  (We are looking for a \intensifier{very} reliable Polish woman as
  domestic help.)\label{exmp:intensifier}
\end{example}
\paragraph{Boundaries.}

Intensifiers are usually expressed by adverbs or adjectives such as
\textit{sehr} (\textit{very}) or \textit{sicherlich}
(\textit{certainly}), but other ways of intensification are still
possible (see Example~\ref{exmp:intensifier-comp}).
\begin{example}
  Dieser Junge ist stark \intensifier{wie ein Pferd}.

  (This boy is strong \intensifier{as a
    horse}.)\label{exmp:intensifier-comp}
\end{example}

\paragraph{Attributes.} An \texttt{intensifier} must always
relate to some \texttt{polar-term}, and you always have to explicitly
show this relation by drawing a \texttt{polar-term-ref} edge from the
\texttt{intensifier} to its modified polar expression.

Further possible attributes of \texttt{intensifier}s are shown in
Table~\ref{tbl:intensifier}.
\begin{center}
  \begin{table}[htb]
    \begin{tabular}{m{0.25\clmnwidth}>{\centering\arraybackslash}m{0.25\clmnwidth}m{0.92\clmnwidth}}\toprule
      \textbf{Attribute} & \textbf{Value} & \multicolumn{1}{c}{\textbf{Meaning}}\\\midrule

      \multirow{-2}{*}[-1.25cm]{degree} & \textit{\shortstack{medium\\(default)}} & the intensifier moderately
      increases the polar sense of the polar term,
      \eg{} \textit{ziemlich (quite), recht (fairly)} etc.\\\cmidrule{2-3}

      & \textit{strong} & the intensifier strongly increases the polar
      sense and stylistic markedness of the polar term, \eg{}
      \textit{sehr (very), super (super), stark (strongly)}
      etc.\\\midrule


      polar-term-ref & \textit{\shortstack{$\longrightarrow$\\(directed
        edge)}} & a directed edge pointing from the intensifier to the
      \texttt{polar-term} whose meaning is being
      intensified\\\bottomrule
    \end{tabular}
    \caption{Attributes of \texttt{intensifier}s}\label{tbl:intensifier}
  \end{table}
\end{center}

\tocless\subsection{diminisher}
\paragraph{Definition.} \emph{Diminisher}s or \emph{downtoner}s are words or phrases
that decrease the polar lexical sense of a \texttt{polar-term}.

\paragraph{Example.} In Example~\ref{exmp:diminisher}, the
diminisher is expressed by the adverb \textit{weniger}
(\textit{less}).
\begin{example}
  \diminisher{Weniger} erfolgreiche Unternehmen verzichten auf externe
  Berater.\label{exmp:diminisher}

  The \diminisher{less} successful companies do not use external
  consultants.
\end{example}
\paragraph{Attributes.} Like intensifiers, diminishers must
always relate to a polar term, and you also have to explicitly show
this relation by using the \texttt{polar-term-ref} attribute; other
attributes of \texttt{diminisher}s mainly coincide with those of
intensifiers and are summarized in Table~\ref{tbl:diminisher}.
\begin{center}
  \begin{table}[hb]
    \begin{tabular}{m{0.25\clmnwidth}>{\centering\arraybackslash}m{0.25\clmnwidth}m{0.92\clmnwidth}}\toprule
      \textbf{Attribute} & \textbf{Value} & \multicolumn{1}{c}{\textbf{Meaning}}\\\midrule

      \multirow{2}{*}[-0.5cm]{degree} & \textit{\shortstack{medium\\(default)}} & diminisher moderately decreases
      the polar sense of its respective \texttt{polar-term},
      \eg{} \textit{wenig (few), bisschen (little)} etc.\\\cmidrule{2-3}

      & \textit{strong} & diminisher strongly
      decreases the polar sense of the \texttt{polar-term},
      \eg{} \textit{kaum (hardly)} etc.\\\midrule

      polar-term-ref & \textit{\shortstack{$\longrightarrow$\\(directed
        edge)}} & see Table~\ref{tbl:intensifier}\\\bottomrule
    \end{tabular}
    \caption{Attributes of \texttt{diminisher}s}\label{tbl:diminisher}
  \end{table}
\end{center}

\tocless\subsection{negation}
\paragraph{Definition.} \emph{Negation}s are elements that turn
the polarity of a \texttt{polar-term} to the opposite.

\paragraph{Example.} In Example~\ref{exmp:negation}, for
instance, the negative article \textit{kein} (\textit{not}) makes the
\emph{contextual} polarity of the word \textit{interessant}
(\textit{interesting}) negative, even though the prior semantic
orientation of this term is positive.
\begin{example}
Diese Geschichte war \"uberhaupt nicht \negation{interessant}!

This story was \negation{not} interesting at all!\label{exmp:negation}
\end{example}

The role of negations is closely related to that of diminishers.  In
order to help you better distinguish between these entities, we have
listed the most obvious differences between the two elements:
\begin{itemize}
  \item\textit{Semantic differences:} diminishers only decrease the
    lexical sense of an polar-term, a but part of its original sense
    still remains active (\ie{} \textit{a hardly understandable
      speech} is still understandable); negations, on the other hand,
    fully deny that meaning and turn it to the complete opposite
    (\textit{a not understandable speech} is absolutely
    unintelligible);

  \item\textit{Part-of-speech differences:} diminishers are usually
    expressed by adjectives or adverbs, whereas negations are
    typically represented by the negative article \textit{kein (no)},
    the negation particle \textit{nicht (not)}, or verbs or
    adjectives, \eg{} \textit{Es ist sehr zweifelhaft, dass die neue
      Version von Windows besser wird} (\textit{It is very doubtful
      that the new Windows version will be any better})

\end{itemize}

\paragraph{Attributes.} The only attribute of negations is the
mandatory edge \texttt{polar-term-ref}.  You have to draw this edge
from the \texttt{negation} to that \texttt{polar-term} that is
negated.  Like intensifiers and diminishers, negations must always
refer to at least one polar item.
\begin{center}
  \begin{table}
    \begin{tabular}{m{0.25\clmnwidth}>{\centering\arraybackslash}m{0.25\clmnwidth}m{0.92\clmnwidth}}\toprule 
      \textbf{Attribute} & \textbf{Value} &
      \multicolumn{1}{c}{\textbf{Meaning}}\\\midrule polar-term-ref &
      \textit{\shortstack{$\longrightarrow$\\(directed edge)}} & an edge from
      \texttt{negation} to the \texttt{polar-term} being
      negated\\\bottomrule
    \end{tabular}
    \caption{Attributes of \texttt{negation}s}\label{tbl:negation}
  \end{table}
\end{center}

\tocless\section{Summary}\label{sec:summary}

Summarizing all of the above, your task in this assignment is to find
subjective evaluative opinions about some entities or events.  You
need to annotate these opinions with the \texttt{sentiment} tags and
also determine the polarity and the intensity of the expressed
attitudes.  After that, you should assign the \texttt{target} tags to
objects or events that are evaluated, and label the holders of these
attitudes as \texttt{source}s.  Both, \texttt{source}s and
\texttt{target}s, can only exist in the presence of a
\texttt{sentiment}.

Another important task is to annotate words and phrases that have a
polar evaluative meaning.  We call these words \texttt{polar-term}s,
and you need to annotate them always, regardless of whether there is a
targeted sentiment or not.  If a \texttt{polar-term} is intensified,
diminished, or negated by another word or phrase, you should also
annotate the modifying element as well.

\tocless\section{Examples}\label{sec:examples}

We conclude these guidelines with a couple of real-world annotation
examples from our corpus, explaining our decisions for these
annotations.

\begin{example}
  \footnotesize WAS HABEN ALLE MIT
  \sentiment[polarity=negative,intensity=strong,sarcasm=false]{IHREN\\
    \emoexpression[polarity=negative,intensity=strong,sarcasm=false]{VERF*CKTEN}\\
    \target{GR\"UNEN AUGEN}}

  {\scriptsize(WHAT DO THEY ALL HAVE WITH
    \sentiment[polarity=negative,intensity=strong,sarcasm=false]{THEIR\\
      \emoexpression[polarity=negative,intensity=strong,sarcasm=false]{F*CKED}\\
      \target{GREEN EYES}})}\label{exmp:sarcasm}
\end{example}

\textbf{Explanation:} In this case, there is an evaluative opinion
about the green eyes of some persons.  It is, however, unclear what is
the author's attitude to the people themselves, we only can see that
she thinks that the eyes of these people are \textit{verf*ckt}
(\textit{f*cked}).  Therefore, the \texttt{target} of this sentiment is
the word \textit{Augen} (\textit{eyes}), and the \texttt{sentiment}
span should enclose the noun phrase comprising that target and its
evaluative term \textit{f*cked}. Since the polar term is an intense
abusive word, we set the polarity of this word and its enclosing
\texttt{sentiment} to \texttt{negative} and the intensity of both tags
to \texttt{strong}.\footnote{In the cases where we do not specify an
  attribute in the example, this attribute is assumed to have the
  default value.}

\begin{example}
  \footnotesize\sentiment[polarity=negative,intensity=medium,sarcasm=true]{Wo
    ist der
    \emoexpression[polarity=positive,intensity=strong,sarcasm=true]{\#Jubel}
    von \target{\#CDU} \target{\#CSU} \& \target{\#FDP} \"uber den Tod
    der Mieterin nach\\
    \#Zwangsr\"aumung?}

  {\scriptsize\sentiment[polarity=negative,intensity=medium,sarcasm=true]{Where
      is the
      \emoexpression[polarity=positive,intensity=strong,sarcasm=true]{\#exultation}
      of \target{\#CDU} \target{\#CSU} \& \target{\#FDP} about the
      death of the renter after
      forced\\ \#eviction?}}\label{exmp:sarcasm}
\end{example}

\textbf{Explanation:} In Example~\ref{exmp:sarcasm}, we have not
labeled \textit{Jubel von \#CDU~\ldots{} \"uber den Tod von \ldots{}}
(\textit{the exultation of the \#CDU~\ldots{} about the death
  of~\ldots{}}) as \texttt{sentiment}, because the truth status of
this statement is unknown.  But, on the other hand, the mere
hypothesis that a political party could experience a glee feeling
because of a renter's death is sarcastic.  We can recognize it from
the \texttt{polar-term} \textit{\#Jubel} (\textit{\#exultation}),
whose prior semantic orientation is positive, but which suggests a
negative attitude to the CSU party in this context, without any
explicit contextual modifiers.  Accordingly, we set the (prior)
\texttt{polarity} of the term to \texttt{positive}, the polarity of
its \texttt{sentiment} to \texttt{negative}, and the \texttt{sarcasm}
attribute of both labels to \texttt{true}.  Apart from having
different polarities, \texttt{sentiment} and \texttt{polar-term} also
have different intensities: since \textit{\#Jubel}
(\textit{\#exultation}) expresses a higher degree of excitement than
the word \textit{Freude} (\textit{joy}), we set its \texttt{intensity}
to \texttt{high}.  On the other hand, the overall sentiment expression
is rather subtle and does not show high exaggeration of the author.
So, we set the \texttt{intensity} of the \texttt{sentiment} to
\texttt{medium} rather than \texttt{high}.

Another non-trivial case is shown in Example~\ref{exmp:nested-2},
which we will analyze step by step:
\begin{example}
  \footnotesize RT @JochenFlasbarth~: Guter \#Spiegel-Titel~, wie
  Welzer~, Sloterdijk und andere Promi \#Nichtw\"ahler die Demokratie
  verspielen~: Tr\"age~, frustriert

  {\scriptsize(RT @JochenFlasbarth~: A good \#Spiegel title~, how
    Welzer~, Sloterdijk, and other celebrity non-voters squander the
    democracy~: Sluggish~, frustrated)}\label{exmp:nested-2}
\end{example}

\textbf{Explanation:} First of all, we have to look at words with
unambiguous lexical polarity (polar-terms), as they are our primary
cues for detecting sentiments.  This tweet features one positive
terms, \textit{guter} (\textit{good}), and there negative polar items,
\textit{verspielen} (\textit{to squander}), \textit{tr\"age}
(\textit{sluggish}), and \textit{frustriert} (\textit{frustrated}).
Since we have two sets of polar-terms with contradicting polarities,
it is most likely that there also are two sentiments---one positive
and one negative.  The positive evaluation obviously pertains to the
suggested \#Spiegel title ``wie Welzer~, Sloterdijk und andere Promi
\#Nichtw\"ahler die Demokratie verspielen: Tr\"age~, frustriert''
(``\textit{how Welzer~, Sloterdijk, and other celebrity non-voters
  squander the democracy: Sluggish~, frustrated}'').  The author finds
this title good, and the annotation of this sentiment then looks as
follows:
\begin{example}
  \sentiment[polarity=positive,intensity=medium,sarcasm=false,id=1]{RT
    \source[sentiment\_ref=1]{@JochenFlasbarth}~:\\ \emoexpression[polarity=positive,intensity=medium,sarcasm=false,
      sentiment\_ref=1]{Guter} \#Spiegel-Titel~,
    \target[sentiment\_ref=1]{wie Welzer~, Sloterdijk und andere Promi
      \#Nichtw\"ahler die Demokratie verspielen~:
      Tr\"age~,\\ frustriert}}

  (\sentiment[polarity=positive,intensity=medium,sarcasm=false,id=1]{RT
    \source[sentiment\_ref=1]{@JochenFlasbarth}~: A\\
    \emoexpression[polarity=positive,intensity=medium,sarcasm=false,
      sentiment\_ref=1]{good} \#Spiegel title~,
    \target[sentiment\_ref=1]{how Welzer~, Sloterdijk, and other
      celebrity non-voters squander the democracy~: Sluggish~,\\
      frustrated}})
\end{example}
The negative opinion, which is expressed by the terms
\textit{verspielen} (\textit{to squander}), \textit{tr\"age}
(\textit{sluggish}), and \textit{frustriert} (\textit{frustrated}),
obviously relates to the celebrity non-voters, \textit{Welzer} and
\textit{Sloterdijk}; so, we annotate this evaluation as:

\begin{example}
  \scriptsize
  \sentiment[{\tiny polarity=negative,intensity=medium,sarcasm=false,id=2}]{RT
    \source[sentiment\_ref=2]{@JochenFlasbarth}~: Guter
    \#Spiegel-Titel~, wie\\\target[sentiment\_ref=2]{Welzer}~,
    \target[sentiment\_ref=2]{Sloterdijk}\\und
    \target[sentiment\_ref=2]{andere Promi \#Nichtw\"ahler} die
    Demokratie\\
    \emoexpression[polarity=negative,intensity=medium,sarcasm=false,sentiment\_ref=2]{verspielen}~:\\
    \emoexpression[polarity=negative,intensity=medium,sarcasm=false,sentiment\_ref=2]{Tr\"age}~,\\ \emoexpression[polarity=negative,intensity=medium,sarcasm=false,sentiment\_ref=2]{frustriert}\\
  }

  {\scriptsize(\sentiment[polarity=negative,intensity=medium,sarcasm=false,id=2]{RT
      \source[sentiment\_ref=2]{@JochenFlasbarth}~: A good \#Spiegel
      title~, how\\\target[sentiment\_ref=2]{Welzer}~,
      \target[sentiment\_ref=2]{Sloterdijk},\\ and
      \target[sentiment\_ref=2]{other celebrity non-voters}\\
      \emoexpression[polarity=negative,intensity=medium,sarcasm=false,sentiment\_ref=2]{squander}
      the democracy~:\\
      \emoexpression[polarity=negative,intensity=medium,sarcasm=false,
        sentiment\_ref=2]{Sluggishly}~,\\
      \emoexpression[polarity=negative,intensity=medium,sarcasm=false,
        sentiment\_ref=2]{frustrated}\\
    })}
\end{example}

In both cases, \textit{@JochenFlasbarth} is the original author of the
cited opinion, so we should label it as a \texttt{source}.  But since
there are two sentiment relations, we assign this tag twice, drawing
an edge (in our example denoted by attribute \texttt{sentiment\_ref})
to the respective \texttt{sentiment} element in each case.}


\chapter{Gradient Computation of the Optimized Projection
  Line}\label{chap:apdx:lex-grad}

In order to prove the correctness of the gradient shown in
Equation~\ref{eq:prj-line-grad}, let us first compute the partial
derivative of the optimized distance function $f =
\sum_{\vec{p}_+}\sum_{\vec{p}_-}\frac{1}{2}{\left(%
\frac{\vec{b}\cdot\left(\vec{p}_+ - \vec{p}_-\right)}{\vec{b}^2}%
\vec{b}\right)}^{2}$ w.r.t.\ to a single element $\vec{b}_j$ of the
projection vector $\vec{b}$.  Assuming that the length of this vector
is normalized at each iteration step prior to calculating the
derivative, we obtain:
{\small
  \begin{align}
    \begin{split}
    \frac{\partial}{\partial\vec{b}_j}f &= %
    \frac{\partial}{\partial\vec{b}_j}\sum_{\vec{p}_+}\sum_{\vec{p}_-}\frac{1}{2}{\left(%
\frac{\vec{b}\cdot\left(\vec{p}_+ - \vec{p}_-\right)}{\vec{b}^2}%
\vec{b}_j\right)}^2\\
&=\sum_{\vec{p}_+}\sum_{\vec{p}_-}\gamma\vec{b}_j%
\frac{\partial}{\partial\vec{b}_j}%
\frac{\vec{b}\cdot\left(\vec{p}_+ - \vec{p}_-\right)}{\vec{b}^2}\vec{b}_j\\
&=\sum_{\vec{p}_+}\sum_{\vec{p}_-}\gamma\vec{b}_j%
\left(\frac{{(\vec{p}_+ - \vec{p}_-)}_{j}\vec{b}^2 - 2\gamma\vec{b}_j}{\vec{b}^4}\vec{b}_j%
+\frac{\gamma}{\vec{b}^2}\right)\\
&=\sum_{\vec{p}_+}\sum_{\vec{p}_-}\gamma\vec{b}_j%
\left({\left(\vec{p}_+ - \vec{p}_-\right)}_j\vec{b}_j - 2\gamma\vec{b}^2_j+\gamma\right)\\
&=\sum_{\vec{p}_+}\sum_{\vec{p}_-}\gamma\left(%
{\left(\vec{p}_+ - \vec{p}_-\right)}_j\vec{b}^2_j - 2\gamma\vec{b}_j\vec{b}^2_j%
+\gamma\vec{b}_j\right),\label{eq:prj-line-partial}%
    \end{split}%
  \end{align}}%
where $\gamma$ is defined as previously:
{\small
  \begin{align*}
    \gamma = \vec{b}\cdot\left(\vec{p}_+ - \vec{p}_-\right).%
  \end{align*}}%
Since Expression~\ref{eq:prj-line-partial} is identical for all $j$,
we can estimate the final form of the gradient as: {\small
  \begin{align}
    \nabla f &= \sum_{\vec{p}_+}\sum_{\vec{p}_-}\gamma\left(%
\left(\vec{p}_+ - \vec{p}_-\right)\vec{b}^2 - 2\gamma\vec{b}\vec{b}^2%
+\gamma\vec{b}\right)\\
&= \sum_{\vec{p}_+}\sum_{\vec{p}_-}\gamma\left(\Delta - \gamma\vec{b}\right),%
\end{align}}%
which is exactly the solution we provide in
Equation~\ref{eq:prj-line-grad}.


\chapter{CRF Training and Inference}\label{chap:apdx:crf-inference}


\tocless\section{Training}

Traditionally, the main objective of CRF's training consists in
finding such feature parameters $\bm{\Theta}$ that maximize the
log-likelihood of a training set
$\mathcal{D}=\{(\bm{x}[m],\bm{y}[m])\}_{m=1}^M$ (where $M$ represents
the total number of training examples, $\bm{x}[m]$ denotes the feature
vector of the $m$-th example, and $\bm{y}[m]$ stands for the vector of
its gold labels):
\begin{equation}\label{eqn:llhood}\scriptstyle
  \begin{split}
    \bm{\Theta} = \argmax_{\bm{\Theta}}
    \sum\limits_{m=1}^{M}\ell_{\bm{Y}|\bm{X}} = \argmax_{\bm{\Theta}}
    \sum\limits_{m=1}^{M}\ln{p_{\bm{\Theta}}(\bm{y}[m]|\bm{x}[m])}.
  \end{split}
\end{equation}
The conditional probability of gold labels for the $m$-th training
instance ($p_{\bm{\Theta}}(\bm{y}[m]|\bm{x}[m])$) is typically
computed as:
\begin{equation*}\label{eqn:prob}\scriptstyle
  \begin{split}
    p_{\bm{\Theta}}(\bm{y}[m]|\bm{x}[m]) =
    \frac{\exp\Big(\sum\limits_{i=1}^n\sum\limits_{k}\theta_{k}\mathit{f}_{k}(\bm{y}[m],
      \bm{x}, i)\Big)}{Z_{\bm{x}}},
  \end{split}
\end{equation*}
where $n$ represents the total length of that instance (in the case of
sentences, $n$ is usually the number of tokens); $\theta_k$ and
$\mathit{f}_k$ are the weight and the value of the $k$-th feature; and
$Z_{\bm{x}}$ is a normalization factor, which is estimated over all
possible features $f$ and label assignments $\mathcal{Y}^n$:
\begin{equation*}\small
  \begin{split}
    Z_{\bm{x}} \defeq \sum\limits_{\bm{y}' \in
      \mathcal{Y}^n}\exp\Big(\sum\limits_{i=1}^n\sum\limits_{k}\theta_{k}\mathit{f}_{k}(\bm{y}',
    \bm{x}[m], i)\Big).
  \end{split}
\end{equation*}
The partial derivatives of feature weights, which are needed for
optimizing the log-likelihood in Equation~\ref{eqn:llhood}, are known
to be equal to the difference between the empirical and model's
expectation of features over the whole training corpus:
\begin{equation}\label{eqn:pderiv}\small
  \begin{split}
    \frac{\partial}{\partial{\theta_k}}\ell_{\bm{Y}|\bm{X}} = &
    \sum\limits_{m=1}^M\sum\limits_{i=1}^n\big(\mathit{f}_k(\bm{y}[m],
    \bm{x}[m], i) - \bm{E}_{\bm{\Theta}}[\mathit{f}_k(\mathcal{Y}^n, \bm{x}, i)] \big)
  \end{split}
\end{equation}

\tocless\subsection{Linear-chain CRFs}

In the case of first-order linear-chain CRFs, these derivatives are
usually estimated with the help of the forward-backward algorithm (a
specific case of the belief-propagation method~[\citeauthor{Pearl:82},
  \citeyear{Pearl:82}]), which can be briefly described as follows.

For each position $i$ of training instance $\bm{x}[m]$ and for each
label $y$ of tagset $\mathcal{Y}$, one first computes the forward
score $\alpha[y][i]$ as:
\begin{equation}\label{eqn:folc-alpha}\small
  \begin{split}
    \alpha[y][i] =& \sum\limits_{y' \in\mathcal{Y}}\alpha[y'][i-1]t(y',y,i-1,i)s(y,i)
  \end{split}
\end{equation}
where $t(y',y,i-1,i)$ is the exponentiated sum of transition features
$\mathit{f}_t(y', y, \bm{x}, i-1, i)$ (which denote the transition
from label $y'$ at position $i-1$ to label $y$ at position $i$)
multiplied with their respective weights $\theta_t$:
\begin{equation*}\label{eqn:sm-trans-score}\small
  \begin{split}
    t(y',y,i-1,i)\defeq\exp\Big(\sum\limits_{t}\theta_t\mathit{f}_t(y',y,\bm{x},i-1,i)\Big).
  \end{split}
\end{equation*}
Similarly, $s(y, i)$ denotes the exponent of the sum of state features
$f_s$ times their weights $\theta_s$:
\begin{equation*}\label{eqn:sm-trans-score}\small
  \begin{split}
    s(y,i)\defeq\exp\Big(\sum\limits_{s}\theta_s\mathit{f}_s(y, \bm{x},
    i)\Big).
  \end{split}
\end{equation*}

The normalizing factor $Z_{\bm{x}}$ is then easily estimated as the
sum of all values from the last column of matrix $\alpha$:
\begin{equation*}\small
  Z_{\bm{x}} = \sum\limits_{y \in \mathcal{Y}}\alpha[y][n].
\end{equation*}

After estimating the forward scores, one compute the backward scores
$\beta$ by applying the same procedure in reverse---from right to
left:
\begin{equation}\label{eqn:folc-beta}\small
  \begin{split}
    \beta[y][i] =& \sum\limits_{y' \in
      \mathcal{Y}}\beta[y'][i+1]t(y,y',i,i+1)s(y',i+1).%
  \end{split}
\end{equation}

The marginal probabilities $p_m$ of state and transition features are
then estimated as:
\begin{equation*}\label{eqn:folc-tmarginal}\small
  \begin{split}
    p_m(\mathit{f}_s(y, \bm{x},i)) =&\frac{1}{Z_{\bm{x}}}%
    \alpha[y][i]\beta[y][i];\\%
    p_m(\mathit{f}_t(y', y, \bm{x}, i-1, i)) &=\frac{1}{Z_{\bm{x}}}%
    \alpha[y'][i-1] \beta[y][i] s(y,i).
  \end{split}
\end{equation*}

Knowing these probabilities, one can easily obtain the gradient of
feature weights using Equation~\ref{eqn:pderiv}.

\tocless\subsection{Semi-Markov CRFs} In contrast to linear-chain
CRFs, semi-Markov conditional random fields do not model transitions
between identical labels (\eg{} \textsc{Source} $\rightarrow$
\textsc{Source}), but instead try to partition the input into
contiguous spans of identical tags and infer the most likely label
assignment for these spans.

In order to do so, the model first determines the maximum possible
length ($L$) of a segment with identical labels that exists in the
training set.  The forward and backward scores are then calculated as:
\begin{equation}\label{eqn:sm-alpha}\small
  \begin{split}
    \alpha[y][i] =&
    \sum\limits_{d=0}^{L-1}\sum\limits_{\{y'\in\mathcal{Y}|y'\ne{}y\}}%
    \alpha[y'][i-d-1]\\%
    &\times t(y',y,i-d-1,i-d)s(y,[i-d,i]);
  \end{split}
\end{equation}

\begin{equation}\label{eqn:sm-beta}\small
  \begin{split}
    \beta[y][i] =& \sum\limits_{d=0}^{L-1}%
    \sum\limits_{\{y'\in\mathcal{Y}|y'\ne{}y\}}\beta[y'][i+d+1]\\%
    &\times t(y,y',\bm{x},i+d,i+d+1)s(y,[i,i+d]);
  \end{split}
\end{equation}
where $s(y,[i,i+d])$ is the exponentiated sum of all state features
$s(y, j)$ that are activated on the interval $[i,\dots,i+d]$.

The marginal probabilities of state and transition features are then
computed as:
\begin{equation*}\label{eqn:sm-marginal}\small
  \begin{split}
    p_m(&\mathit{f}_s(y,\bm{x},[i-d,i])) =\frac{1}{Z_{\bm{x}}}%
    s(y,[i-d,i])\\%
    &\times\sum\limits_{\{y'\in\mathcal{Y}|y'\ne{}y\}}%
    \alpha[i-d-1][y']t(y',y,i-d-1,i-d)\\%
    &\times\sum\limits_{\{y''\in\mathcal{Y}|y''\ne{}y\}}%
    \beta[i+1][y'']t(y,y'',i,i+1);\\%
  \end{split}
\end{equation*}
and
\begin{equation*}\small
  \begin{split}
    p_m(\mathit{f}_t(y',y,\bm{x},[i-d,i]))&=\frac{1}{Z_{\bm{x}}}\alpha[y'][i-1]\beta[y][i]\\%
    &\times{}t(y',y,i-1,i).%
  \end{split}
\end{equation*}


\tocless\subsection{Higher-order CRFs}

In contrast to first-order models, which only consider the scores of
one immediate label to the left when computing the $\alpha$ values or
one immediate label to the right when estimating the $\beta$ scores,
higher-order CRFs keep separate track of each \emph{sequence of
  labels} that might precede or follow the currently analyzed token.

In particular, instead of simply computing the scores for each tagset
label $y \in \mathcal{Y}$ at each sentence position $i$, higher-order
conditional random fields estimate these values for complete sequence
of tags $y_{1}, \ldots, y_{d}$, where $d$ is the order of the model.

This extension is possible for both linear-chain- and semi-Markov
CRFs, and the way of estimating forward and backward scores as well as
computing marginal probabilities $p_m$ is almost identical to the
respective original implementations.  The only differences in the
higher-order case are that
\begin{itemize}
  \item it now becomes possible to use state and transition features
    that are associated with label chains up to length $d$ and not
    only single tags (\eg{} instead of using a state feature which
    tells that verbs starting with ``mis'' are likely to be
    \texttt{sentiment}s, one can refine the feature function and say
    that such words very probably represent \texttt{sentiment}s
    preceded by \texttt{source}s, \ie{} are associated with the label
    sequence <\texttt{source}, \texttt{sentiment}>);
  \item secondly, when estimating the scores $\alpha[y_{1}, \ldots,
    y_{d}][i]$ and $\beta[y_{1}, \ldots, y_{d}][i]$, one does not
    simply iterate over all cells of the previous or next column of
    the corresponding matrix, but only considers those preceding or
    following states that allow the label sequence $y_{1}, \ldots,
    y_{d}$ at the $i$-th position.  That is,
    Equations~\ref{eqn:folc-alpha} and~\ref{eqn:folc-beta} become:
\begin{equation*}
  \begin{split}
    \alpha[y_{1}, \ldots, y_{d}][i] =& \sum\limits_{\{y'_{1},\ldots,y'_{d}\in\mathcal{Y}^d|y'_{2},\ldots,y'_{d}=y_{1},\ldots,y_{d-1}\}}\alpha[y'_{1},\ldots,y'_{d}][i-1]\\
    &\times t\left(\left(y'_{1},\ldots,y'_{d}\right),\left(y_{1},\ldots,y_{d}\right),i-1,i\right)s\left((y_{1},\ldots,y_{d}),i\right)
  \end{split}
\end{equation*}
and
\begin{equation*}
  \begin{split}
    \beta[y_{1},\ldots,y_{d}][i] =& \sum\limits_{\{y'_{1},\ldots,y'_{d}\in\mathcal{Y}^d|y_{2},\ldots,y_{d} = y'_{1},\ldots,y'_{d-1}\}}\beta[y'_{1},\ldots,y'_{d}][i+1]\\
    &\times t\left(\left(y_{1},\ldots,y_{d}\right),\left(y'_{1},\ldots,y'_{d}\right),i,i+1\right)s\left((y'_{1},\ldots,y'_{d}),i+1\right),
  \end{split}
\end{equation*}
respectively.  The same change also applies to
Equations~\ref{eqn:sm-alpha} and~\ref{eqn:sm-beta} in the case of
semi-Markov models.
\end{itemize}

\tocless\subsection{Tree-structured CRFs}

The main difference between applying the belief-propagation algorithm
to trees instead of linear chains is that the inference flow happens
in a ``vertical'' way---from tree's leaves to its root and vice
versa---whereas in the standard forward-backward setting, we typically
compute the scores ``horizontally''---from the left-most word of a
sequence to the right-most one and then in the opposite direction.

More precisely, the $\alpha$ and $\beta$ scores for trees are
estimated as:
\begin{equation*}\label{eqn:tree-alpha-beta}\small
  \begin{split}
    \alpha[y][p] =& \prod\limits_{c \in \text{children}(p)}\left(%
    \sum\limits_{y' \in \mathcal{Y}} \alpha[y'][c] t(y',y,\bm{x},c, p)\right)%
    s(y, p);\\%
    \beta[y][c] =& \sum \limits_{y' \in \mathcal{Y}} %
    \frac{\alpha[y'][p]\beta[y'][p]}{\alpha_{c\rightarrow{}p}}%
    t(y, y', \bm{x},c,p);
  \end{split}
\end{equation*}
where $p$ is the index of the parent node of token $c$, and
$\alpha_{c\rightarrow{}p}$ is the part of the $\alpha$ score of token
$p$ that has been previously propagated to it from its child $c$:
\begin{equation*}\small
  \begin{split}
 \alpha_{c \rightarrow{} p} &\defeq \sum\limits_{y'' \in \mathcal{Y}} \alpha[y''][c] t(y'',y,\bm{x},c,p)
  \end{split}
\end{equation*}

The normalizing factor $Z_{\bm{x}}$ and marginal probabilities of
state features are calculated in the same way as for the linear-chain
models with the only difference that the partition factor $Z$ is
computed as the sum of the $\alpha$-scores of the root word $r$ and
not of the last word $n$ of the instance.

The marginal probabilities of transition features are computed using
the following equation:
\begin{equation*}\label{eqn:tree-tmarginal}\small
  \begin{split}
    p_m(\mathit{f}_t(y', y, \bm{x},c,p)) =&%
    \frac{\alpha[y'][c]t(y', y, c, p)\alpha[y][p]\beta[y][p]}{\alpha_{c\rightarrow{}p}Z_{\bm{x}}}.
  \end{split}
\end{equation*}

\tocless\section{Inference}

Once model parameters have been learned, one applies the optimized
model to new, unseen instances in order to predict their most probable
labels---a task which is commonly refered to as \emph{inference}.

For CRFs, inference actually boils down to computing the matrix
$\alpha$ with the following minor modifications:
\begin{itemize}
\item First of all, instead of taking the sum over all previous labels
  $y'\in\mathcal{Y}$ when computing $\alpha[y][i]$, one only estimates
  the maximum possible score for that cell that is possible
  w.r.t.\ the probabilities of its preceding labels;

\item Second, apart from storing the maximum (unnormalized)
  probability of label $y$ at the $i$-th position, one also stores the
  label at the previous position ($i-1$) that has lead to the maximum
  value of $\alpha[y][i]$.
\end{itemize}
In other words, in the case of linear-chain CRFs, we transform
Equation~\ref{eqn:folc-alpha} into:
\begin{equation}\label{eqn:folc-viterbi}\small
  \begin{split}
    \alpha[y][i] =& <\max_{y' \in\mathcal{Y}}\left(a(y',y,i-1,i)\right),
    \argmax_{y' \in\mathcal{Y}}\left(a(y',y,i-1,i)\right)>
  \end{split}
\end{equation}
where
\begin{equation*}\small
  \begin{split}
    a(y',y,i-1,i) \defeq \alpha[y'][i-1][0]t(y',y,i-1,i)s(y,i).
  \end{split}
\end{equation*}

We similarly modify the $\alpha$-computation in the semi-Markov case,
but this time, apart from remembering the highest possible probability
of the $y$-th label at the $i$-th position and its most likely
predecessor, we also need to store the most probable length of the tag
span $y$, \ie{}:
\begin{equation*}\small
  \begin{split}
    \alpha[y][i] =& <\max_{y' \in\mathcal{Y}, d \in [1,\ldots,L]}\left(a(y',y,d,i)\right),
    \argmax_{y' \in\mathcal{Y}, d \in [1,\ldots,L]}\left(a(y',y,d,i)\right)>
  \end{split}
\end{equation*}
with $a$ now defined as:
\begin{equation*}\small
  \begin{split}
    a(y',y,d,i) \defeq \alpha[y'][i-d-1][0]t(y',y,i-d-1,i-d)s(y,[i-d,i]).
  \end{split}
\end{equation*}

Finally, in the case of tree-structured CRFs, we could have basically
completely re-used the formula from Equation~\ref{eqn:folc-viterbi} if
each tree node only had one child.  But since, most of the time, this
is rarely the case, we need to circumvent the need for storing
multiple child labels in a single $\alpha$ cell because this
significantly slows down the inference due to additional memory
allocation on the fly.  The way we do that is by applying the
following trick: instead of storing in each cell $\alpha[y][i]$ the
maximum possible score for the $y$-th tag at the $i$-th position and
the labels of its children that have lead to this score, we store the
score and the most likely tag of the $i$-th node that yielded the
maximum possible value for the $y$-th tag at the parent position,
\ie{}:
\begin{equation}\small
  \begin{split}
    \alpha[y][c] =& <\max_{y' \in\mathcal{Y}}\left(a(y',y,c,p)\right),
    \argmax_{y' \in\mathcal{Y}}\left(a(y',y,c,p)\right)>
  \end{split}
\end{equation}
where
\begin{equation*}\small
  \begin{split}
    a(y',y,i-1,i) \defeq \alpha[y'][c][0]t(y',y,c,p)s(y,p).
  \end{split}
\end{equation*}
with $c$ denoting the index of the child, and $p$ standing for the
index of the parent.

After computing the scores for the final node, we scan the last (root)
column of the $\alpha$ matrix for the maximum value and trace back the
complete sequence of labels that has yielded this score---a procedure
which is commonly known as the Viterbi algorithm.

\bibliographystyle{apalike}
\bibliography{bibliography}

\end{document}